\pdfoutput=1

\documentclass[11pt]{article}
\usepackage[]{naacl2021}

\usepackage{times}
\usepackage{latexsym}
\usepackage{amsmath, amssymb}
\usepackage{resizegather}
\usepackage{graphicx}
\usepackage{subcaption}
\usepackage{booktabs}
\usepackage{amsthm}
\usepackage{cleveref}
\usepackage{xcolor}
\crefformat{section}{\S#2#1#3}
\crefformat{subsection}{\S#2#1#3}
\crefformat{subsubsection}{\S#2#1#3}
\crefrangeformat{section}{\S\S#3#1#4 to~#5#2#6}
\crefmultiformat{section}{\S\S#2#1#3}{ and~#2#1#3}{, #2#1#3}{ and~#2#1#3}
\usepackage[labelformat=simple]{subcaption}

\theoremstyle{definition}
\newtheorem{definition}{Definition}
\usepackage{multirow}

\usepackage[linesnumbered,ruled,vlined]{algorithm2e}
\SetKwInput{KwInput}{Input}
\SetKwInput{KwOutput}{Output}
\usepackage[T1]{fontenc}
\usepackage[utf8]{inputenc}

\usepackage{microtype}

\title{Low Anisotropy Sense Retrofitting (LASeR) : Towards Isotropic and Sense Enriched Representations}

\author{Geetanjali Bihani \and Julia Taylor Rayz \\
        Department of Computer and Information Technology \\ Purdue University \\ \texttt{\{gbihani,jtaylor1\}@purdue.edu}}

\begin{document}
\maketitle
\begin{abstract}

Contextual word representation models have shown massive improvements on a multitude of NLP tasks, yet their word sense disambiguation capabilities remain poorly explained. To address this gap, we assess whether contextual word representations extracted from deep pretrained language models create distinguishable representations for different senses of a given word. We analyze the representation geometry and find that most layers of deep pretrained language models create highly anisotropic representations, pointing towards the existence of representation degeneration problem in contextual word representations. After accounting for anisotropy, our study further reveals that there is variability in sense learning capabilities across different language models. Finally, we propose \textbf{LASeR}, a ‘Low Anisotropy Sense Retrofitting’ approach that renders off-the-shelf representations isotropic and semantically more meaningful, resolving the representation degeneration problem as a post-processing step, and conducting sense-enrichment of contextualized representations extracted from deep neural language models. 

\end{abstract}

\section{Introduction}

Distributional word representations, developed using large-scale training corpora, form an integral part of the modern NLP methodological paradigm. The advent of deep pre-trained neural language models such as BERT \citep{devlin2018bert} and GPT-2 \citep{radford2019language} has led the shift towards the development of contextualized word representations. Unlike static word representation models, such as \textit{word2vec} \citep{mikolov2013efficient} and \textit{fastText} \cite{bojanowski2017enriching}, which conflate multiple senses of a word within a single representation, contextual word representation models assign as many representations to a word as the number of contexts it appears in. The preference for contextual word representations can be attributed to the significant improvements they have achieved in a wide variety of NLP tasks including question answering, textual entailment, sentiment analysis \citep{peters2018deep, devlin2018bert} and commonsense reasoning \citep{da2019cracking, sap-etal-2020-commonsense}, to name a few.

To utilize contextual word representations as knowledge resources, it is necessary to determine their ability to mirror the linguistic relations employed in language \citep{schnabel2015evaluation}. There is a growing body of literature that assesses whether contextual representations encode information about word-senses, where each word-sense portrays an aspect of the meaning of a given word in a given context \citep{jurafsky2019speech}. A recent analysis by \citet{nair2020contextualized} reported that contextual word representations can learn human-like word sense knowledge, where they compared cosine relatedness between homonyms and polysemous word senses against human sense-related judgements. When calculating cosine relatedness, such studies assume the encoded vector space to be isotropic in nature. Geometrically, isotropy in a vector space is defined as vectors being uniformly distributed across all directions, instead of occupying a narrow cone \citep{ethayarajh2019contextual, mu2018all}. Recent studies point towards anisotropy (lack of isotropy) in contextual word representations \citep{ethayarajh2019contextual, zhang2020revisiting}, which affects prior conclusions regarding word-sense information encoded in vector spaces. For example, in an isotropic vector space, if cosine relatedness between word representations $A$ and $B$ is $0.9$, we conclude them to be highly similar. But, if the vector space is anisotropic, where cosine relatedness between randomly sampled words is $0.95$, then the representations $A$ and $B$ are deemed less similar than randomly sampled words. This shows that the existence and the extent of anisotropy in the vector space affects conclusions regarding whether representations are actually similar or merely a product of representation degeneration. Hence, when evaluating the sense learning capabilities of deep pretrained language models through vector relatedness measures, accounting and adjusting for vector space anisotropy becomes necessary. 

In this regard, our work presents three key contributions. First, we analyze and adjust for anisotropy across contextual representations extracted from all layers of four language models (BERT, GPT-2, XLNet and ELECTRA). The representation space for each model encodes anisotropy, varying in terms of number and strength of common directions in model representations. We find that models learning unidirectional context create more anisotropic representations than models learning bidirectional context. Second, we observe that sense information is not equally encoded in all models, where (pseudo) bidirectional models learn to disambiguate word senses better than others. Moreover, sense information is better retained in the lower layers and significantly reduces in the upper model layers due to the representations getting more contextualized. Third, to address these preliminary findings and to contribute towards the creation of sense-coherent representations, we propose \textbf{LASeR}, a ‘Low Anisotropy Sense Retrofitting’ approach, bringing word representations closer to the goal of mirroring lexical semantic relations present in natural language while removing artifacts of representation degeneration from learned representations. Thus, we combine vector space transformation and knowledge-based vector specialization methods to create more isotropic and sense enriched representations, ensuring that we retain the distributional properties learnt during pretraining, while aligning and grounding the representation geometry towards better sense learning.

\section{Related Work}

Prior works which modify off-the-shelf embeddings to improve their lexical-semantic representation can be divided into two primary categories: (1) Anisotropy treatment methods and (2) Retrofitting methods. Anisotropy treatment methods focus on improving the isotropy of word vectors, promoting uniform distribution of information across all directions \citep{mu2018all, raunak2019effective, wang2019evaluating}. Isotropy in contextual vector spaces is regarded valuable, especially when utilizing vector geometry and relatedness measures in downstream analyses \citep{ethayarajh2019contextual}. Prior methods that focus on creating more isotropic vector spaces have suggested principle component manipulation (removal, extension) of vector spaces \citep{mu2018all, jo2018extrofitting}. To our knowledge, these methods have been proposed for static word representations, but are yet to be extended to contextual word representations extracted from a wide variety of language models.

On the other hand, retrofitting methods are focused on enhancing the representation geometry, by encoding lexical semantic relations through semantic specialization, a post-processing approach that enforces linguistic constraints on vector spaces by relying on external linguistic knowledge databases \citep{vulic2018specialising, faruqui2015retrofitting, jo2018extrofitting, vulic2018injecting}. Semantic specialization as a post processing step (retrofitting) is currently limited to static word representations \citep{mu2018all, vulic2018specialising} where they have yielded impressive performance improvements over raw embeddings \citep{lauscher2020specializing}. Existing methods towards semantic specialization of contextual representations primarily focus on retraining the model from scratch \citep{lauscher-etal-2020-specializing} or post-hoc fine-tuning the model \citep{zhang-etal-2019-ernie, peters-etal-2019-knowledge, wang2020k}. These methods are (1) resource-intensive (retraining or fine-tuning) and (2) do not address the representation degeneration problem in vector representations \citep{gao2018representation}.

\section{Methodology}

\subsection{Contextual Word Representation Models}
In this work, we focus on contextual word representations generated from four transformer-based model architectures, i.e., BERT \citep{devlin2018bert}, GPT-2 \citep{radford2019language}, XLNet \citep{yang2019xlnet} and ELECTRA \citep{clark2019electra}. These models have been selected to assess the impact of variation in context learning and pre-training over the quality of generated representations, while keeping the number of hidden layers and dimensionality identical (\textit{layers = $13$ ($0+12$); dimensions = $768$}). BERT and ELECTRA are both bidirectional learners, but they differ in terms of the pre-training objectives used to train the models: BERT uses a masked language modeling objective, limiting its learning to a small subset of word tokens; ELECTRA uses replaced token detection and is able to learn across a wider range of words tokens. On the other hand, GPT-2 and XLNet are both unidirectional learners, where GPT-2 learns only left-to-right context, while XLNet learns over all possible permutations of the given input. A comparison over these models in a uniform setting allows us to relate the behavior of representations to the context learning and pre-training choices of the respective models.

\begin{table}[b!]
\small
\centering
\begin{tabular}{cccc}
\hline
\textbf{Corpus}                    & \textbf{Nouns} & \textbf{Verbs} & \textbf{Adjectives} \\ \hline
\verb|S3-T1|         & 428   & 635   & 166        \\
\verb|S2-TA| & 292   & 307   & 344        \\
\verb|S13-T12|      & 338   & 164   & 46         \\
\verb|S7-T7|       & 512   & 814   & 380        \\
\verb|S15-T13|      & 110   & 156   & 127        \\ \hline
\textbf{Total}                     & 1680  & 2076  & 1063       \\ \hline
\end{tabular}
\caption{Data Summary.}
\label{tab:data_desc} 
\end{table}
\subsection{Data}
Contextual word representations for individual words are generated by feeding sentences into the language model. In order to generate representations, we use sense annotated corpora from various SemEval and SenseEval tasks, including SensEval 3 task 1 \verb|(S3-T1)| \citep{snyder2004english}, SensEval 2 all-words task \verb|(S2-TA)| \citep{edmonds2001senseval}, SemEval 2013 task 12 \verb|(S13-T12)|  \citep{navigli2013semeval}, SemEval 2007 task 7 \verb|(S7-T7)| \citep{navigli2007semeval} and SemEval 2015 task 13 \verb|(S15-T13)| \citep{moro2015semeval}. To ensure that the Wordnet sense keys are unified across corpora, we utilize the Wordnet 3.0 sense annotated data  \citep{vial2018ufsac} and summarized in Table \ref{tab:data_desc}. Since we want to evaluate sense-learning, we limit our analyses to multi-sense words, retaining nouns, adjectives and verbs that appear within the corpora as more than one sense.

\subsection{Sense Learning Measures}\label{sense_learn_meas}
In order to compute how sense information is encoded with the word representations, we define two word-sense specific cosine relatedness measures. 

\begin{definition}[Sense Similarity]
Let $w_{s}$ be a sense of the word $w$, appearing in $m$ different contexts. Let $v_{l}$ be the vector that maps the each word sense  occurrence $w_{s_{i}}$ to the vector space. Then, the average sense similarity between all $m$ instances of the word sense $w_s$ for layer $\ell$ is
\begin{equation}
\scriptstyle
SenSim_{\ell}(w_{s})=\frac{1}{m} \sum_{j} \sum_{k \neq j} \cos (v_{\ell}(w_{s_j}), v_{\ell}(w_{s_k}))
\end{equation}
This metric calculates the average cosine similarity between contextual representations of the same sense of a word. 
\end{definition}

\begin{definition}[Inter Sense Similarity]
Let the word $w$ have $S$ different word senses, where $w_{a}$ and $w_{b}$ are a pair of different senses of $w$, appearing in $m$ and $n$ different contexts respectively and $a, b \in S$. Let $v_{l}$ be the vector that maps each word sense occurrence $w_{s_{i}}$ to the vector space. Then, the average inter sense similarity between the representations of all instances of the word $w$ for layer $\ell$ is
\begin{equation}
\scriptstyle
InterSim_{\ell}(w)=  \mathop{\mathbb{E}}_{a, b \in  S} \left[\frac{1}{mn}\sum_{ j =1}^m \sum_{i=1}^n \cos(v_{\ell}(w_{a_{i}}), v_{\ell}(w_{b_{j}})) \right]
\end{equation}
This metric calculates the average cosine similarity between contextual representations of different senses of a word. 
\end{definition}
\normalsize
\begin{figure}
\centering
\subfloat[Original Representations\label{fig:1(a)}]{%
  \includegraphics[clip,width=0.7\columnwidth]{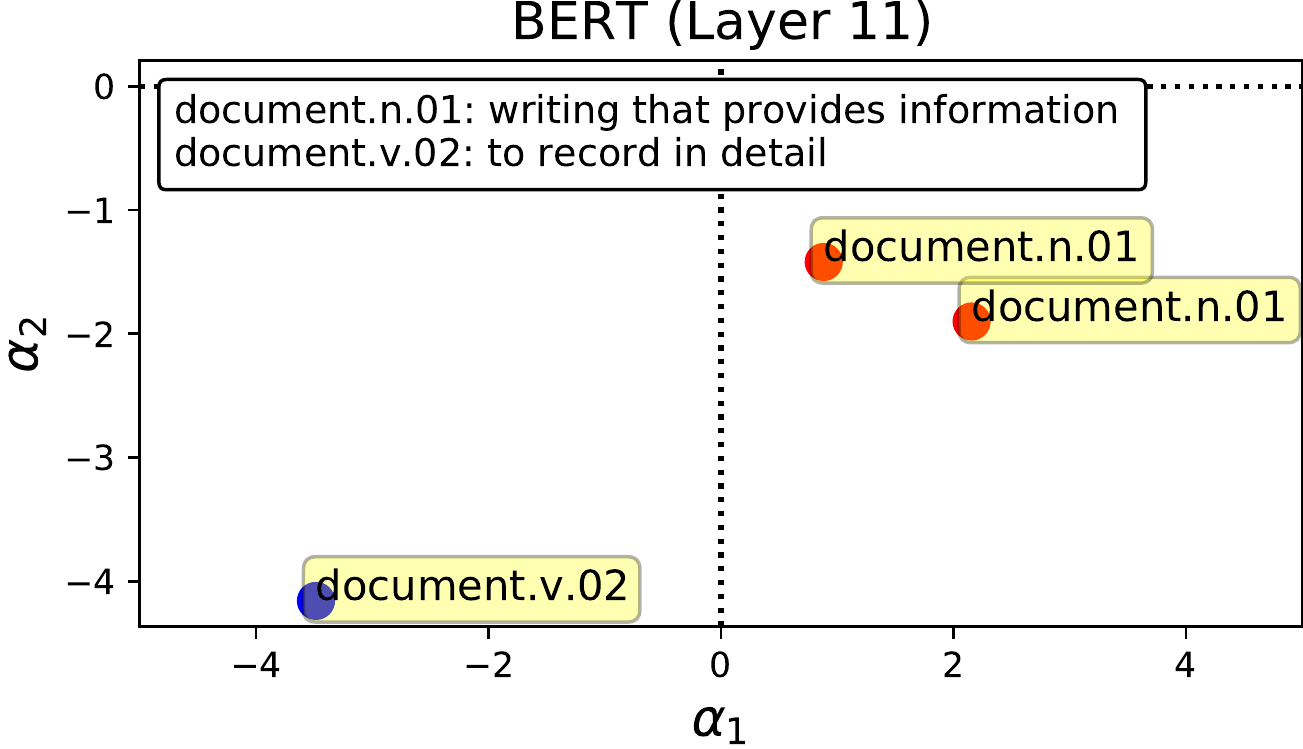}%
}

\subfloat[LASeR Representations\label{fig:1(b)}]{%
  \includegraphics[clip,width=0.7\columnwidth]{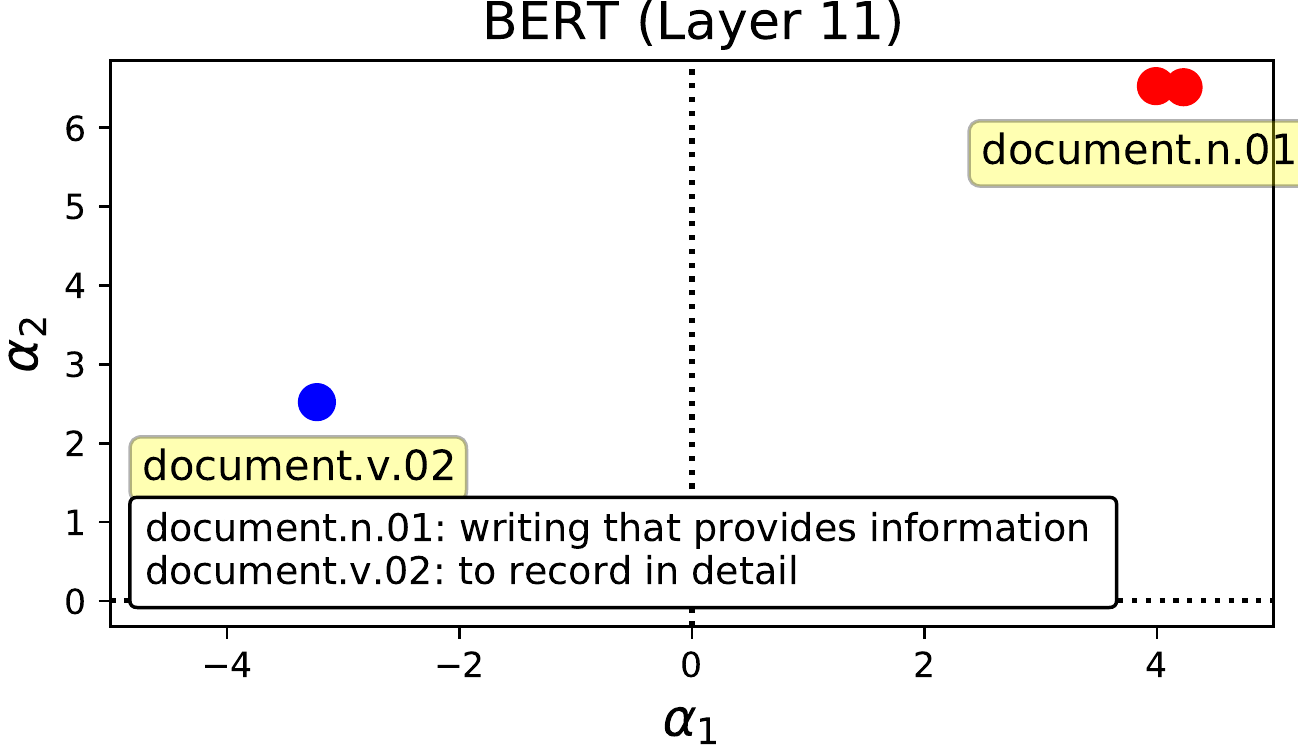}%
}
\caption{Representations of different senses of the word ‘document’ (BERT Layer 11).}
\label{fig:1} 
\end{figure}

Thus, if a word $w$ has $SenSim_{l}(w_{s})>InterSim_{l}(w)$, it suggests that the representations for the same sense of a given word lie much closer together within the vector space, as compared to the representations of different senses of the same word. For example, a given word \textit{`document'} can refer to multiple senses. According to WordNet 3.0, two senses of the word \textit{`document'} are: (1) \textit{document.n.01} - writing that provides information and (2) \textit{document.v.02} - to record in detail.

As an example, we have visualized the representations of these two senses as encoded within the vector space of BERT (Layer 11), shown in Figure \ref{fig:1}. The `original' representations, shown in Figure \ref{fig:1(a)}, of the word sense \textit{document.n.01} lie slightly close to each other, and farther away from the \textit{document.v.02} representation. Thus, if a model is able to encode similar representations for same sense of a word, and distinguishable representations for different senses of a word, we claim that the model encodes sense information. 

\subsection{Anisotropy Adjusted Sense Similarity}
In order to assess whether contextual word representations encode sense information, we measure the sense similarity and inter sense similarity for multi-sense words (polysemes and homonyms) in our datasets, across model layers. Given that contextual word representations encode anisotropy, we calculate anisotropy adjusted sense relatedness measures as follows.
\begin{equation}
\scriptstyle
B(v_{\ell})=\mathop{\mathbb{E}}_{a, b \sim U}\left[\cos(v_{\ell}(a), v_{\ell}(b))\right] \\ \scriptstyle SenSim_{\ell}(w_{s})^{*} = SenSim_{\ell}(w_{s}) - B(v_{\ell}) \\ \scriptstyle
InterSim_{\ell}(w)^{*} = InterSim_{\ell}(w) - B(v_{\ell})
\end{equation}
This baseline calculation utilizes the theory from prior works examining contextualization in word representations \citep{ethayarajh2019contextual}. Here, $B(v_{\ell})$ is the average cosine similarlity between $n$ randomly sampled words, $U$ is the set of all word occurrences, and $v_{\ell}(.)$ maps a word occurrence to the respective word representation in layer $\ell$.

\subsection{Low Anisotropy Sense Retrofitting}

In this subsection, we describe LASeR, a post-processing approach to render off-the-shelf representations more isotropic and sense-enriched. Our approach builds upon the work on anisotropy  reduction \citet{mu2018all} and retrofitting \citet{faruqui2015retrofitting}. \citet{mu2018all} suggests that anisotropy can be reduced by removing primary components to make the representations more distinct and uniformly distributed within the vector space. We extend this to contextual word representations, evaluating the efficacy of removing primary components on anisotropy reduction in contextual representations. Turning towards retrofitting methods, we extend the retrofitting approach proposed by \citet{mu2018all}, which targets static word representations and brings synonyms closer together in the vector space. Our work extends this retrofitting goal to contextual representations, where we aim to bring representations of same word-senses closer in the vector space, ensuring better sense disambiguation capabilities for representations.
\begin{algorithm}

\SetAlgoLined
\KwInput{Raw word representation $\left\{v(w_{i}), w_{i} \in V\right\}$}

Perform mean centering of vector: $\mu \leftarrow \frac{1}{|v|} \Sigma_{w_{i} \in V} v(w_{i}); \tilde{v}\left(w_{i}\right) \leftarrow v\left(w_{i}\right)-\mu$

Compute the PCA components: $u_{i1}, \ldots, u_{iD} \leftarrow PCA\left(\left\{\tilde{v}(w_{i}), w_{i} \in \mathcal{V}\right\}\right) $

Remove top \textit{d} principal components: $ v^{\prime}\left(w_{i}\right) \leftarrow \tilde{v}\left(w_{i}\right)-\Sigma_{j=1}^{d}\left(u_{i j}^{\top} v\left(w_{i}\right)\right) u_{i j}$

Apply retrofitting update: $ \hat{v}(w_{i})=\frac{\sum_{j:(i, j) \in E} \beta_{i j} v(w_{j})+\alpha_{i} v(w_{i})}{\sum_{j:(i, j) \in E} \beta_{i j}+\alpha_{i}}$

\KwOutput{Processed word representation $\hat{v}\left(w_{i}\right) $ } 
 \caption{\textbf{LASeR} (Low Anisotropy Sense Retrofitting).}
\end{algorithm}
\setlength{\textfloatsep}{1pt}

Let $v(w_i)$ be the original contextual representation, $v^{\prime}(w_i)$ be the \textit{low anisotropy }contextual representation and $\hat{v}(w_i)$ be the \textit{sense enriched} contextual representation of $i^{th}$ occurrence of a word sense $w$. We simulate an undirected knowledge graph $\Omega (V,E)$, where $V$ represents the vocabulary of word tokens, each word token representing a vertex, and $E$ represents all the edges connecting respective vertices. Finally, $Q$ represent the matrix of post-processed representations $[\hat{v}(w_1), \hat{v}(w_2),\ldots,\hat{v}(w_n)]$. The approach works on achieving dual objectives, described as follows:
\noindent
\textbf{Objective 1 (Lower Anisotropy) :} Remove top \textit{d} common directions across all $v(w_i)$, to create $v^{\prime}(w_i)$, creating more uniformly distributed word vectors and lowering anisotropy in representations.
\noindent
\textbf{Objective 2 (Sense Retrofitting) :} Learn $ \hat{v}(w_i) $ such that same sense representations lie closer together in vector space as well as close to the original embedding

The algorithm takes the original representations as input. These representations undergo mean centering and removal of dominant primary components (\small{1,2,3 }) \normalsize to reduce the anisotropy in the vector space. This is followed by a sense-retrofitting update (\small{4}) \normalsize. Here, for each word token representation $v(w_{i})$, we define its neighbours as $v(w_{j}), \forall j $ where $sense(w_{i}) = sense(w_{j})$, and hyper-parameters $\beta_{ij}$ and $\alpha_i = 1$ represent the reciprocal of the node degree of the word token $w_{i}$ and edge weights respectively. 

\section{Results}

\begin{figure}[b!]
\centering
\includegraphics[scale = 0.5]{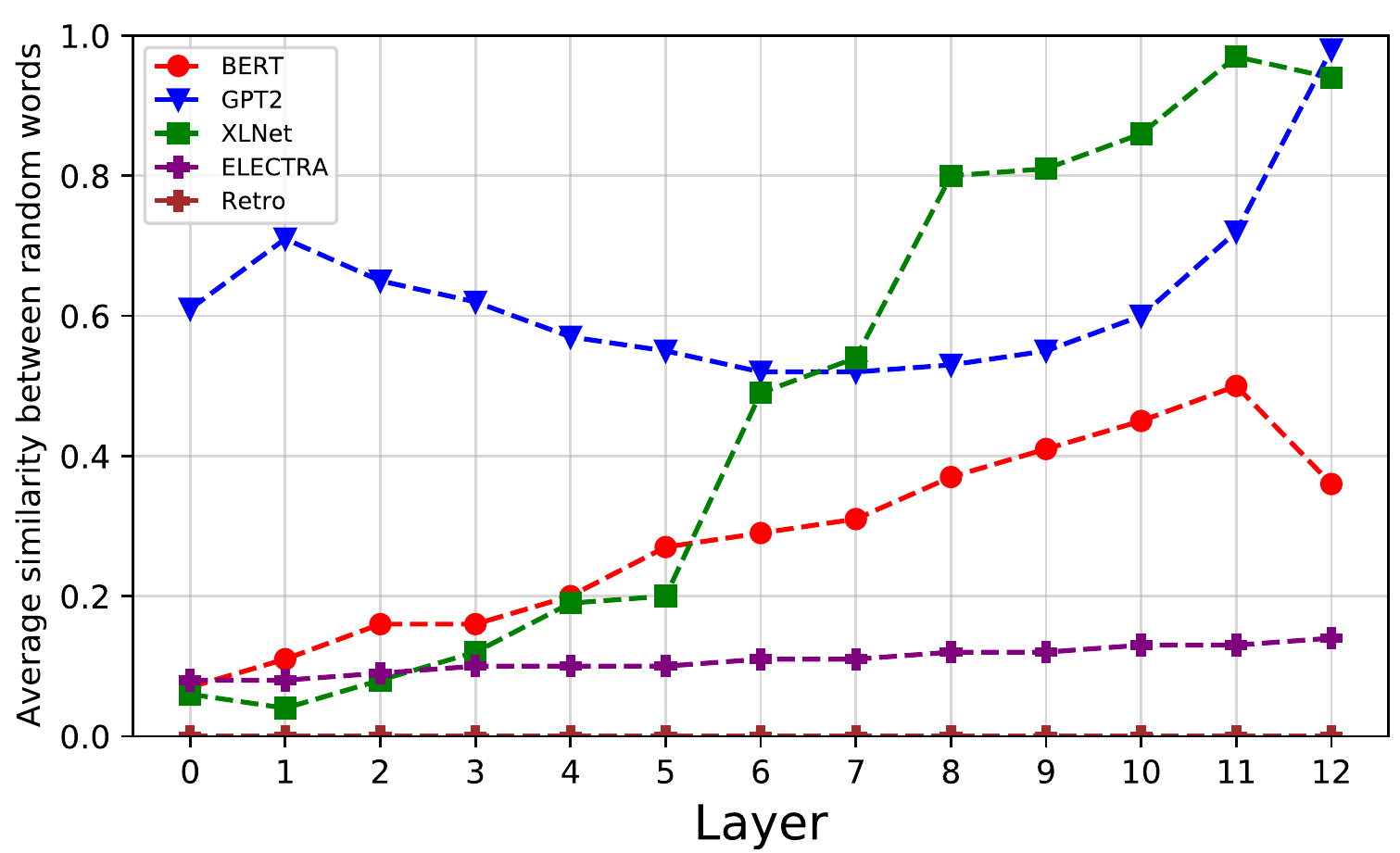}
\caption{Average similarity between representations of randomly sampled words (1K) across model layers.}
\label{fig:2}       
\end{figure}

We first show anisotropy analysis results (\cref{aniso_ana}), further evaluating sense learning in contextual representations (\cref{sense_learn_0}). Finally, we present improvements in isotropy and lexical-semantic capabilities of the post-processed representations (\cref{sense_learn_1}).
\subsection{Anisotropy Analysis}\label{aniso_ana}
\subsubsection{Similarity between Random Words}\label{rand_word_sim}

We first assess the amount of anisotropy encoded within contextual word vector spaces. We plot the average cosine similarity between 1K randomly sampled words, across different layers of language models, as seen in Figure \ref{fig:2}. If a vector space is isotropic, the average cosine similarity between uniformly randomly sampled words would be $0$ \citep{ethayarajh2019contextual}. Thus, the closer this measure is to $1$, the more anisotropic the vector space. It can be seen that anisotropy evolves very differently across different models. Unidirectional language models (XLNet, GPT-2) portray far more anisotropy in word representations as compared to bidirectional language models (BERT, ELECTRA). Thus, language models learning one-directional context (L-to-R or R-to-L) encode more common directions in the representations as compared to those learnt from bidirectional context. Moreover, anisotropy monotonically increases across layers for BERT and XLNet, where both models have been trained on masked language modeling tasks. This shows that anisotropy accumulates in the upper layers of masked language models. The rate of increase in anisotropy in XLNet is higher than BERT representations, showing that permutation language modeling propagates higher amounts of anisotropy than traditional MLM. These results are consistent with the results obtained for all multi-sense words in the corpora (Appendix \ref{sec:appendix_A}). 

\begin{figure}[t!]
\centering
\includegraphics[scale = 0.5]{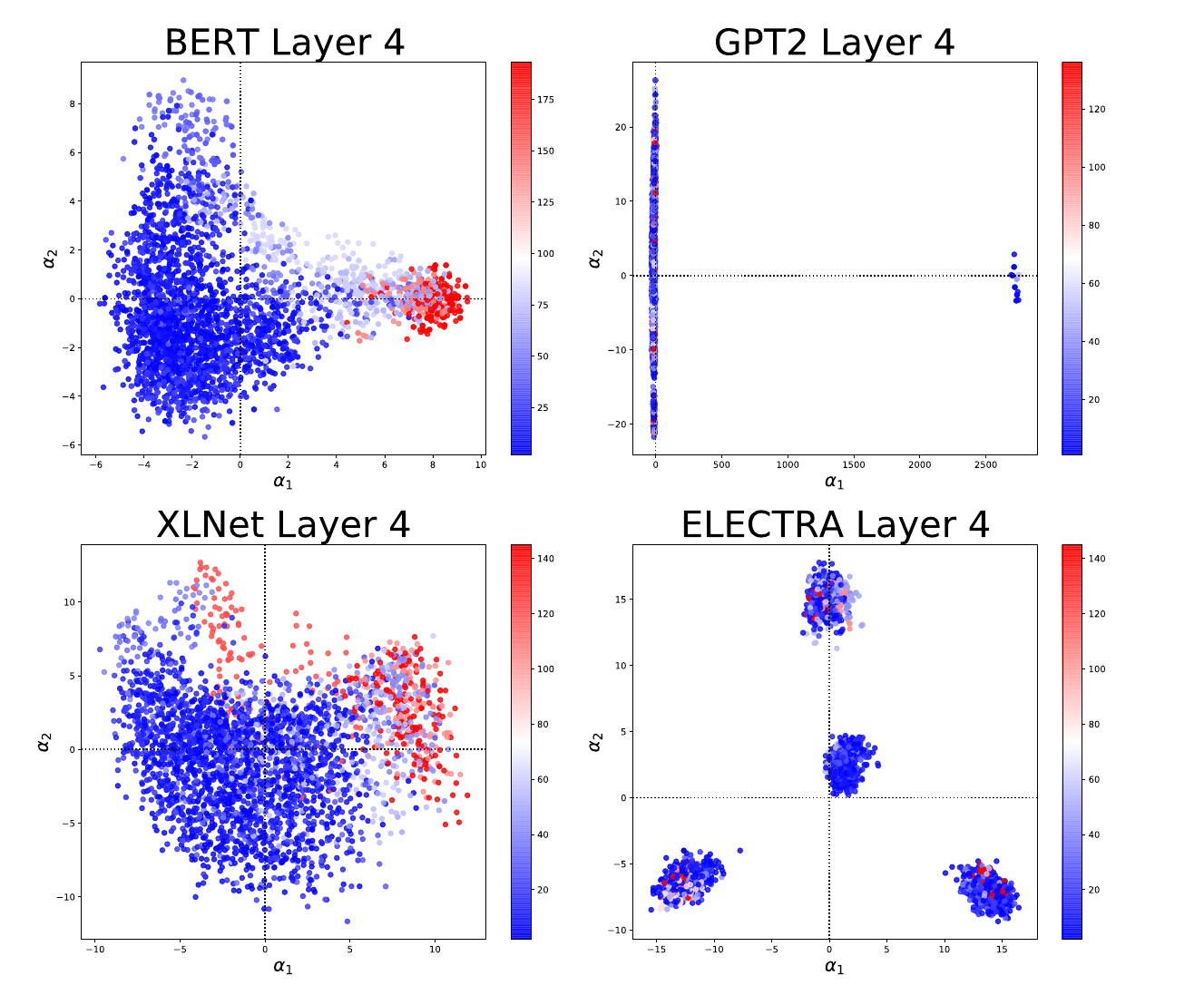}
\caption{PCA plots of original word representations across top 2 primary components; Blue:Low frequency word tokens, Red:High frequency word tokens.}
\label{fig:3}       
\end{figure}

\subsubsection{Analysis of Principal Components}\label{pca_analysis}
\begin{figure*}
\centering
    \centering
    \begin{subfigure}[t]{0.49\textwidth}
        \centering
        \includegraphics[scale=0.5]{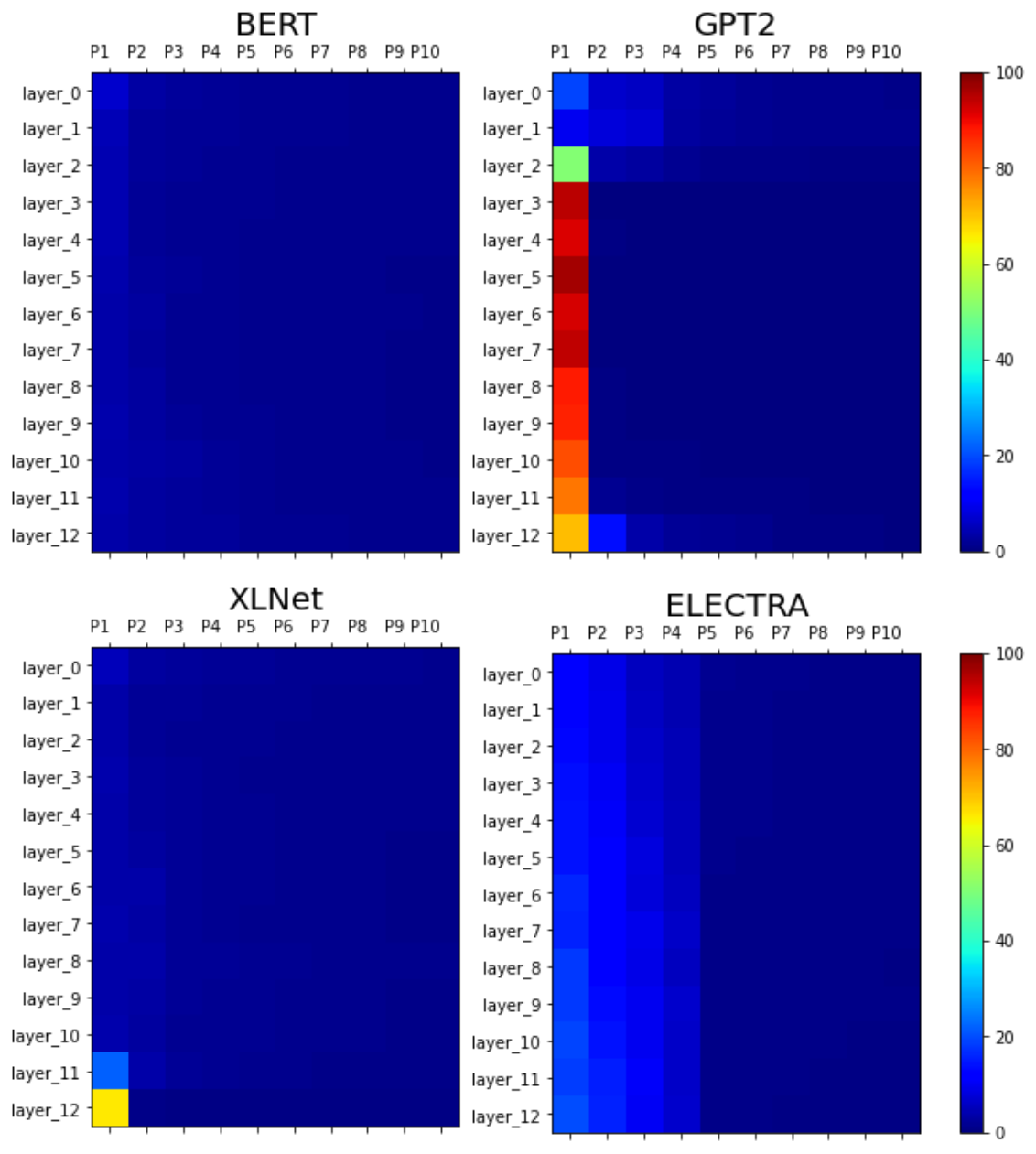}
        \caption{Vanilla Representations.}
        \label{fig:4(a)}
    \end{subfigure}%
    ~ 
    \begin{subfigure}[t]{0.49\textwidth}
        \centering
        \includegraphics[scale=0.5]{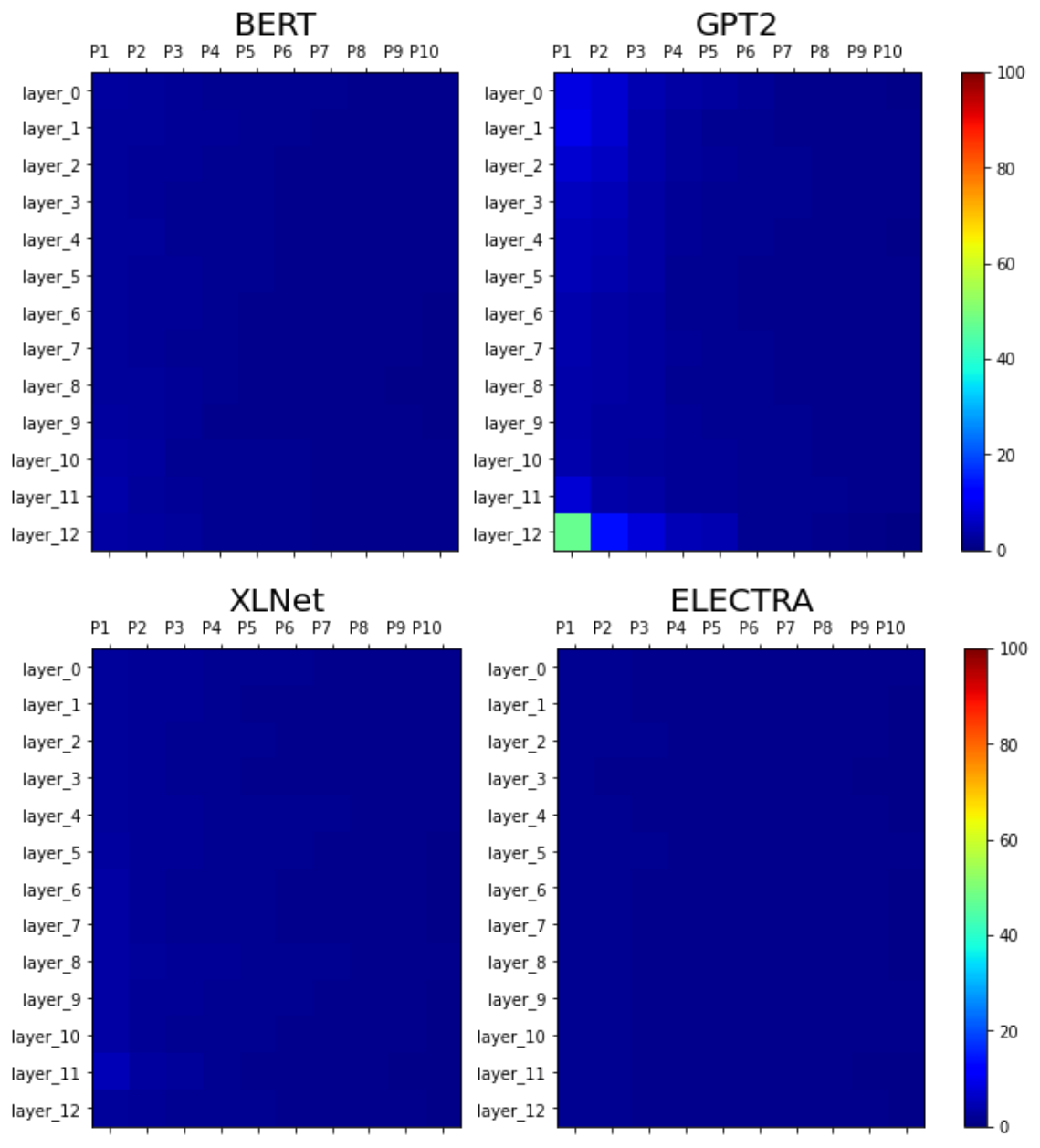}
        \caption{Retrofitted Representations.}
        \label{fig:4(b)}
    \end{subfigure}
    \caption{Plots of proportion of variance encoded within the top $d=10$ dominant principal components of the contextual representations across model layers. The horizontal labels (P1-P10) represent each of the ten principal components, and the vertical labels (layer\_0 - layer\_12) represent each of the 12 model layers.}
\label{fig:4}
\end{figure*}

High anisotropy leads to word vectors being distributed within a very narrow cone in the vector space \citep{mimno2017strange}, further signifying that the word representations encode common directions \citep{mu2018all}. We plot the top two dominating directions for word representations, across each model's layers, as shown in Figure \ref{fig:3}. These plots reveal that contextual word representations extracted from different language models are encoded extremely differently within the vector space. It can be seen that BERT and XLNet embeddings are more spread across the vector space, as compared to GPT-2 and ELECTRA embeddings. Moreover, ELECTRA embeddings form highly concentrated, yet separated regions of anisotropy, thus leading to an overall low score on the average similarity between randomly sampled words. Moreover, GPT-2 embeddings reveal extreme anisotropy, where most of the embeddings encode a singular common direction. The plots in Figure \ref{fig:3} also reveal that word frequency is significantly encoded in the top two principal components of BERT and XLNet embeddings. We cannot claim the same for GPT-2 and ELECTRA embeddings, where all embeddings cluster within highly dense regions of anisotropy.

We also evaluate anisotropy across model layers by assessing the explained variance across common directions encoded across all word representations. We plot the proportion of variance encoded within the top $d=10$ dominant principal components of the original contextual representations across model layers, shown in Figure \ref{fig:4(a)}. While bidirectional models such as BERT and ELECTRA encode multiple common directions, unidirectional models like GPT-2 and XLNet embeddings primarily encode a singular common direction. For BERT embeddings, the top 10 primary components only contribute to 17-24\% of the explained variance, showing that the embeddings are more uniformly distributed across the vector space, as compared to other models. GPT-2 provides a stark contrast, where the top 10 principal components contribute to up to ~97\% of the explained variance, highly concentrated within the first principal component, especially for the middle layers (Layer 3-8). XLNet embeddings capture comparatively lower common directions across model layers, apart from the final model layer (Layer 12), where 66.1\% of the explained variance is concentrated within the first principal component. Thus, representations learnt through the goal of predicting the next word yields all representations extremely similar.

\subsection{Sense Learning in Original Representations}\label{sense_learn_0}
A model differentiates between different word senses if it encodes representations of the same sense of a word to be more similar than the representations of other senses of the same word. We utilize the sense learning measures, defined in (\cref{sense_learn_meas}) to assess whether original representations encode word-sense information. To examine overall learning across model layers, we calculate average sense similarity ($\overline{SenSim_{\ell}(w)}$) and mean difference between average sense similarity and inter sense similarity for a word token $w$ ($\Delta$).
\begin{equation}
\scriptstyle
\Delta = \overline{SenSim_{\ell}(w)} - InterSim_{\ell}(w); \enspace \Delta \in [-1,1]
\end{equation}
Ideally, a language model being able to capture distinction between all word senses should have $\overline{SenSim_{\ell}(w)} = 1$ and $\Delta>>0$. Here, higher sense similarities correspond to similar senses being encoded closer in the vector space and $\Delta > 0$ shows that on an average, same sense representations are more cohesive and well separated from the representations of other senses.

\begin{figure}[b!]
\centering
\includegraphics[scale = 0.55]{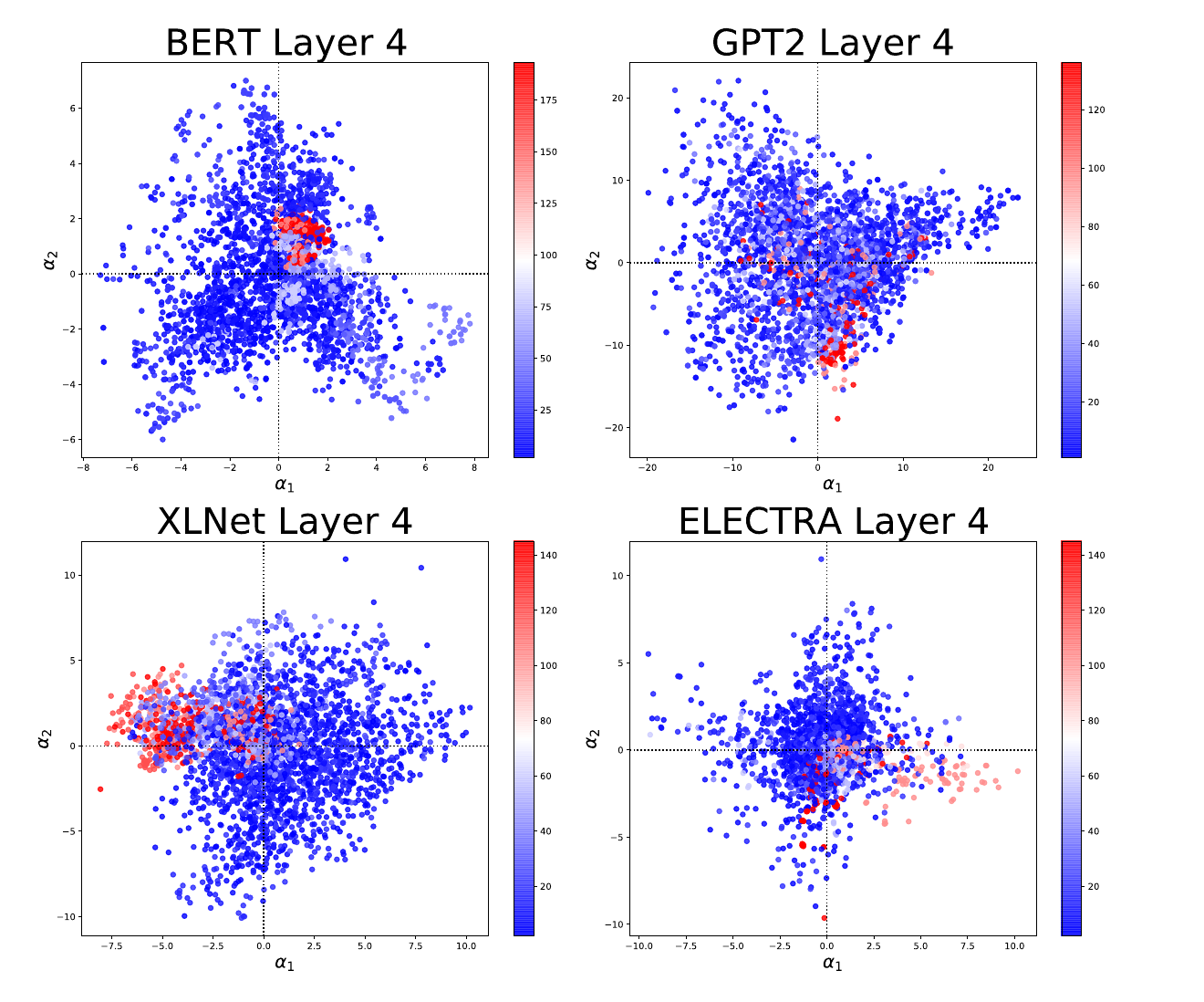}
\caption{PCA plots of post-processed word representations across top two primary components, for each model.}
\label{fig:5}       
\end{figure}

The evolution of sense learning over different models and their layers is portrayed using sense similarity measures, aggregated in Table \ref{tab:sense_sim}. The reported vanilla sense similarity scores have been adjusted for anisotropy. Prior to retrofitting, BERT and XLNet embeddings for the same word senses show increasing dissimilarity across model layers, signifying a loss of sense information as the model gets more contextualized. The similarity between same sense word representations from the GPT-2 model is close to $0$, showing that GPT-2 captures almost no sense information within the embedding space. ELECTRA embeddings remain consistent in terms of sense learning, not varying significantly across model layers. Furthermore, $\Delta\sim0$ across all models shows that the original representations do not significantly distinguish between different senses of a given word. We visualize an example in Figure \ref{fig:1(a)}, where representations of the word \textit{document} lie close together, regardless of the different senses associated with each occurrence. This finding signifies that the sole reliance on word form to learn representations does not suffice in helping the model distinguish between multiple senses of a given word.

\subsection{Low Anisotropy Sense Retrofitting}\label{sense_learn_1}
\begin{table*}
\small
\centering
\begin{tabular}{|c|cc|cc|cc|cc|}
\hline
\multirow{2}{*}{\textbf{Layer}} & \multicolumn{2}{c|}{\textbf{BERT}} & \multicolumn{2}{c|}{\textbf{GPT-2}} & \multicolumn{2}{c|}{\textbf{XLNet}} & \multicolumn{2}{c|}{\textbf{ELECTRA}} \\
                       & \textit{vanilla} ($\Delta$)   & \textit{retro} ($\Delta$)   & \textit{vanilla}($\Delta$)     & \textit{retro} ($\Delta$)   & \textit{vanilla} ($\Delta$)      & \textit{retro} ($\Delta$)   & \textit{vanilla} ($\Delta$)      & \textit{retro} ($\Delta$)    \\ \hline
0                      & 0.82 (0.00)     & 0.93 (0.04)        & 0.08 (0.05)      & 0.49 (0.38)        & 0.08 (-0.27)     & 0.85 (-0.02)         & 0.49 (0.11)      & 0.64 (0.25)          \\
1                      & 0.74 (0.02)     & 0.90 (0.08)        & 0.06 (0.03)     & 0.51 (0.38)        & 0.80 (0.02)     & 0.90 (0.08)         & 0.50 (0.11)      & 0.65 (0.26)          \\
2                      & 0.66 (0.05)    & 0.88 (0.13)       & 0.06 (0.02)    & 0.51 (0.38)        & 0.65 (0.07)     & 0.83 (0.18)        & 0.50  (0.12)     & 0.65 (0.26)          \\
3                      & 0.62 (0.07)     & 0.85 (0.18)        & 0.08 (0.04)     & 0.52 (0.39)        & 0.58 (0.10)     & 0.80 (0.23)         & 0.50 (0.12)      & 0.65 (0.26)          \\
4                      & 0.54 (0.09)     & 0.82 (0.23)        & 0.09 (0.05)     & 0.53 (0.39)        & 0.46 (0.09)     & 0.76 (0.27)        & 0.51 (0.12)      & 0.65 (0.26)          \\
5                      & 0.43  (0.10)    & 0.77 (0.28)        & 0.09 (0.05)     & 0.53 (0.39)       & 0.40 (0.09)     & 0.72 (0.30)         & 0.52 (0.13)      & 0.65 (0.27)          \\
6                      & 0.36 (0.12)     & 0.72 (0.33)        & 0.11 (0.06)     & 0.54 (0.40)        & 0.23 (0.06)     & 0.68 (0.31)         & 0.51 (0.12)      & 0.65 (0.27)          \\
7                      & 0.31 (0.11)    & 0.68 (0.35)        & 0.12 (0.06)     & 0.55 (0.41)        & 0.20 (0.06)     & 0.66 (0.33)         & 0.52 (0.13)      & 0.65 (0.27)           \\
8                      & 0.25 (0.10)     & 0.65 (0.37)        & 0.11 (0.07)     & 0.55 (0.41)        & 0.08 (0.03)     & 0.64 (0.33)        & 0.53 (0.13)      & 0.64 (0.26)          \\
9                      & 0.22 (0.09)     & 0.63 (0.39)        & 0.11 (0.07)     & 0.56 (0.42)        & 0.07 (0.02)     & 0.63 (0.34)         & 0.53 (0.13)      & 0.64 (0.26)          \\
10                     & 0.22 (0.09)     & 0.63 (0.38)        & 0.09 (0.06)     & 0.56 (0.43)        & 0.05 (0.01)     & 0.65 (0.34)         & 0.53 (0.13)      & 0.64 (0.27)          \\
11                     & 0.20 (0.07)     & 0.63 (0.36)        & 0.08 (0.04)      & 0.55 (0.43)        & 0.01 (0.00)    & 0.65 (0.32)         & 0.53 (0.13)      & 0.64 (0.27)          \\
12                     & 0.24 (0.09)     & 0.62 (0.37)        & 0.00 (0.01)    & 0.51 (0.40)        & 0.02 (0.00)     & 0.63 (0.32)         & 0.53 (0.13)      & 0.64 (0.27)           \\ \hline
\end{tabular}
\caption{Average sense similarity scores across model layers; $\Delta = SenSim_{\ell}(w) - InterSim_{\ell}(w)$.}
\label{tab:sense_sim}
\end{table*}

We evaluate the efficacy of the proposed \textbf{LASeR} approach by comparing improvements in vector space isotropy and improved disambiguation of different word senses, as captured by retrofitted word representations.

\subsubsection{Improvements in Isotropy}

We conduct experiments by removing the most dominant common direction ($d = 1$) across generated embeddings across each model layer. This step yields significantly better isotropy in the resulting representations, where average similarity between randomly sampled words ($k=1000$) is 0, across all models and model layers. Improvements can also be observed from the reduced proportions of explained variance in Figure \ref{fig:4(b)}. Overall, most of the anisotropy in the vector space is treated by removing one dominating direction. The retrofitted GPT-2 embeddings still show high anisotropy in the $12^{th}$ layer, showing that more common directions remain to be addressed and possibly removed. These results show that high anisotropy effects can be reduced by removing the primary common directions across representations. The effect of this step is also visualized in Figure \ref{fig:5}, where the representations are significantly less anisotropic and more uniformly spread across the vector space, encoding fewer artifacts of word frequency in the vector space, as compared to the original representations. For visualizations across all model layers, refer to Appendix \ref{sec:appendix_B}. In most cases, removal of the most dominant common direction can yield significant improvements in isotropy, as seen for BERT, XLNet and ELECTRA. In other cases, where representations share more than one significant common directions, such as for GPT-2, we can remove $d>1$ common directions to treat anisotropy.

\subsubsection{Improvements in Sense Representation}
The retrofitting update applied to model representations enforces lexical-semantic constraints, bringing same sense representations closer together (increase same-sense cohesion) and pushing different sense representations farther apart (increase inter-sense separation).

Results from Table \ref{tab:sense_sim} show the efficacy of our retrofitting update ($\alpha_{i} = 1$), where average sense similarity between word vectors increases significantly, and similarity between same sense representations is significantly higher than similarity between representations of different senses. This portrays that the retrofitted representations encode same sense representations closer together and different sense representations farther apart. An example of how retrofitting changes the distribution of representations in the vector space is given in Figure \ref{fig:1(b)}, where inter-sense separation between two different senses of the word \textit{document} increases and same-sense cohesion between representations of the same word sense increases. 
\begin{figure}[b!]
\centering
\includegraphics[scale = 0.37]{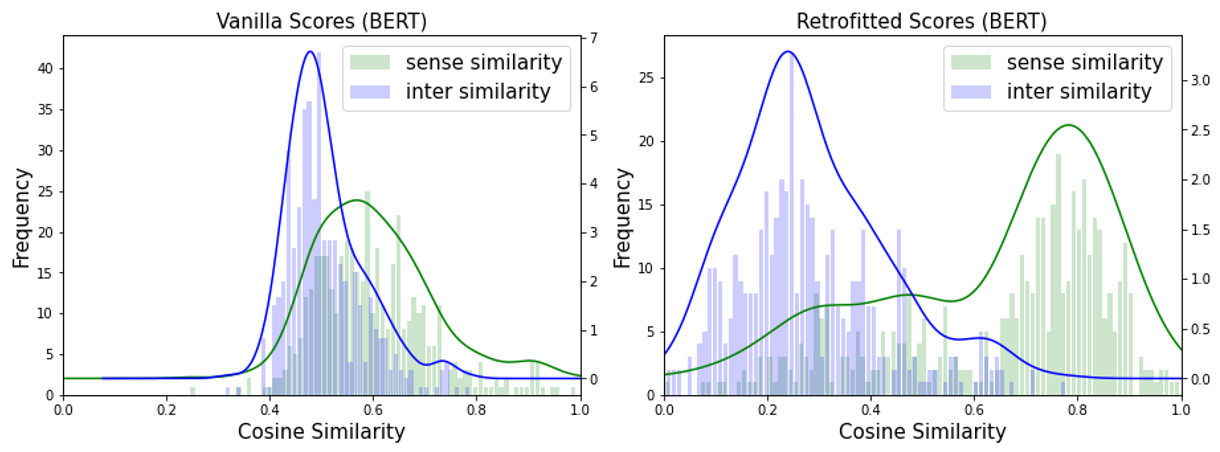}
\caption{Effect of retrofitting on sense relatedness in contextual embeddings. Here, retrofitted embeddings portray higher same-sense similarity and lower inter-sense similarity.}
\label{fig:6}       
\end{figure}

Across the model layers, the retrofitting significantly increases sense similarity and $\Delta$. The improved similarity scores can be seen in Figure \ref{fig:6}, which show that retrofitting moves same sense representations to be more similar than different sense representations. For BERT embeddings, the improvements are more visible in the upper model layers, as they create more separated different sense representations, and more cohesive same sense representations. The slight drop in cohesion ($SenseSim$) is due to the model's upper layers being more contextualized than the lower layers, also suggested in prior works on contextualization \citep{ethayarajh2019contextual}. Retrofitting is extremely effective for GPT-2 embeddings. This can been from the drastic increase in sense similarity ($SenseSim$) and $\Delta$, showing that same sense representations lie closer and different sense representations lie farther apart in the retrofitted vector space. 

While originally, the representations were highly anisotropic and held no sense learning, the retrofitted embeddings capture better sense distinction. XLNet embeddings, much like BERT, encode representations of the same word form closer together, especially in lower model layers, regardless of the respective word-sense distinction. Post-retrofitting, XLNet embeddings show higher similarity between same word senses and lower similarity between different word senses, revealing better sense disambiguation. Compared to the other three models, original ELECTRA embeddings are able to capture more distinction between different sense representations and more similarity between same sense representations. Our retrofitting update further improves these lexical-semantic relations in the representation space.

\section{Discussion}

Recent works have discussed whether contextualized word representations extracted from deep pretrained language models encode word sense knowledge within the representation space. Studies suggest that while lower layer BERT embeddings encode more semantic information \citep{NEURIPS2019_159c1ffe}, the upper layer embeddings become increasingly contextual \citep{ethayarajh2019contextual}. Works exploring semantic capabilities of representations have also used nearest neighbour classifier probes to assess whether same-sense representations are classified together \citep{NEURIPS2019_159c1ffe, nair2020contextualized}. Since these classifiers show slightly better accuracy than classifying as the most frequent sense, they claim that the representation space encodes sense information. Although our work supports this conclusion, we additionally argue that after accounting for anisotropy, the cohesion between same sense representations and separation between different sense representations is not significant. Here, the principal premise of the removal of anisotropy prior to injecting sense information is based on creating an embedding space geometry where the effects of representation degeneration are reduced. The representation degeneration of embeddings reduces their representational power \citep{gao2018representation}. Thus, to improve the representation ability of embeddings, we deem it important to create methods that promote representations that are not only lexico-semantic relation enriched but also isotropic. Our method reveals that the additional step of lowering anisotropy renders improved representation geometry, where word vectors are not constricted within a narrow cone, and are uniformly distributed within the vector space. Further, sense-retrofitting on contextualized word representations render same sense representations more similar and different sense representations more different, increasing the word sense disambiguation capabilities of the encoded representations.

Our work presents a novel intrinsic evaluation of sense information in word embeddings, required to understand the sense geometry encoded by various models. In the future, we will focus on integrating sense information in contextual word representations by extending this approach to words that are unseen to the LASeR model, and further perform extrinsic analyses of the embeddings.

\section{Conclusion}
In this work, we investigated the geometry of contextual word representations for isotropy and sense disambiguation capabilities. We further proposed a post-processing approach for anisotropy treatment and semantic enrichment of contextual word representations, by transforming the vector space using principal component manipulation and lexical semantic knowledge-based sense-retrofitting. Our method significantly reduced the impact of representation degeneration problem, improving isotropy within the vector space and rendered off-the-shelf contextual word vectors semantically more meaningful. In the future work, we will study the impact of changes in retrofitting hyperparameters and variable removal of primary components on representation quality. Further, we will focus on extrinsic evaluation of the impact of anisotropy removal and sense retrofitting on downstream word-sense disambiguation tasks.

\section*{Acknowledgements}
This work\footnote{This is a preprint of the accepted manuscript: Geetanjali Bihani and Julia Taylor Rayz, Low Anisotropy Sense Retrofitting (LASeR) : Towards Isotropic and Sense Enriched Representations, to appear at the Deep Learning Inside Out workshop at NAACL 2021.} is partially supported by National Science Foundation grant number 1737591. We would like to thank the three anonymous reviewers for their helpful comments and suggestions. We would also like to thank the members of the AKRaNLU lab at Purdue University for their comments and suggestions.

\bibliography{custom}
\bibliographystyle{acl_natbib}

\newpage
\appendix

\section{Anisotropy Across All Words}
\label{sec:appendix_A}
\begin{figure}[h!]
\includegraphics[scale = 0.43]{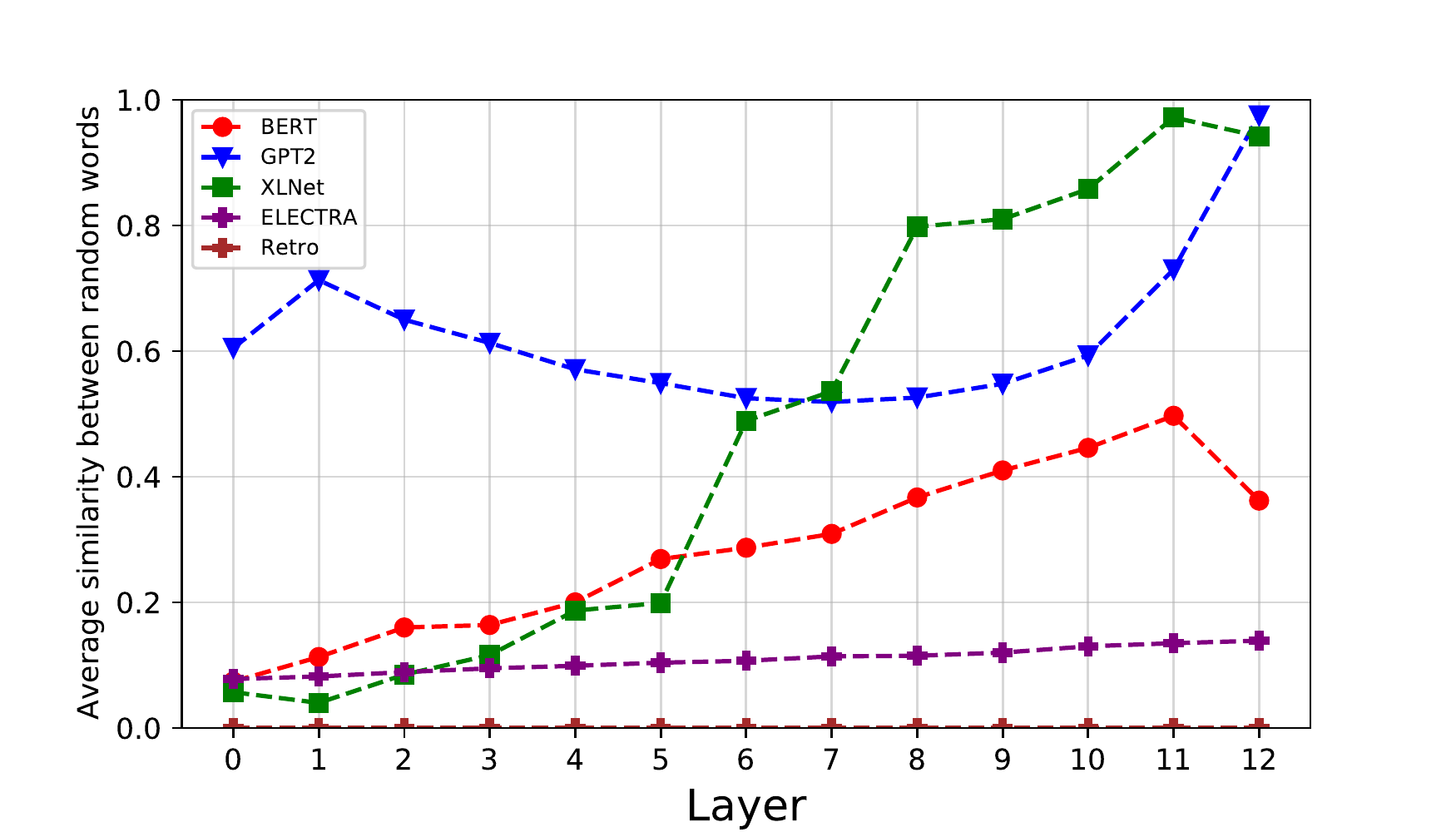}
\caption{Average similarity between representations of randomly sampled words across model layers.}
\label{fig:7}  
\end{figure}

We plot the average similarity between all words (multi-sense nouns, verbs and adjectives) extracted from the annotated corpora, across model layers, as shown in Figure \ref{fig:7}.

\section{PCA Plots of Word Representations}
\label{sec:appendix_B}

\begin{figure*}
\centering
\subfloat[original]{\label{fig:a}\includegraphics[width=0.2\linewidth]{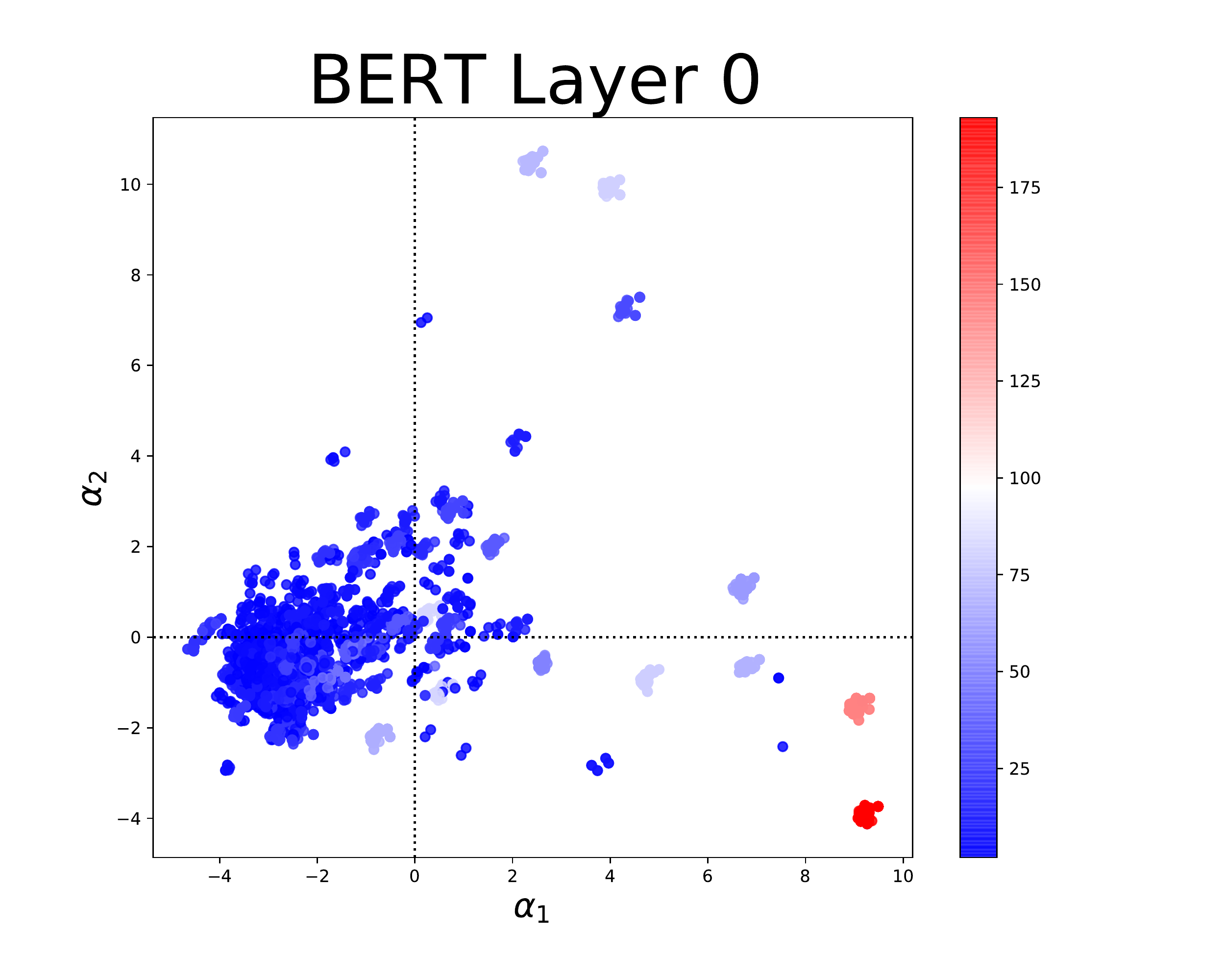}}
\subfloat[retrofitted]{\label{fig:b}\includegraphics[width=0.2\linewidth]{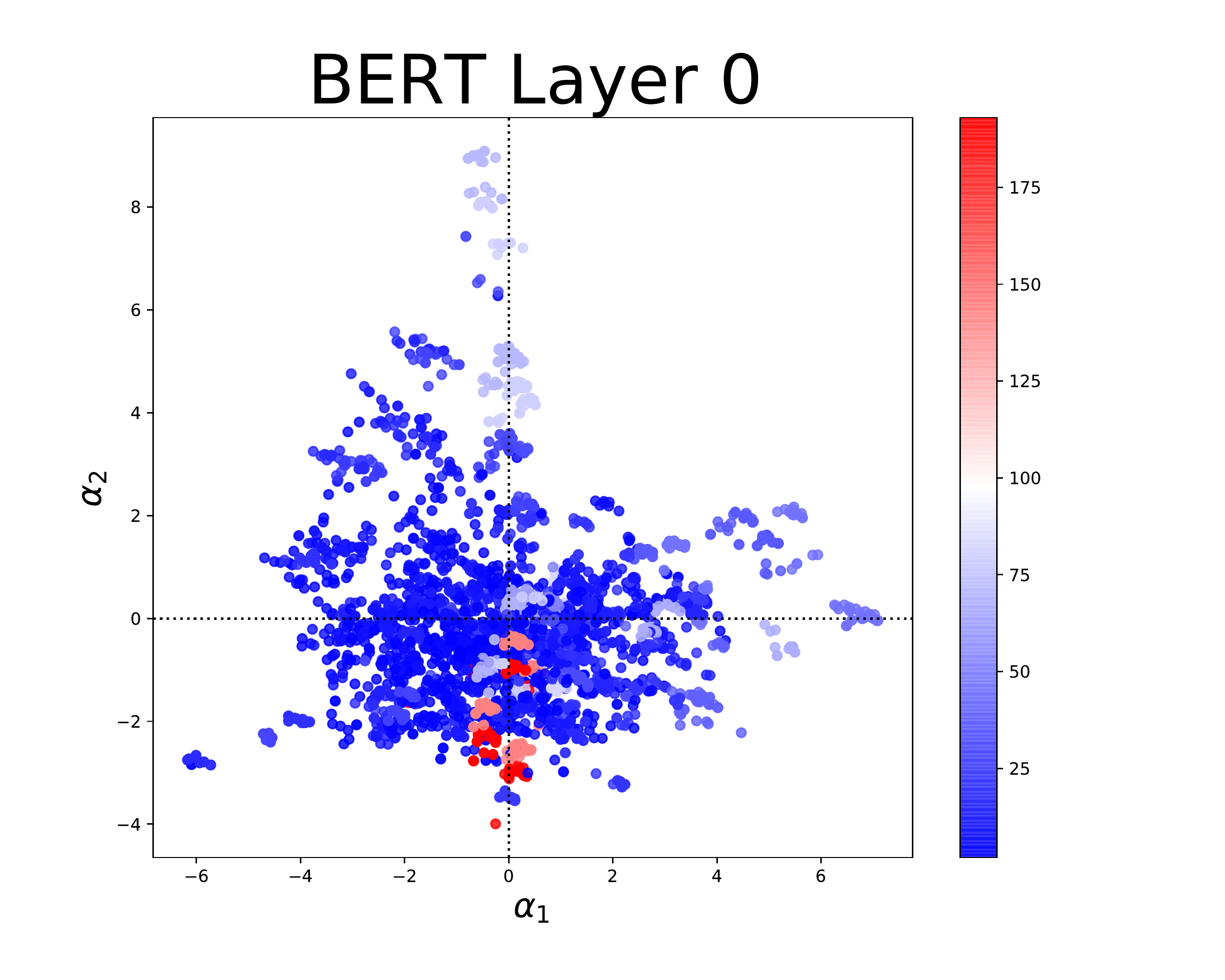}}
\subfloat[original]{\label{fig:c}\includegraphics[width=0.2\linewidth]{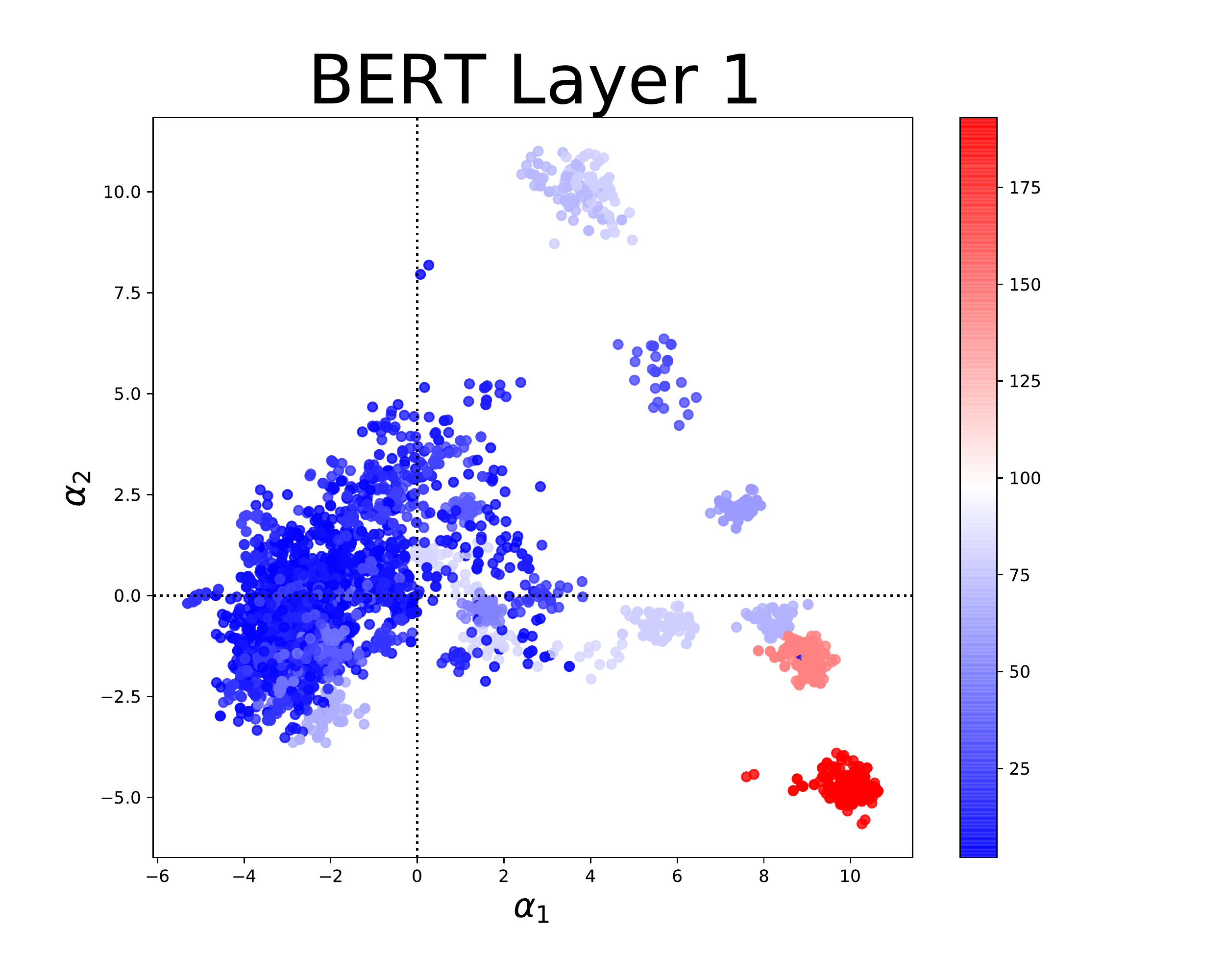}}
\subfloat[retrofitted]{\label{fig:d}\includegraphics[width=0.2\linewidth]{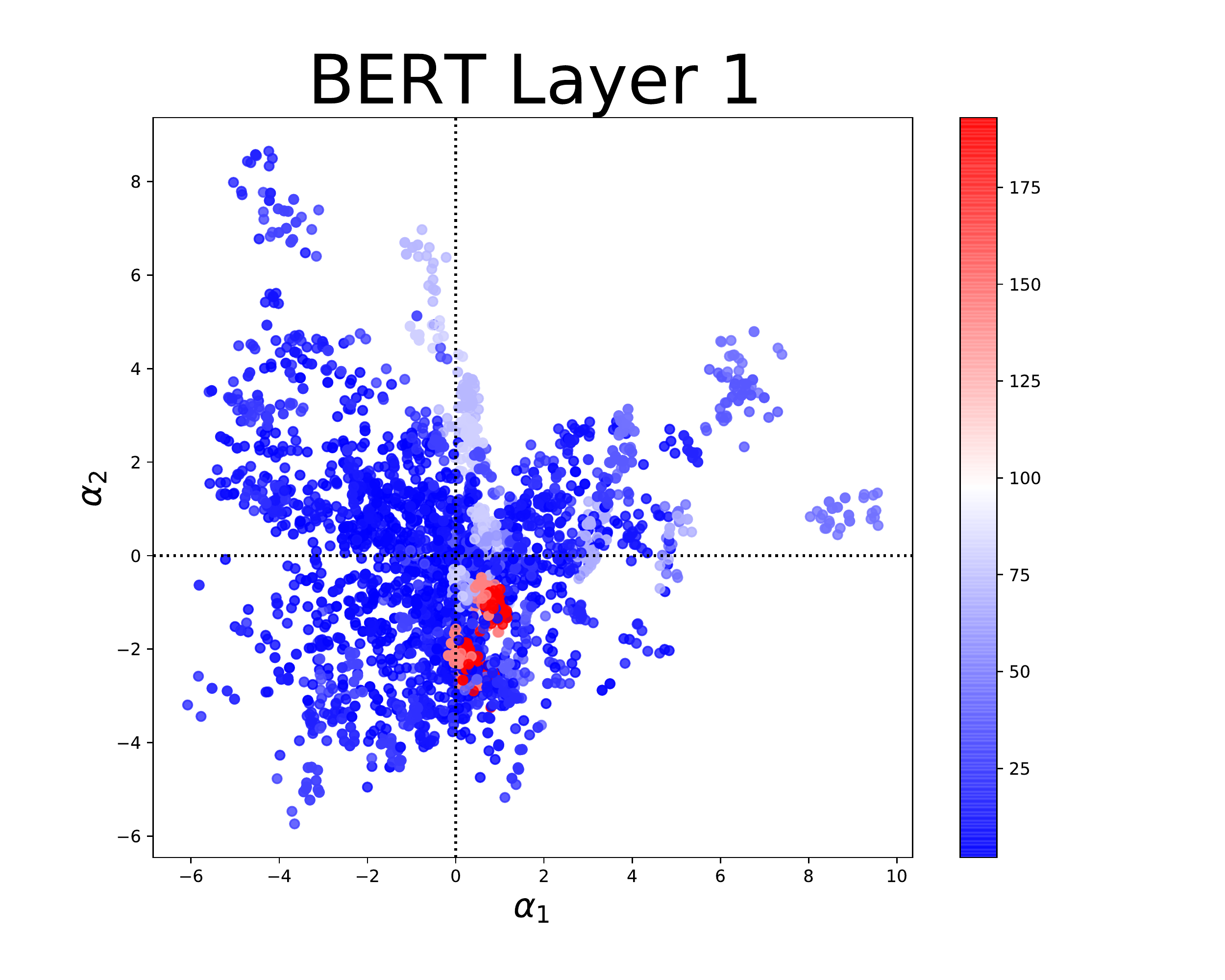}}\\
\subfloat[original]{\label{fig:e}\includegraphics[width=0.2\linewidth]{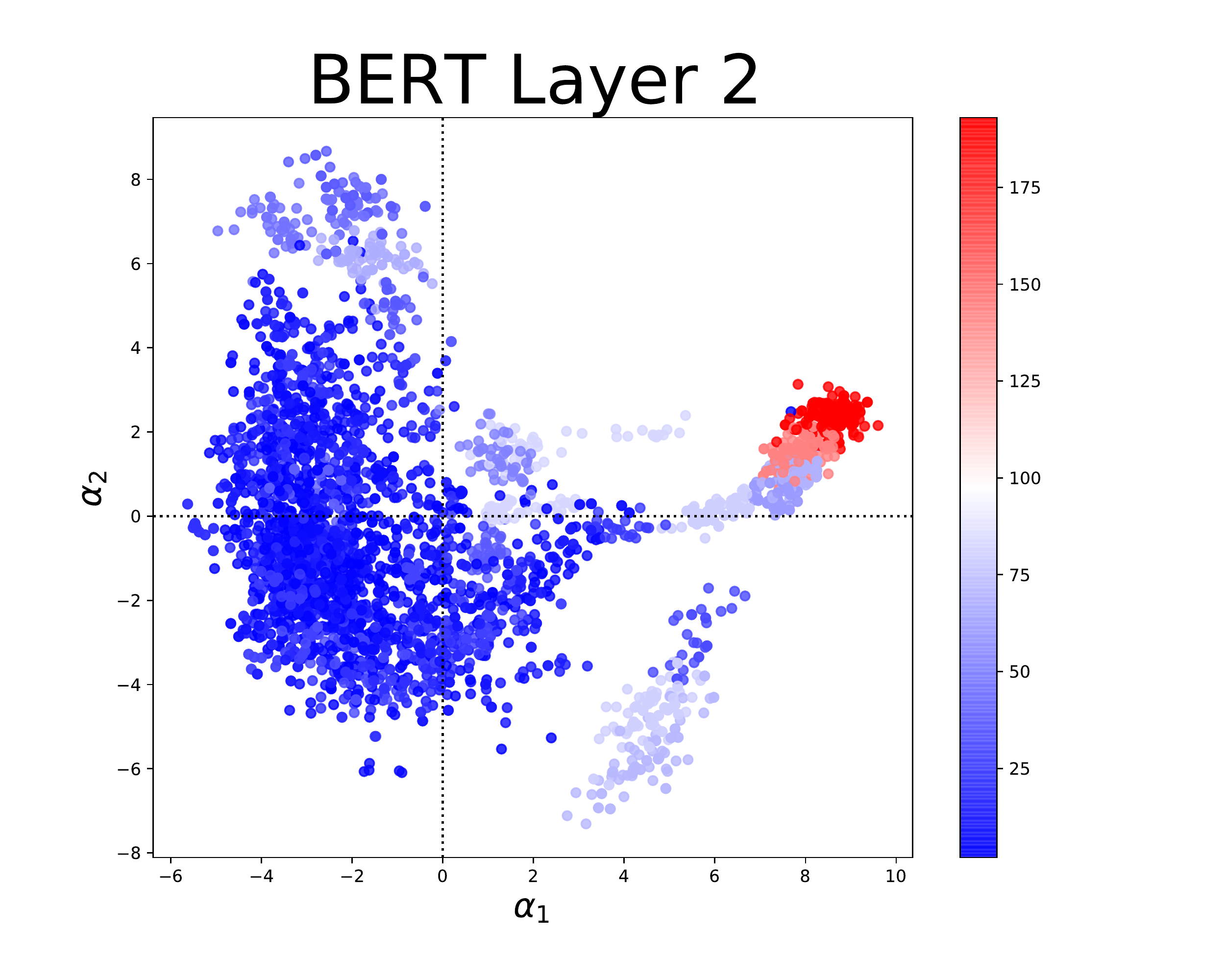}}
\subfloat[retrofitted]{\label{fig:f}\includegraphics[width=0.2\linewidth]{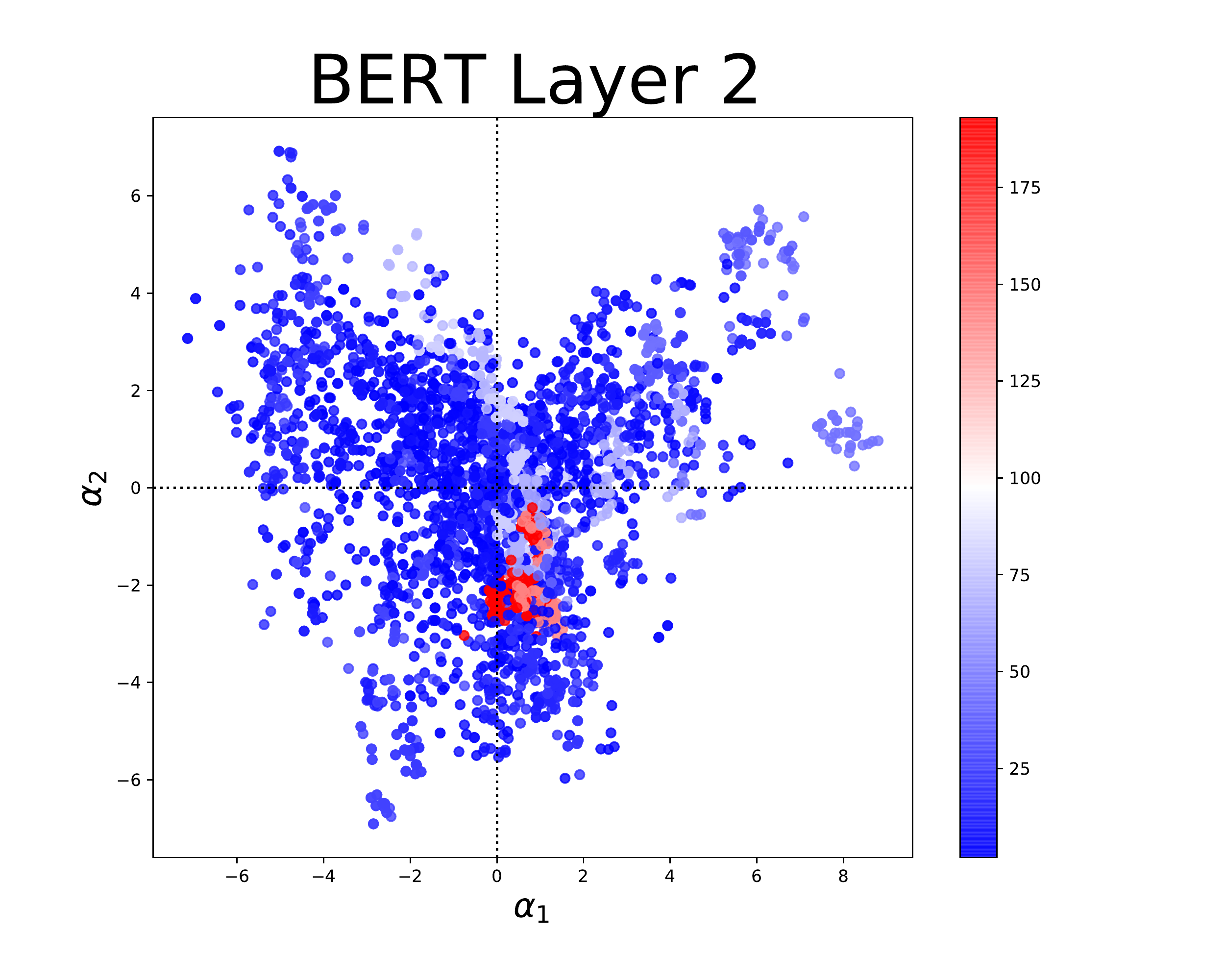}}
\subfloat[original]{\label{fig:g}\includegraphics[width=0.2\linewidth]{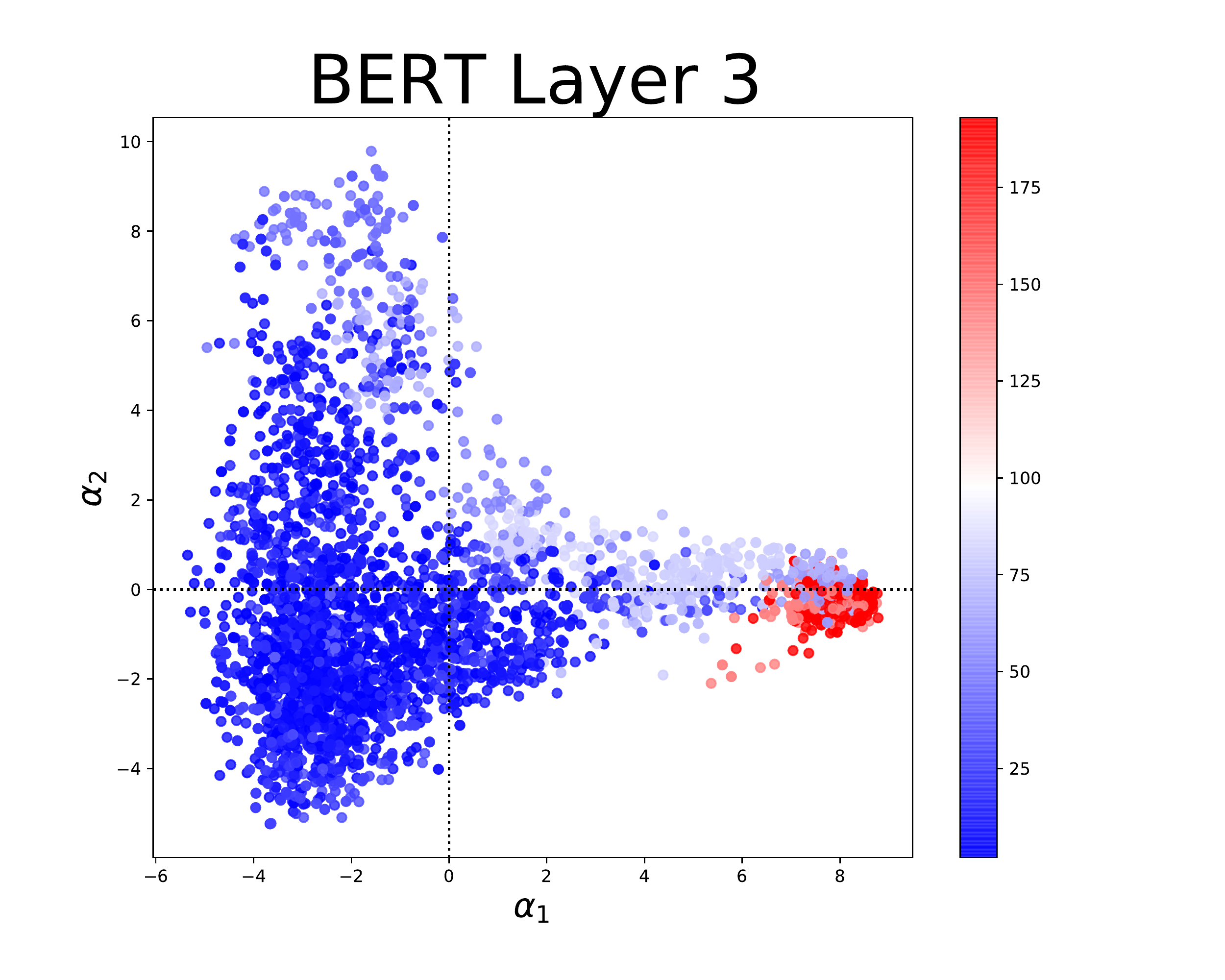}}
\subfloat[retrofitted]{\label{fig:h}\includegraphics[width=0.2\linewidth]{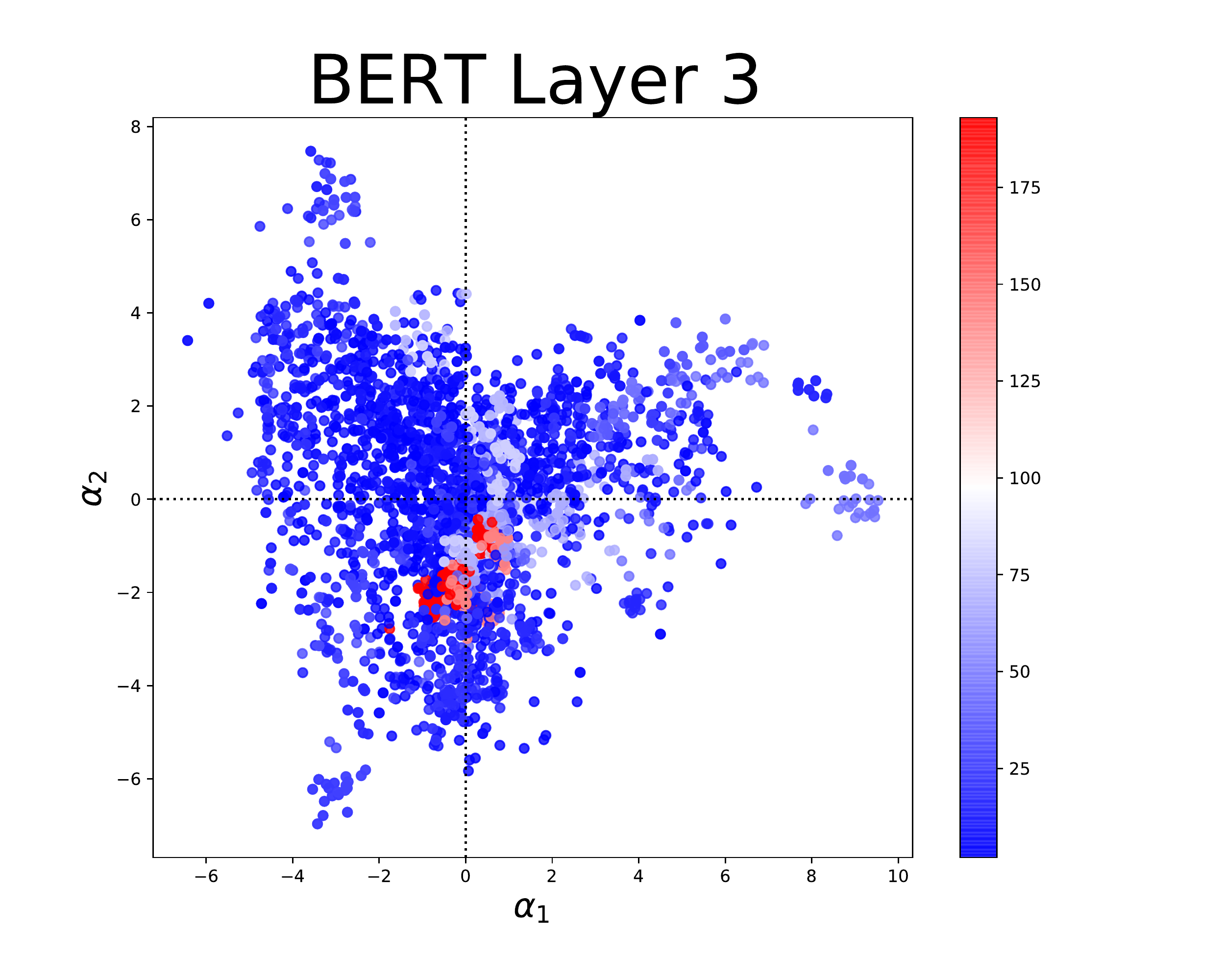}}\\
\subfloat[original]{\label{fig:i}\includegraphics[width=0.2\linewidth]{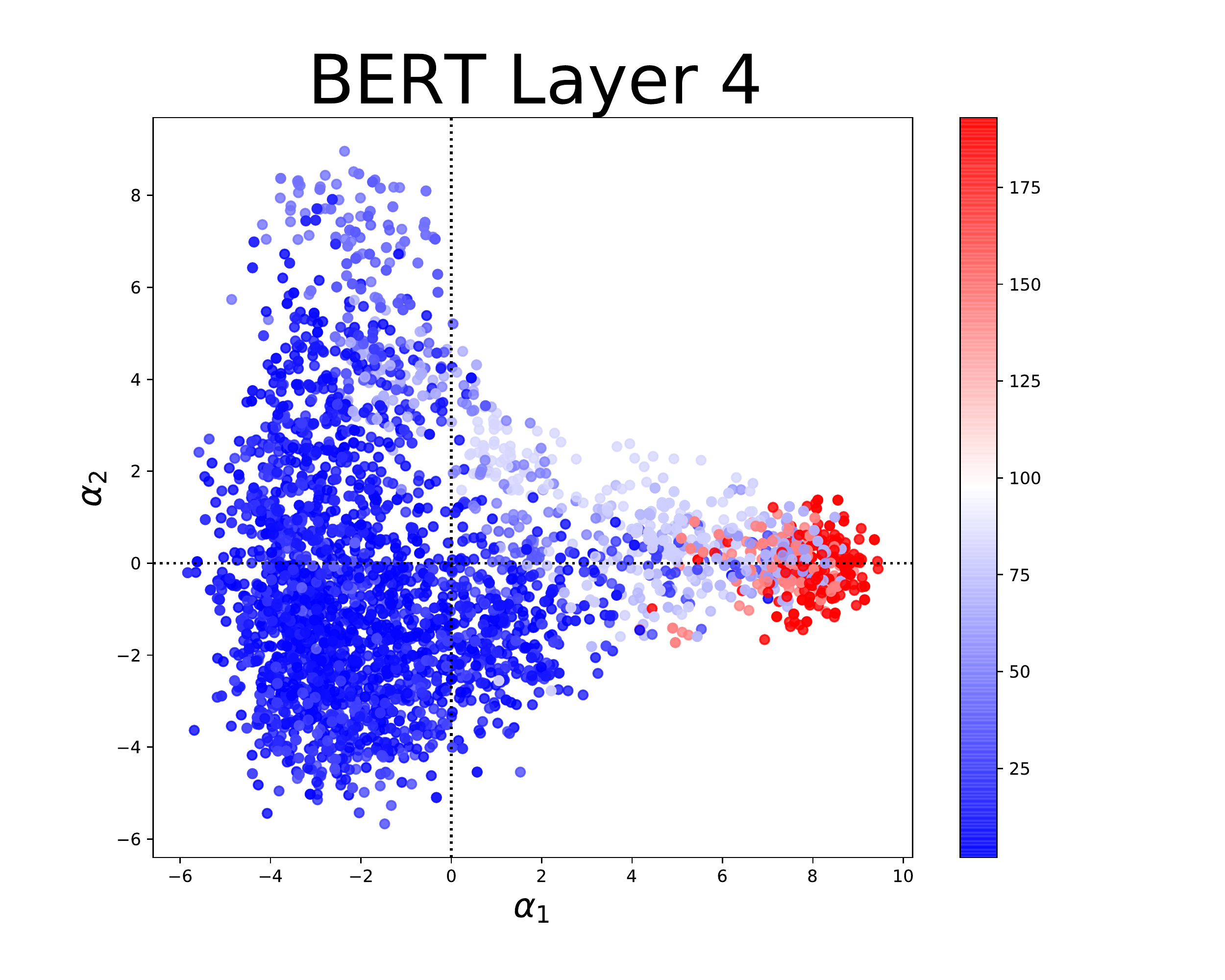}}
\subfloat[retrofitted]{\label{fig:j}\includegraphics[width=0.2\linewidth]{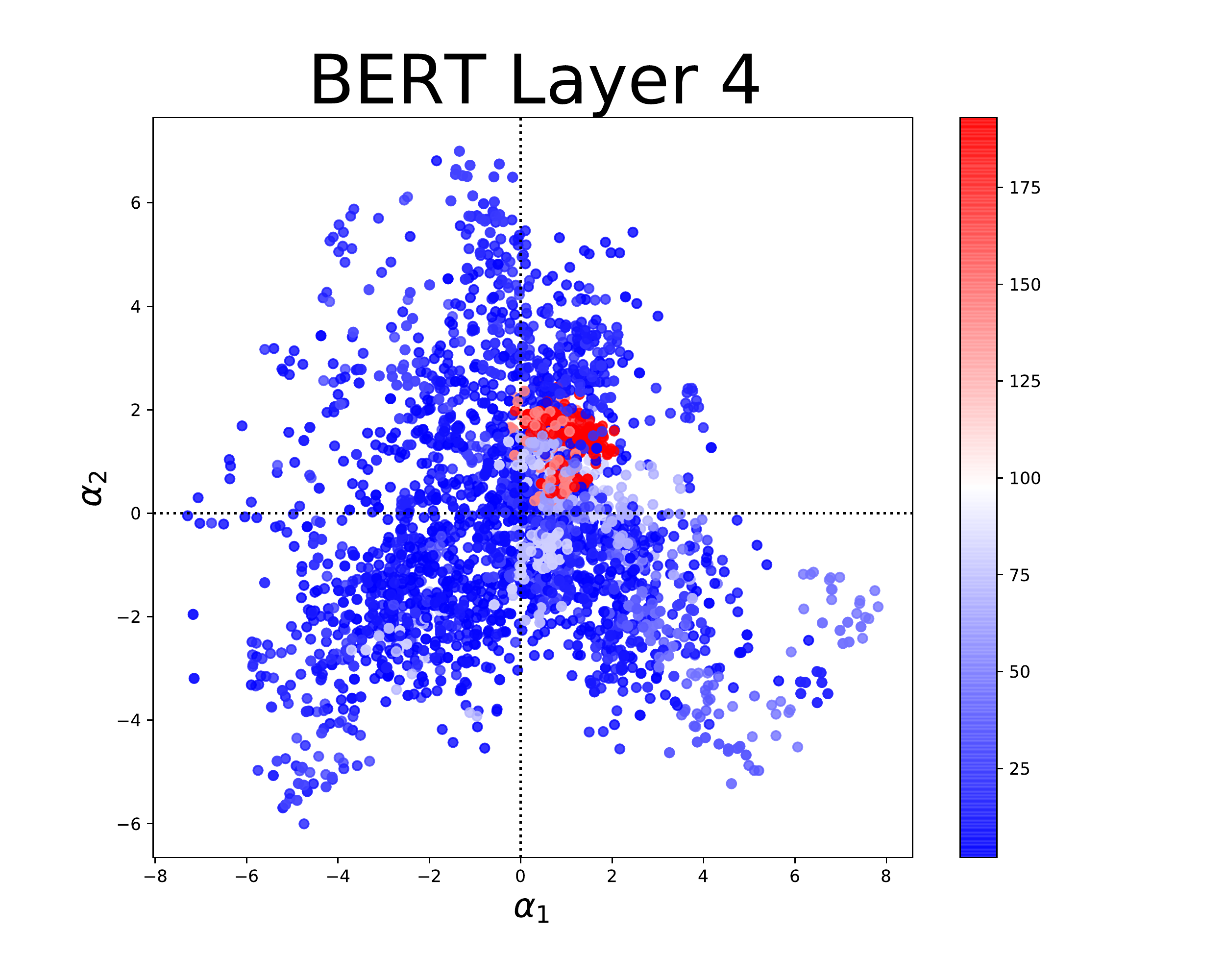}}
\subfloat[original]{\label{fig:k}\includegraphics[width=0.2\linewidth]{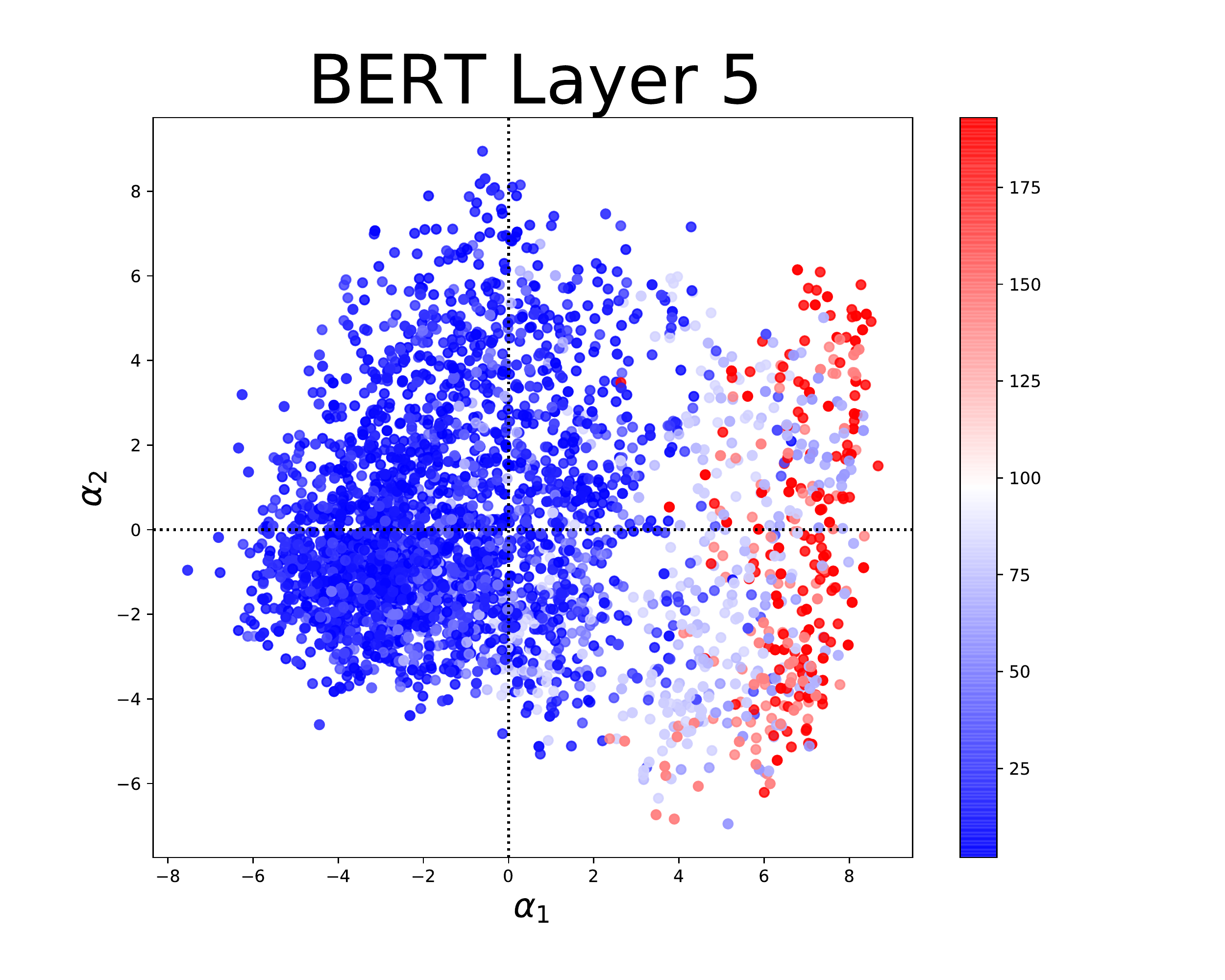}}
\subfloat[retrofitted]{\label{fig:l}\includegraphics[width=0.2\linewidth]{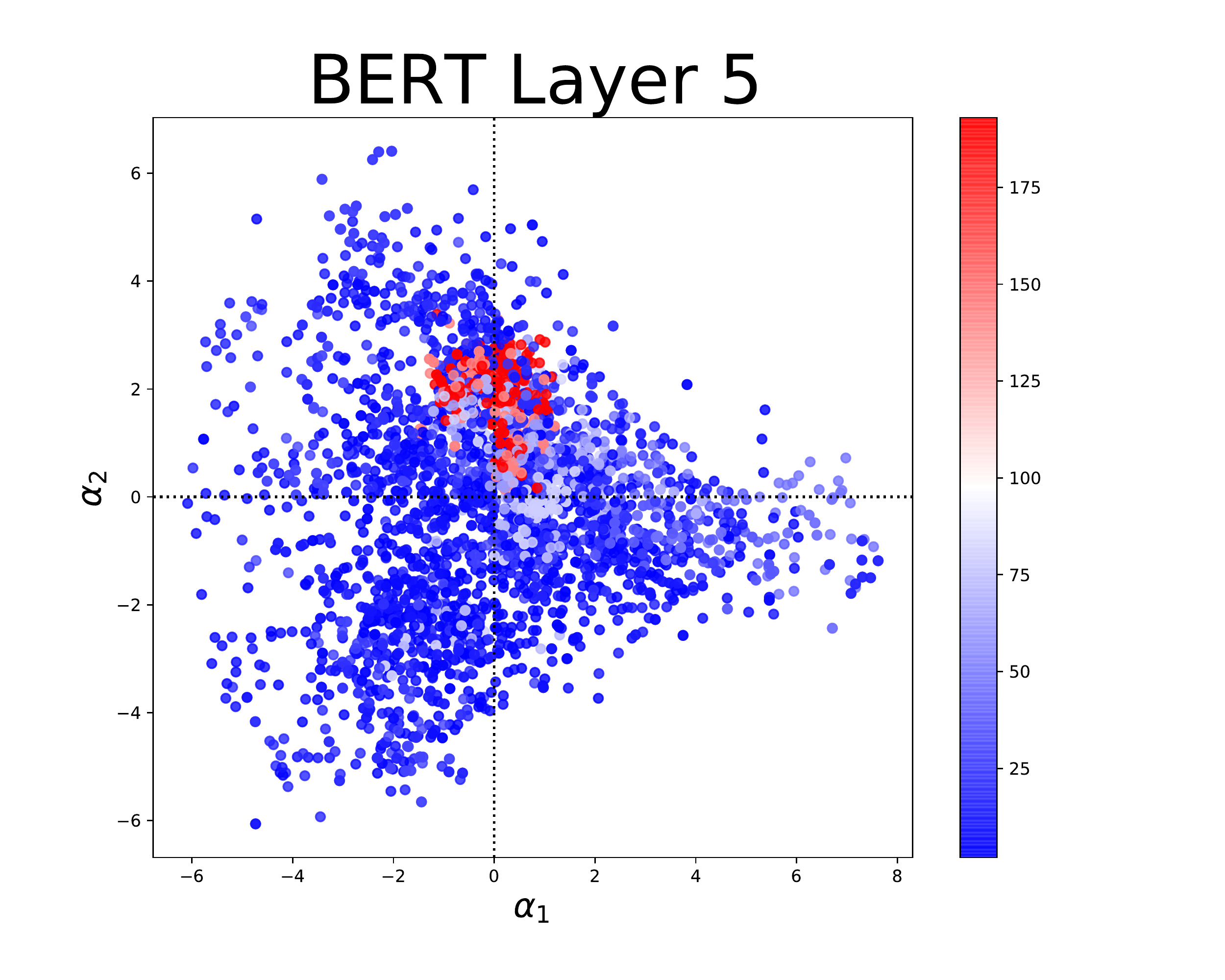}}\\
\subfloat[original]{\label{fig:m}\includegraphics[width=0.2\linewidth]{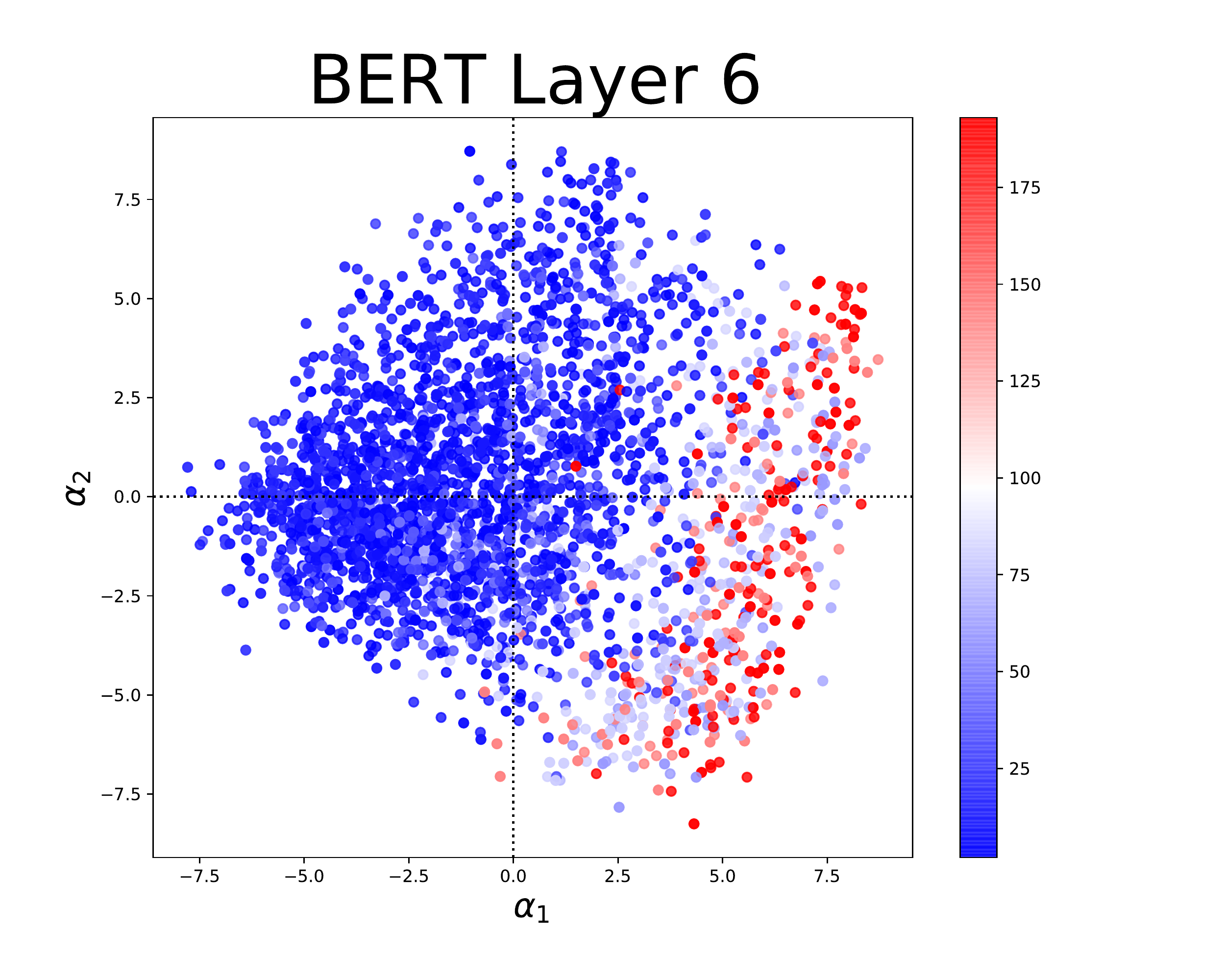}}
\subfloat[retrofitted]{\label{fig:n}\includegraphics[width=0.2\linewidth]{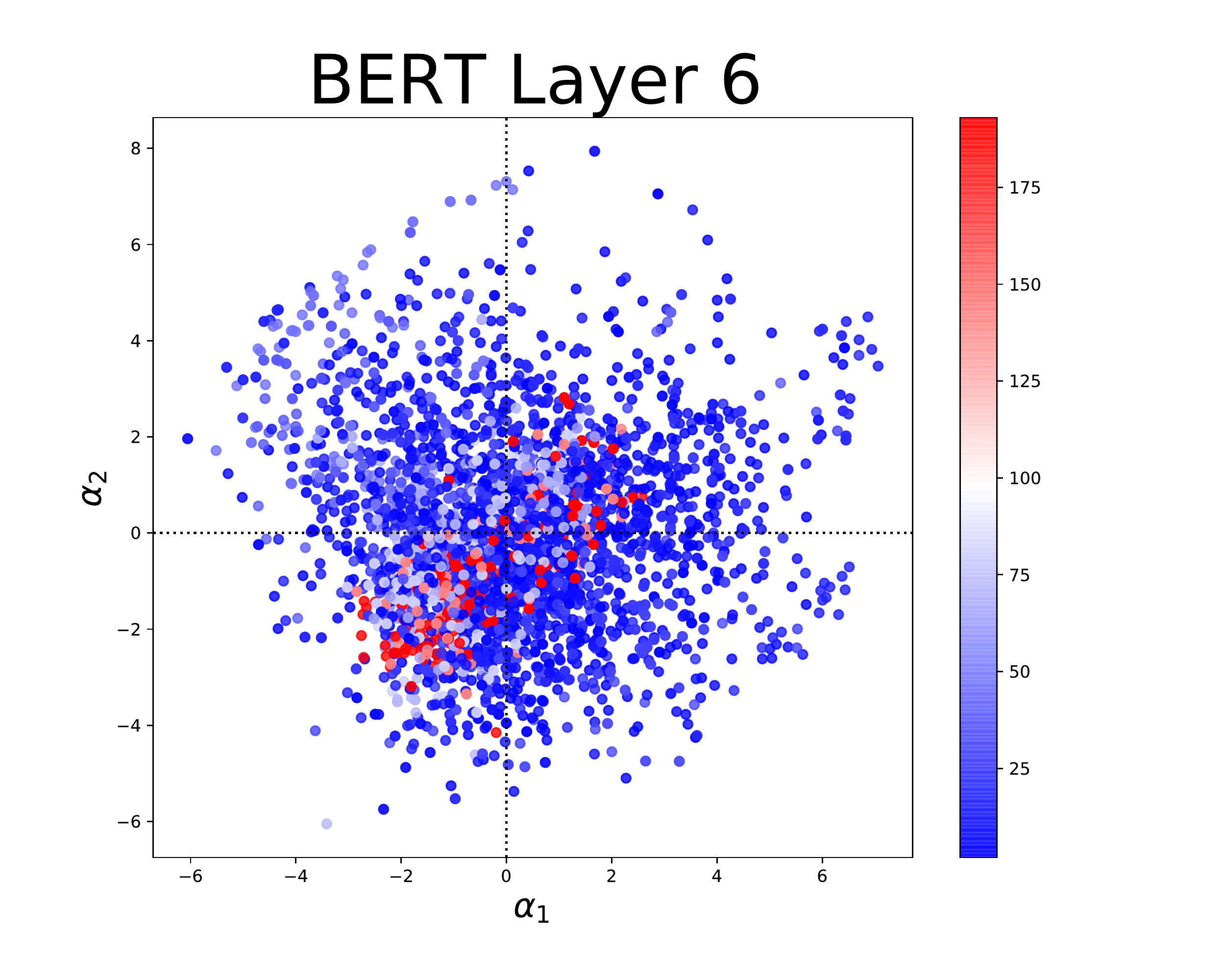}}
\subfloat[original]{\label{fig:o}\includegraphics[width=0.2\linewidth]{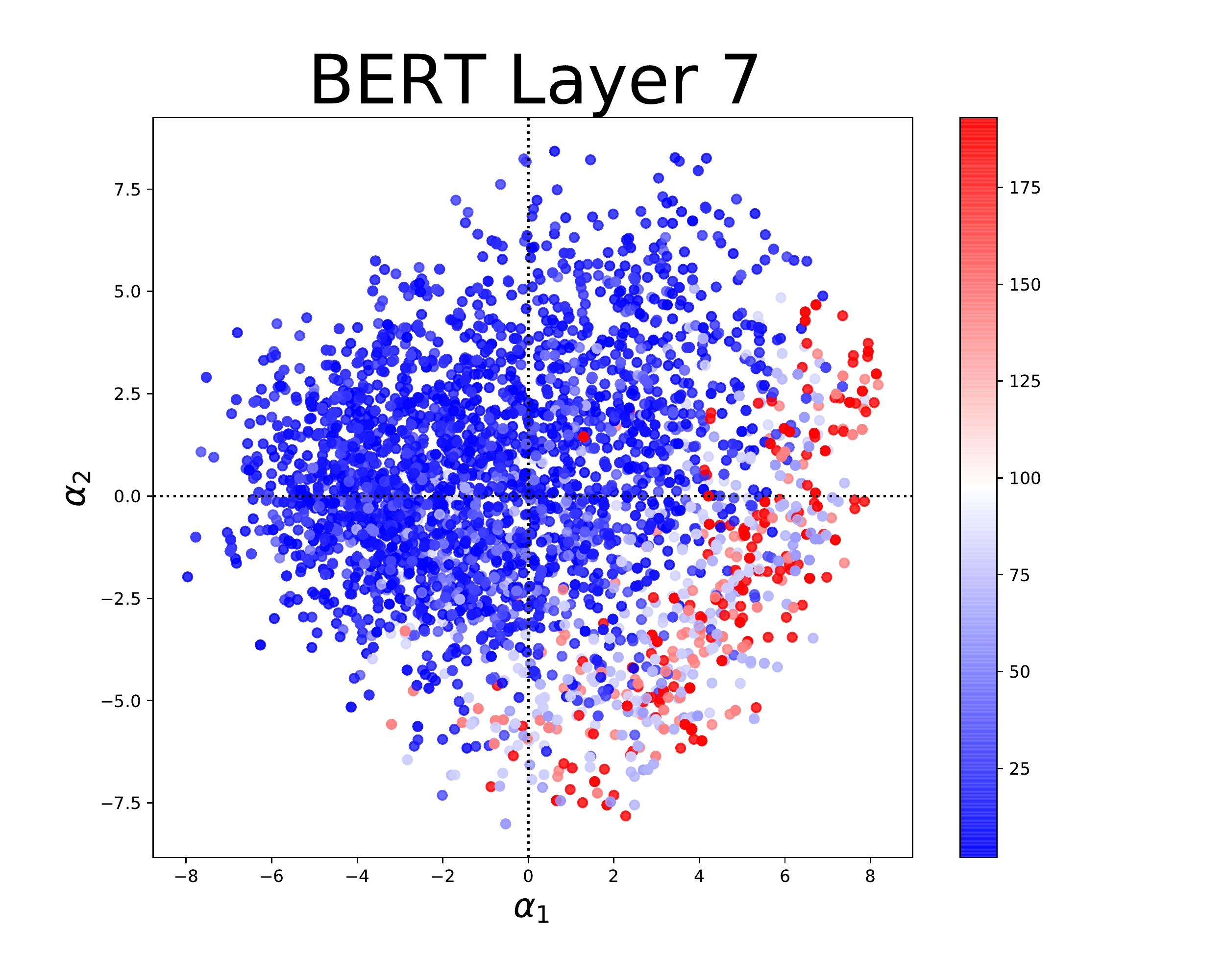}}
\subfloat[retrofitted]{\label{fig:p}\includegraphics[width=0.2\linewidth]{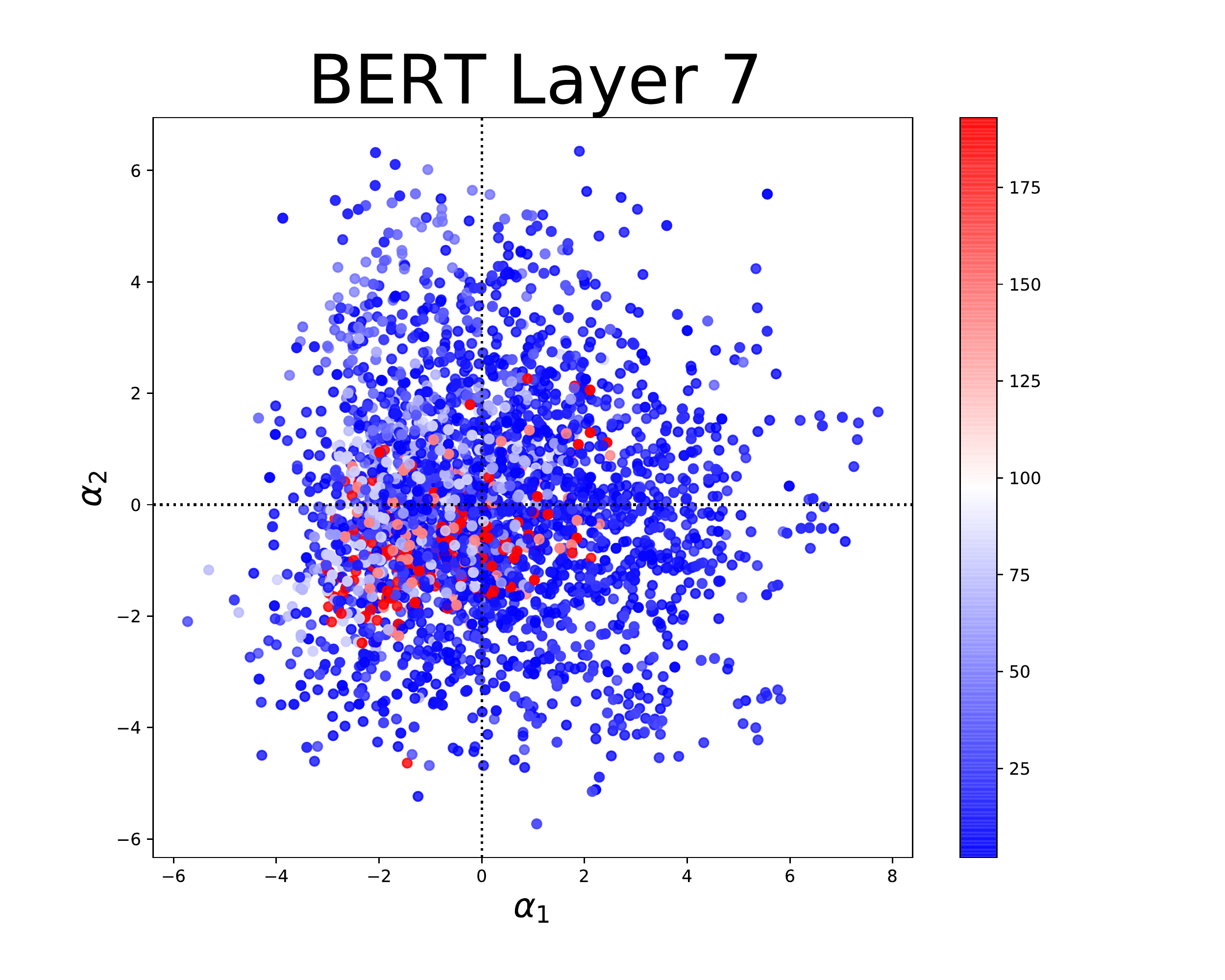}}\\
\subfloat[original]{\label{fig:q}\includegraphics[width=0.2\linewidth]{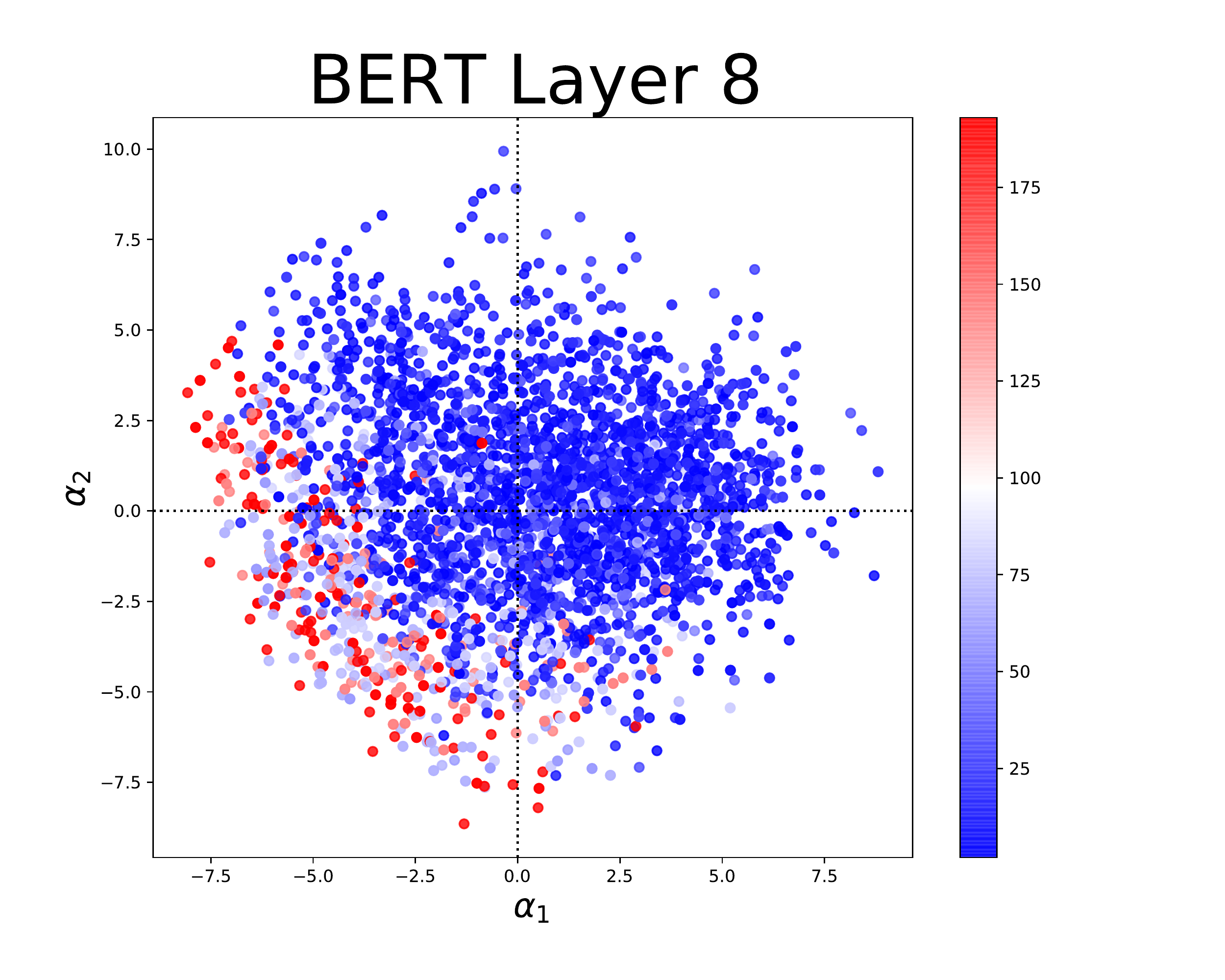}}
\subfloat[retrofitted]{\label{fig:r}\includegraphics[width=0.2\linewidth]{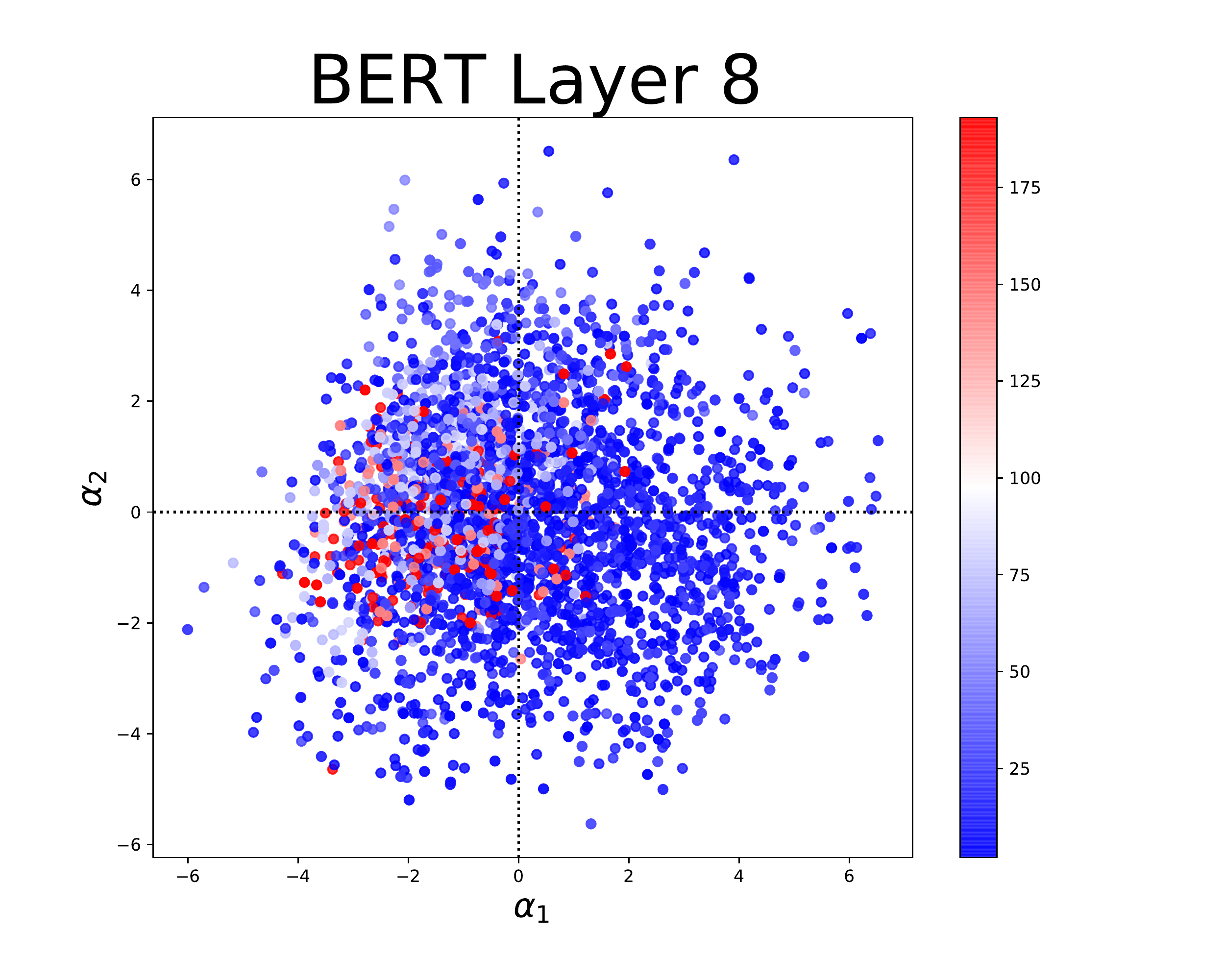}}
\subfloat[original]{\label{fig:s}\includegraphics[width=0.2\linewidth]{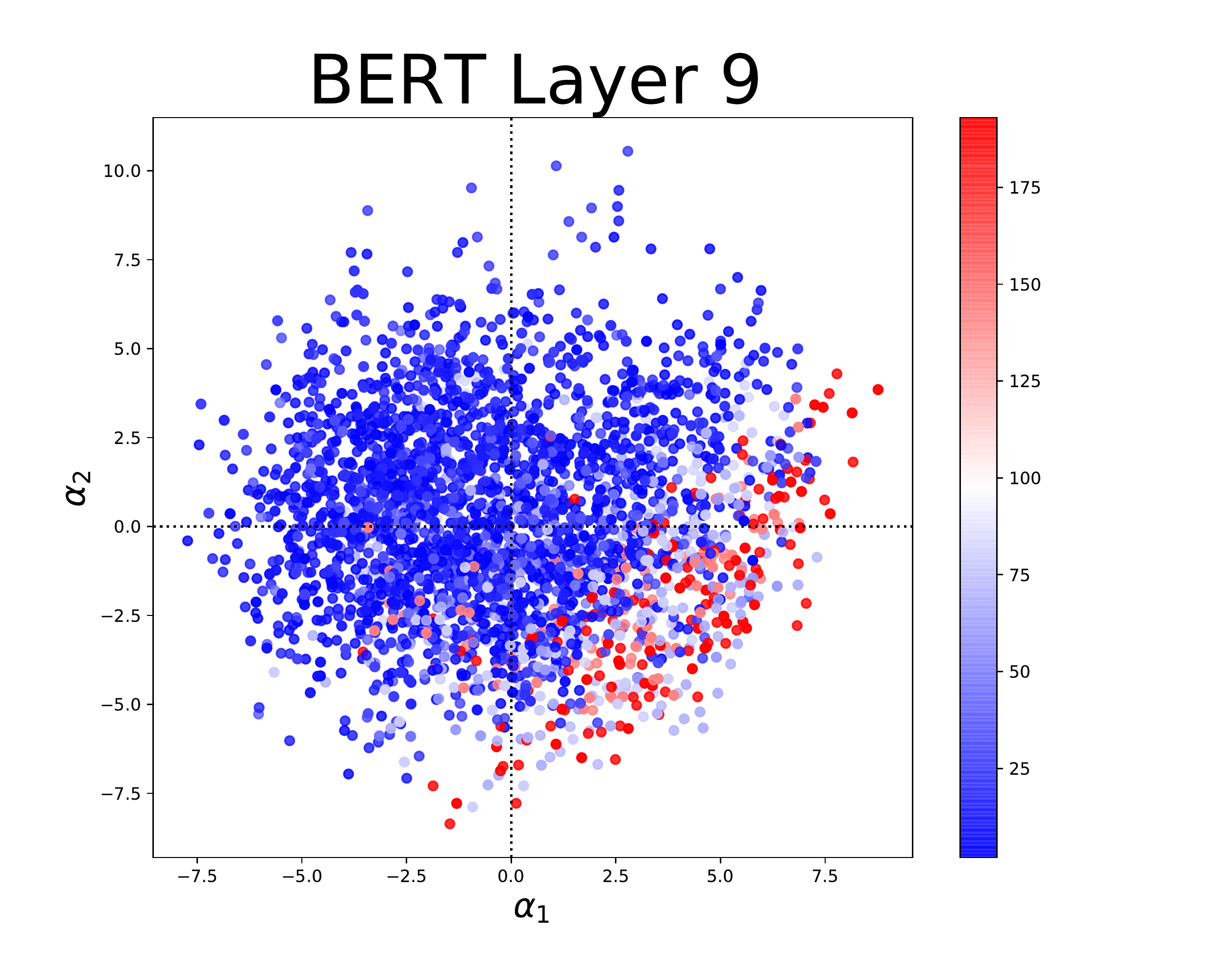}}
\subfloat[retrofitted]{\label{fig:t}\includegraphics[width=0.2\linewidth]{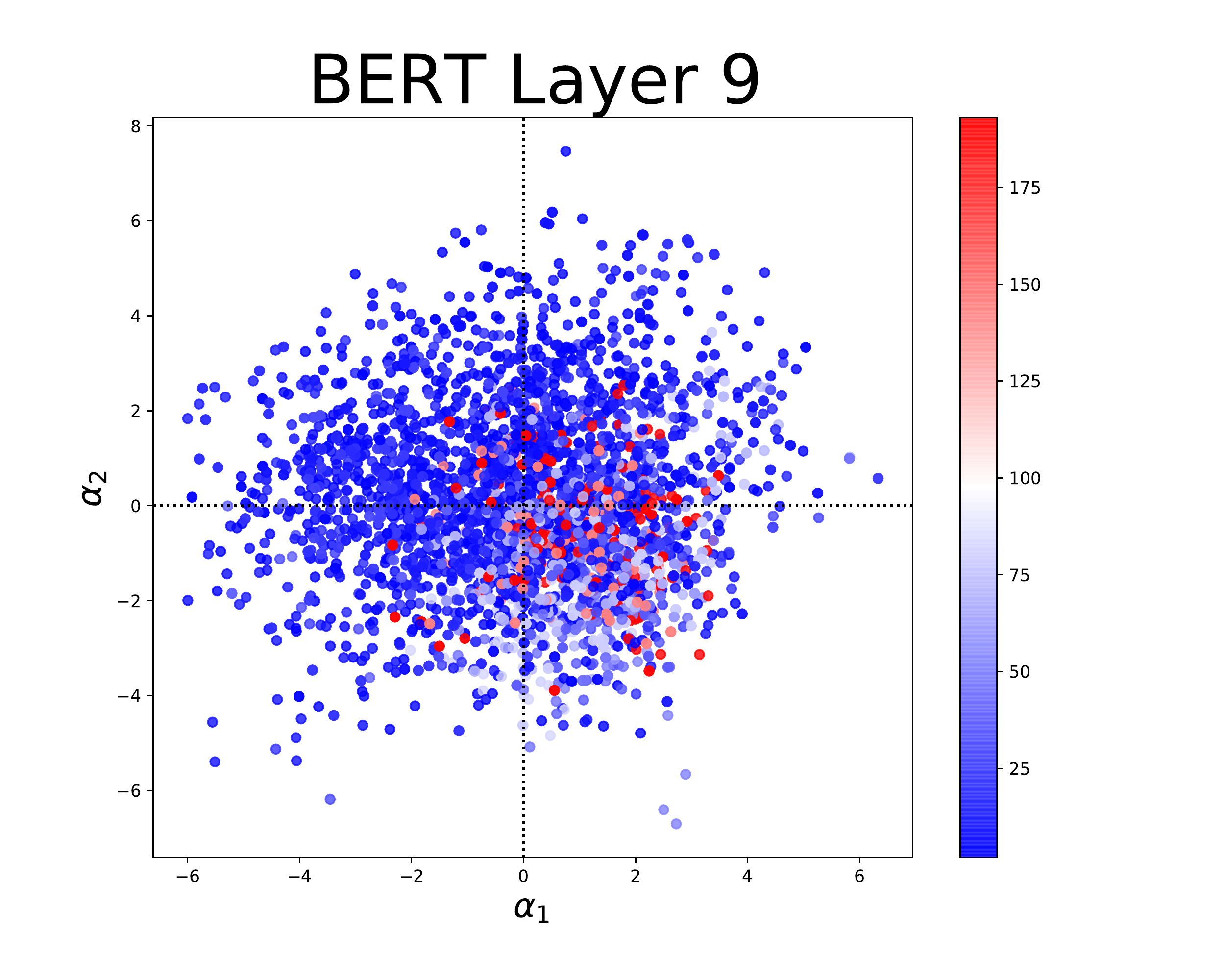}}\\
\subfloat[original]{\label{fig:u}\includegraphics[width=0.2\linewidth]{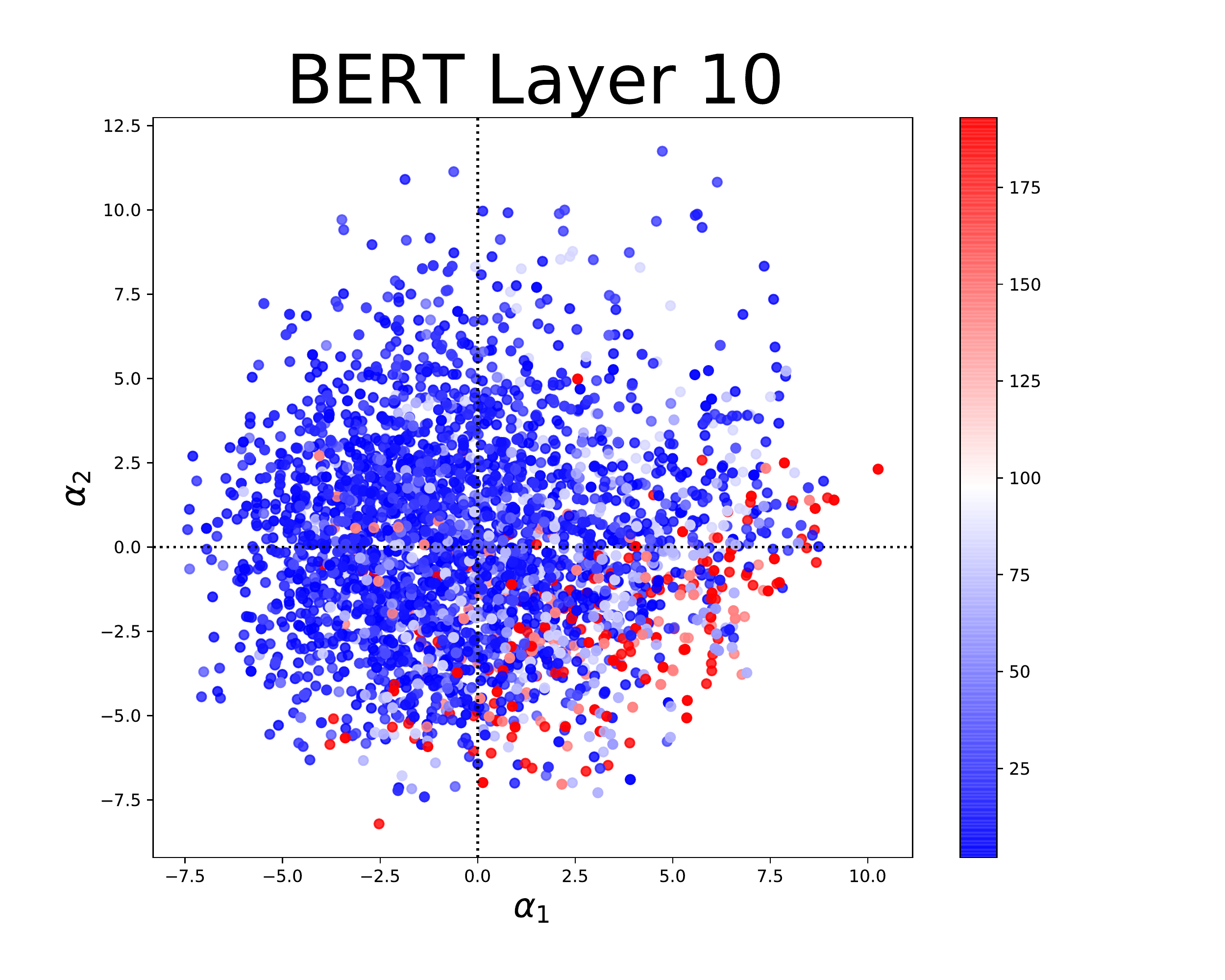}}
\subfloat[retrofitted]{\label{fig:v}\includegraphics[width=0.2\linewidth]{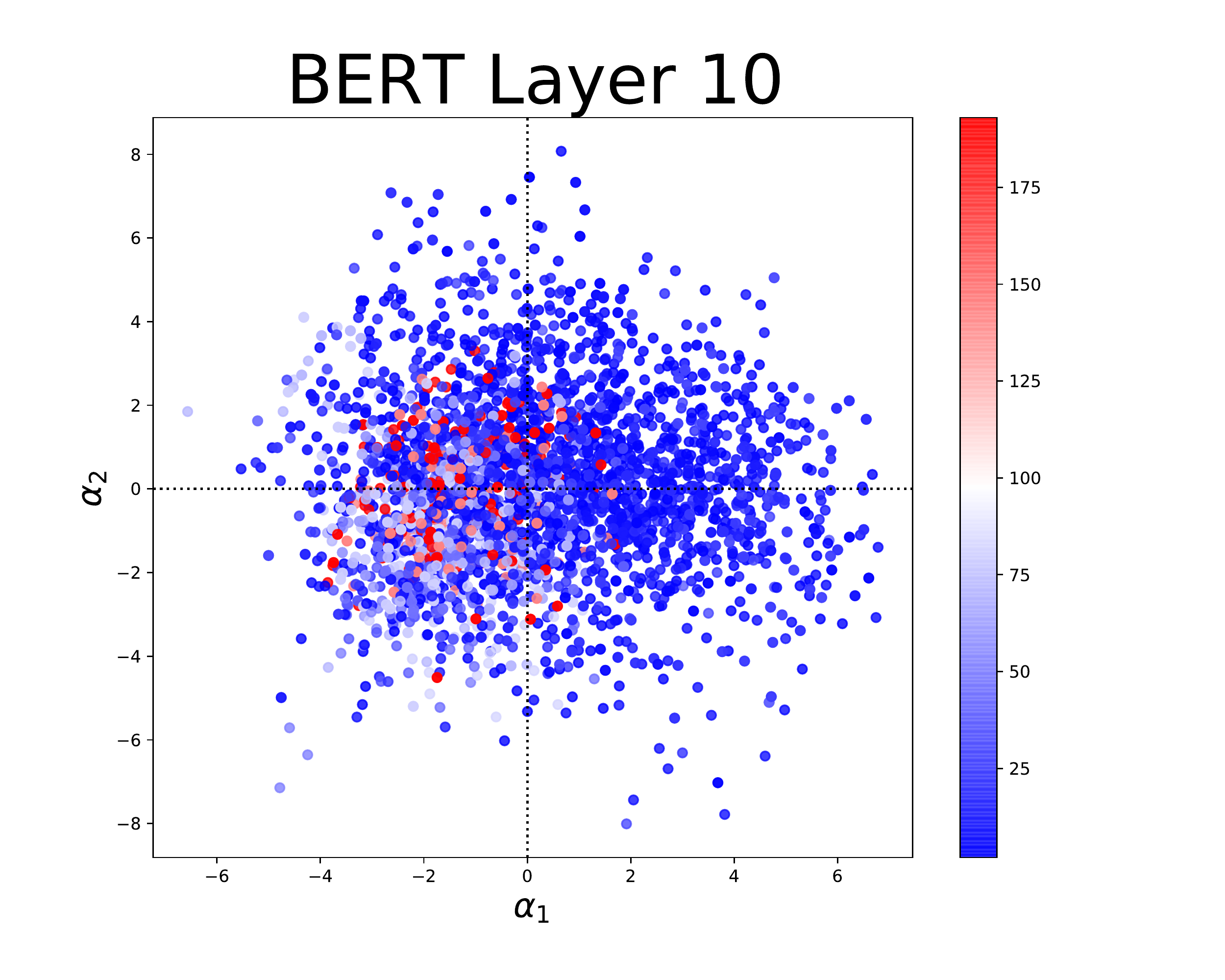}}
\subfloat[original]{\label{fig:w}\includegraphics[width=0.2\linewidth]{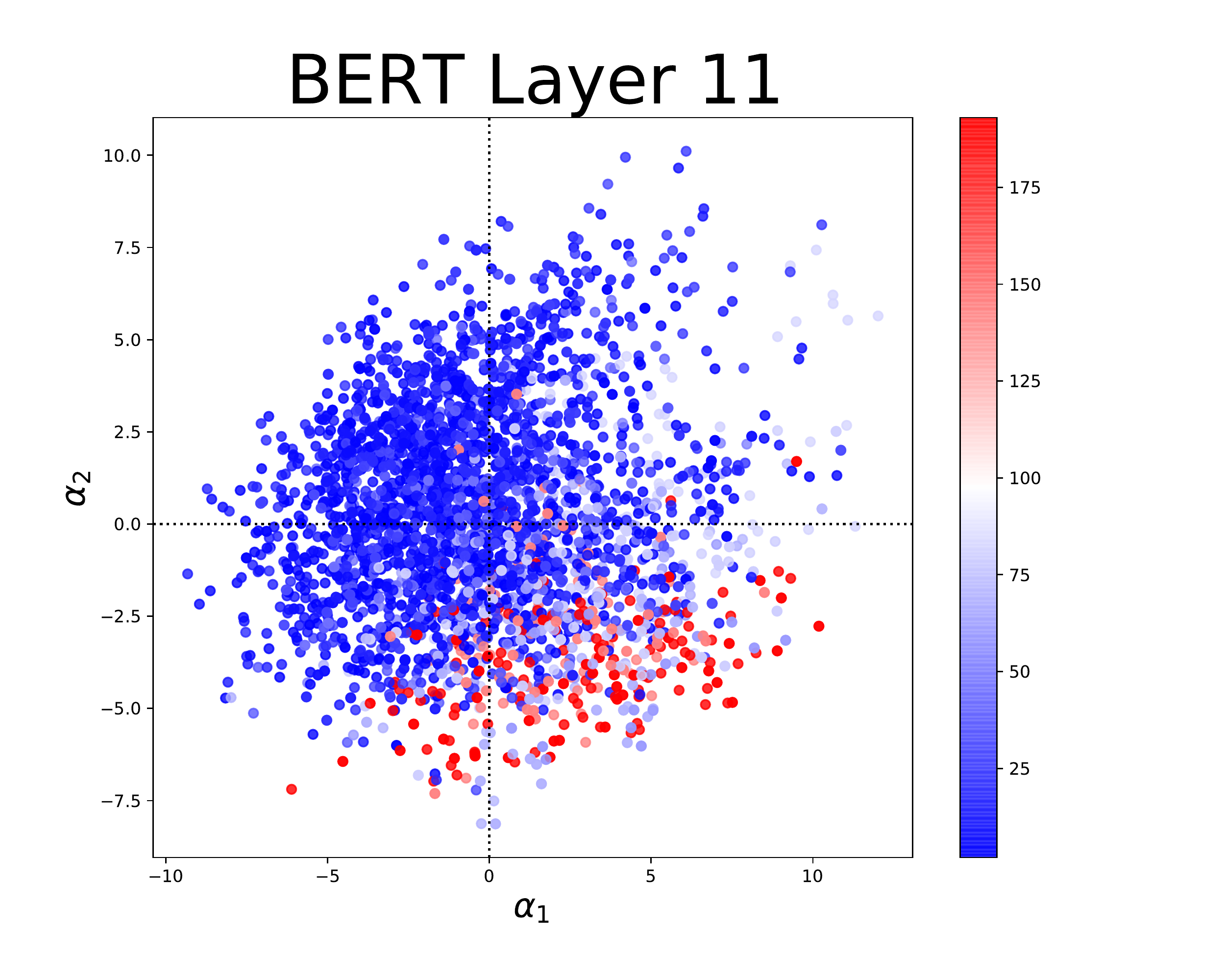}}
\subfloat[retrofitted]{\label{fig:x}\includegraphics[width=0.2\linewidth]{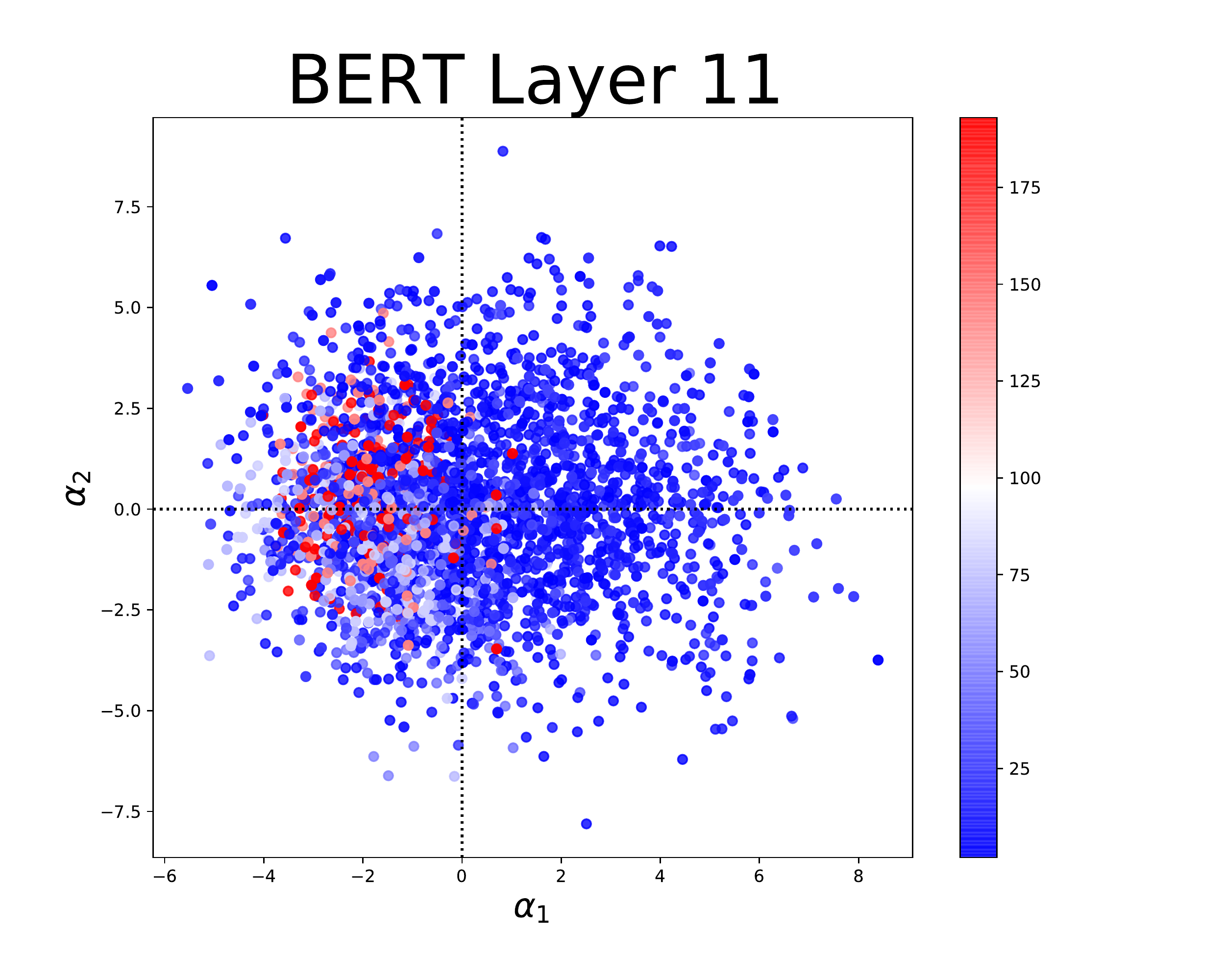}}\\
\subfloat[original]{\label{fig:y}\includegraphics[width=0.2\linewidth]{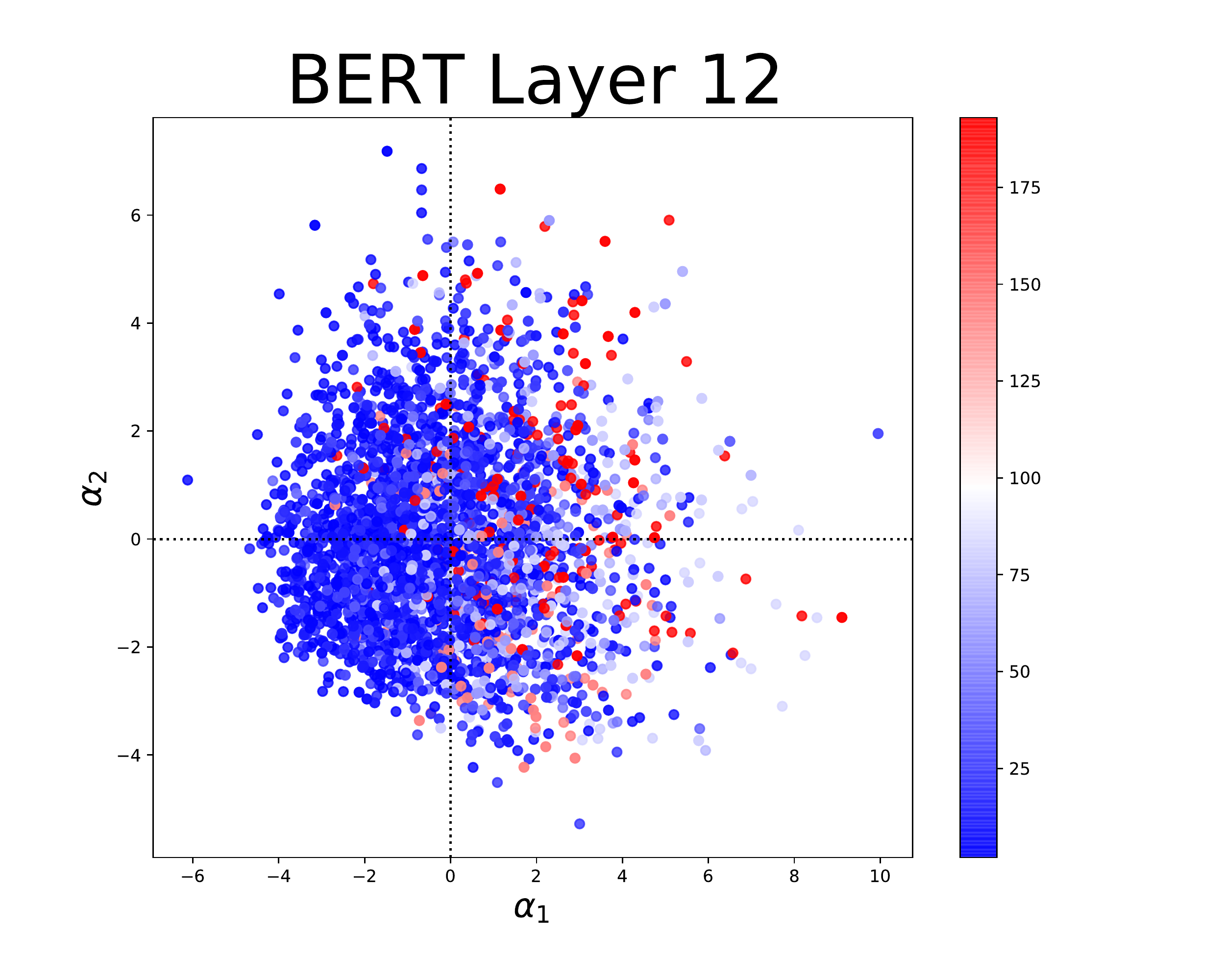}}
\subfloat[retrofitted]{\label{fig:z}\includegraphics[width=0.2\linewidth]{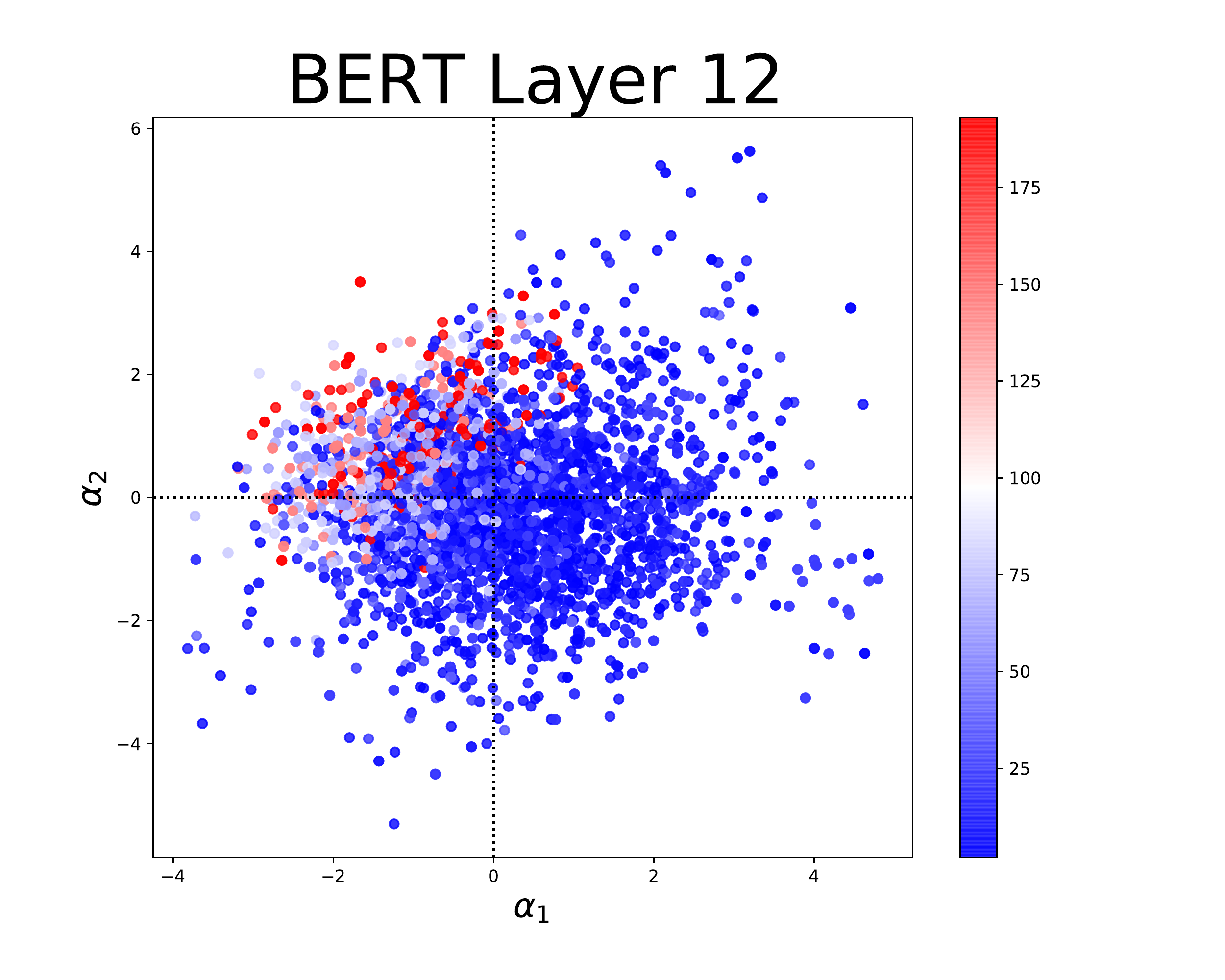}}
\caption{PCA Plots of BERT Word Representations.}
\label{fig:bert_fig}
\end{figure*}

\begin{figure*}
\centering
\subfloat[original]{\label{fig:a0}\includegraphics[width=0.2\linewidth]{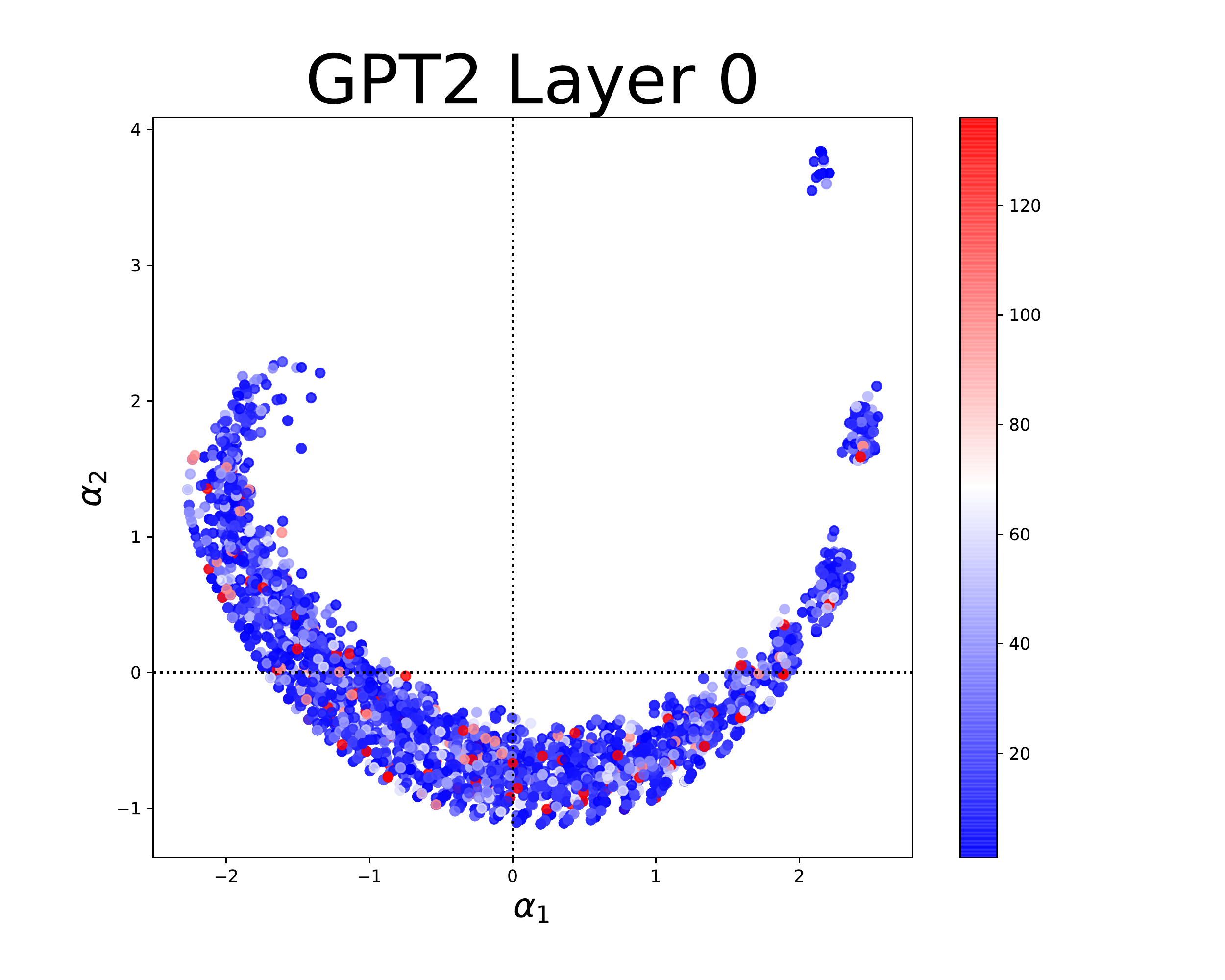}}
\subfloat[retrofitted]{\label{fig:0}\includegraphics[width=0.2\linewidth]{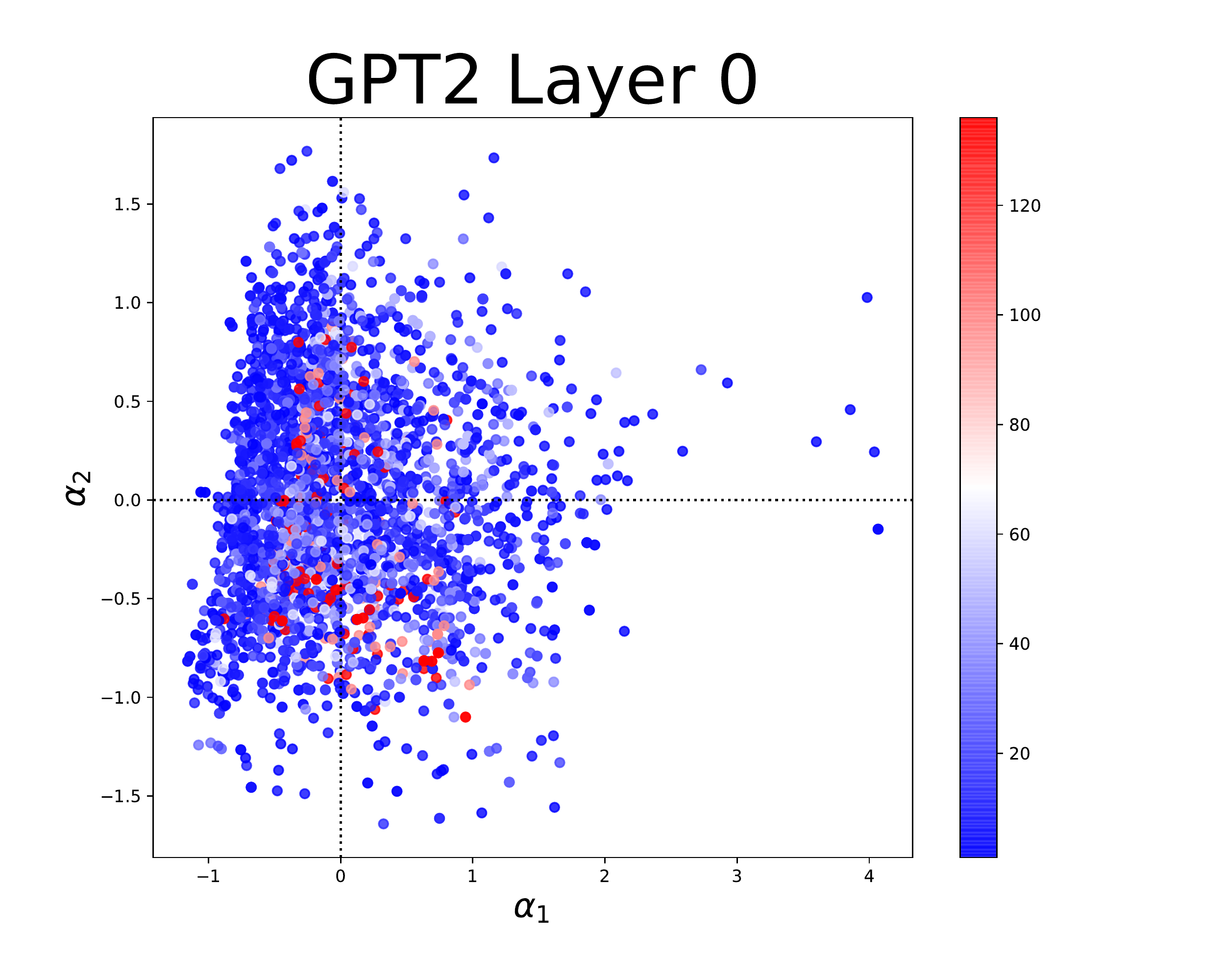}}
\subfloat[original]{\label{fig:c0}\includegraphics[width=0.2\linewidth]{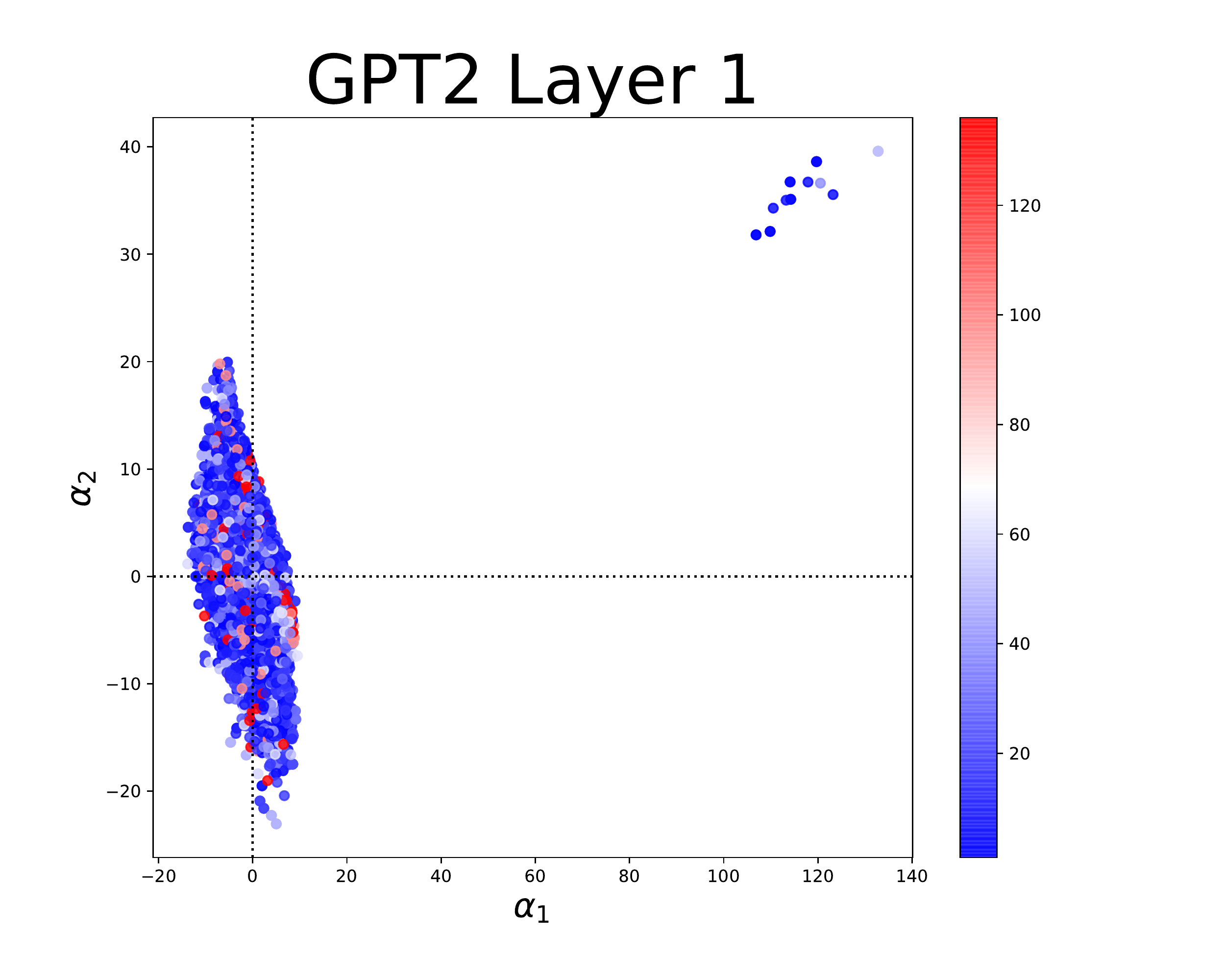}}
\subfloat[retrofitted]{\label{fig:d0}\includegraphics[width=0.2\linewidth]{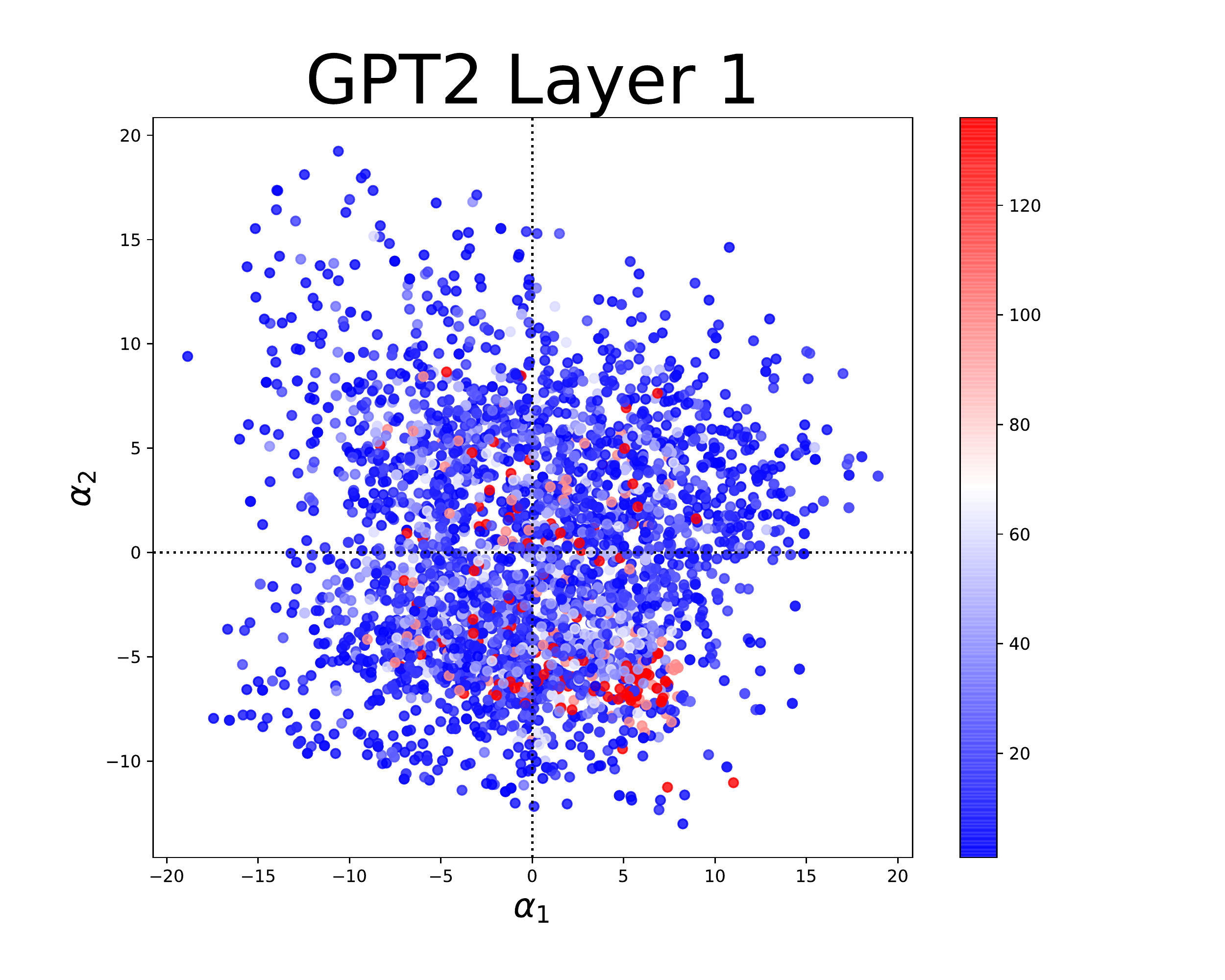}}\\
\subfloat[original]{\label{fig:e0}\includegraphics[width=0.2\linewidth]{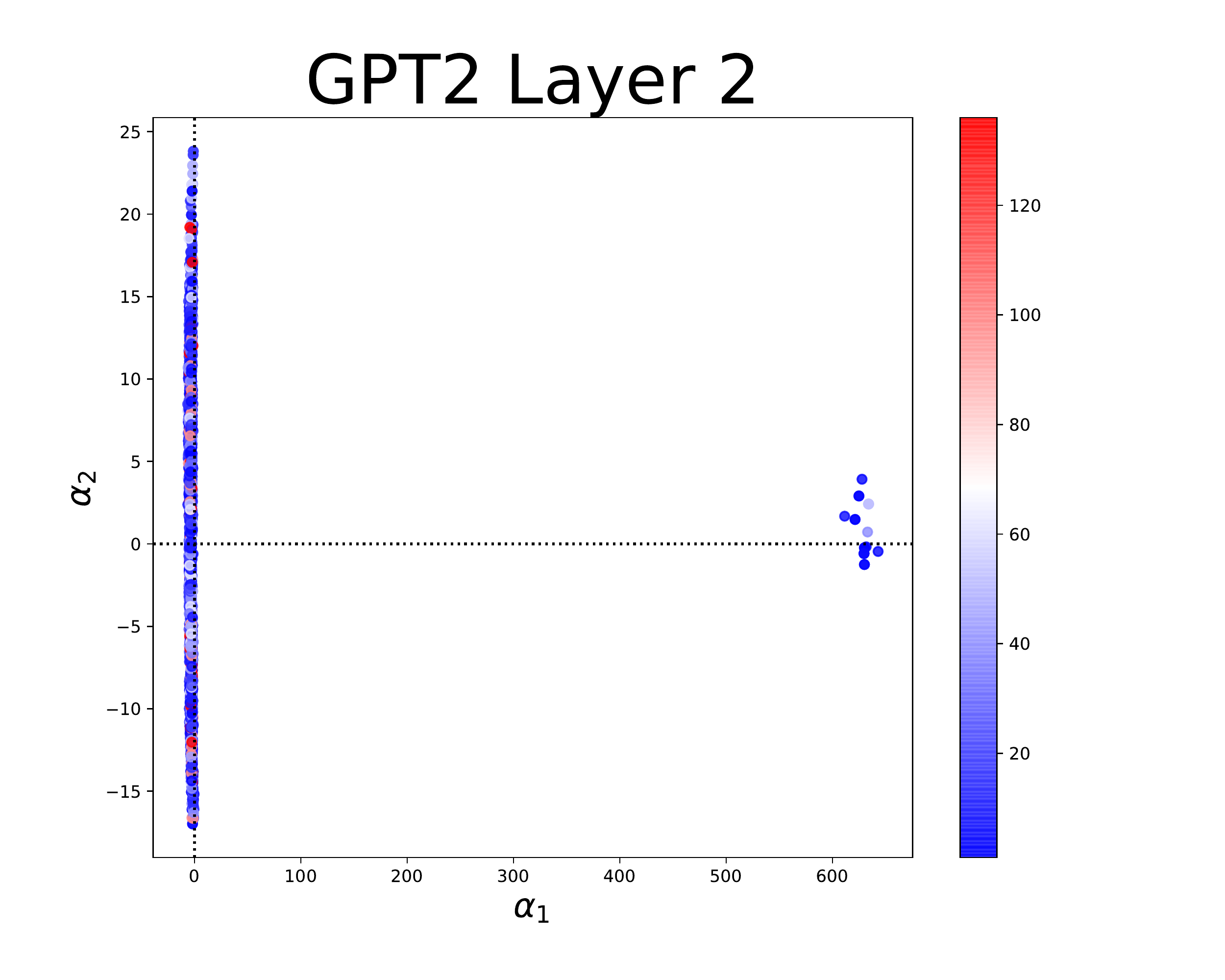}}
\subfloat[retrofitted]{\label{fig:f0}\includegraphics[width=0.2\linewidth]{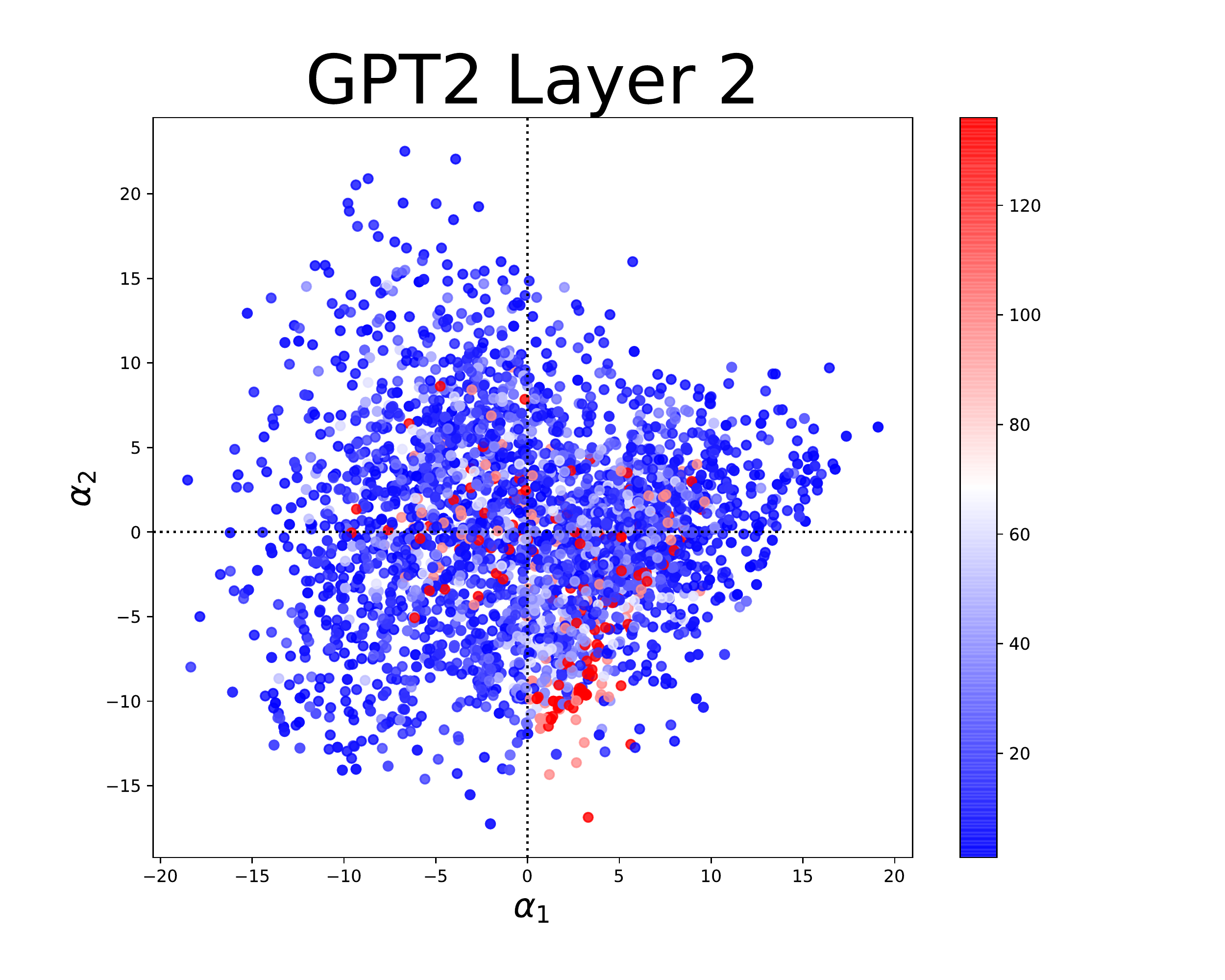}}
\subfloat[original]{\label{fig:g0}\includegraphics[width=0.2\linewidth]{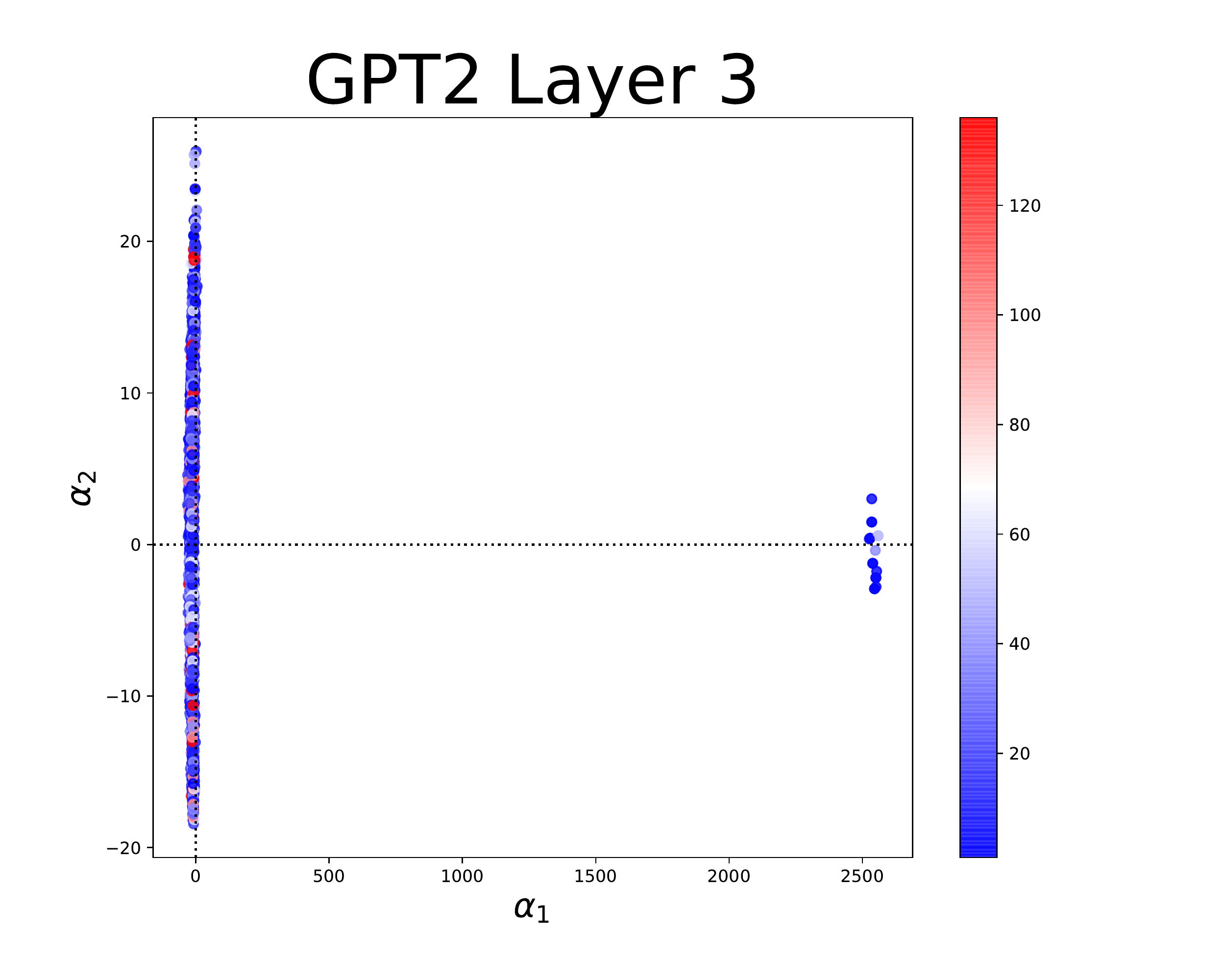}}
\subfloat[retrofitted]{\label{fig:h0}\includegraphics[width=0.2\linewidth]{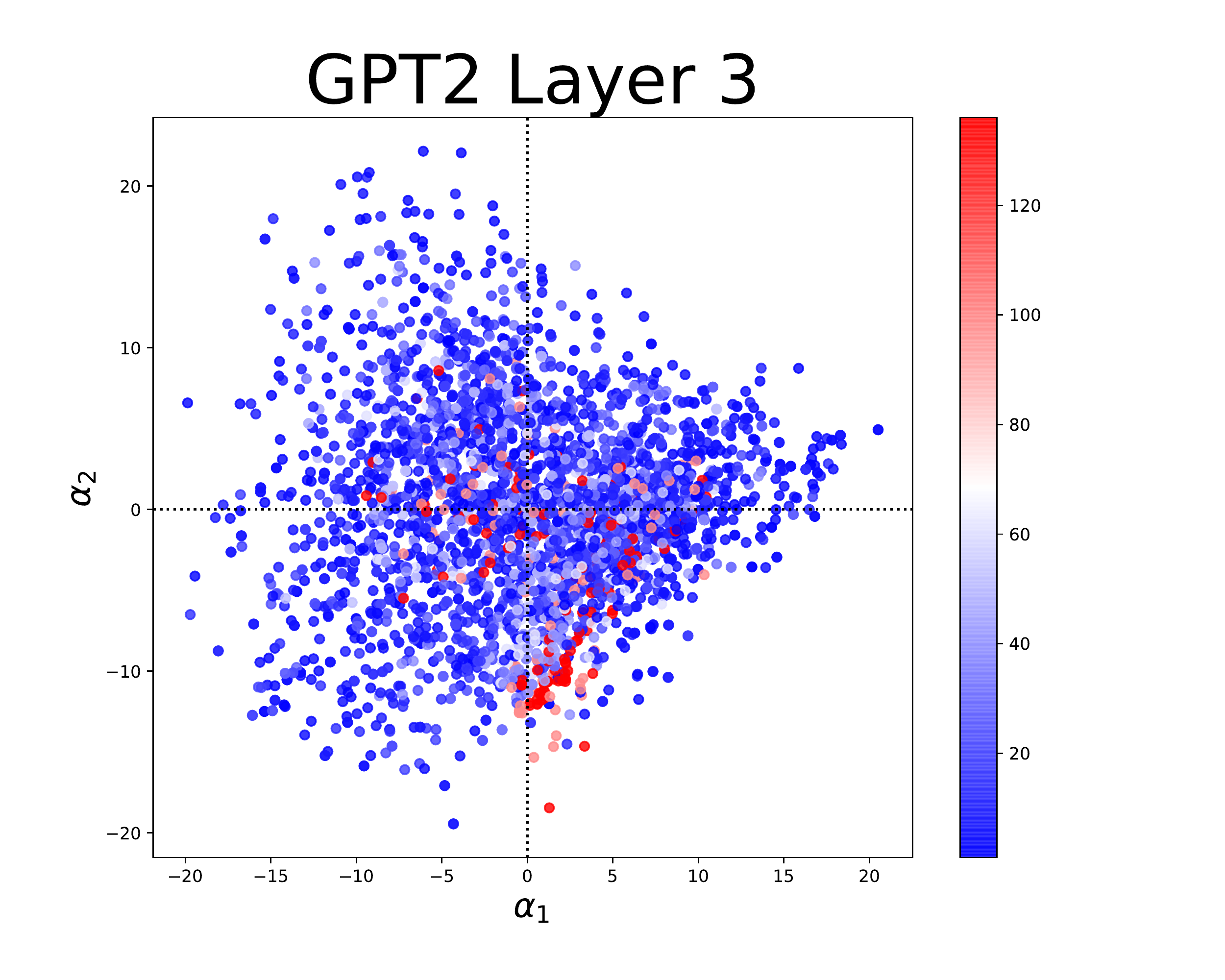}}\\
\subfloat[original]{\label{fig:i0}\includegraphics[width=0.2\linewidth]{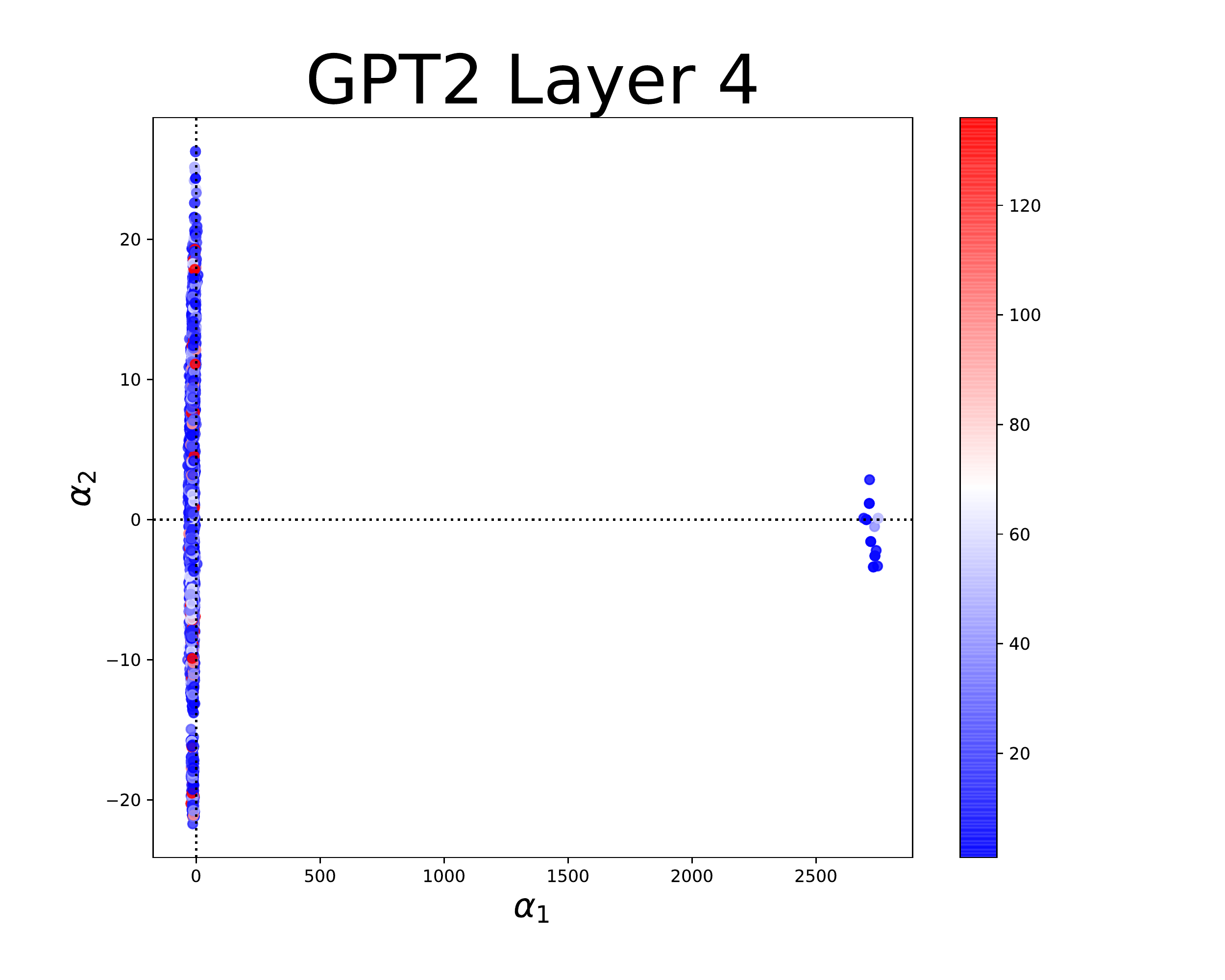}}
\subfloat[retrofitted]{\label{fig:j0}\includegraphics[width=0.2\linewidth]{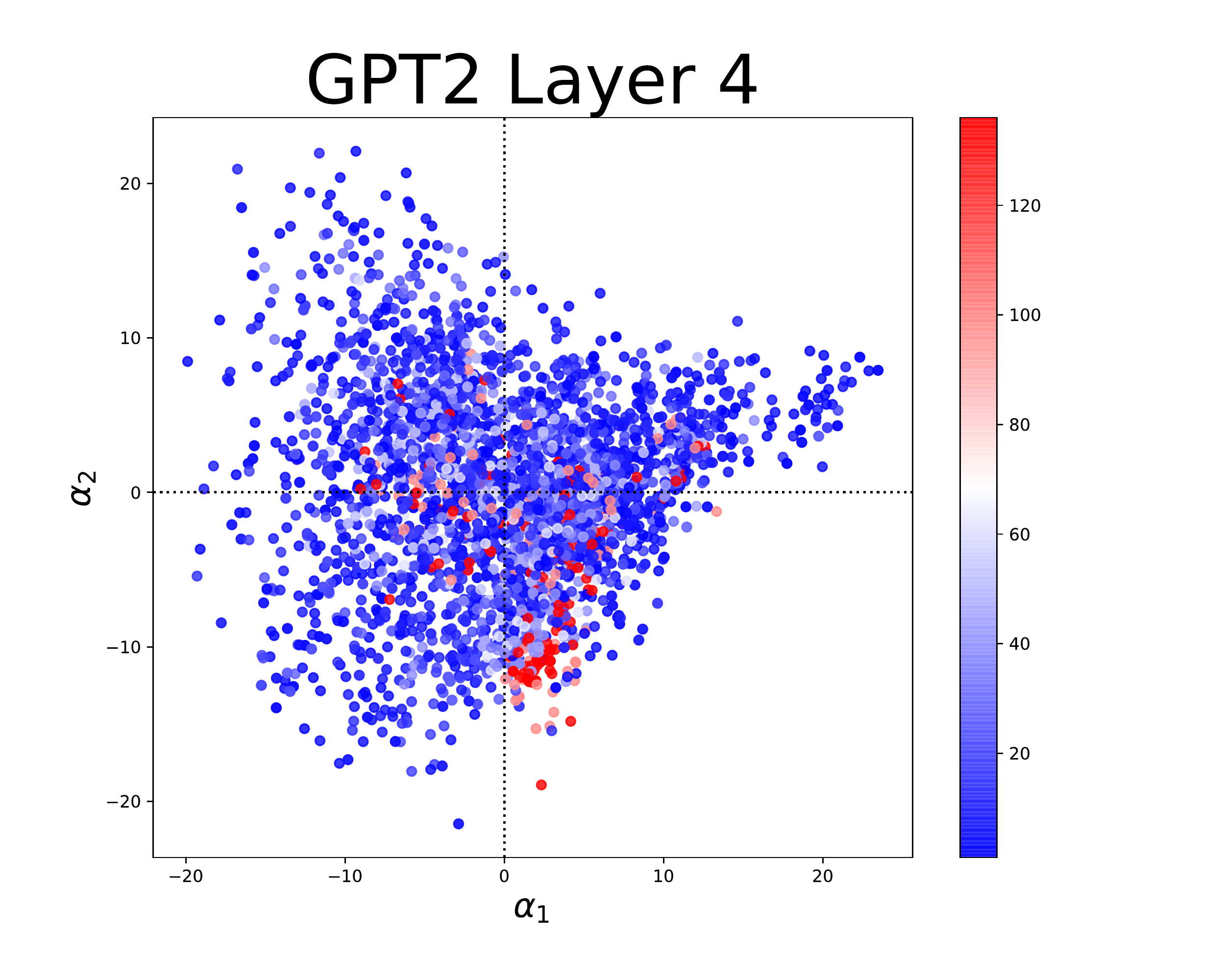}}
\subfloat[original]{\label{fig:k0}\includegraphics[width=0.2\linewidth]{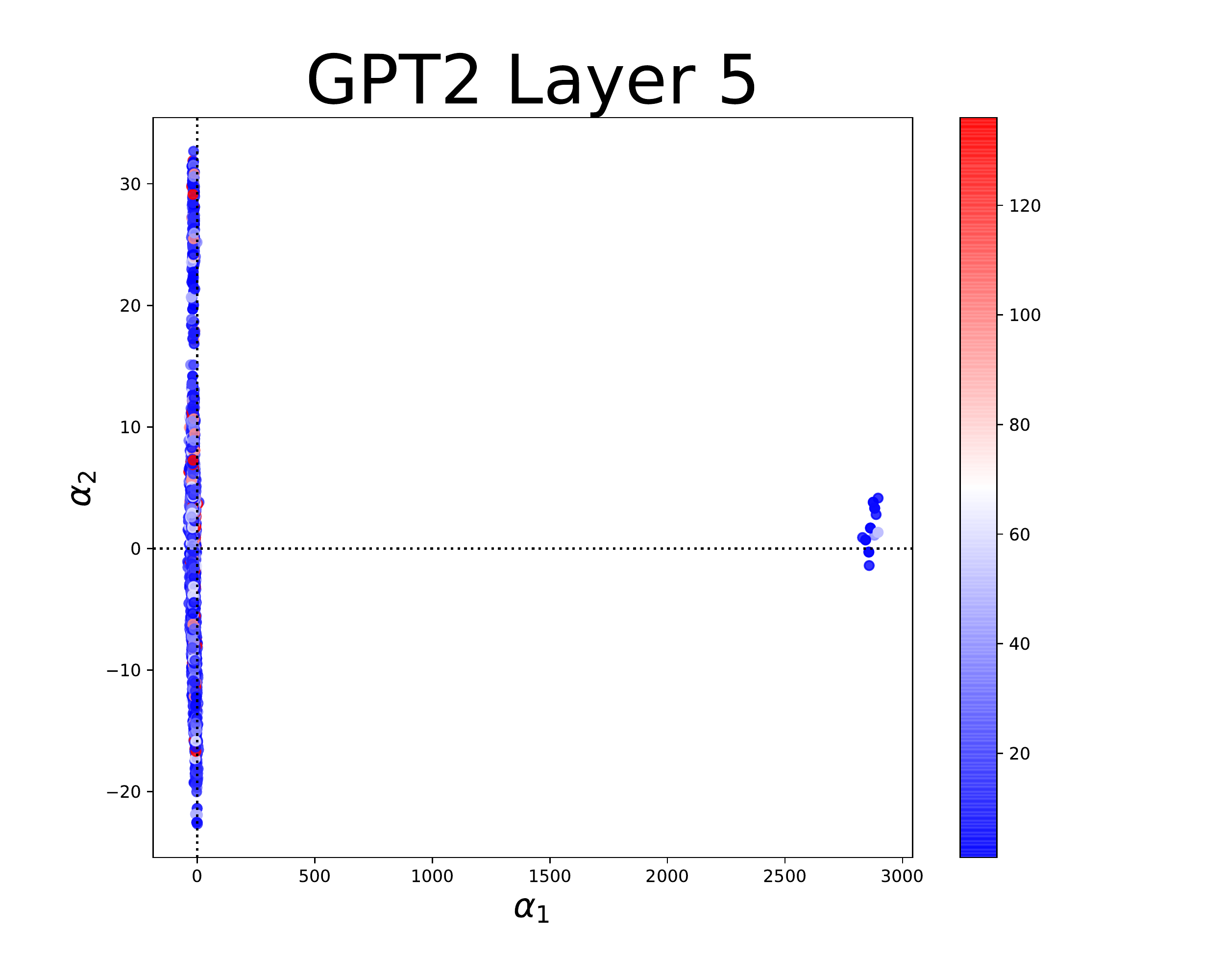}}
\subfloat[retrofitted]{\label{fig:l0}\includegraphics[width=0.2\linewidth]{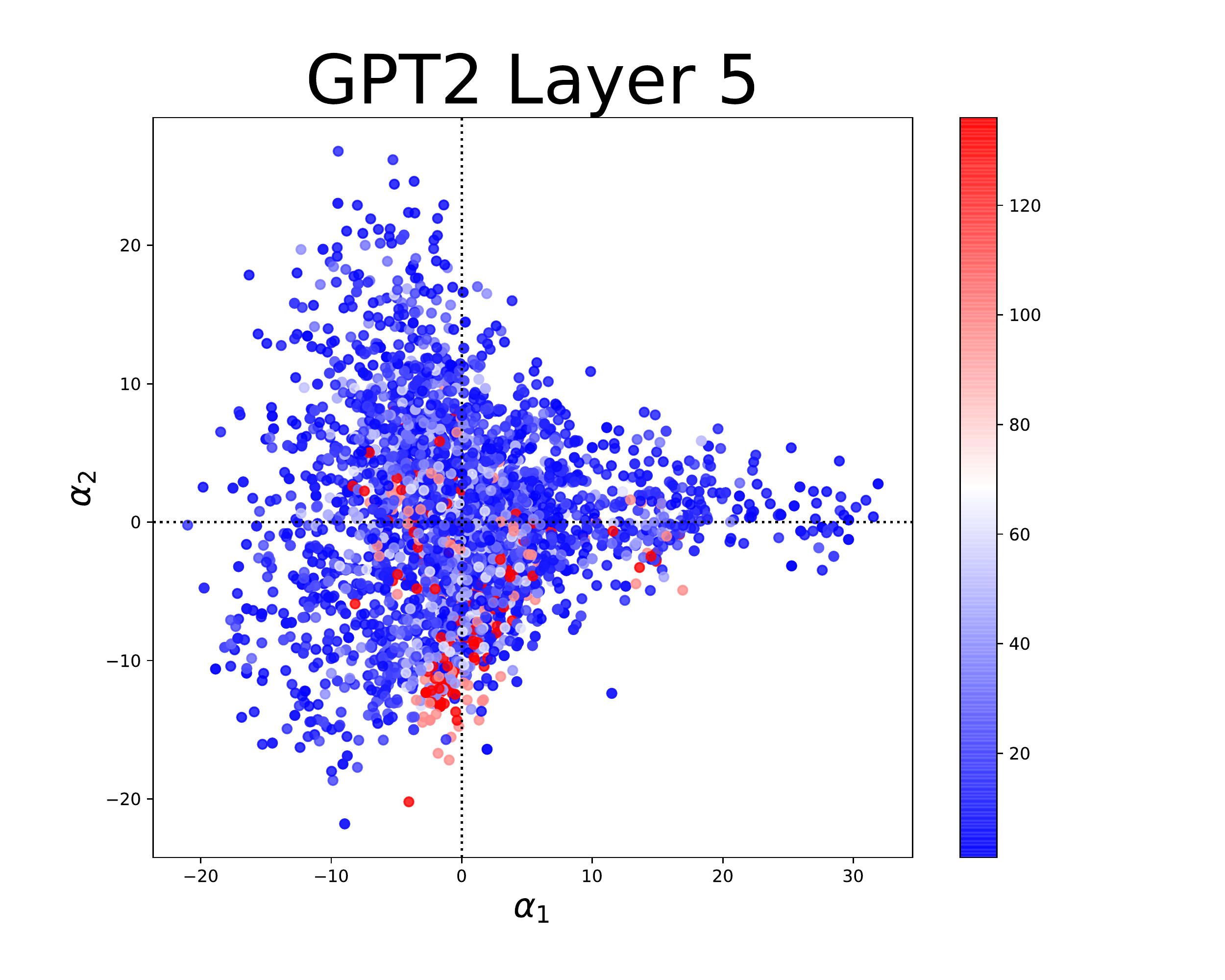}}\\
\subfloat[original]{\label{fig:m0}\includegraphics[width=0.2\linewidth]{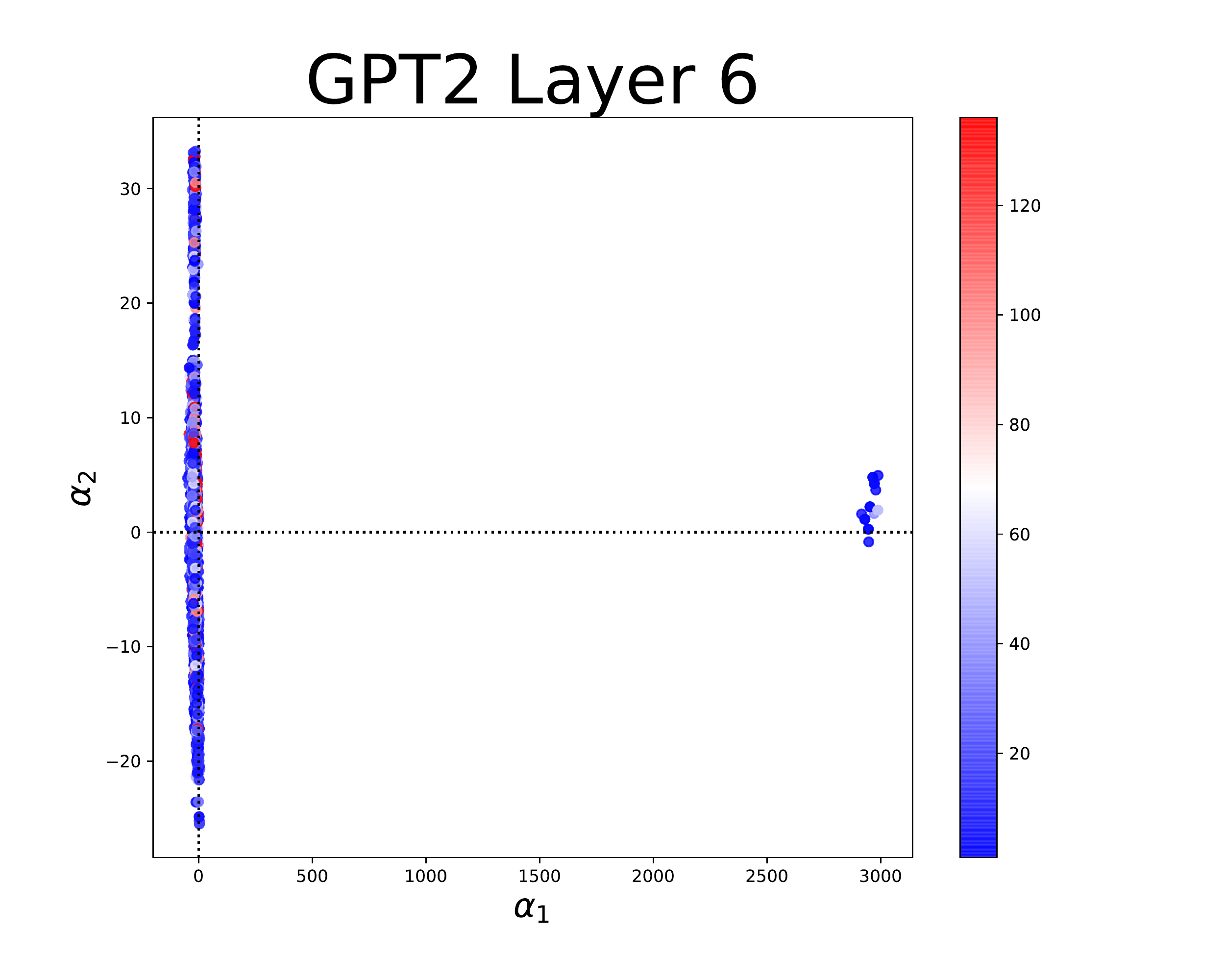}}
\subfloat[retrofitted]{\label{fig:n0}\includegraphics[width=0.2\linewidth]{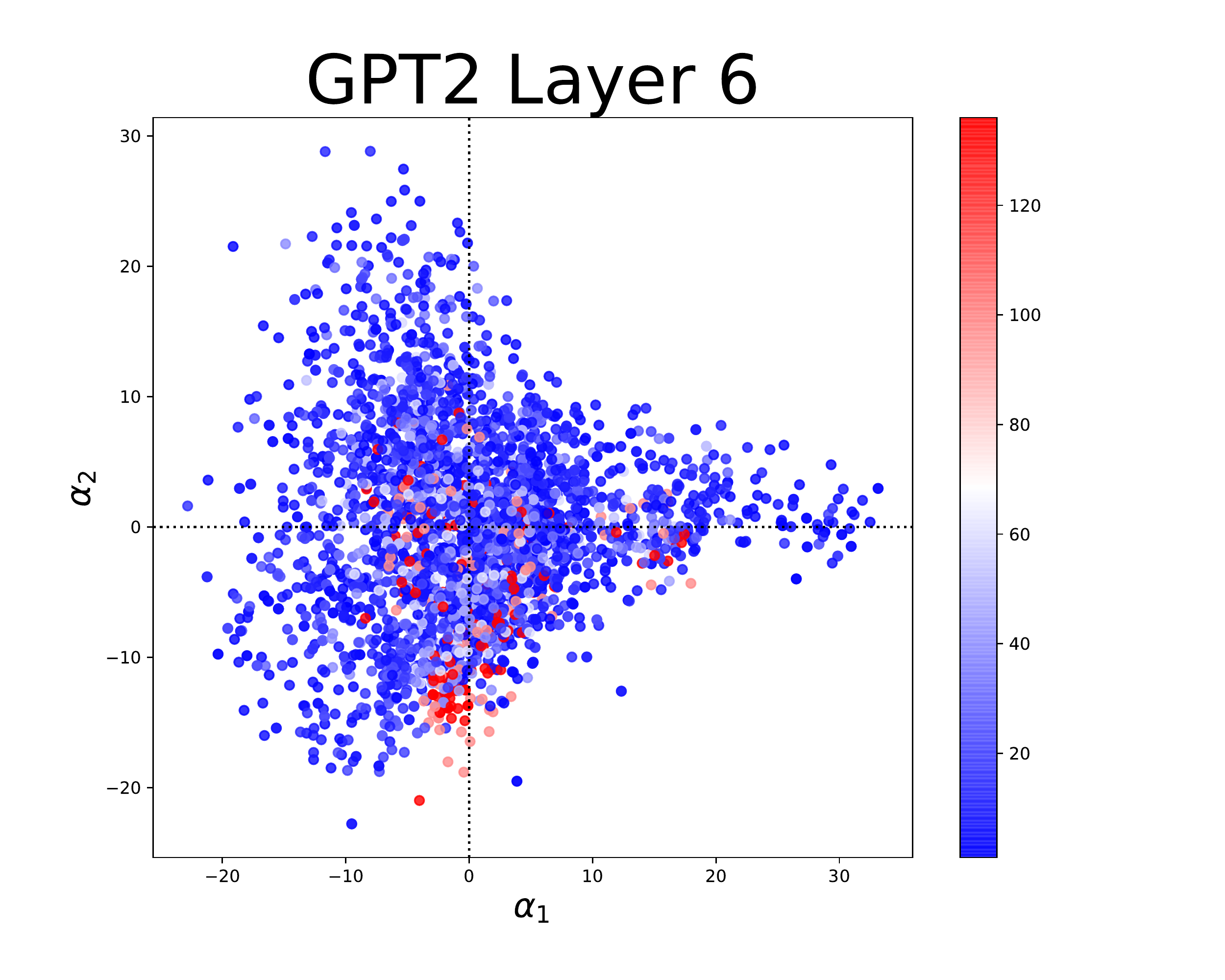}}
\subfloat[original]{\label{fig:o0}\includegraphics[width=0.2\linewidth]{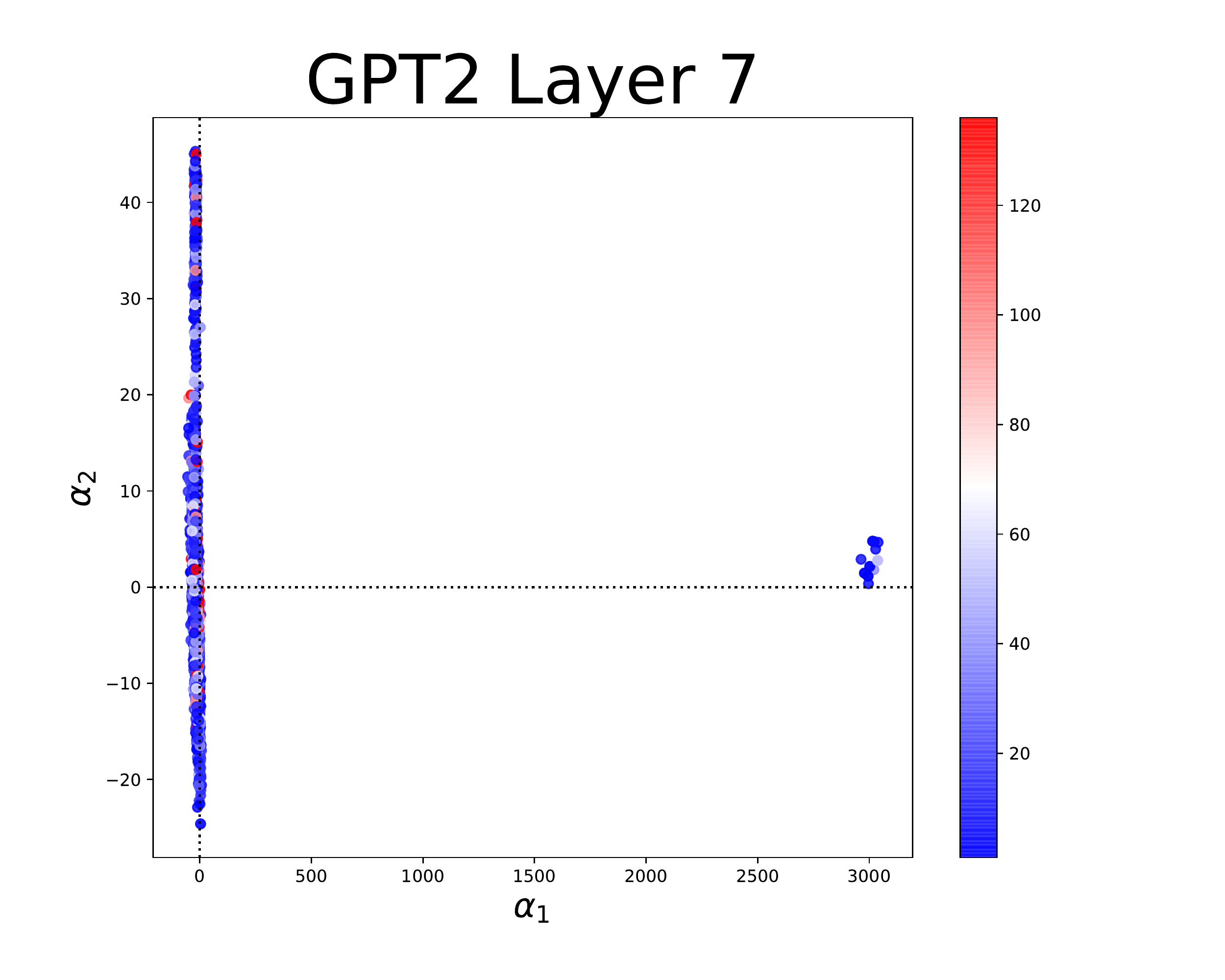}}
\subfloat[retrofitted]{\label{fig:p0}\includegraphics[width=0.2\linewidth]{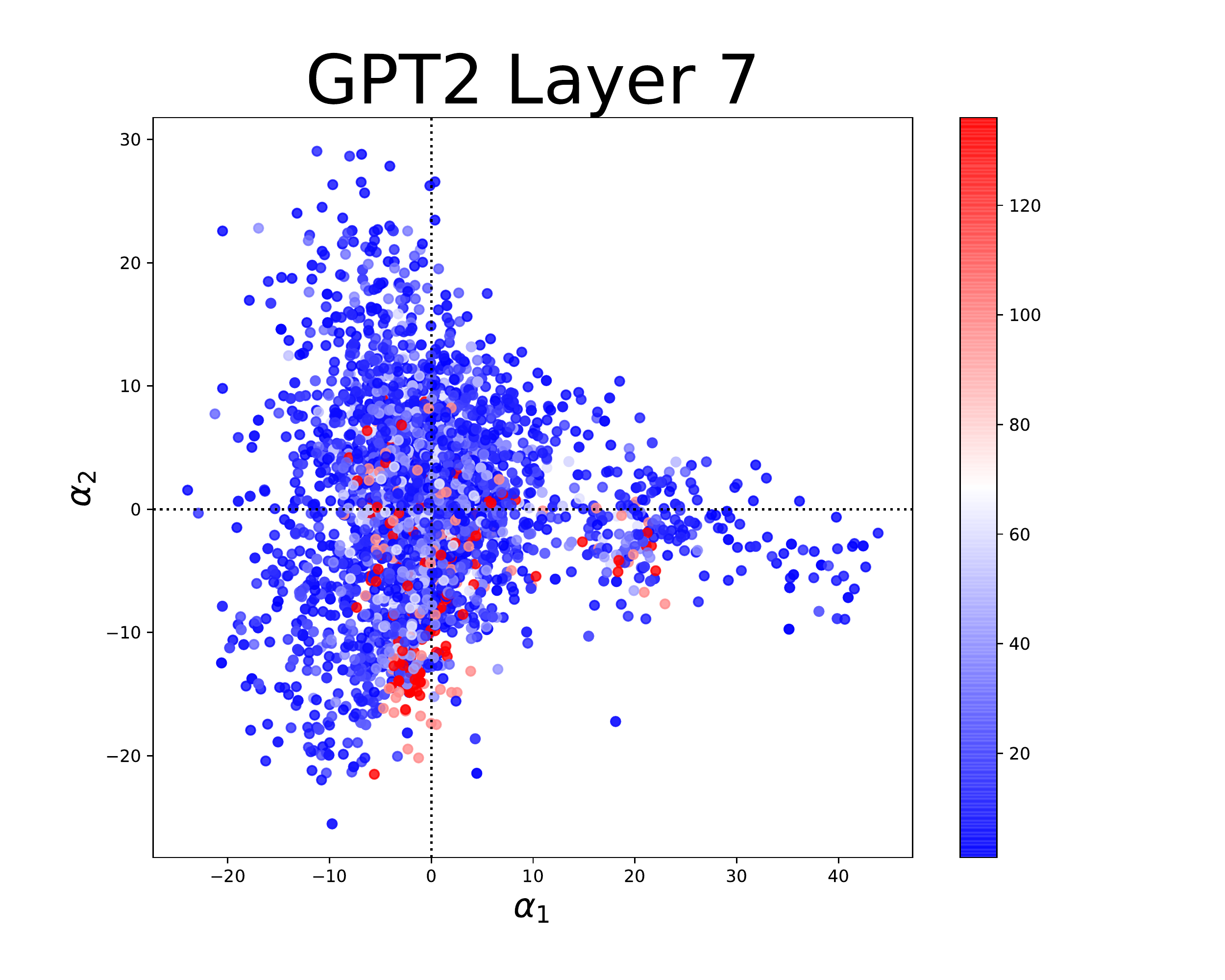}}\\
\subfloat[original]{\label{fig:q0}\includegraphics[width=0.2\linewidth]{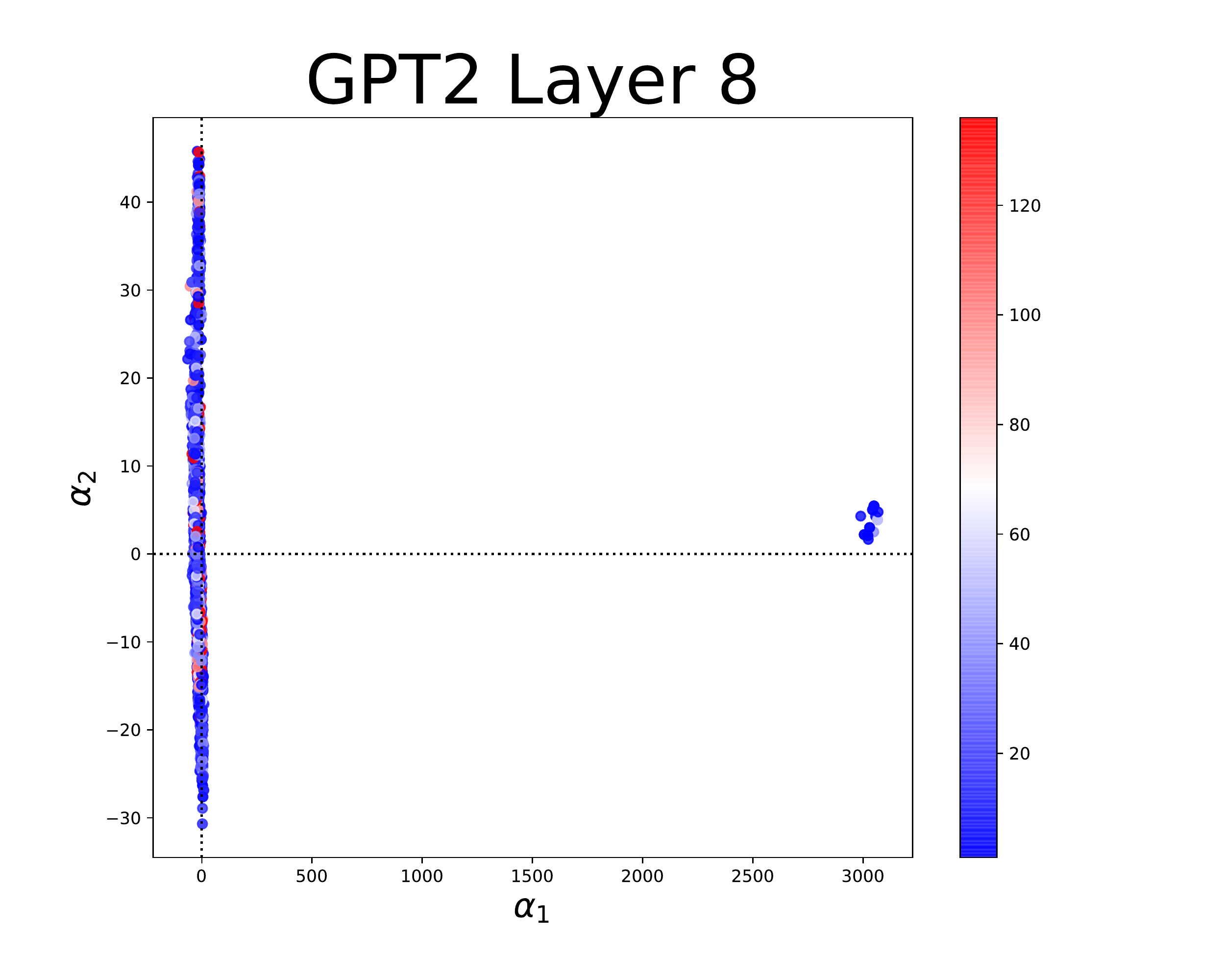}}
\subfloat[retrofitted]{\label{fig:r0}\includegraphics[width=0.2\linewidth]{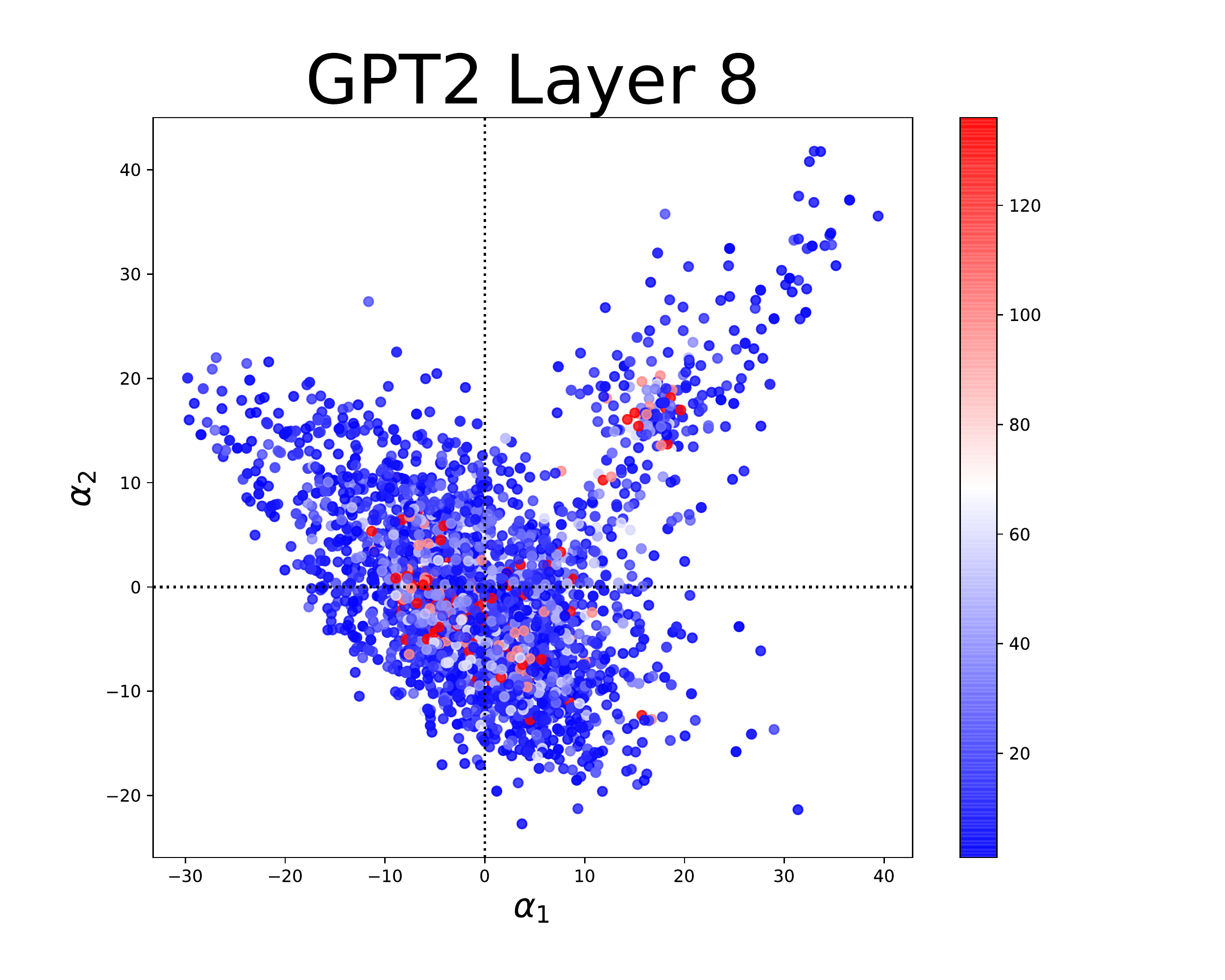}}
\subfloat[original]{\label{fig:s0}\includegraphics[width=0.2\linewidth]{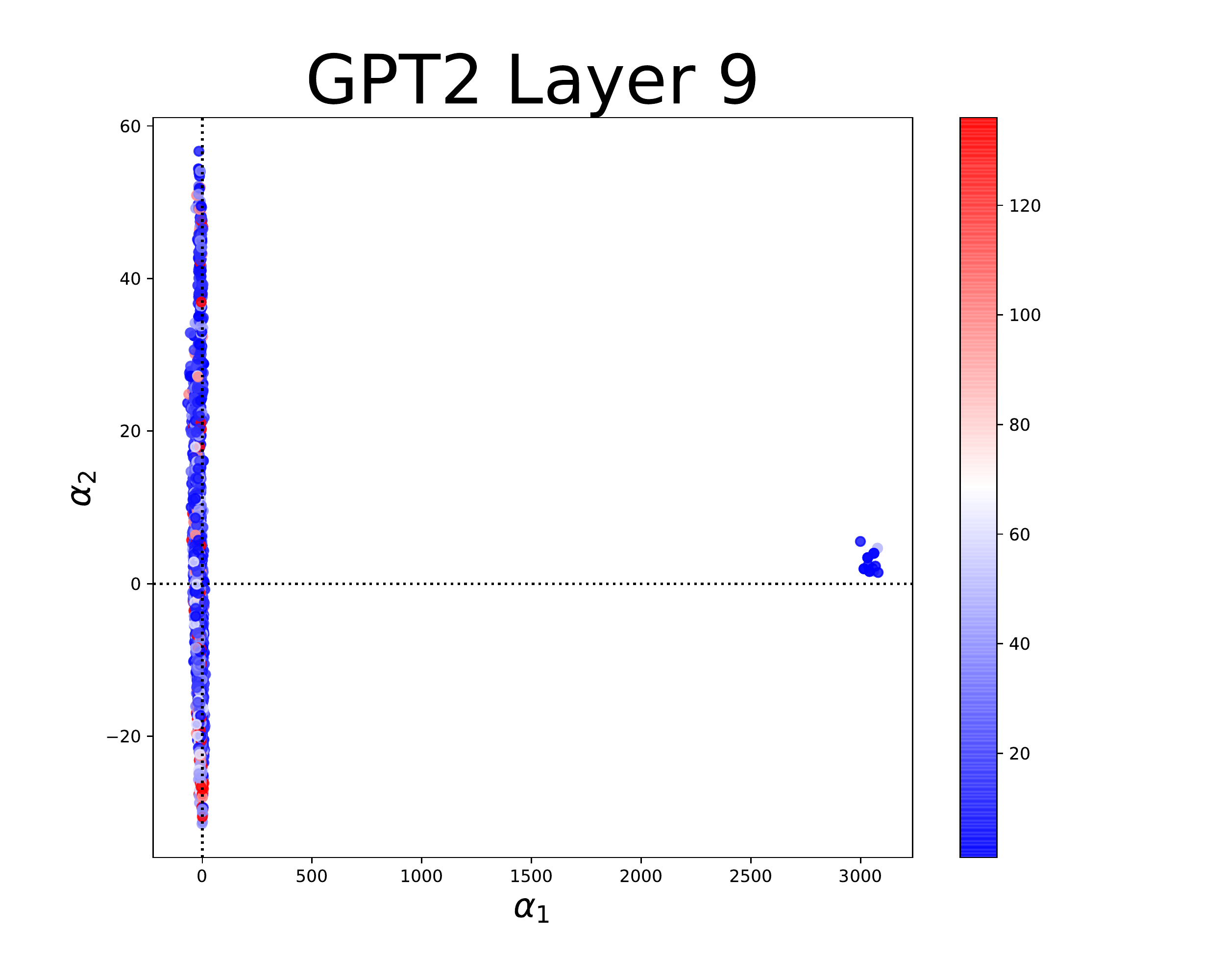}}
\subfloat[retrofitted]{\label{fig:t0}\includegraphics[width=0.2\linewidth]{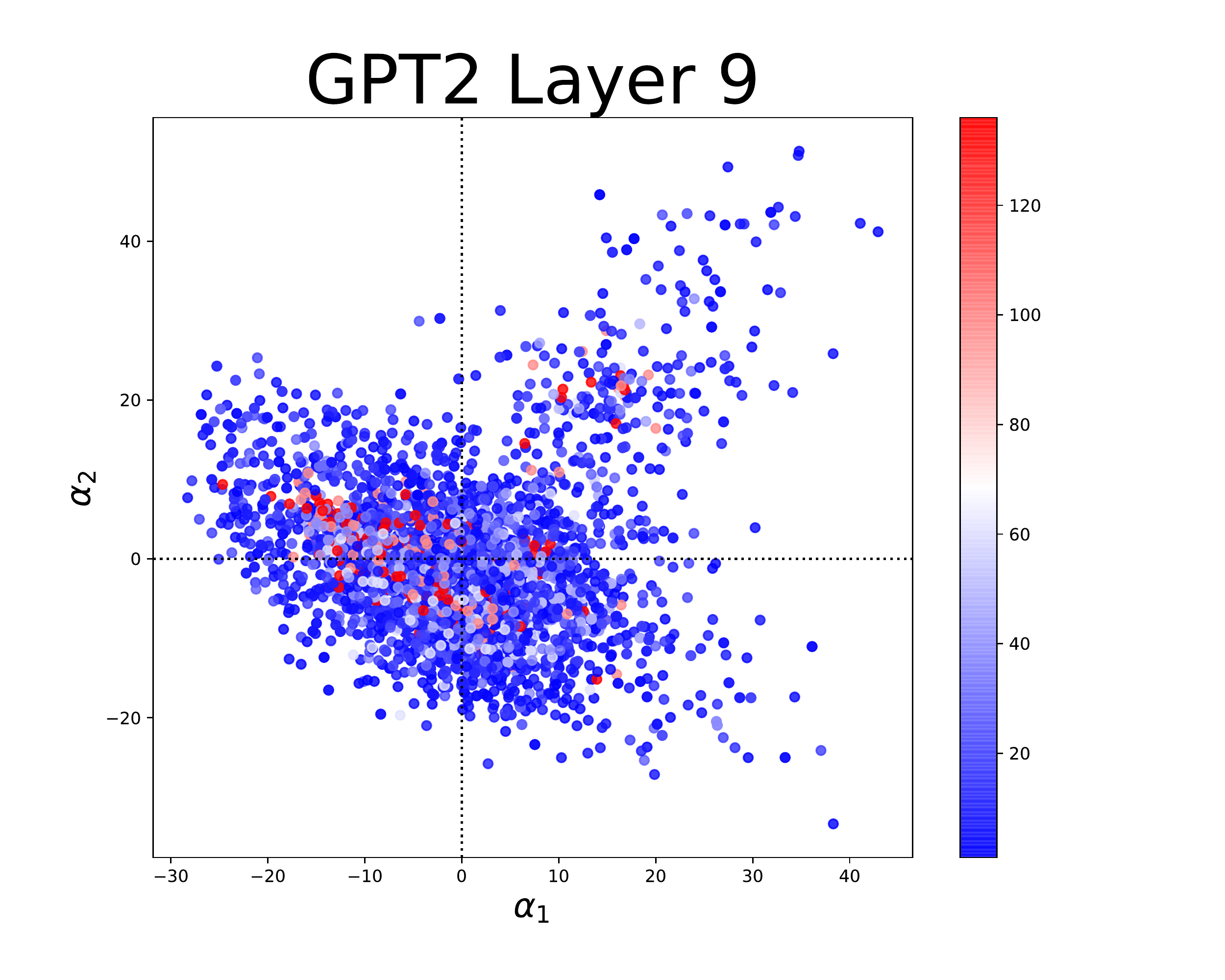}}\\
\subfloat[original]{\label{fig:u0}\includegraphics[width=0.2\linewidth]{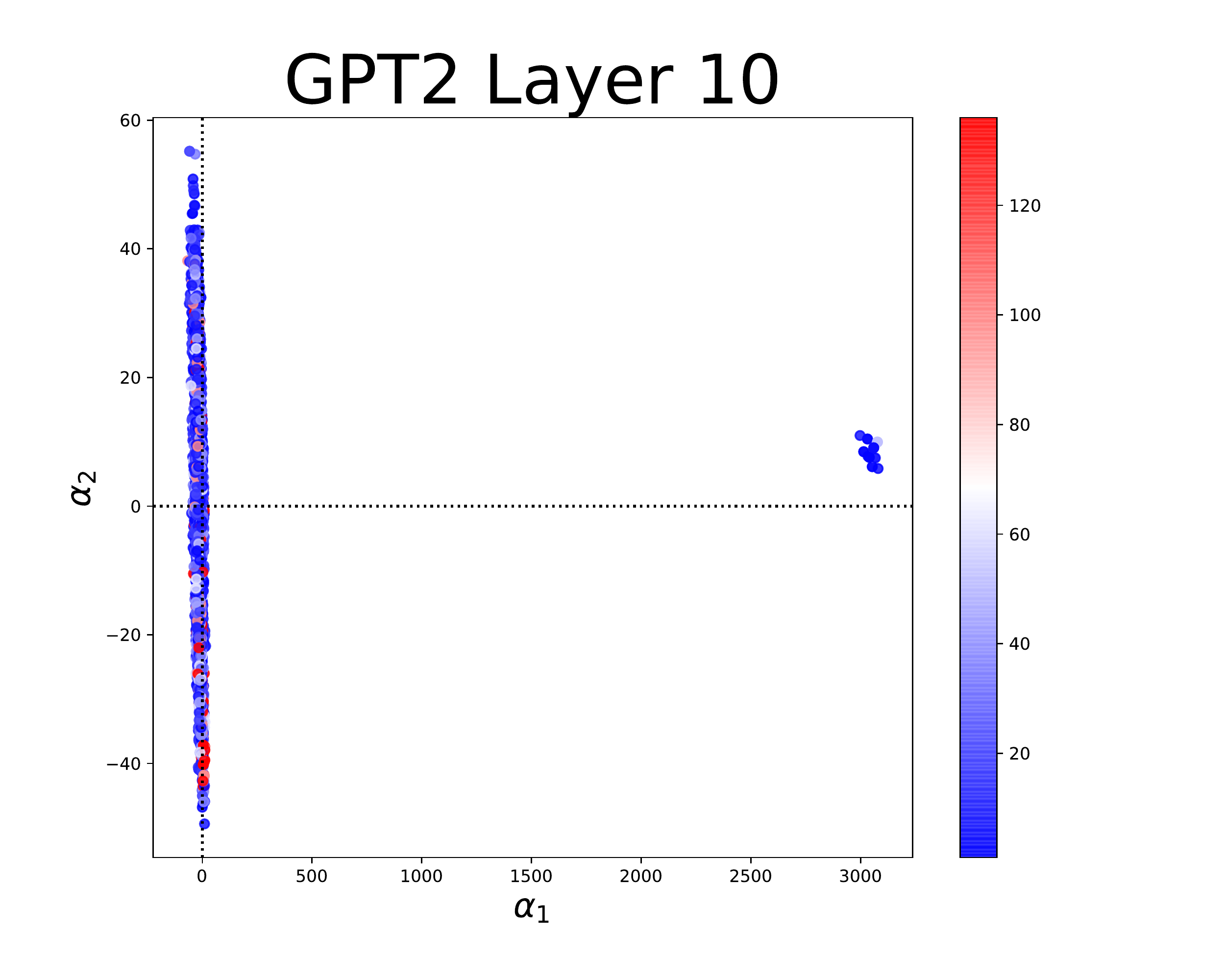}}
\subfloat[retrofitted]{\label{fig:v0}\includegraphics[width=0.2\linewidth]{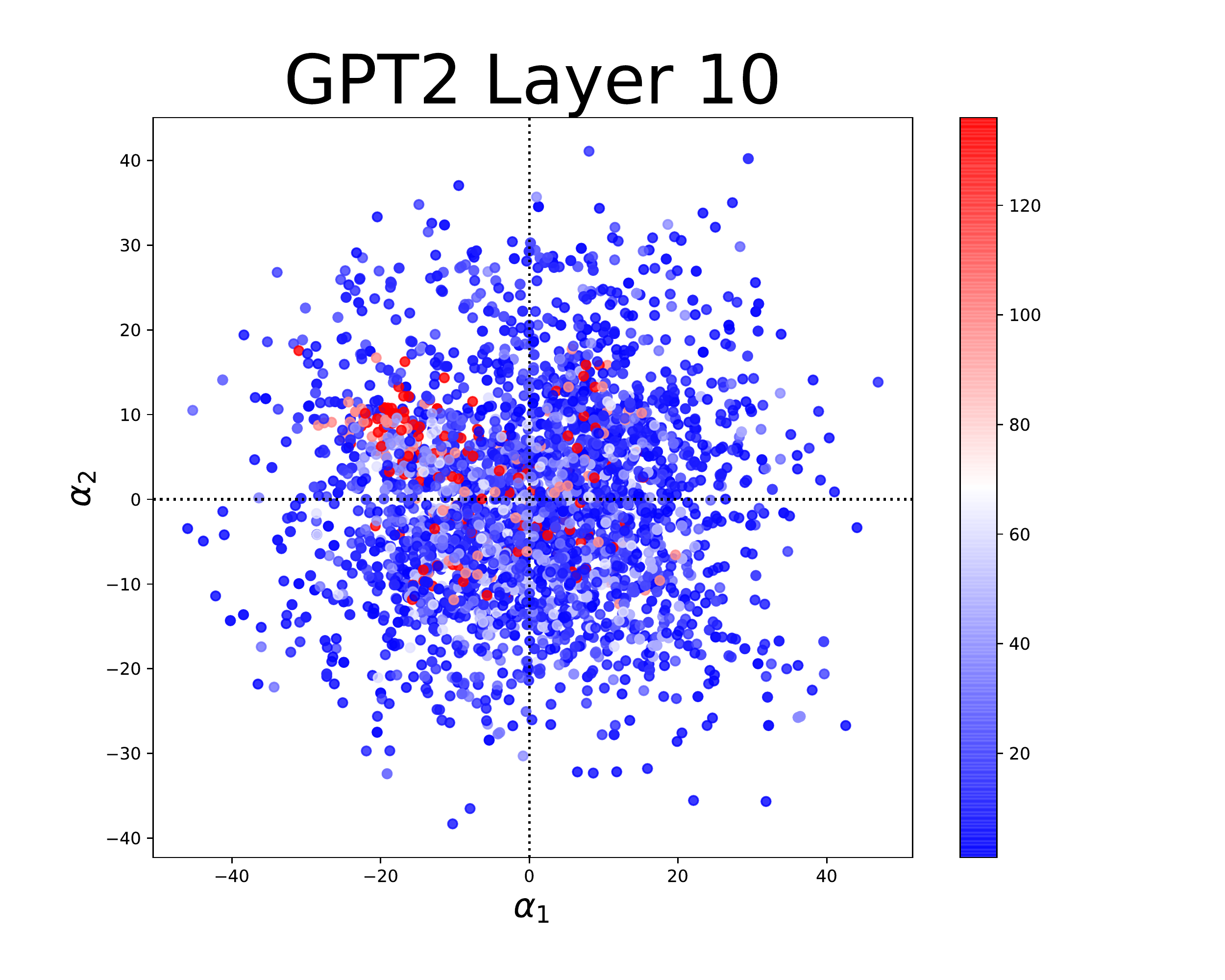}}
\subfloat[original]{\label{fig:w0}\includegraphics[width=0.2\linewidth]{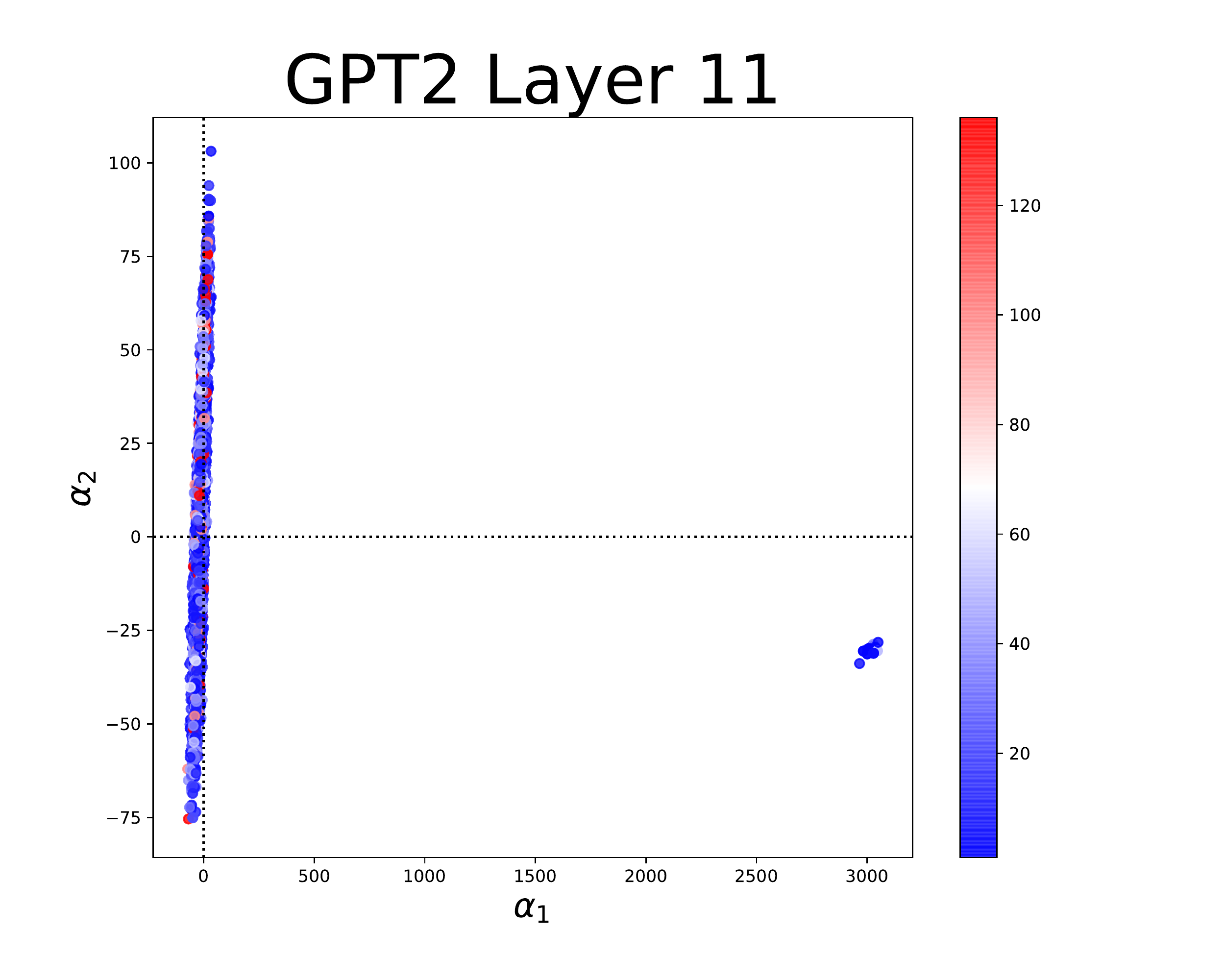}}
\subfloat[retrofitted]{\label{fig:x0}\includegraphics[width=0.2\linewidth]{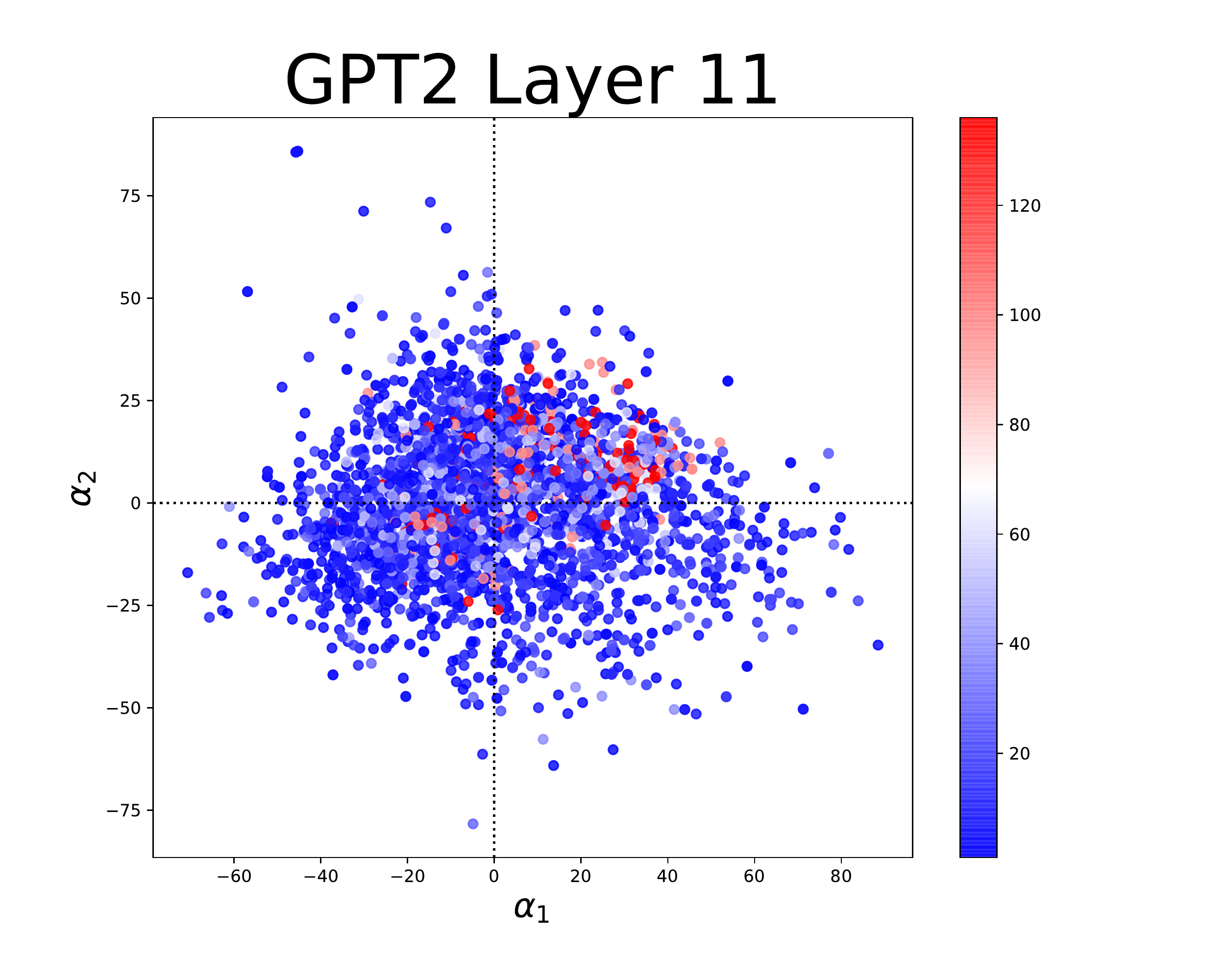}}\\
\subfloat[original]{\label{fig:y0}\includegraphics[width=0.2\linewidth]{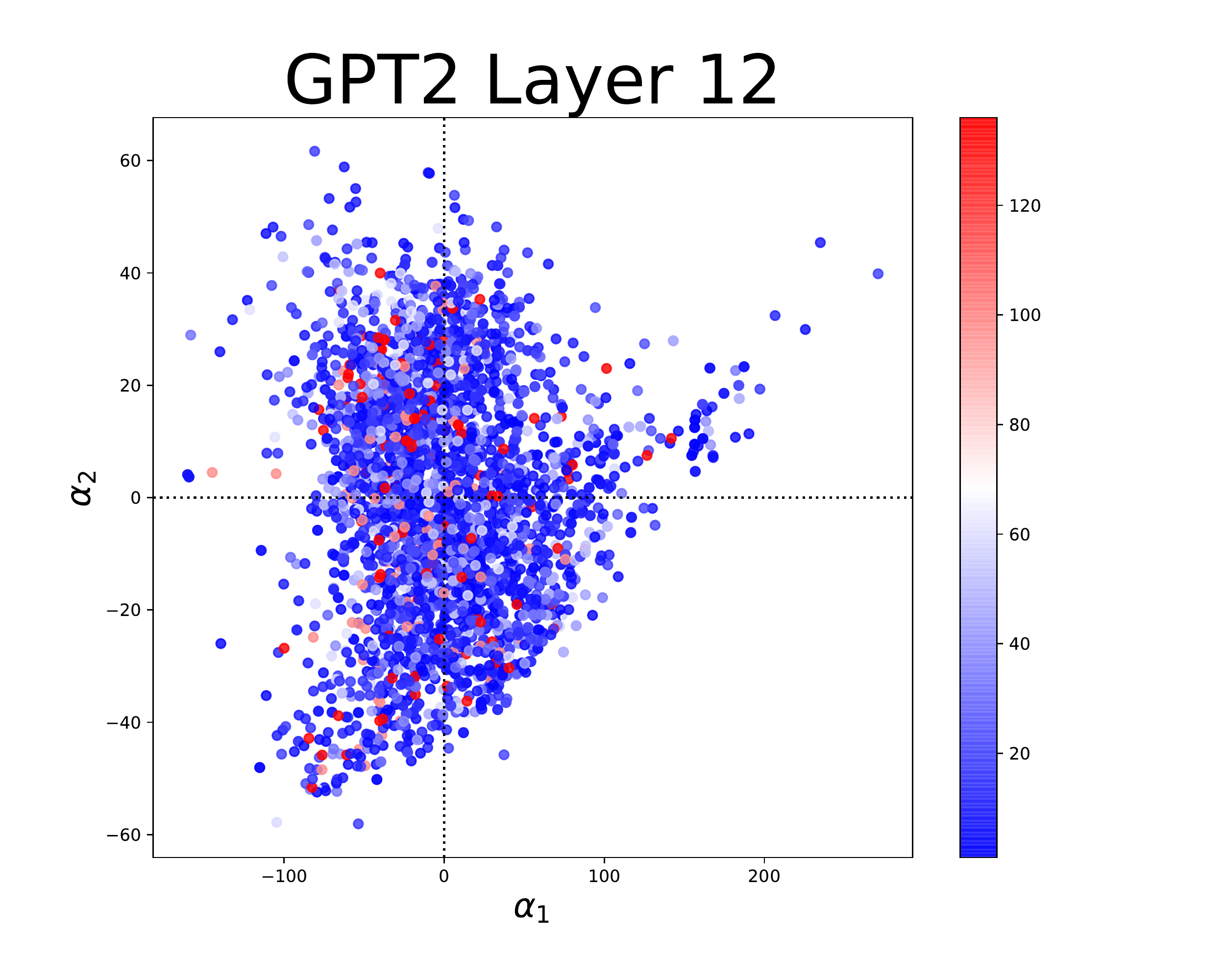}}
\subfloat[retrofitted]{\label{fig:z0}\includegraphics[width=0.2\linewidth]{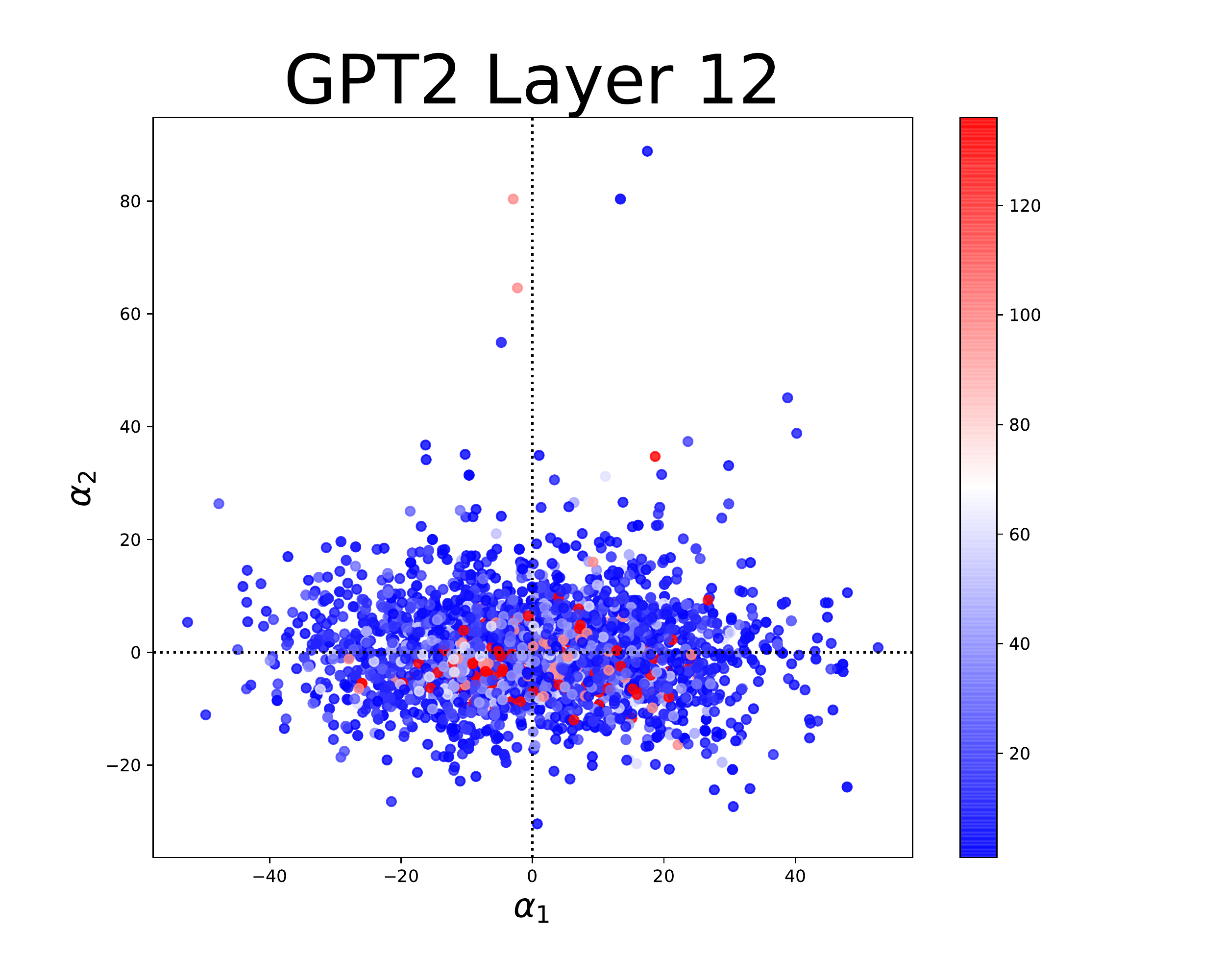}}
\caption{PCA Plots of GPT2 Word Representations.}
\label{fig:gpt2_fig}
\end{figure*}

\begin{figure*}
\centering
\subfloat[original]{\label{fig:a1}\includegraphics[width=0.2\linewidth]{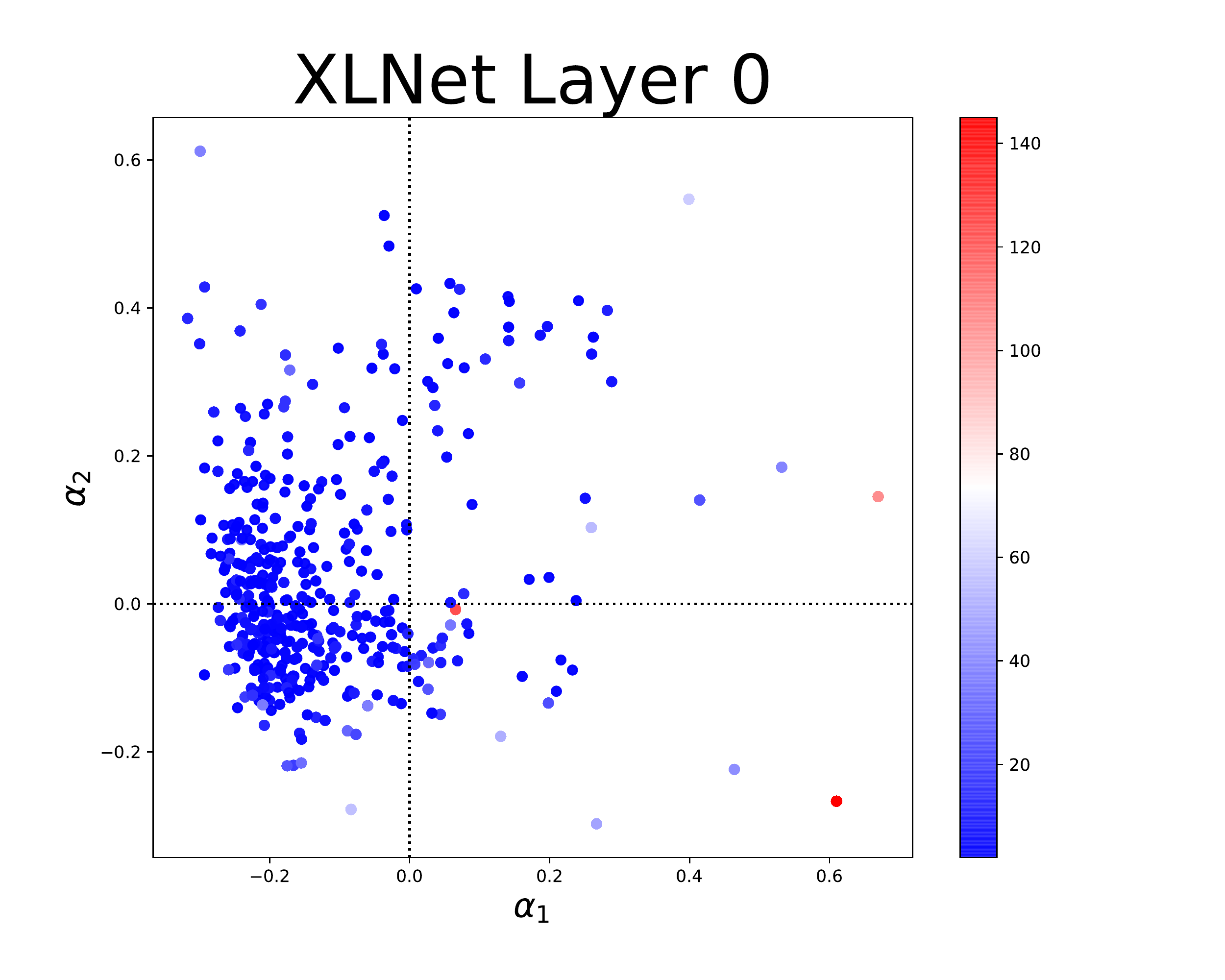}}
\subfloat[retrofitted]{\label{fig:b1}\includegraphics[width=0.2\linewidth]{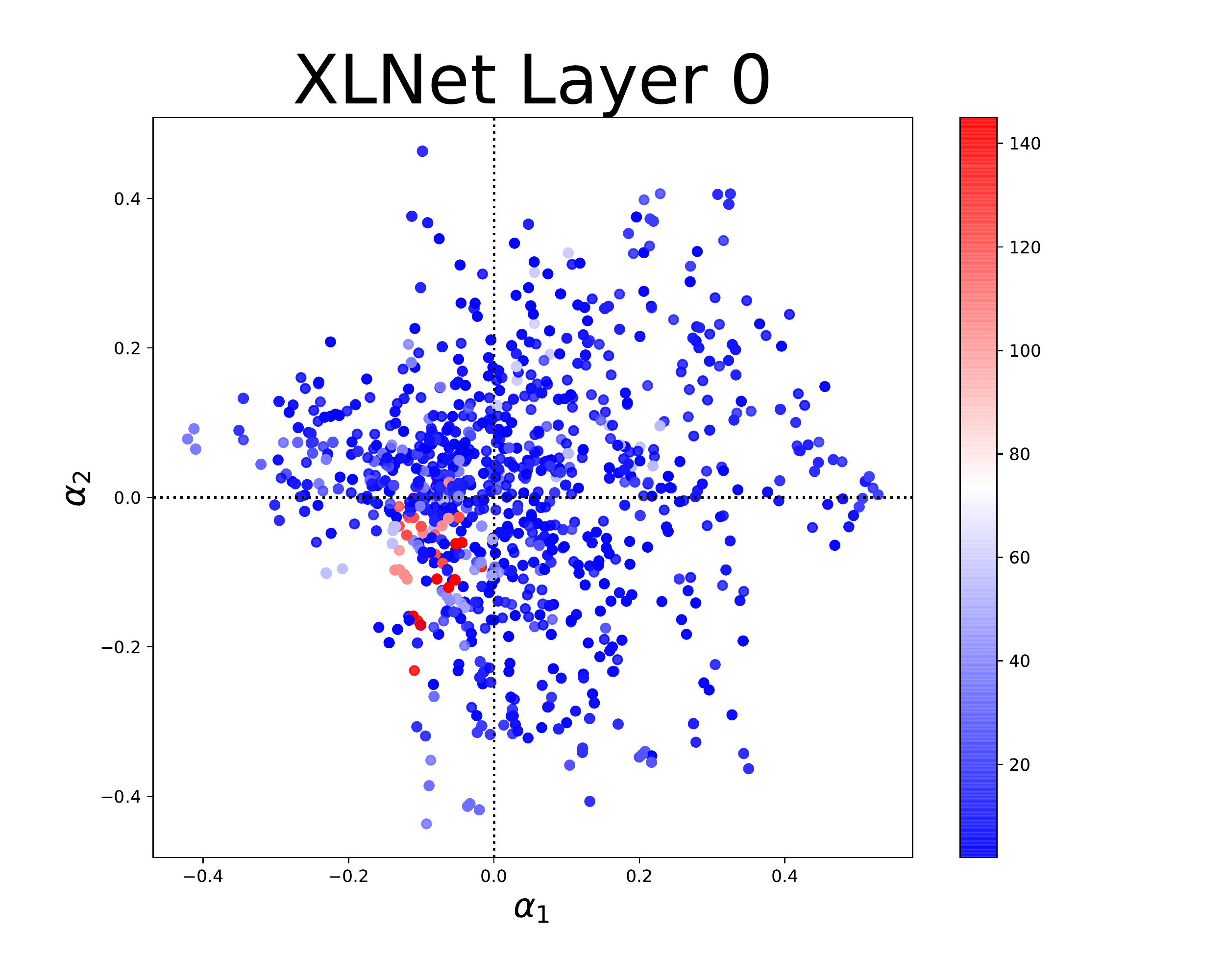}}
\subfloat[original]{\label{fig:c1}\includegraphics[width=0.2\linewidth]{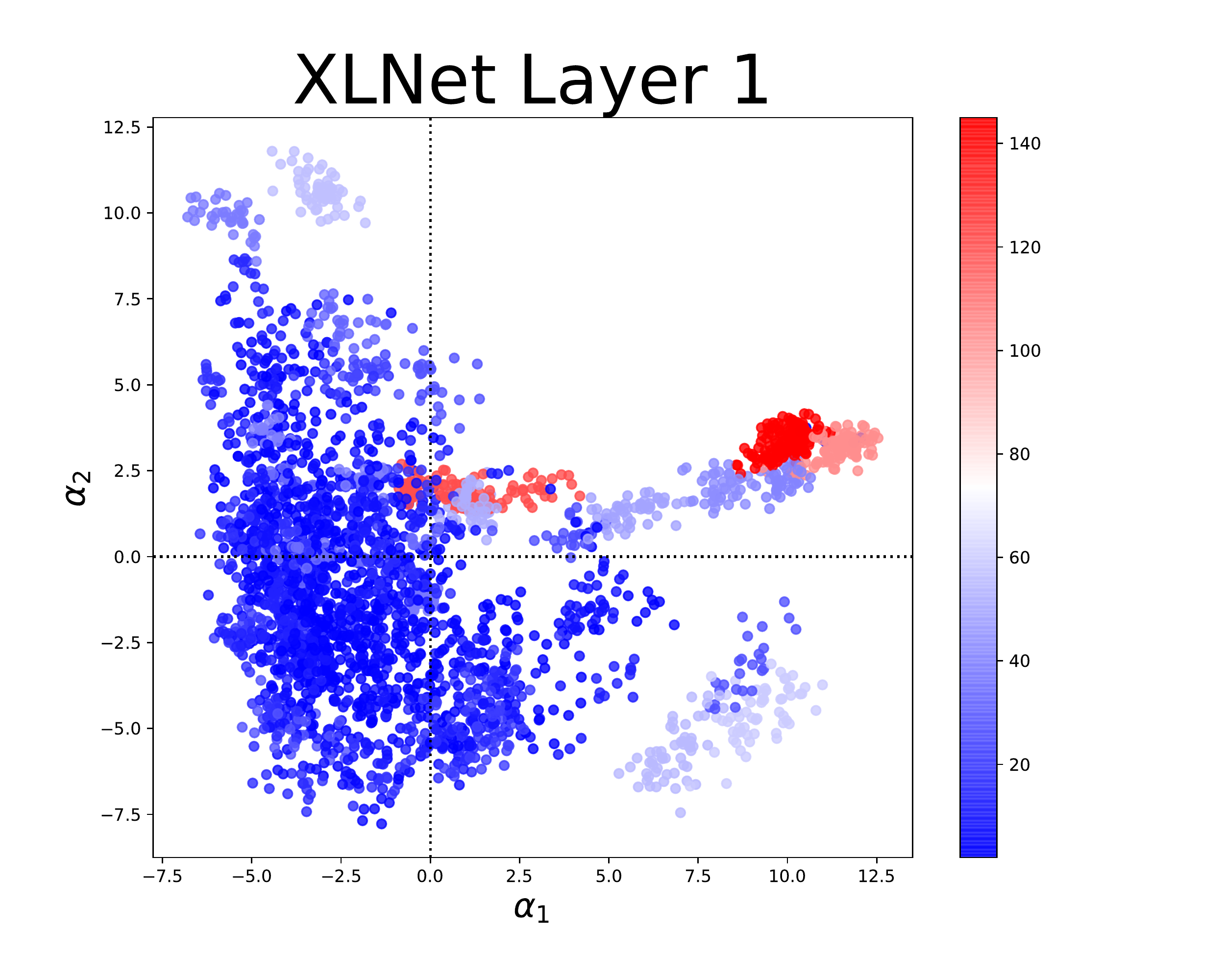}}
\subfloat[retrofitted]{\label{fig:d1}\includegraphics[width=0.2\linewidth]{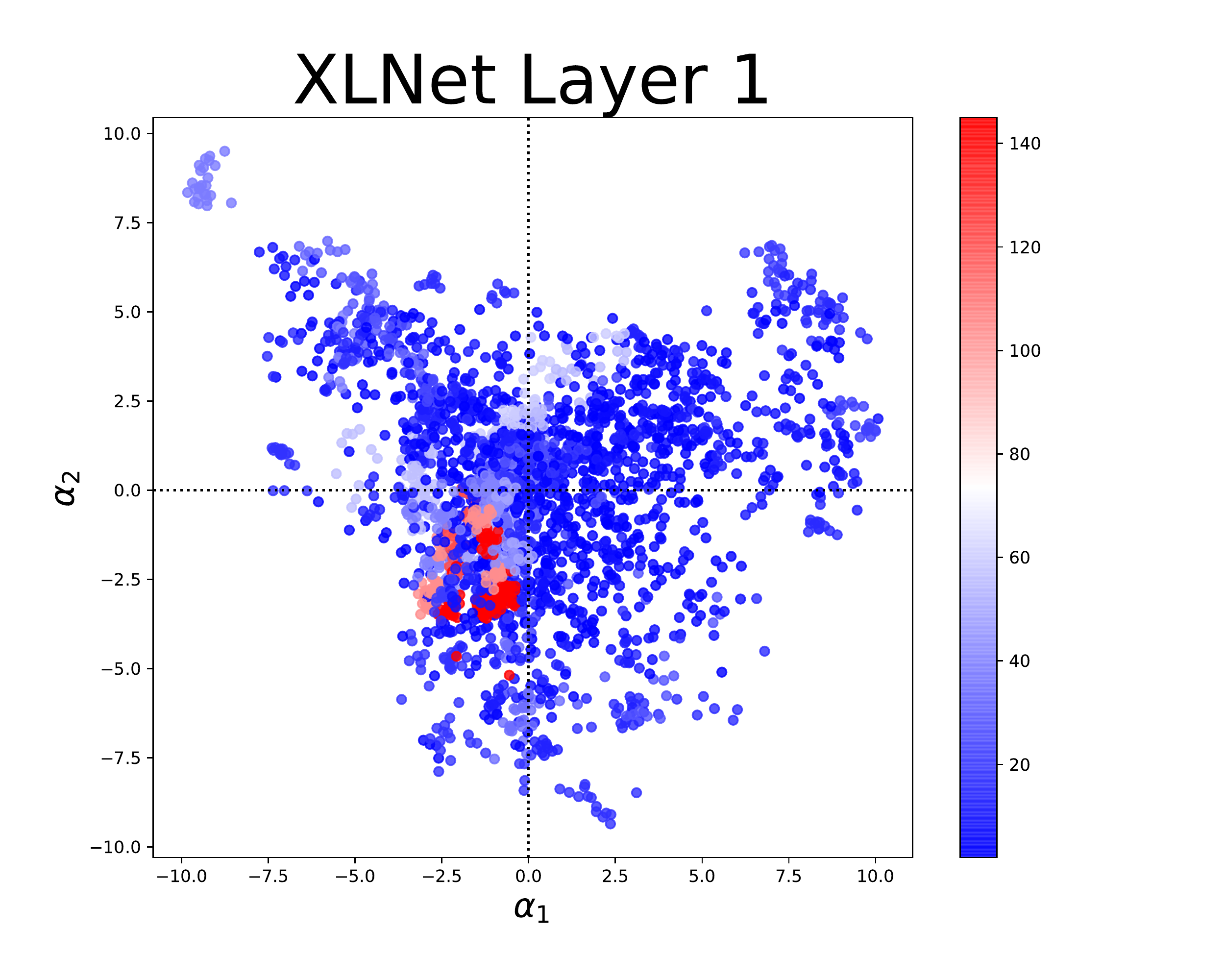}}\\
\subfloat[original]{\label{fig:e1}\includegraphics[width=0.2\linewidth]{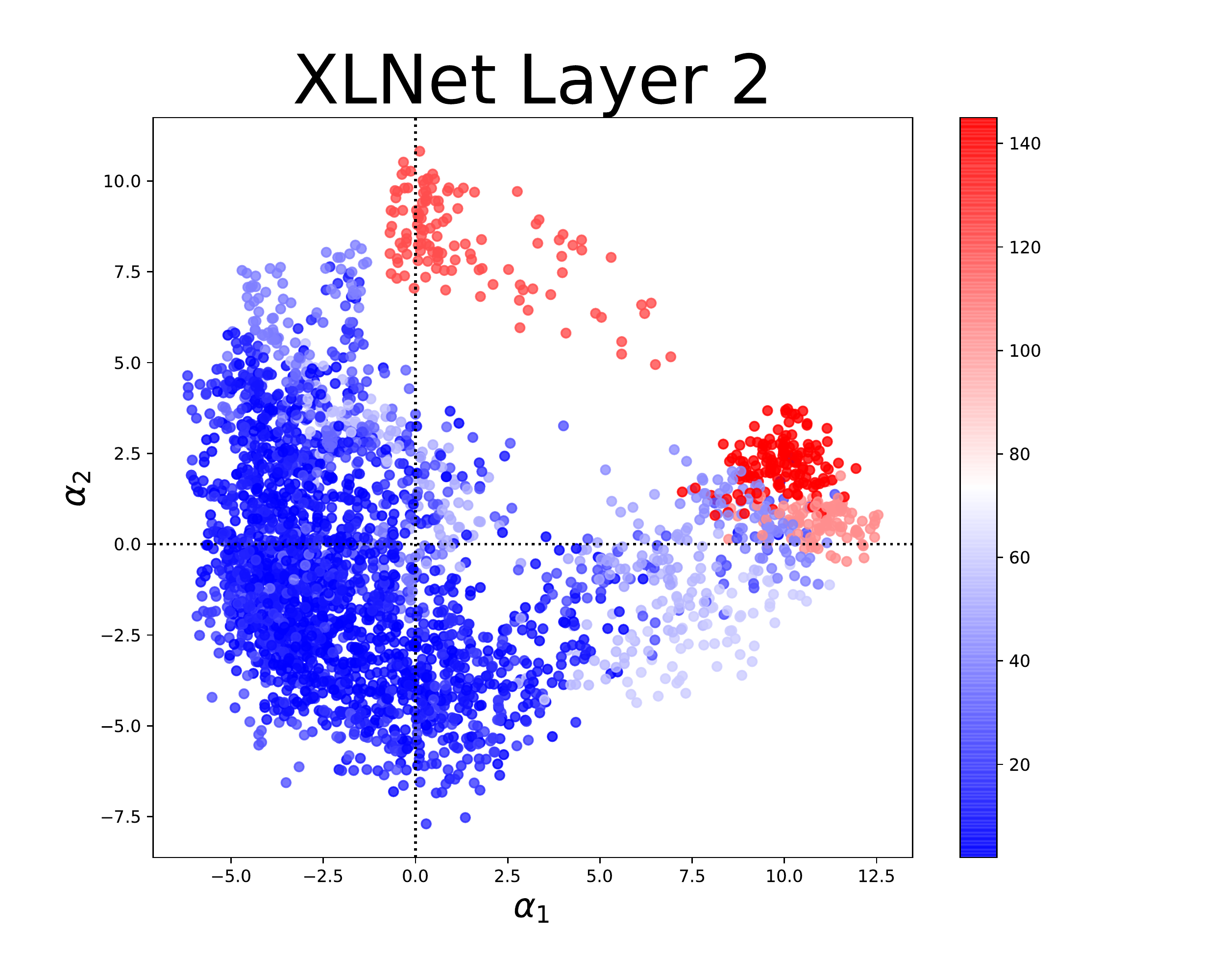}}
\subfloat[retrofitted]{\label{fig:f1}\includegraphics[width=0.2\linewidth]{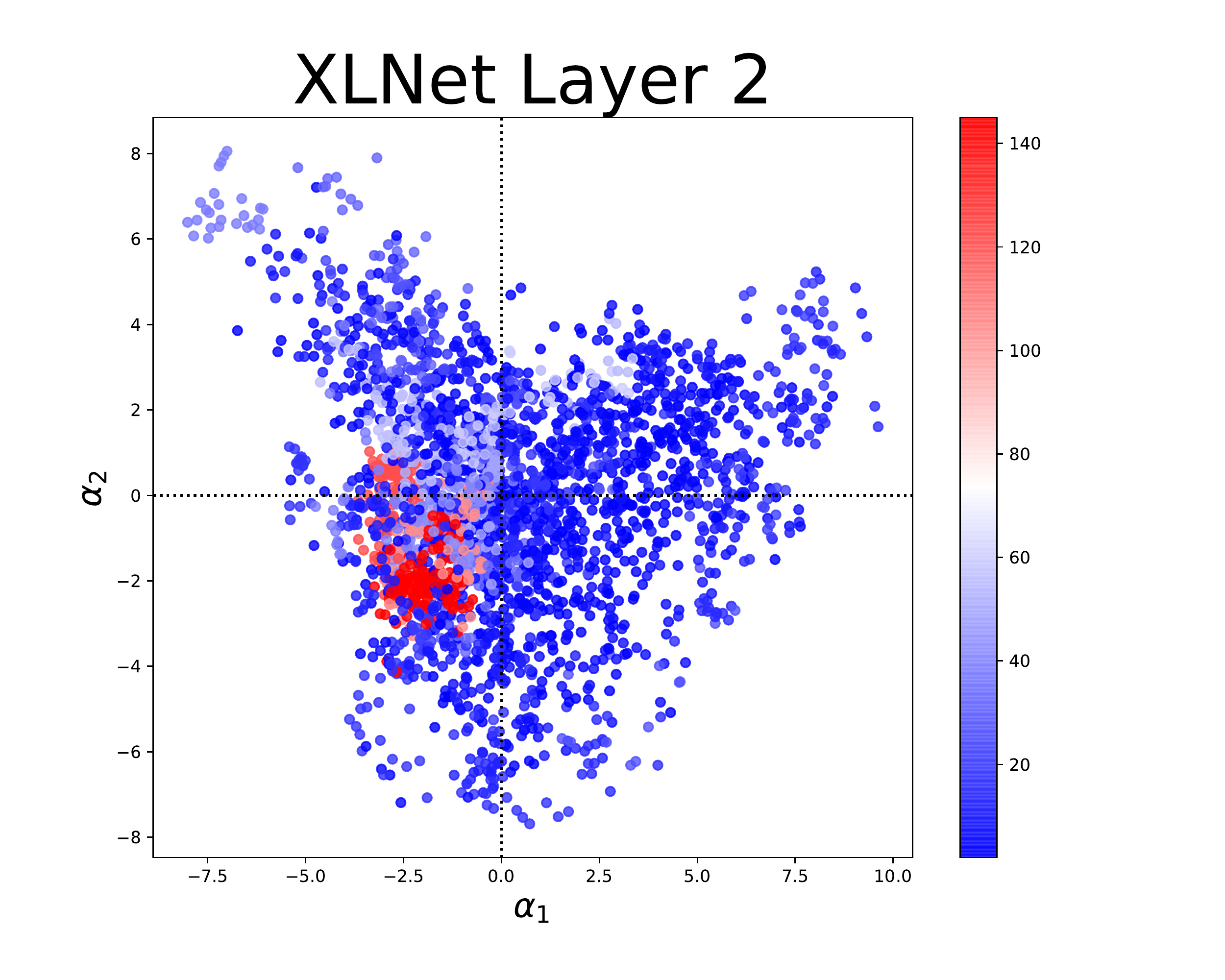}}
\subfloat[original]{\label{fig:g1}\includegraphics[width=0.2\linewidth]{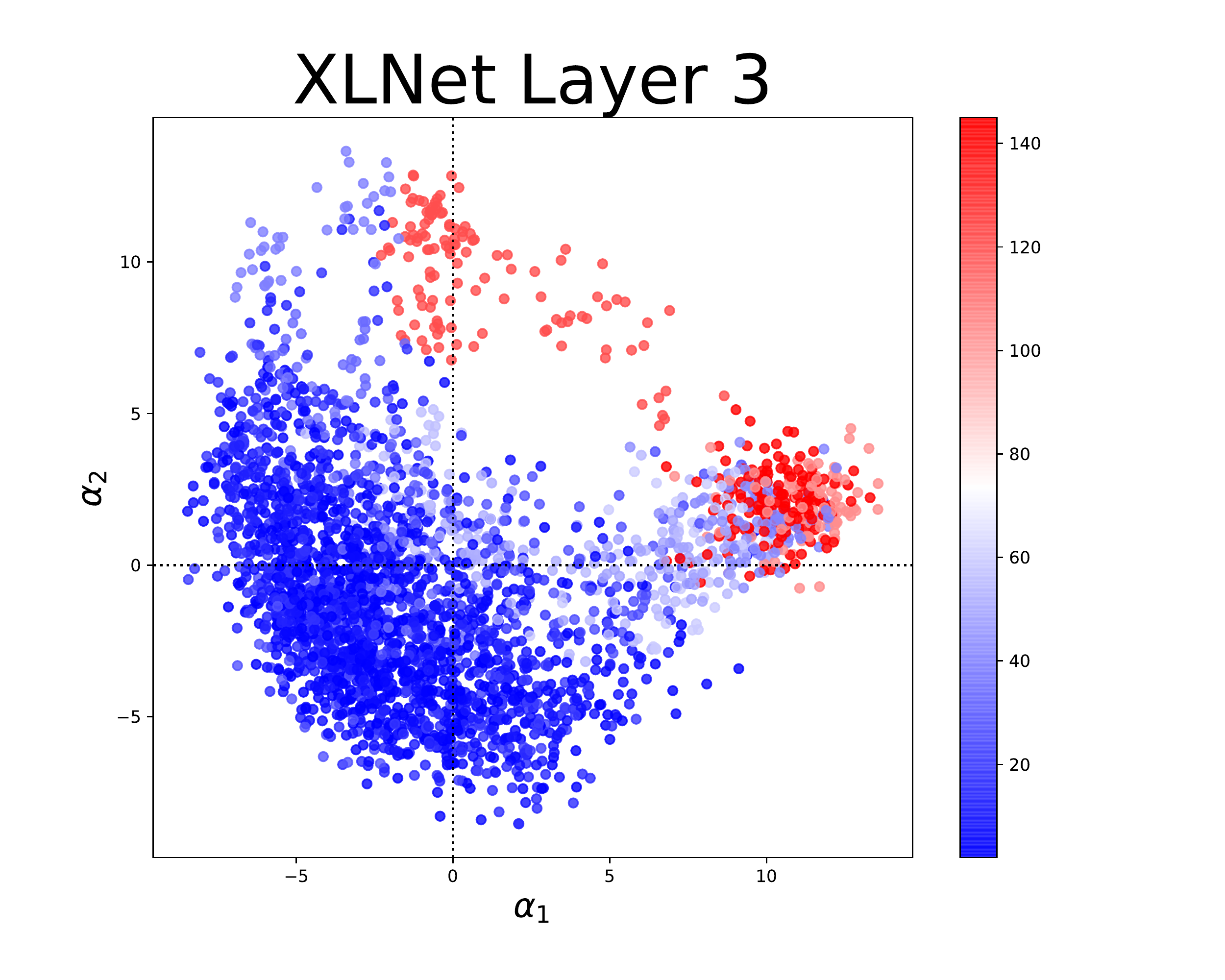}}
\subfloat[retrofitted]{\label{fig:h1}\includegraphics[width=0.2\linewidth]{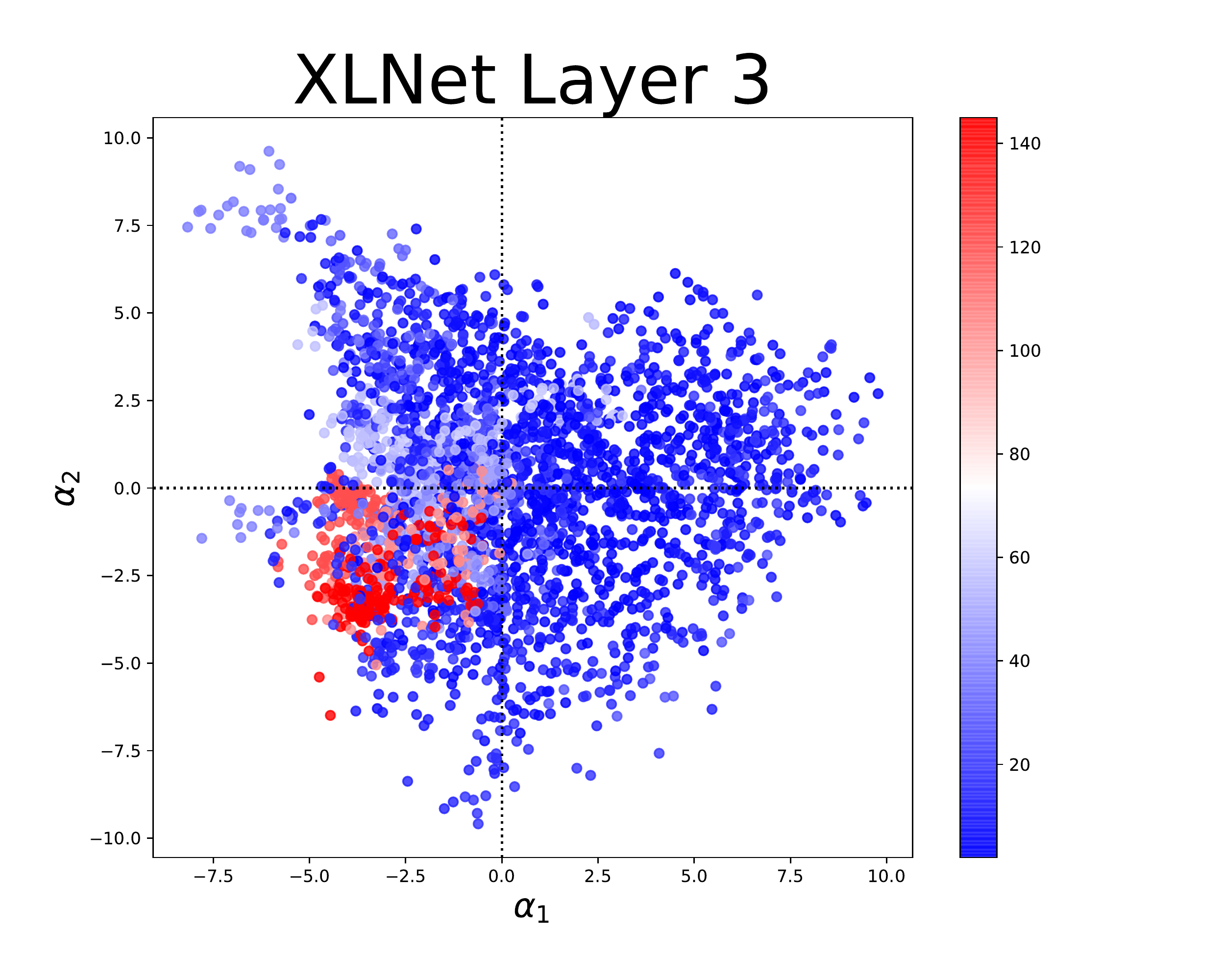}}\\
\subfloat[original]{\label{fig:i1}\includegraphics[width=0.2\linewidth]{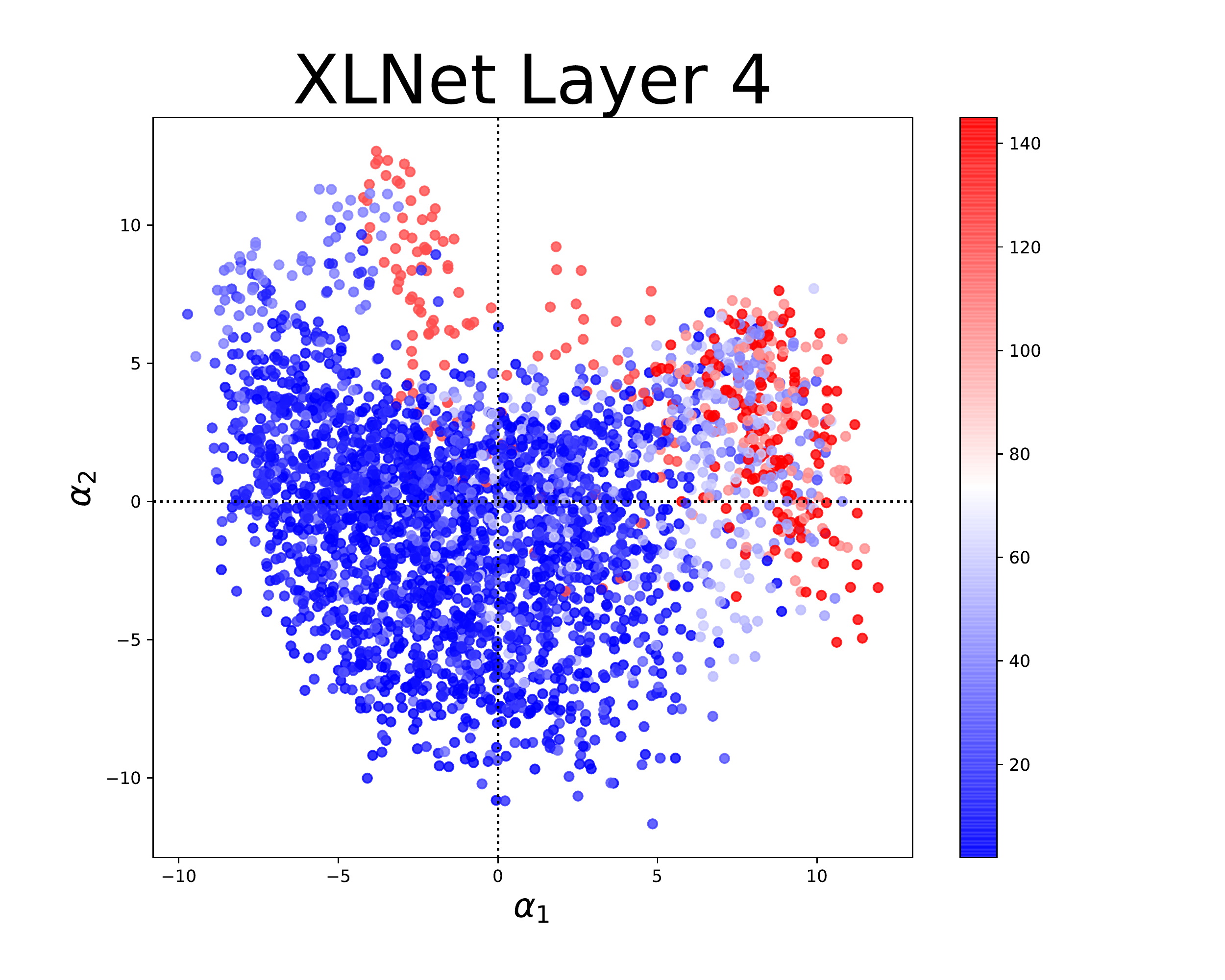}}
\subfloat[retrofitted]{\label{fig:j1}\includegraphics[width=0.2\linewidth]{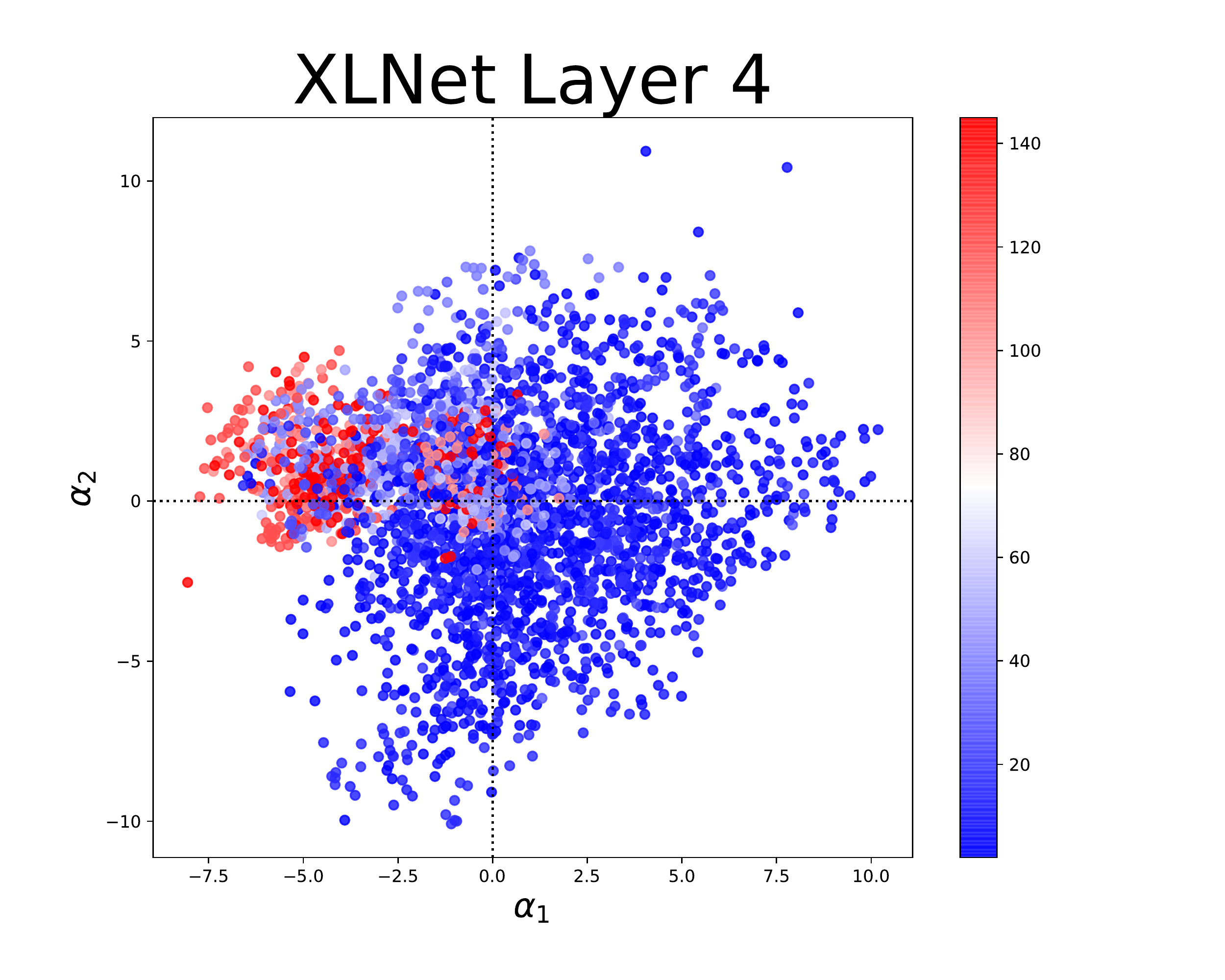}}
\subfloat[original]{\label{fig:k1}\includegraphics[width=0.2\linewidth]{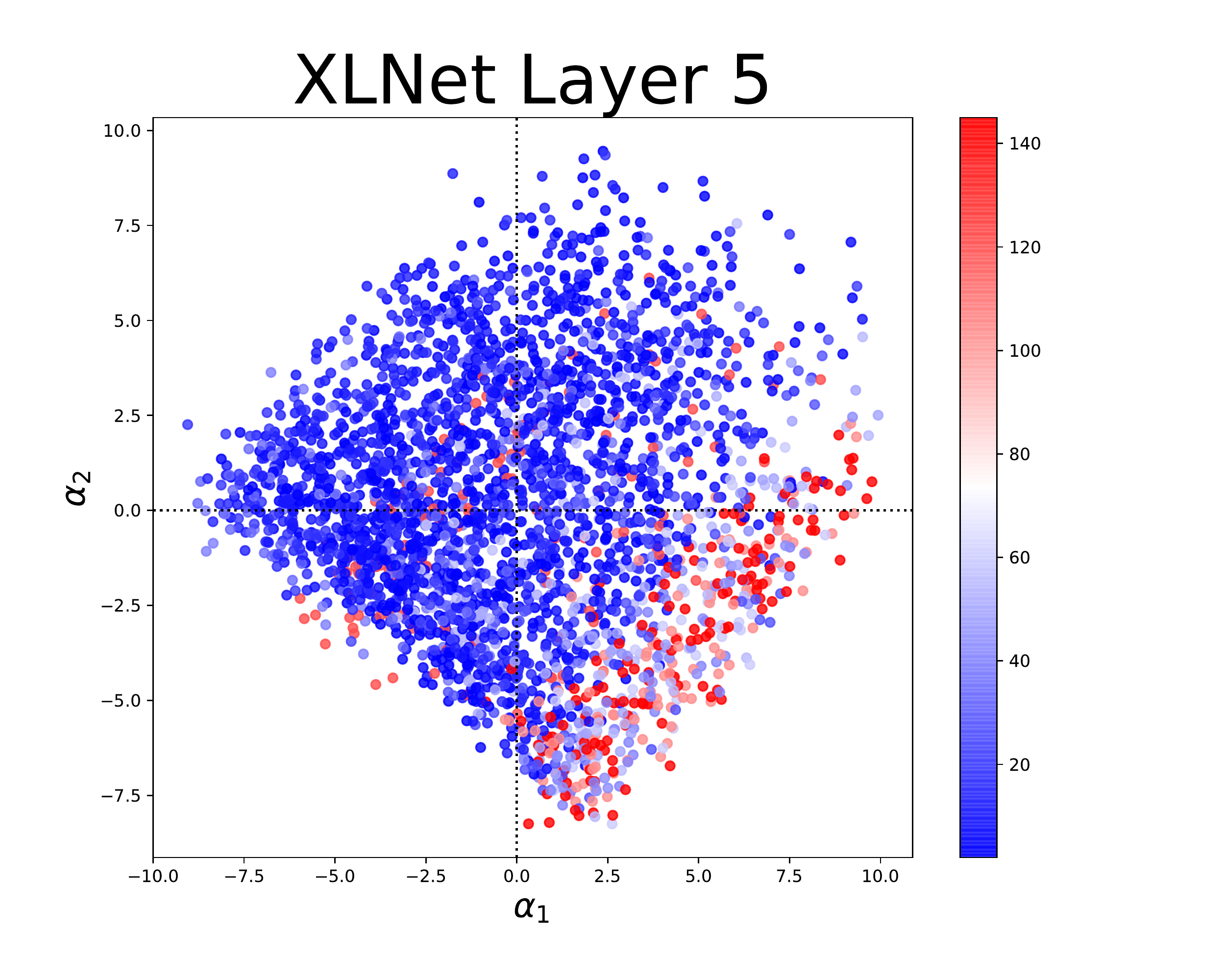}}
\subfloat[retrofitted]{\label{fig:l1}\includegraphics[width=0.2\linewidth]{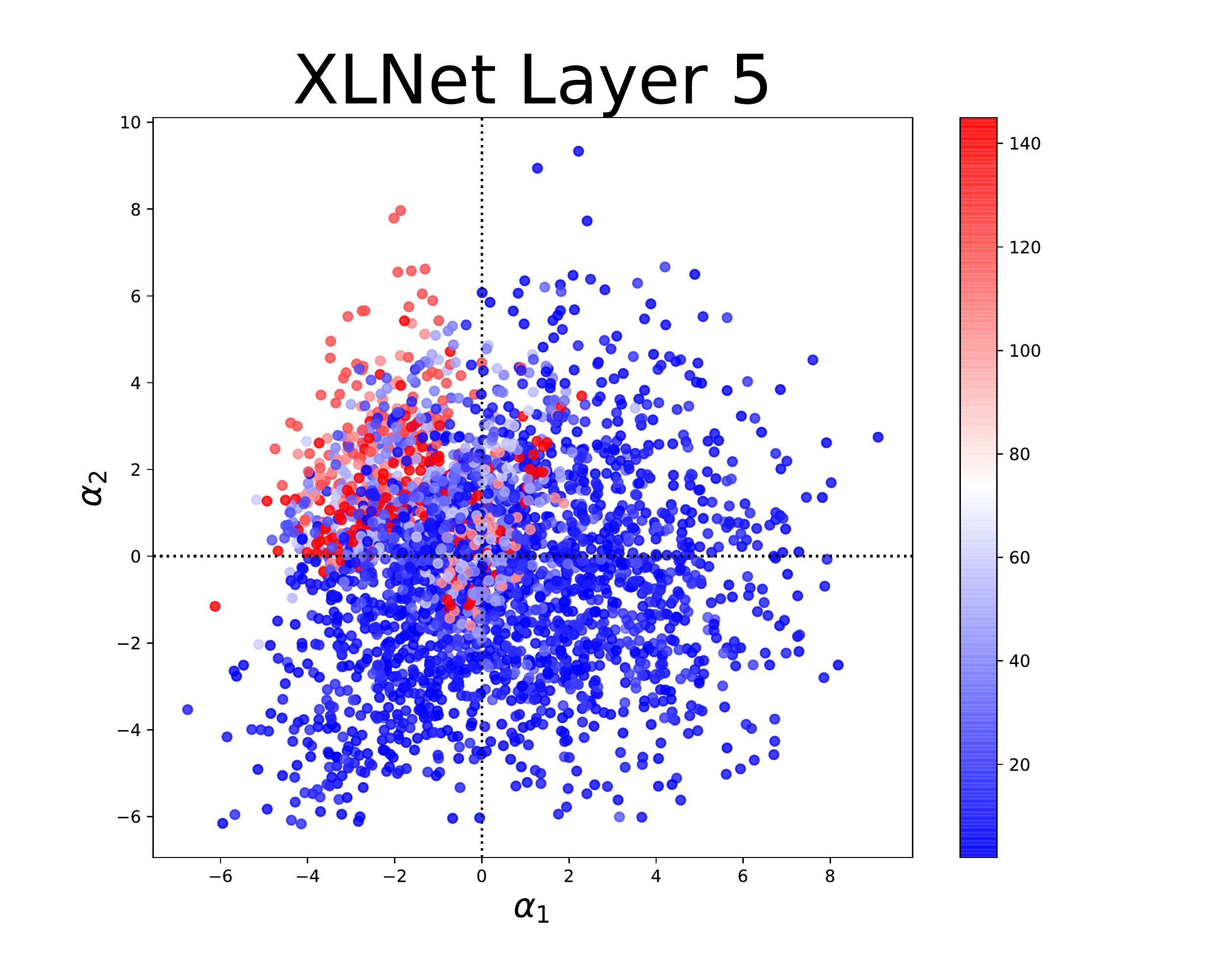}}\\
\subfloat[original]{\label{fig:m1}\includegraphics[width=0.2\linewidth]{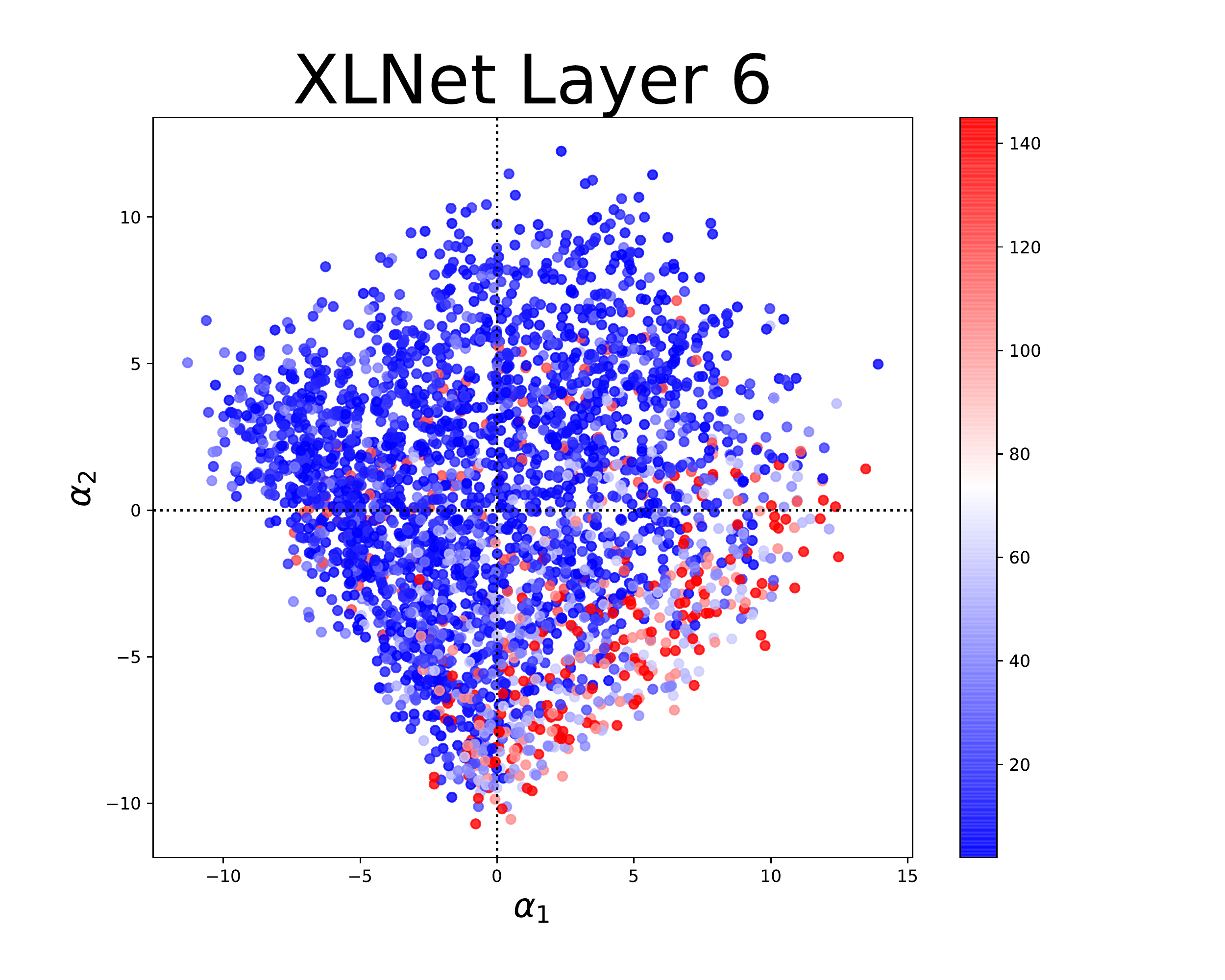}}
\subfloat[retrofitted]{\label{fig:n1}\includegraphics[width=0.2\linewidth]{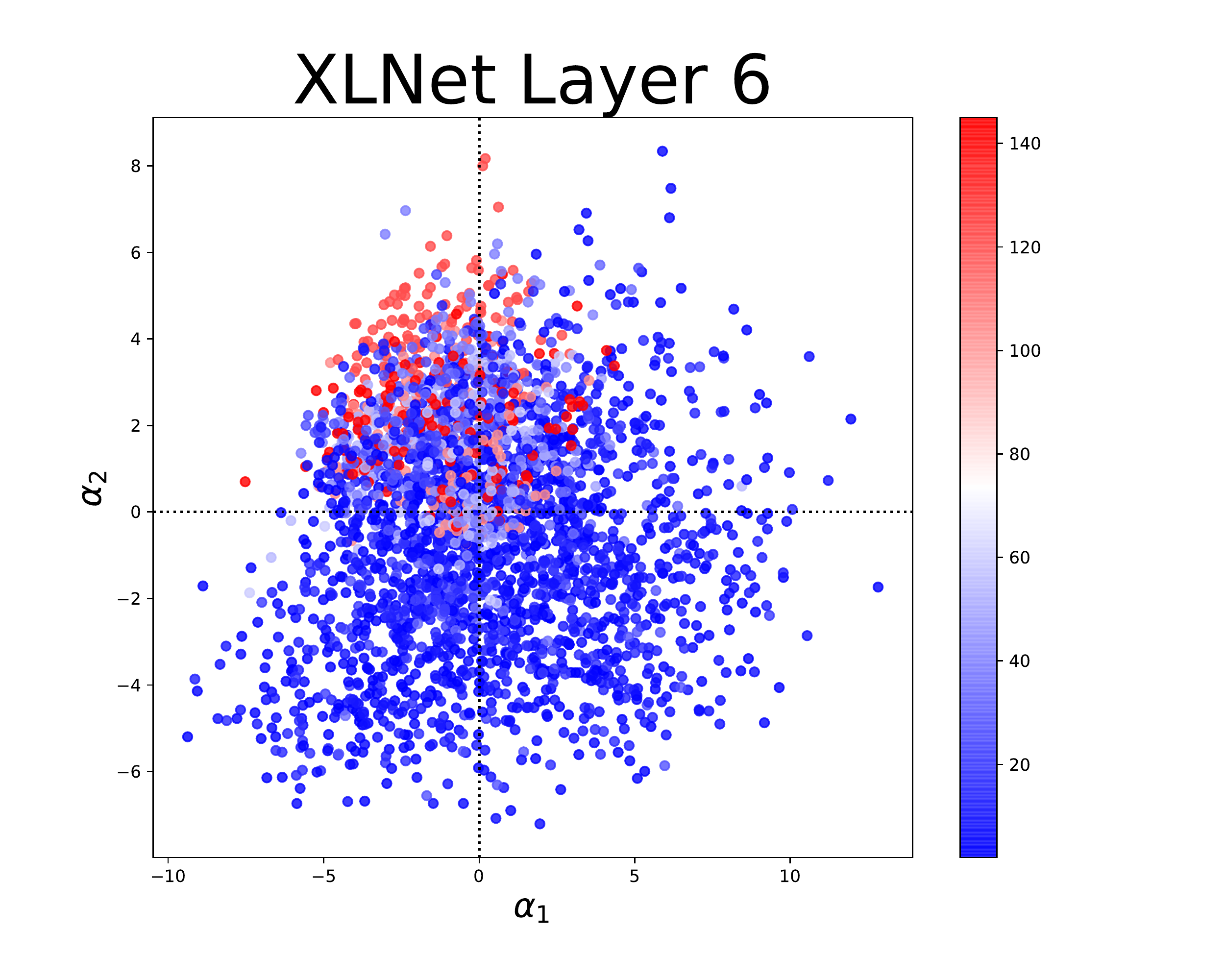}}
\subfloat[original]{\label{fig:o1}\includegraphics[width=0.2\linewidth]{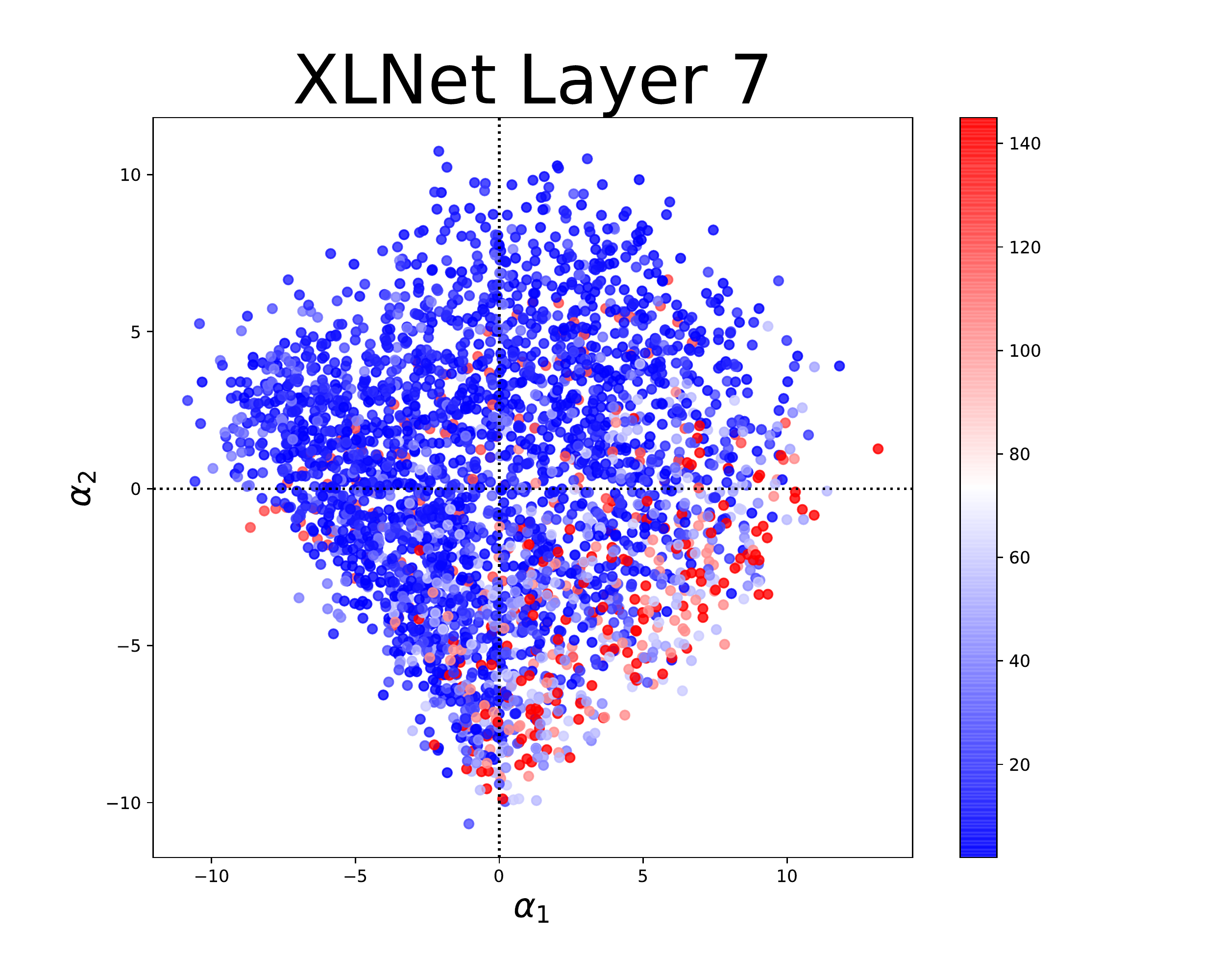}}
\subfloat[retrofitted]{\label{fig:p1}\includegraphics[width=0.2\linewidth]{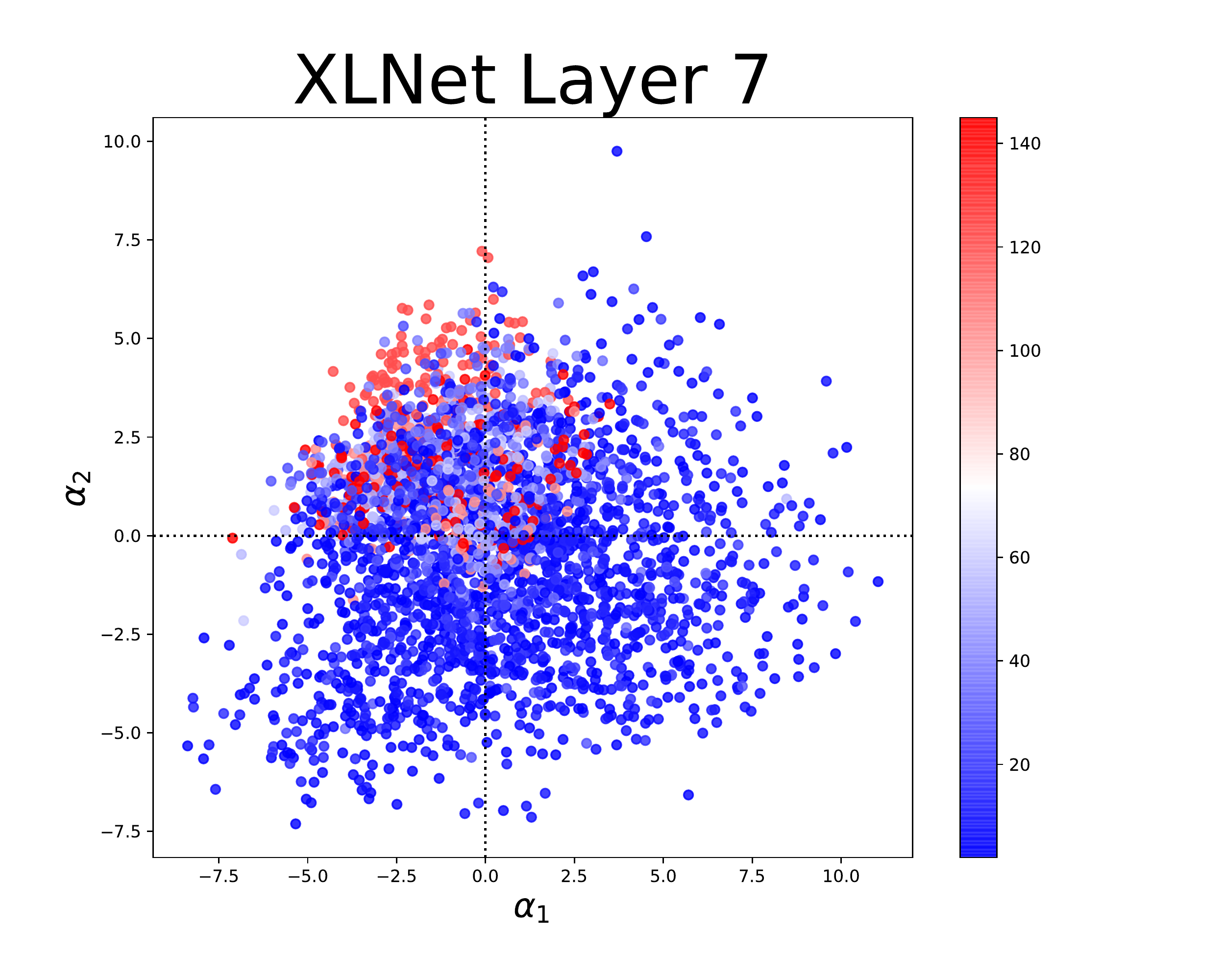}}\\
\subfloat[original]{\label{fig:q1}\includegraphics[width=0.2\linewidth]{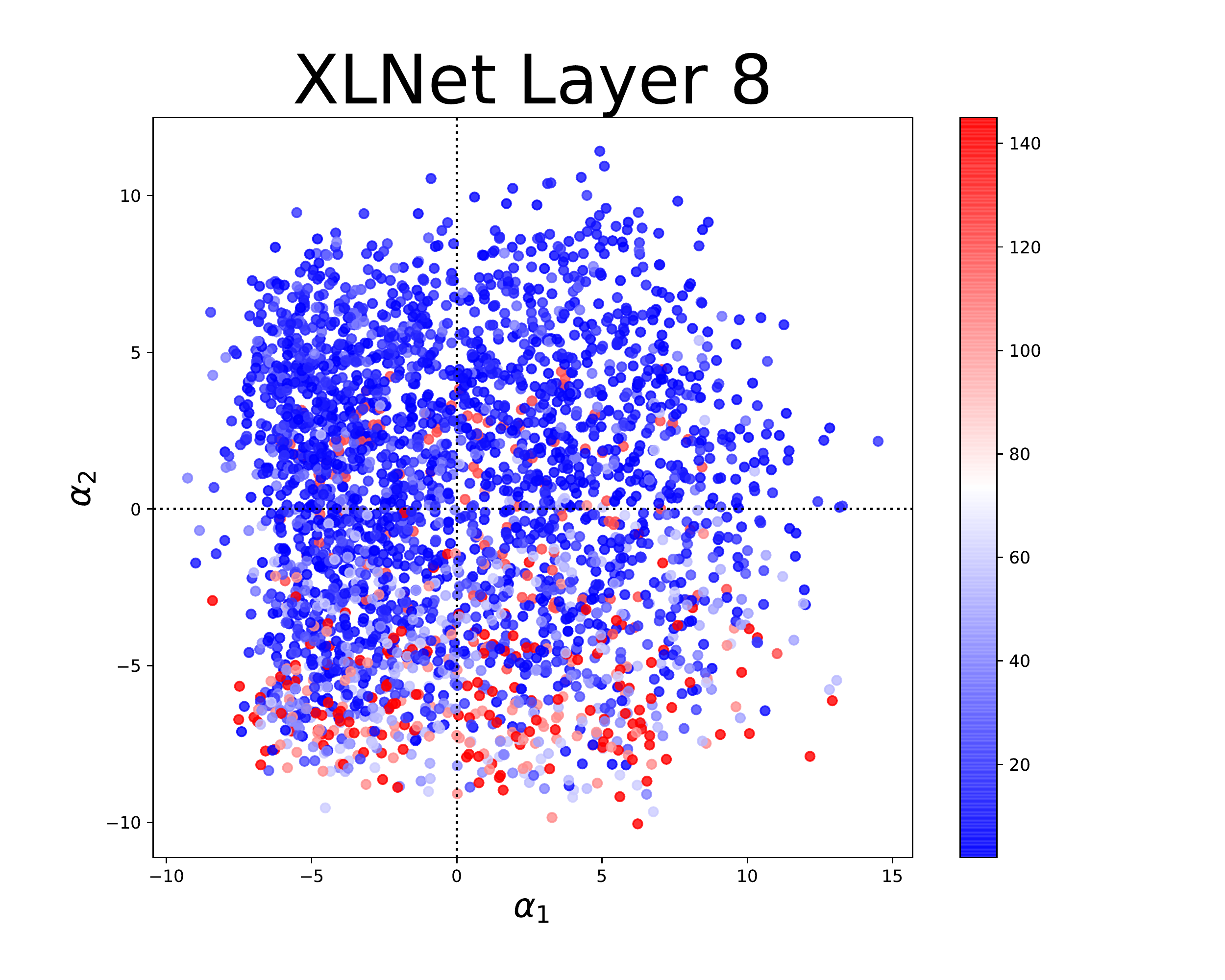}}
\subfloat[retrofitted]{\label{fig:r1}\includegraphics[width=0.2\linewidth]{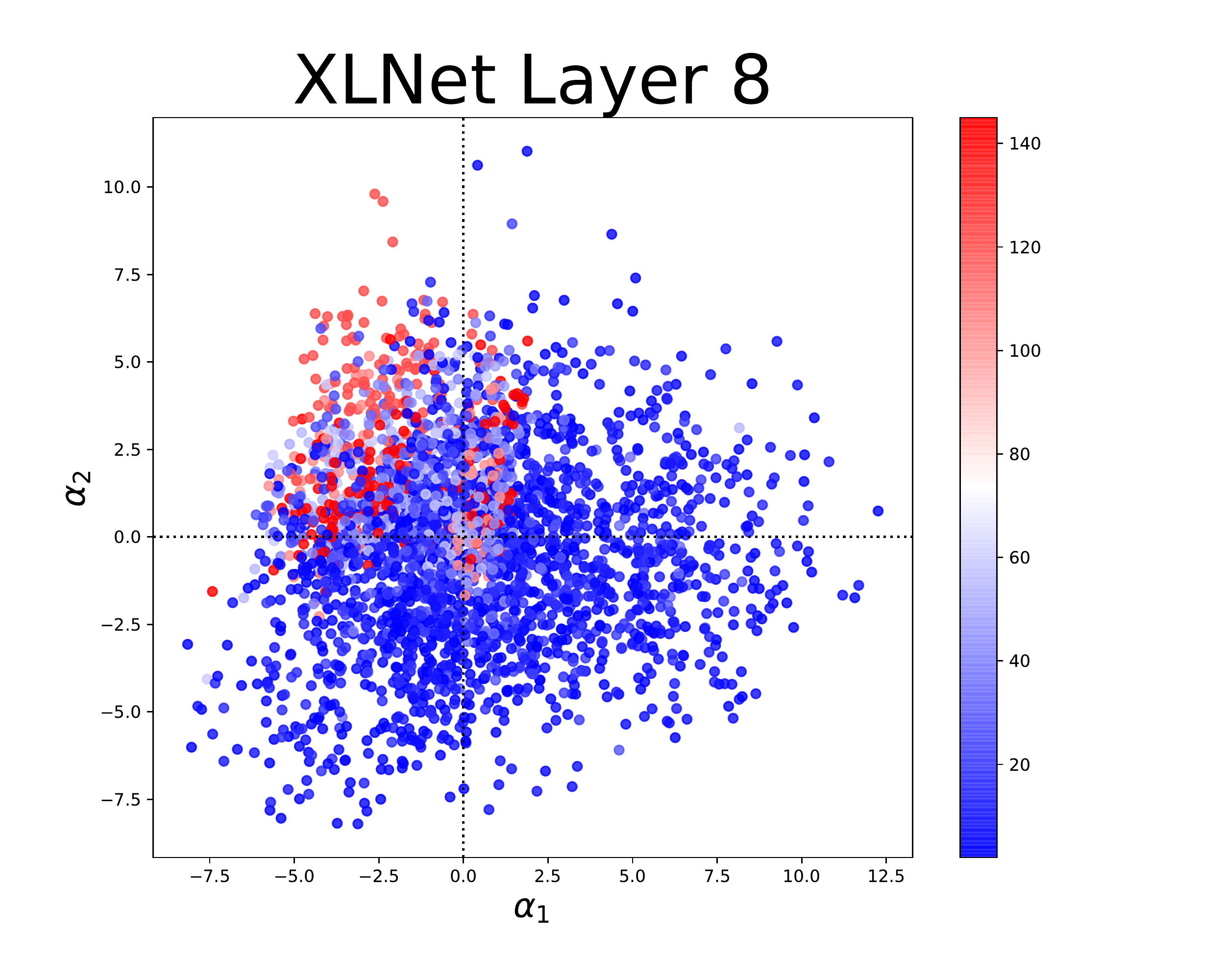}}
\subfloat[original]{\label{fig:s1}\includegraphics[width=0.2\linewidth]{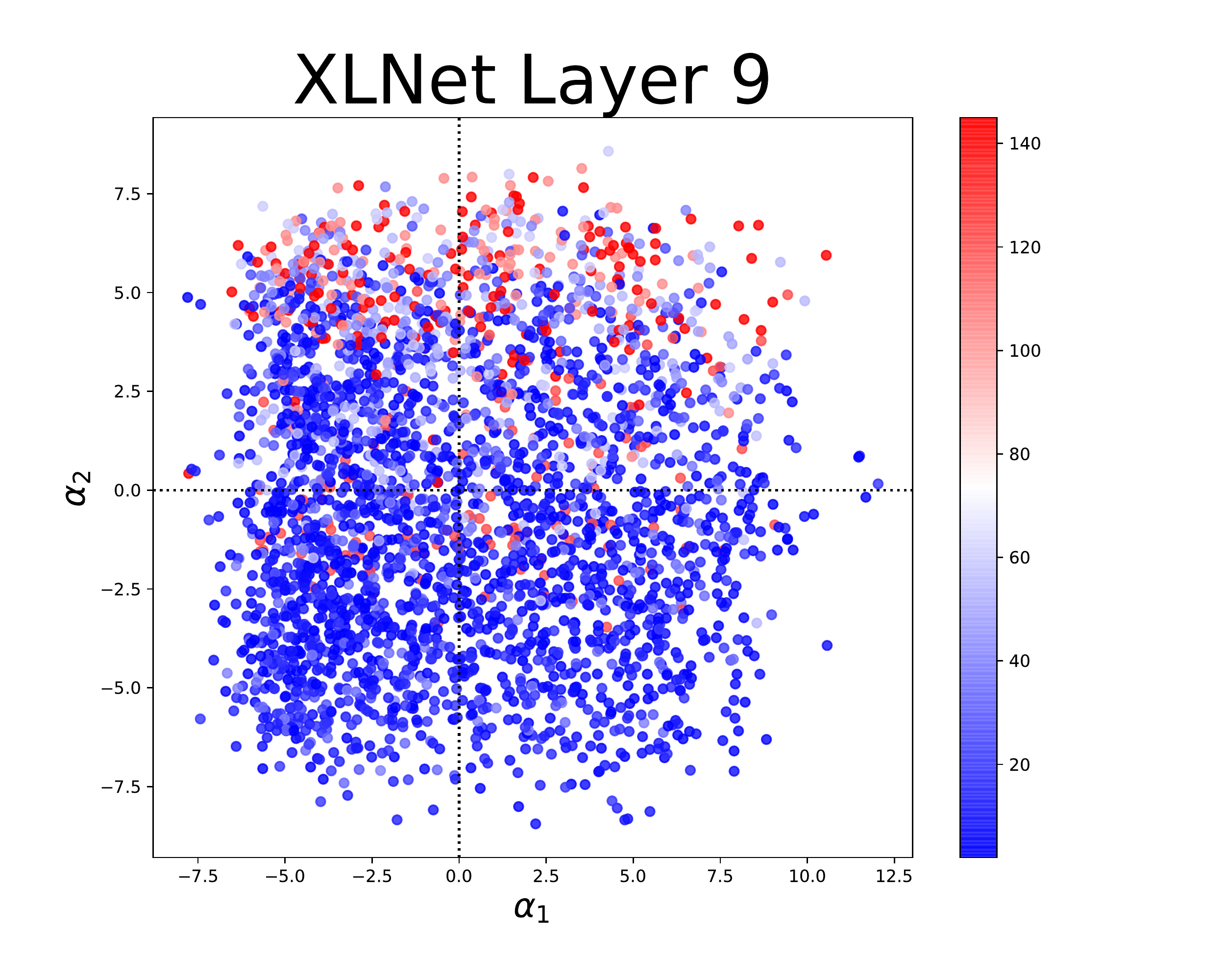}}
\subfloat[retrofitted]{\label{fig:t1}\includegraphics[width=0.2\linewidth]{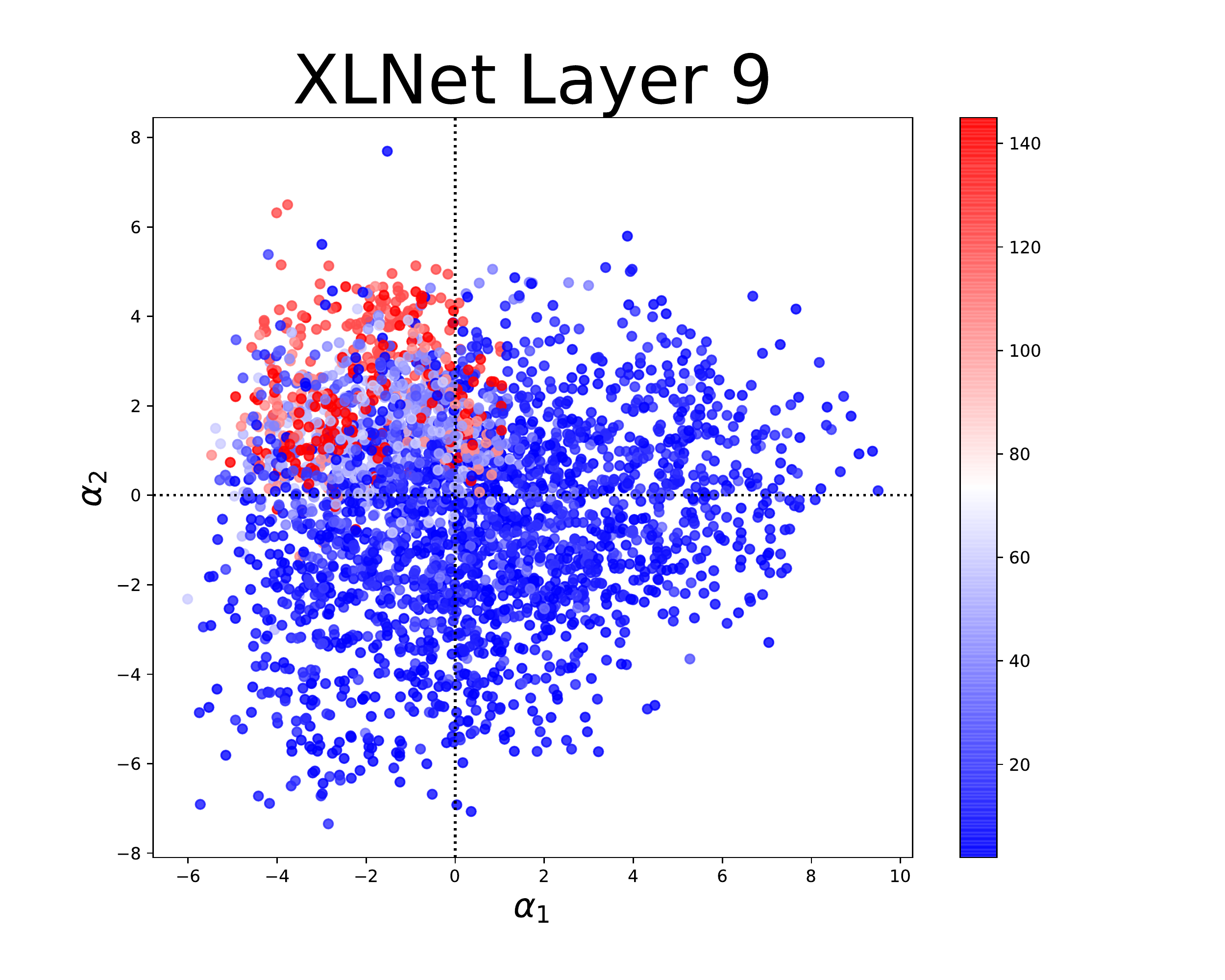}}\\
\subfloat[original]{\label{fig:u1}\includegraphics[width=0.2\linewidth]{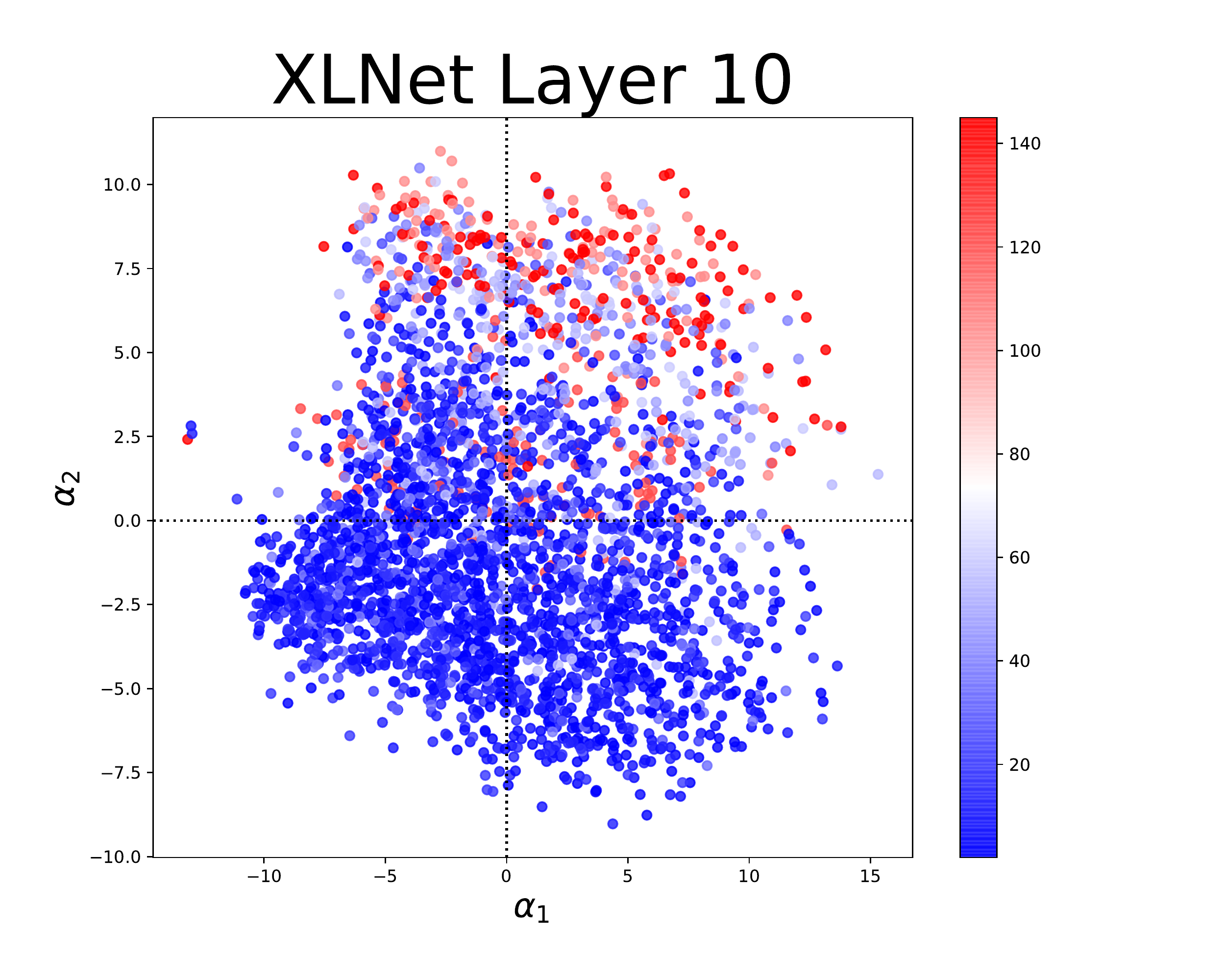}}
\subfloat[retrofitted]{\label{fig:v1}\includegraphics[width=0.2\linewidth]{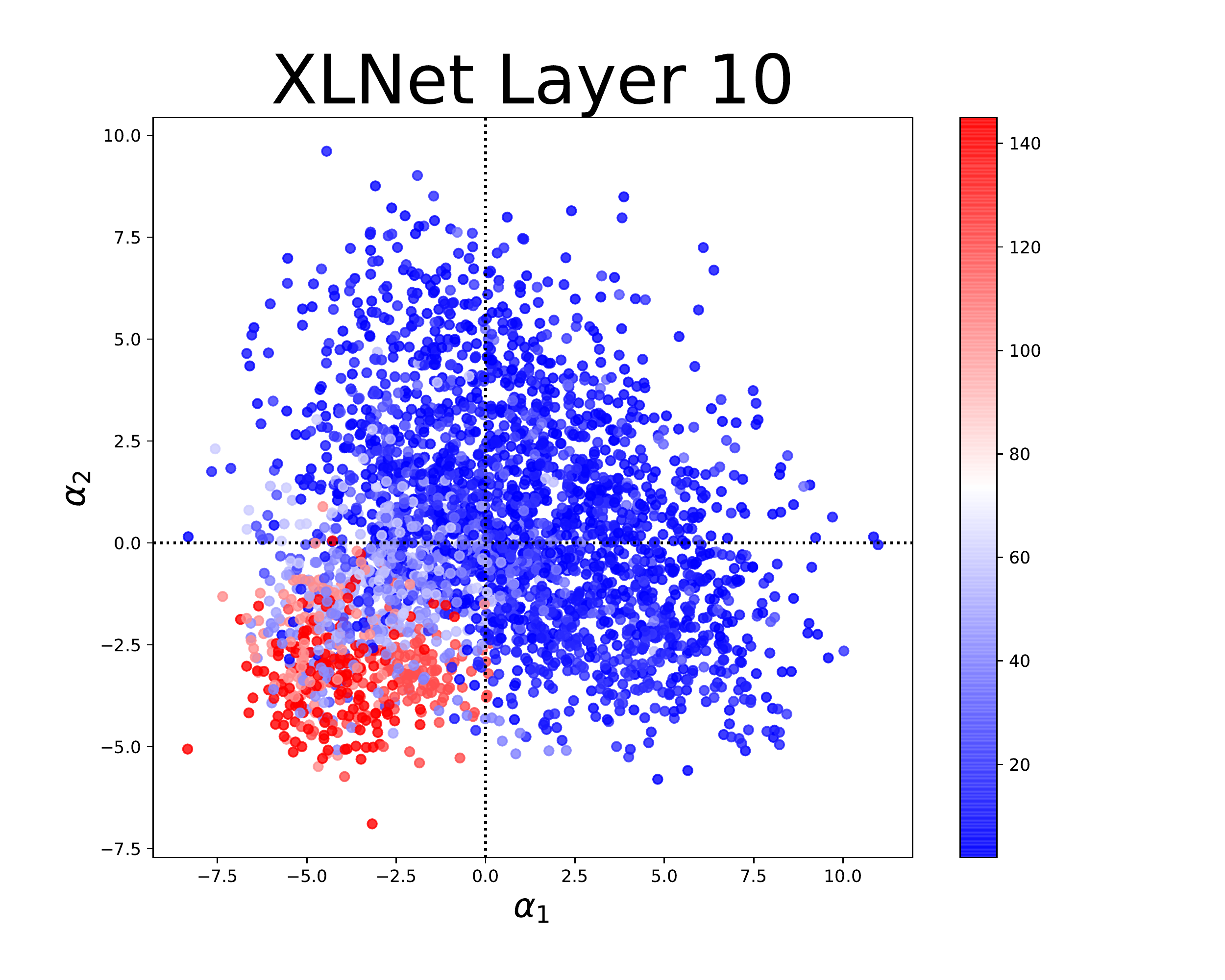}}
\subfloat[original]{\label{fig:w1}\includegraphics[width=0.2\linewidth]{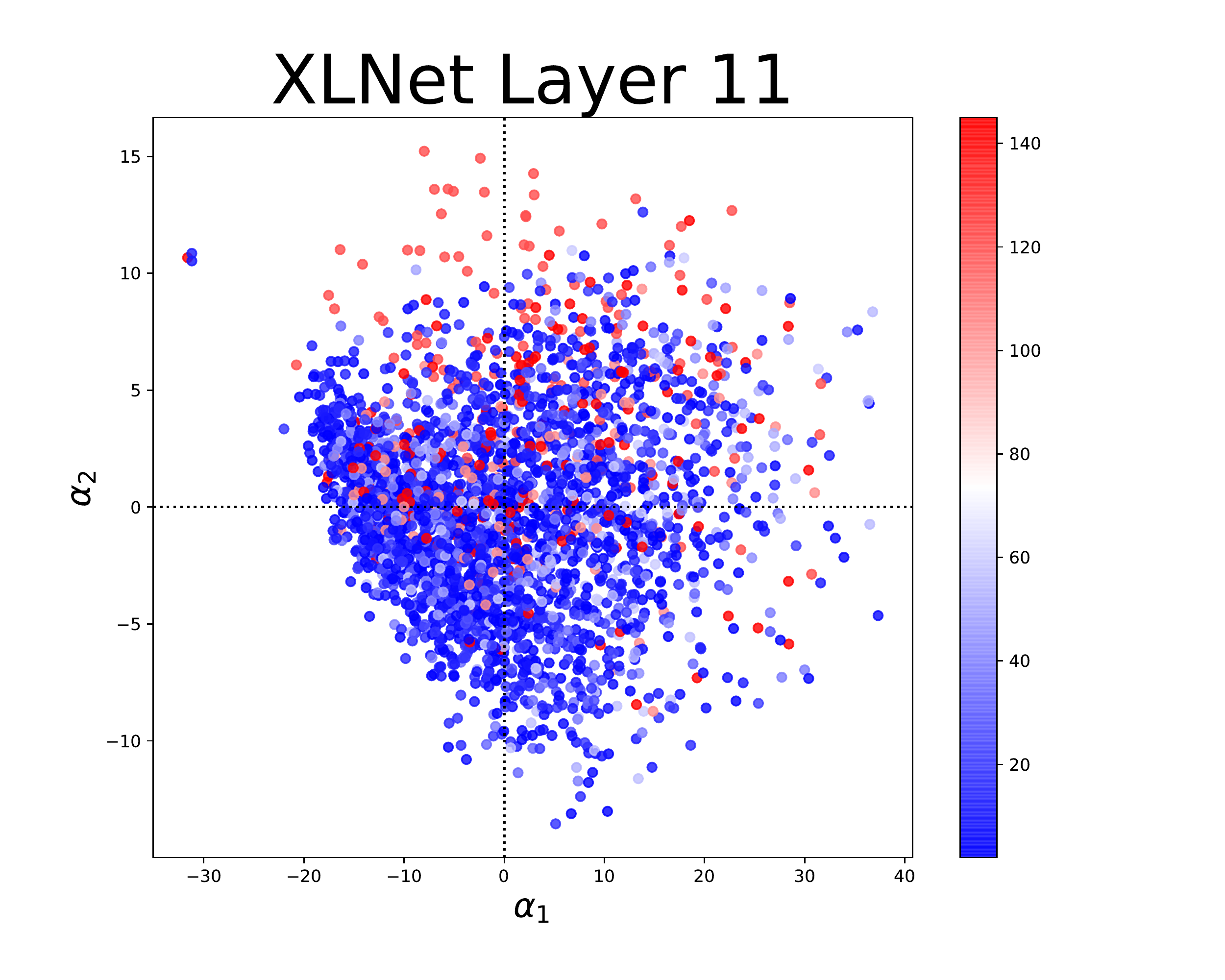}}
\subfloat[retrofitted]{\label{fig:x1}\includegraphics[width=0.2\linewidth]{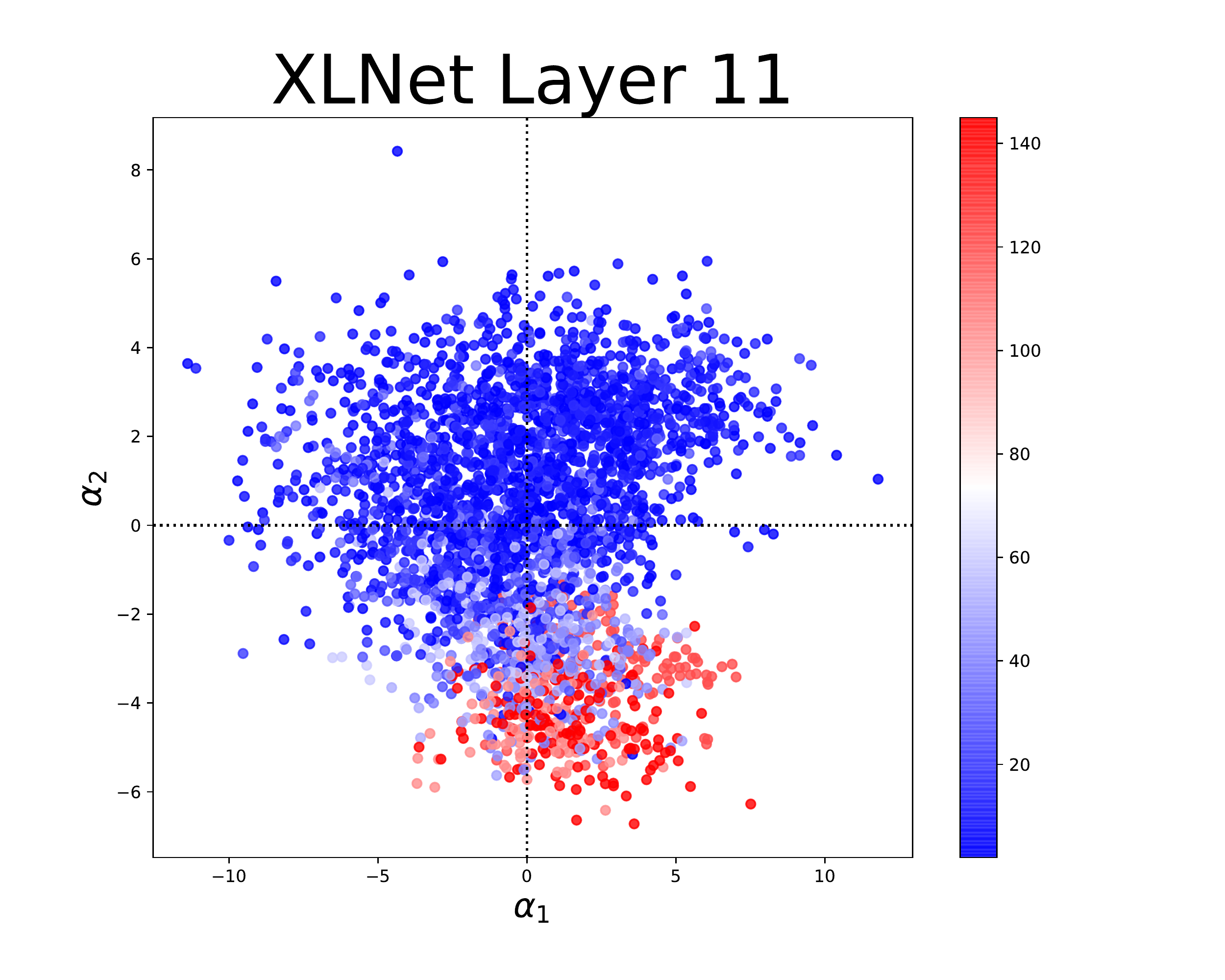}}\\
\subfloat[original]{\label{fig:y1}\includegraphics[width=0.2\linewidth]{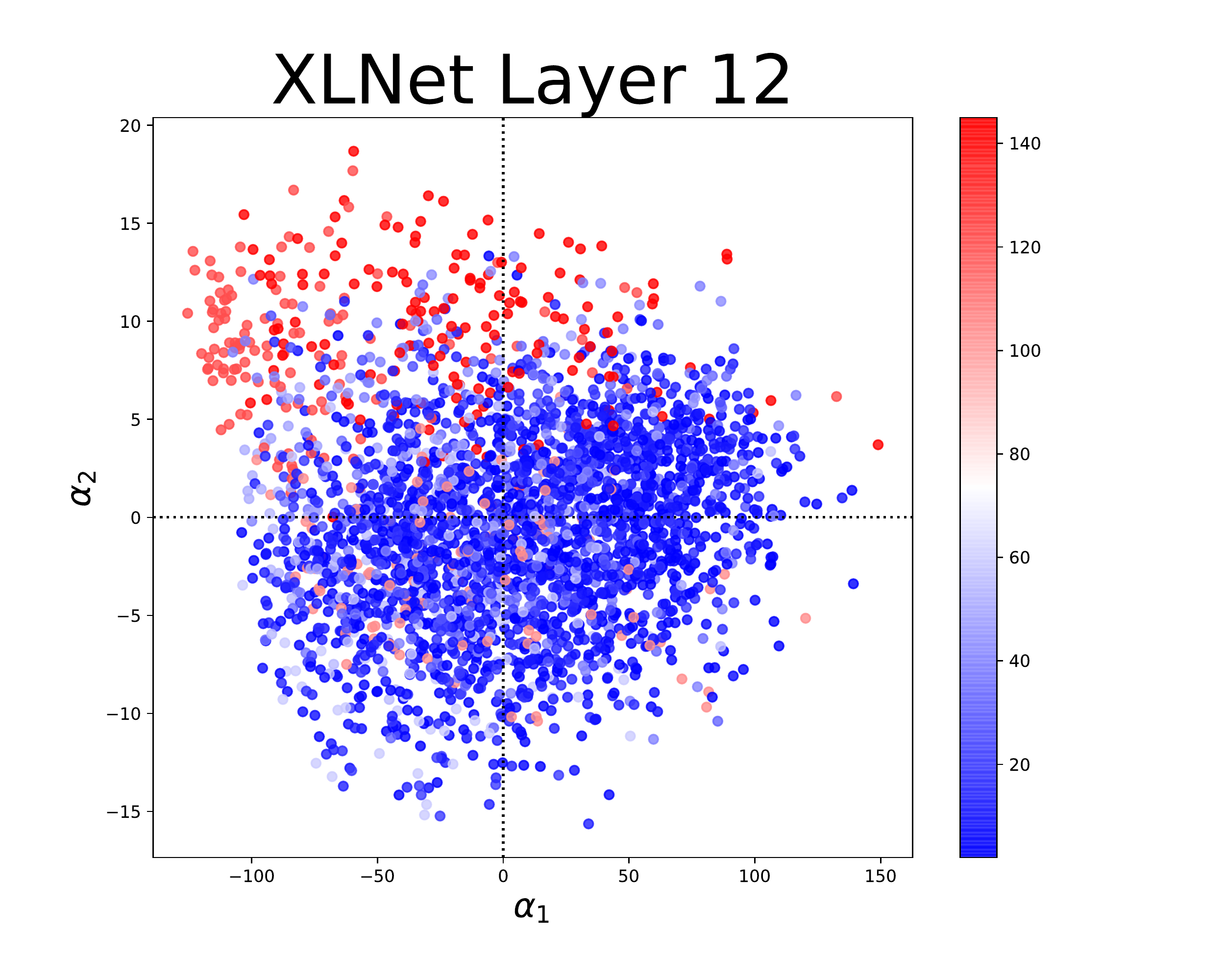}}
\subfloat[retrofitted]{\label{fig:z1}\includegraphics[width=0.2\linewidth]{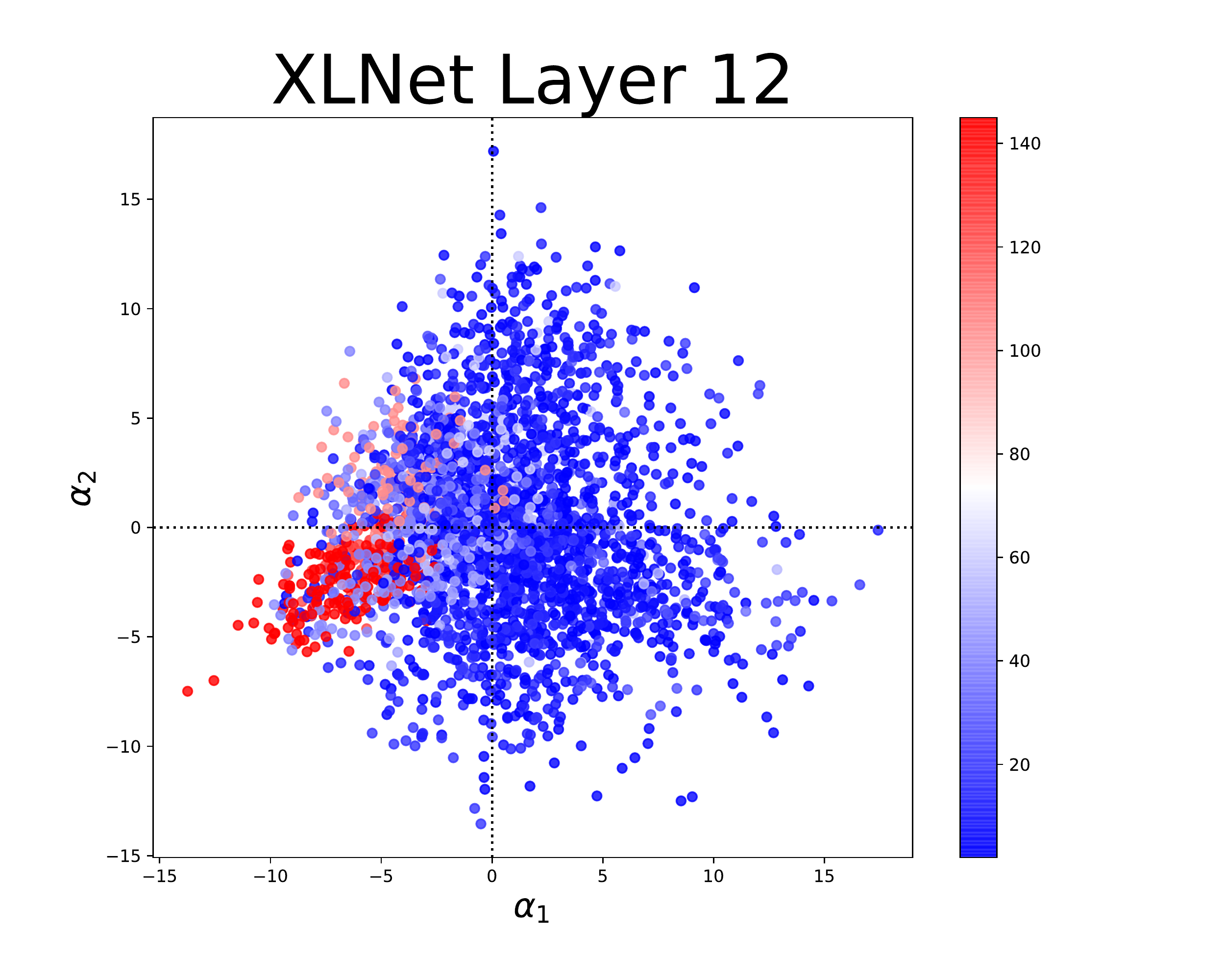}}
\caption{PCA Plots of XLNet Word Representations.}
\label{fig:xlnet_fig}
\end{figure*}

\begin{figure*}
\centering
\subfloat[original]{\label{fig:a2}\includegraphics[width=0.2\linewidth]{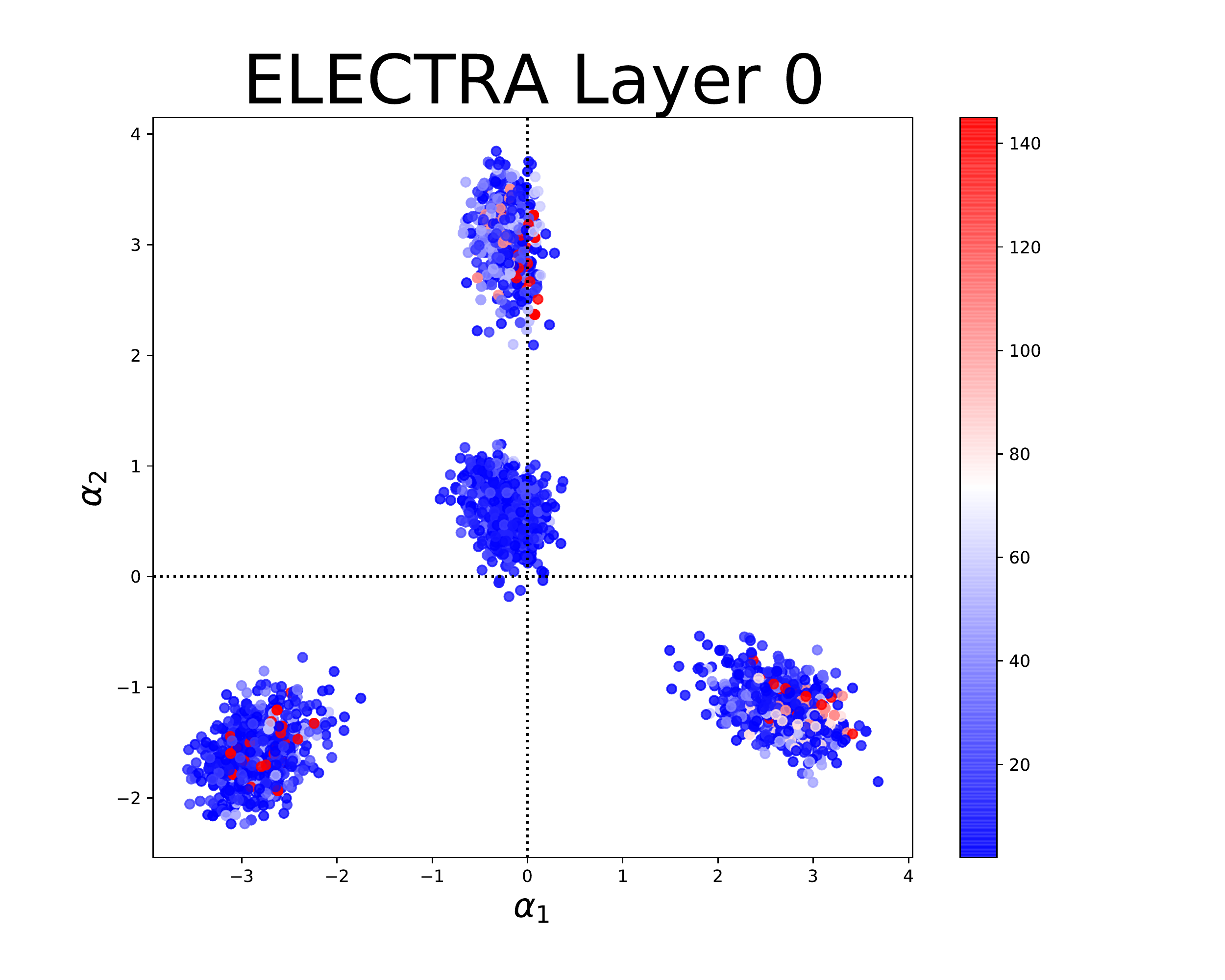}}
\subfloat[retrofitted]{\label{fig:b2}\includegraphics[width=0.2\linewidth]{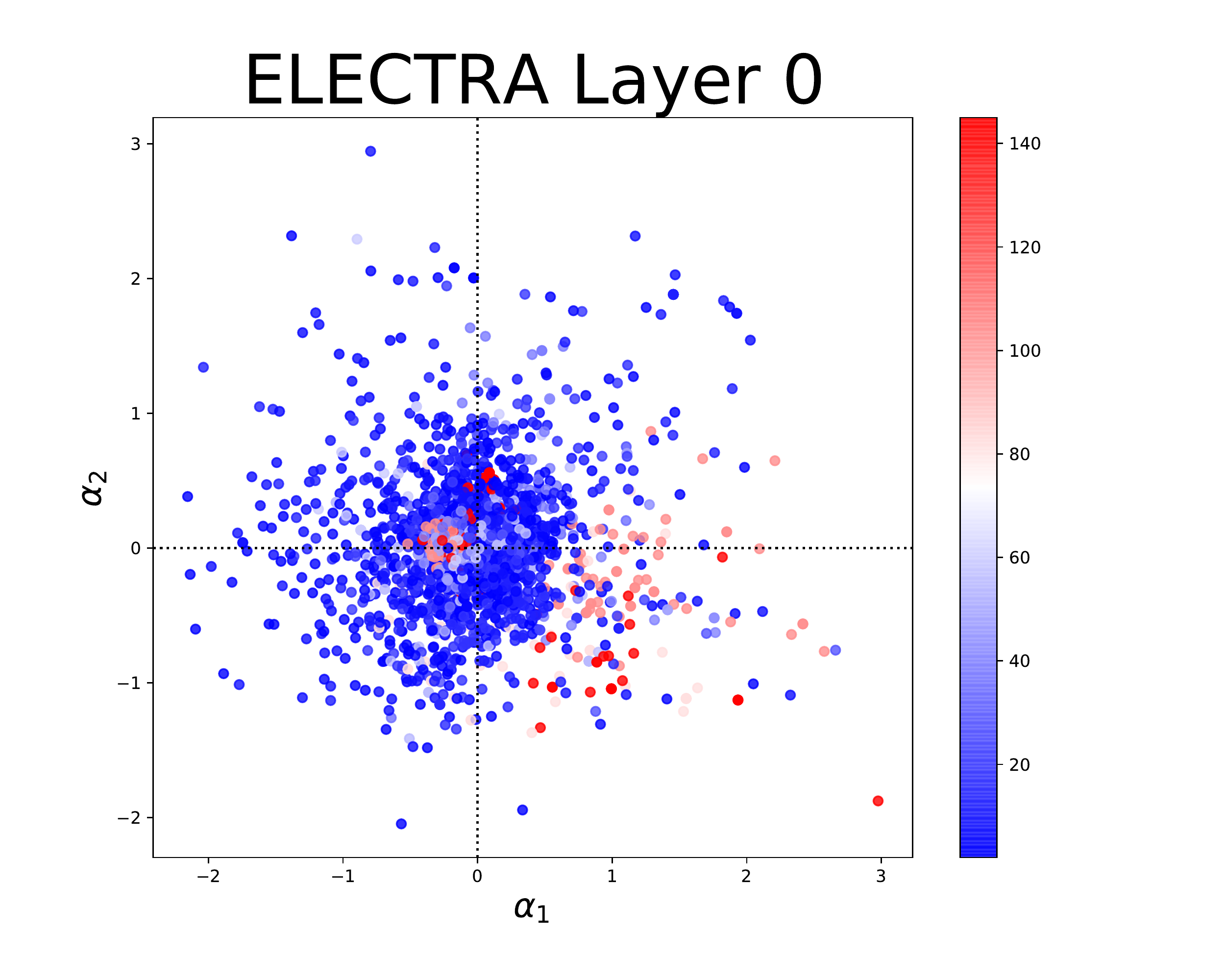}}
\subfloat[original]{\label{fig:c2}\includegraphics[width=0.2\linewidth]{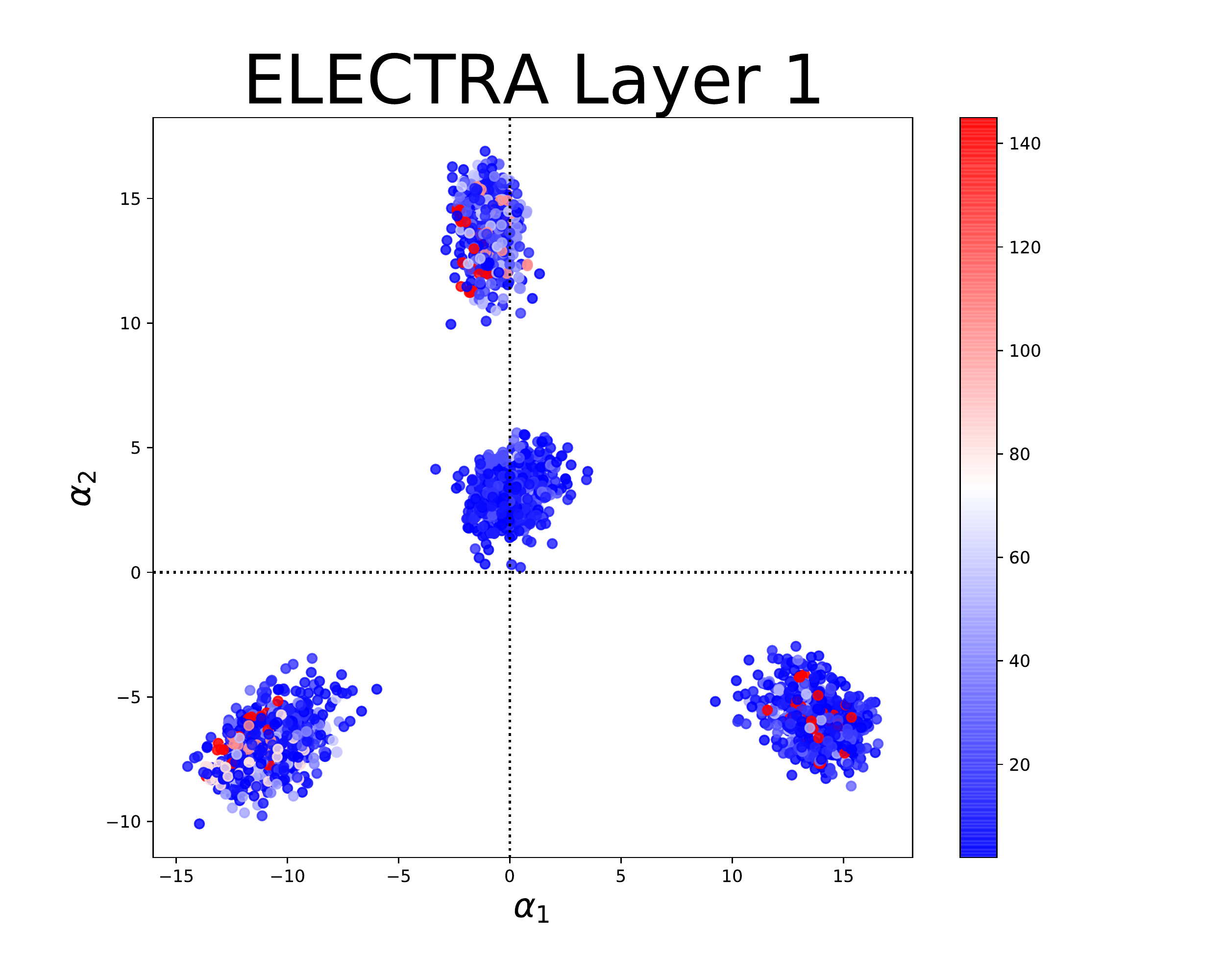}}
\subfloat[retrofitted]{\label{fig:d2}\includegraphics[width=0.2\linewidth]{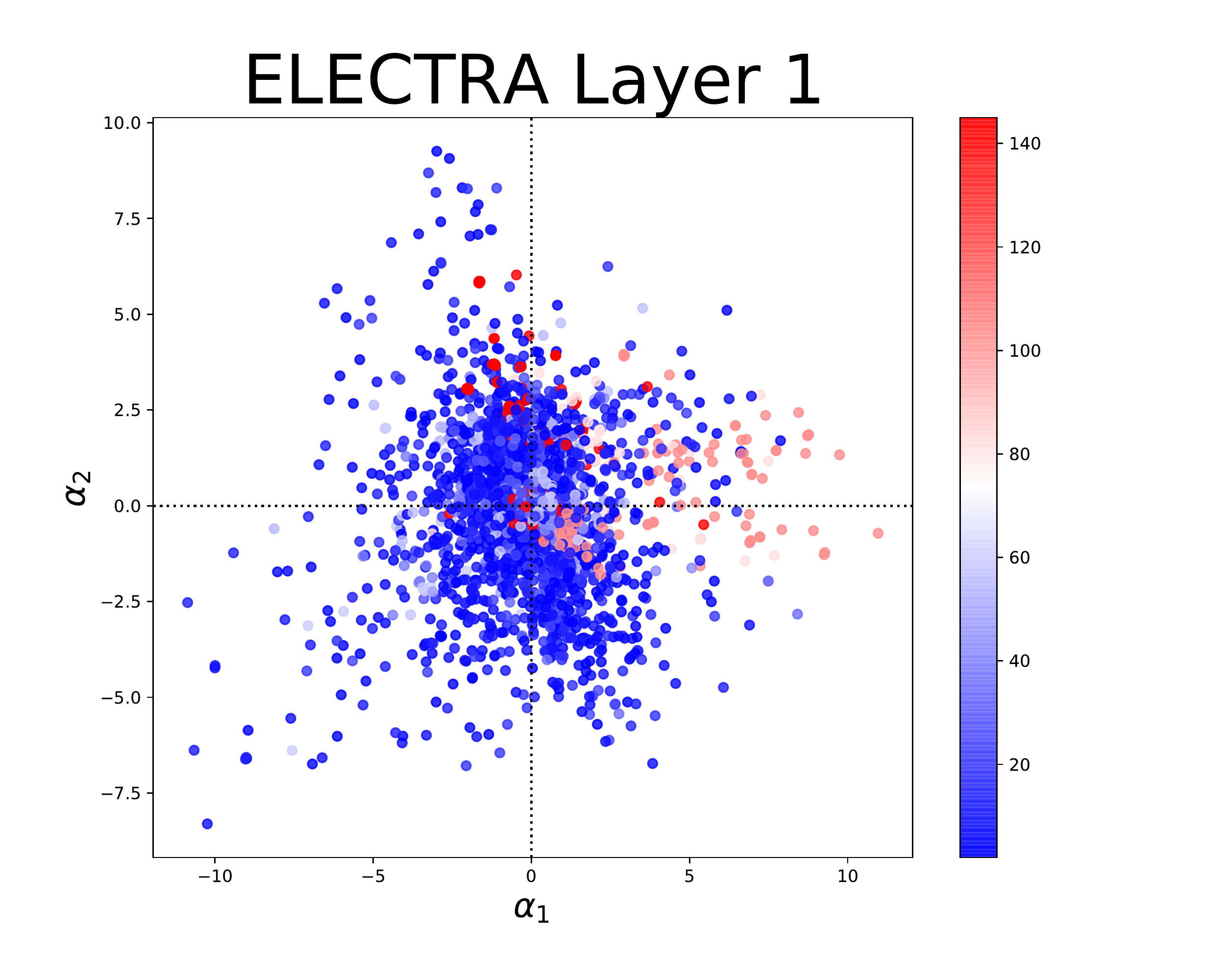}}\\
\subfloat[original]{\label{fig:e2}\includegraphics[width=0.2\linewidth]{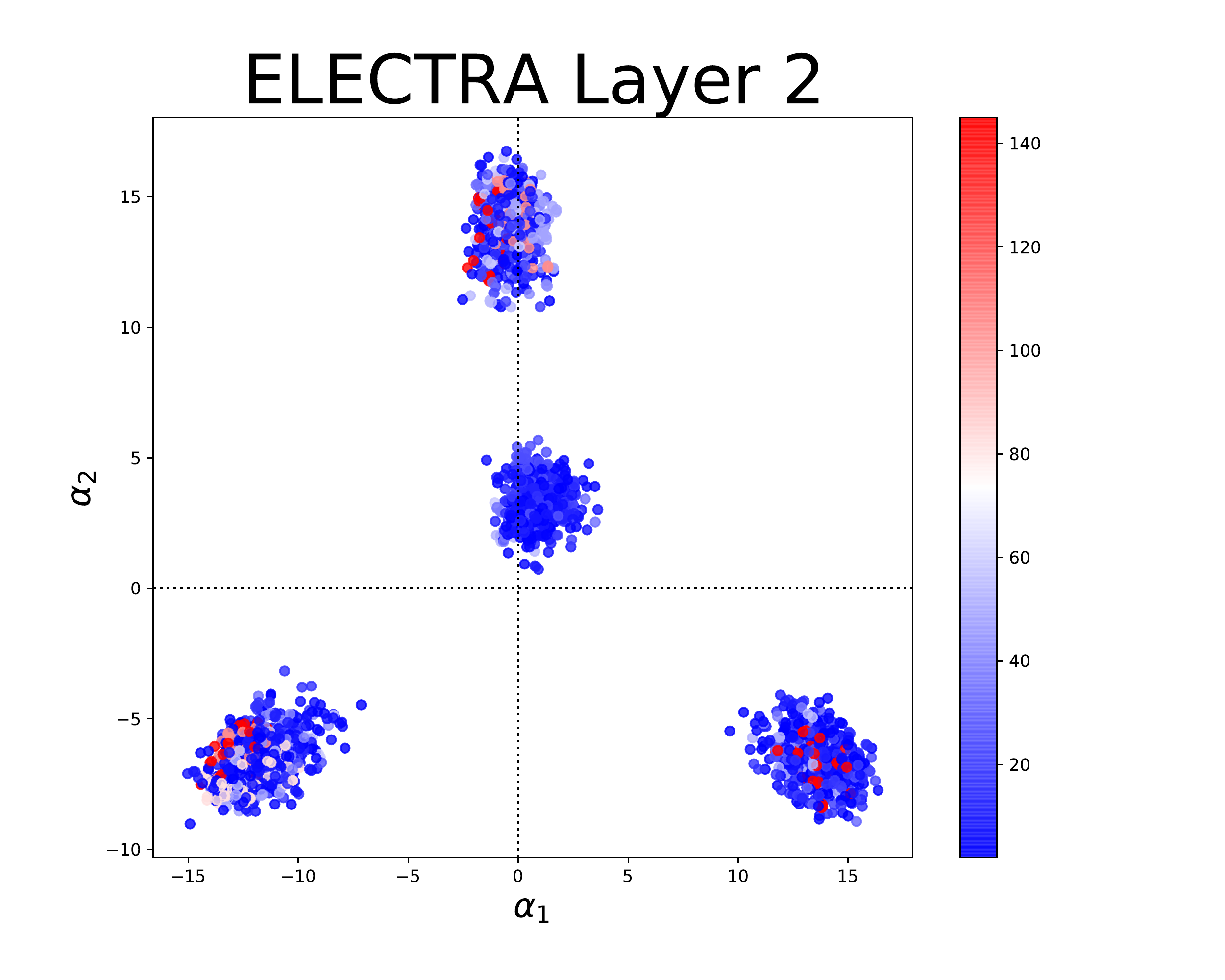}}
\subfloat[retrofitted]{\label{fig:f2}\includegraphics[width=0.2\linewidth]{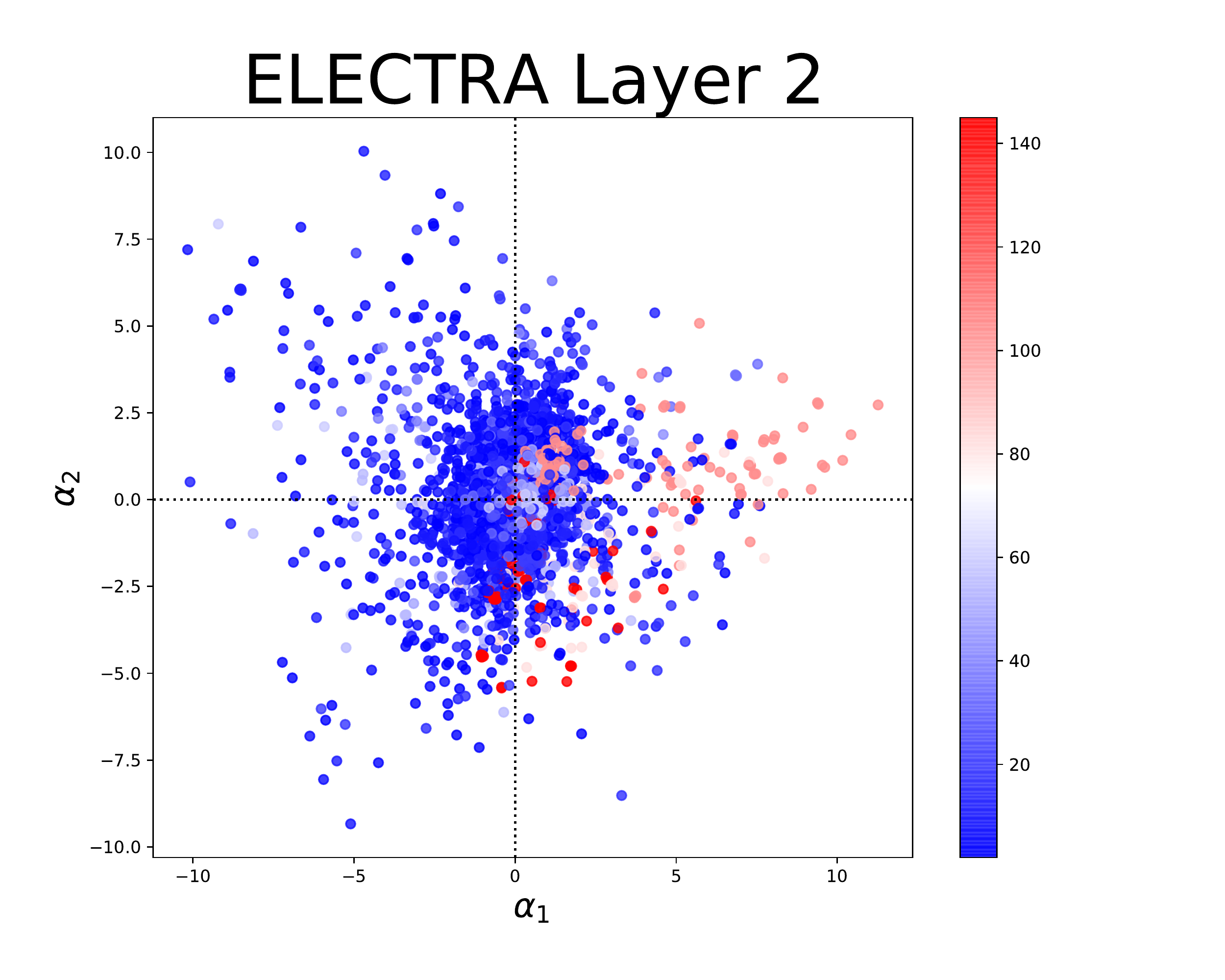}}
\subfloat[original]{\label{fig:g2}\includegraphics[width=0.2\linewidth]{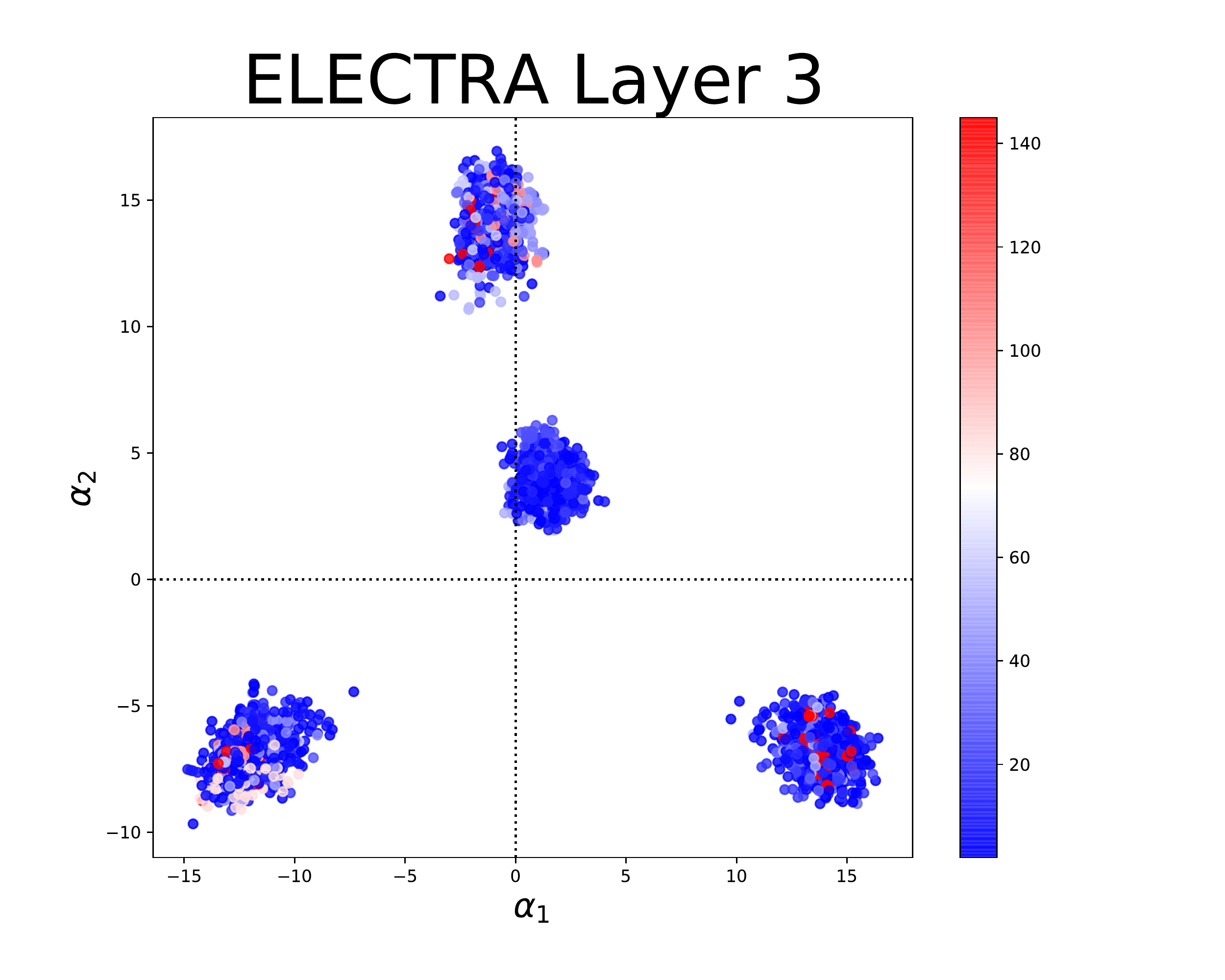}}
\subfloat[retrofitted]{\label{fig:h2}\includegraphics[width=0.2\linewidth]{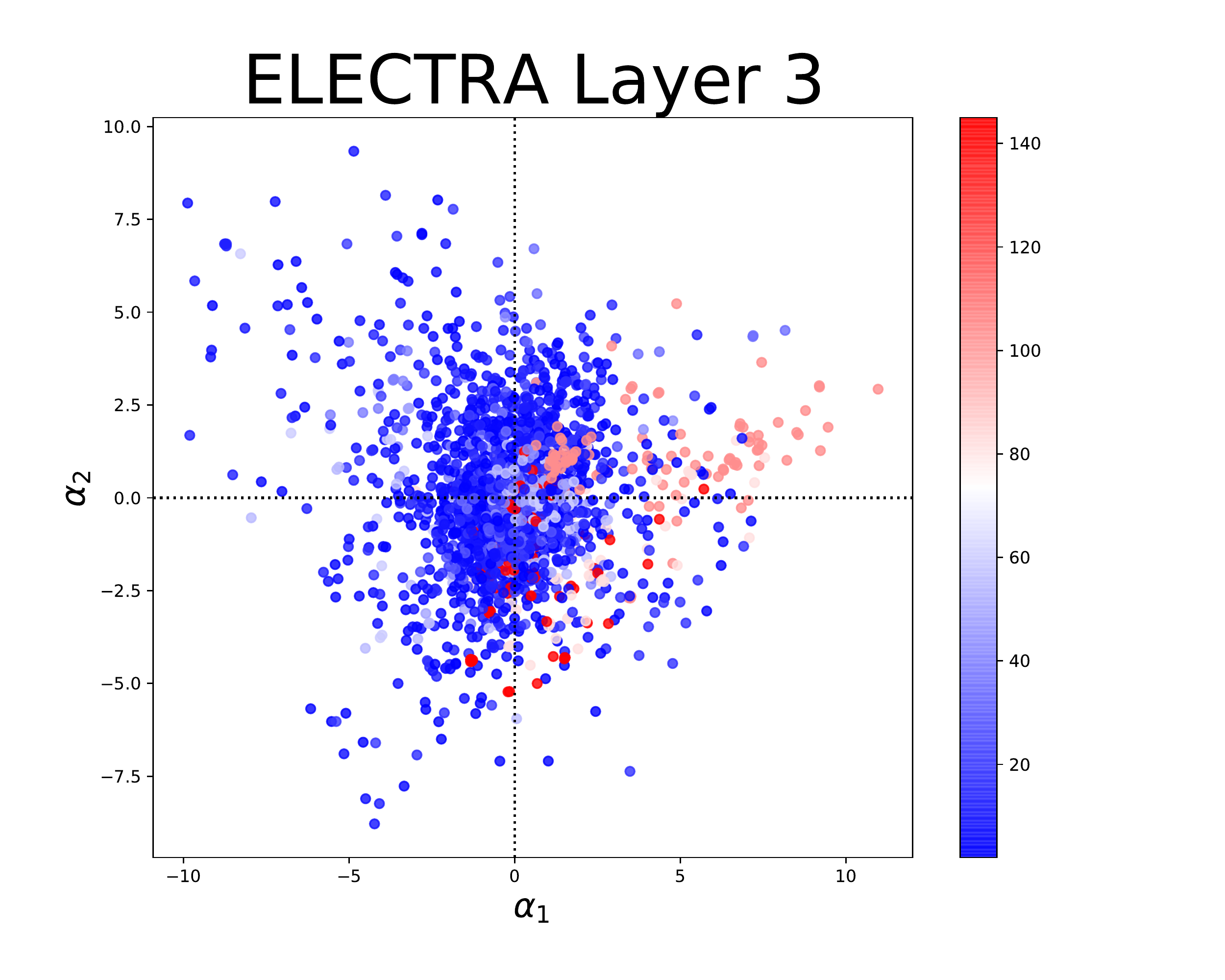}}\\
\subfloat[original]{\label{fig:i2}\includegraphics[width=0.2\linewidth]{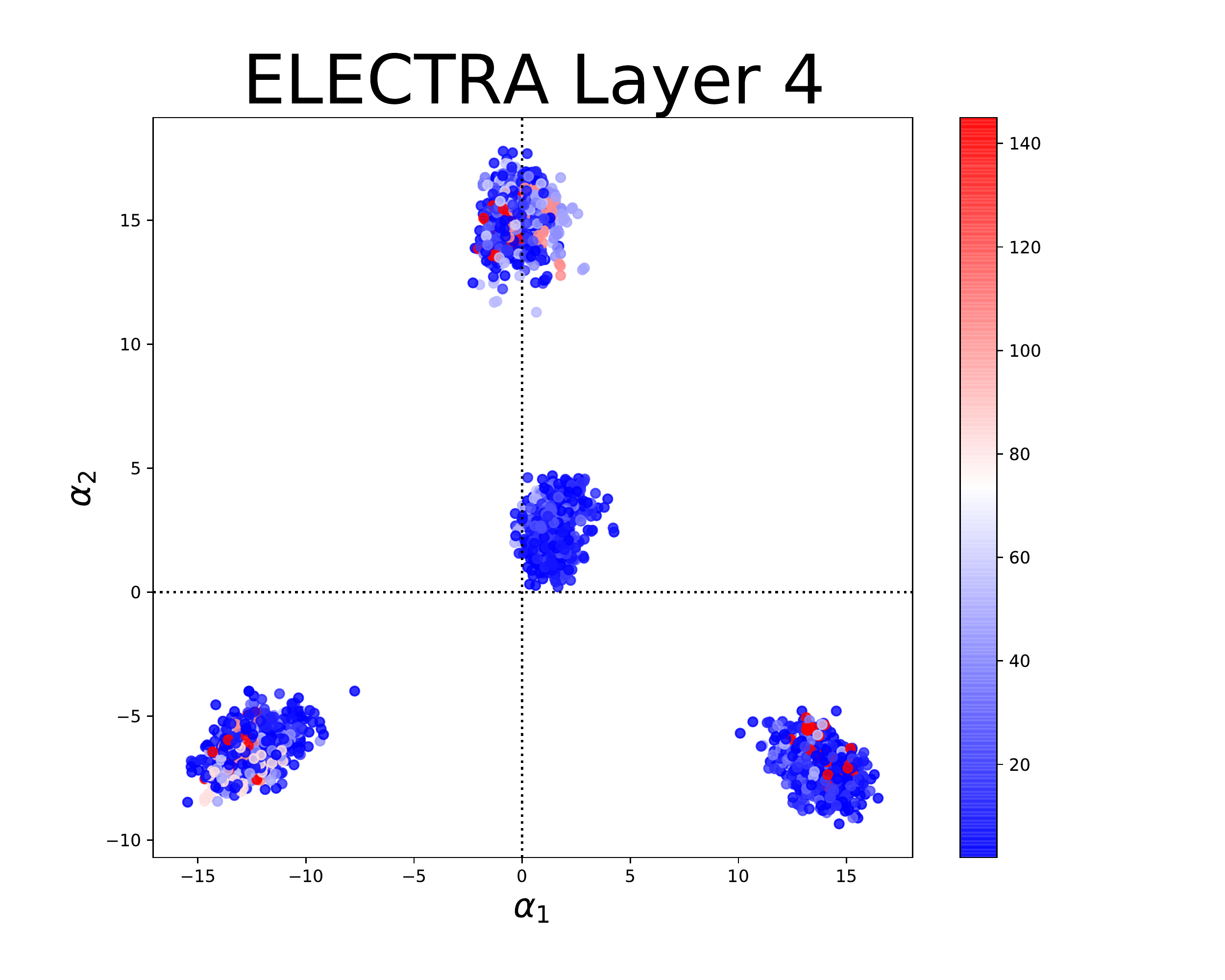}}
\subfloat[retrofitted]{\label{fig:j2}\includegraphics[width=0.2\linewidth]{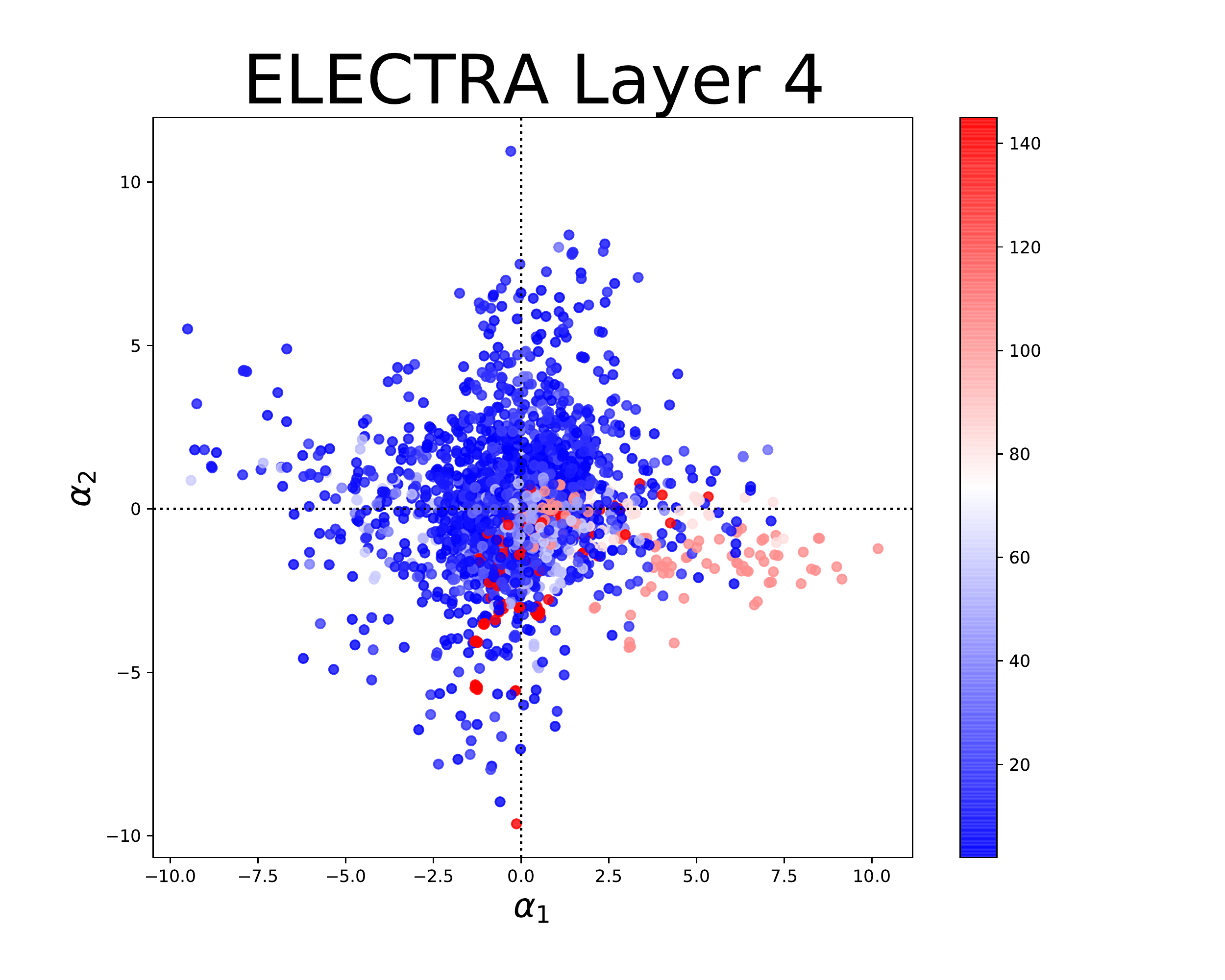}}
\subfloat[original]{\label{fig:k2}\includegraphics[width=0.2\linewidth]{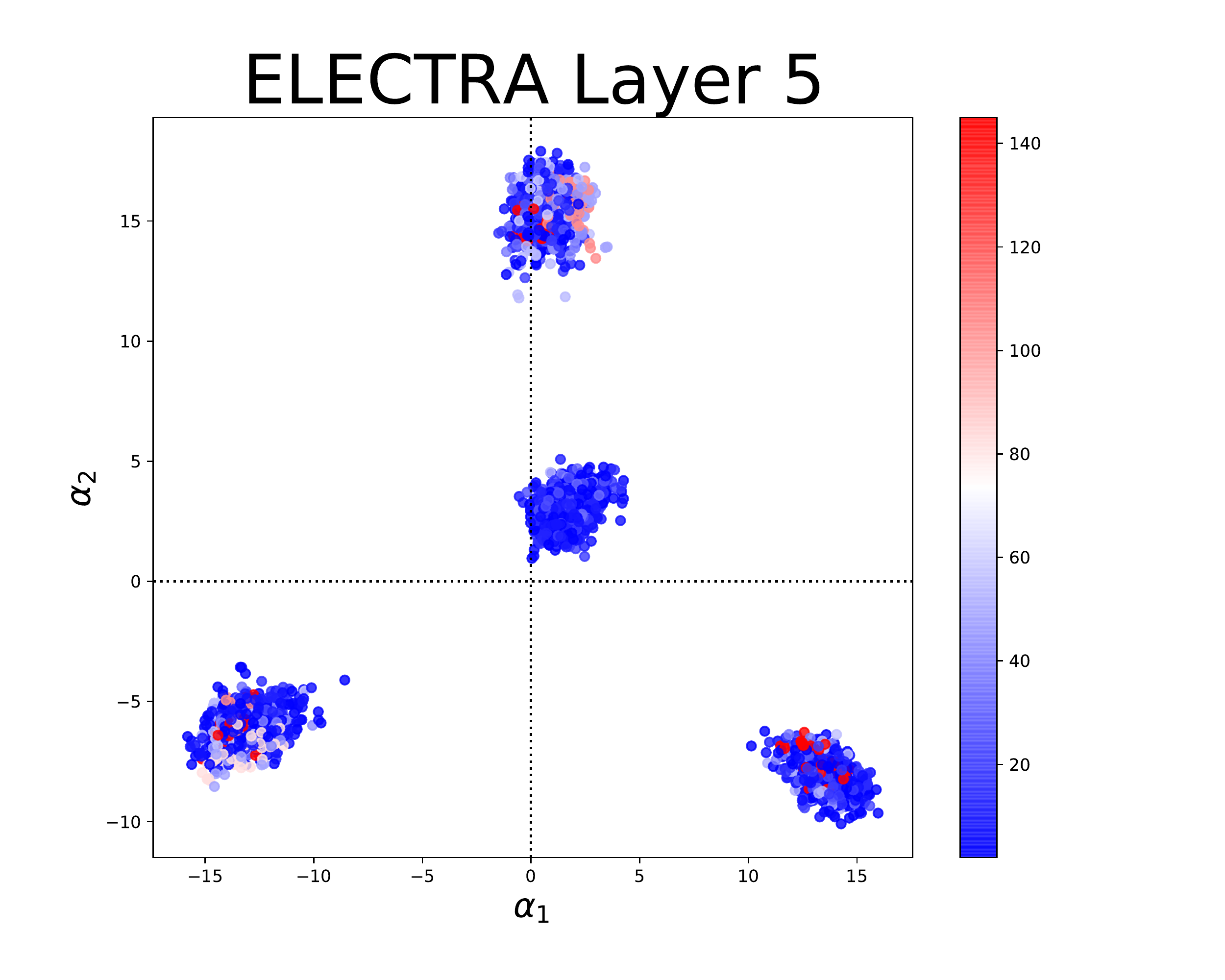}}
\subfloat[retrofitted]{\label{fig:l2}\includegraphics[width=0.2\linewidth]{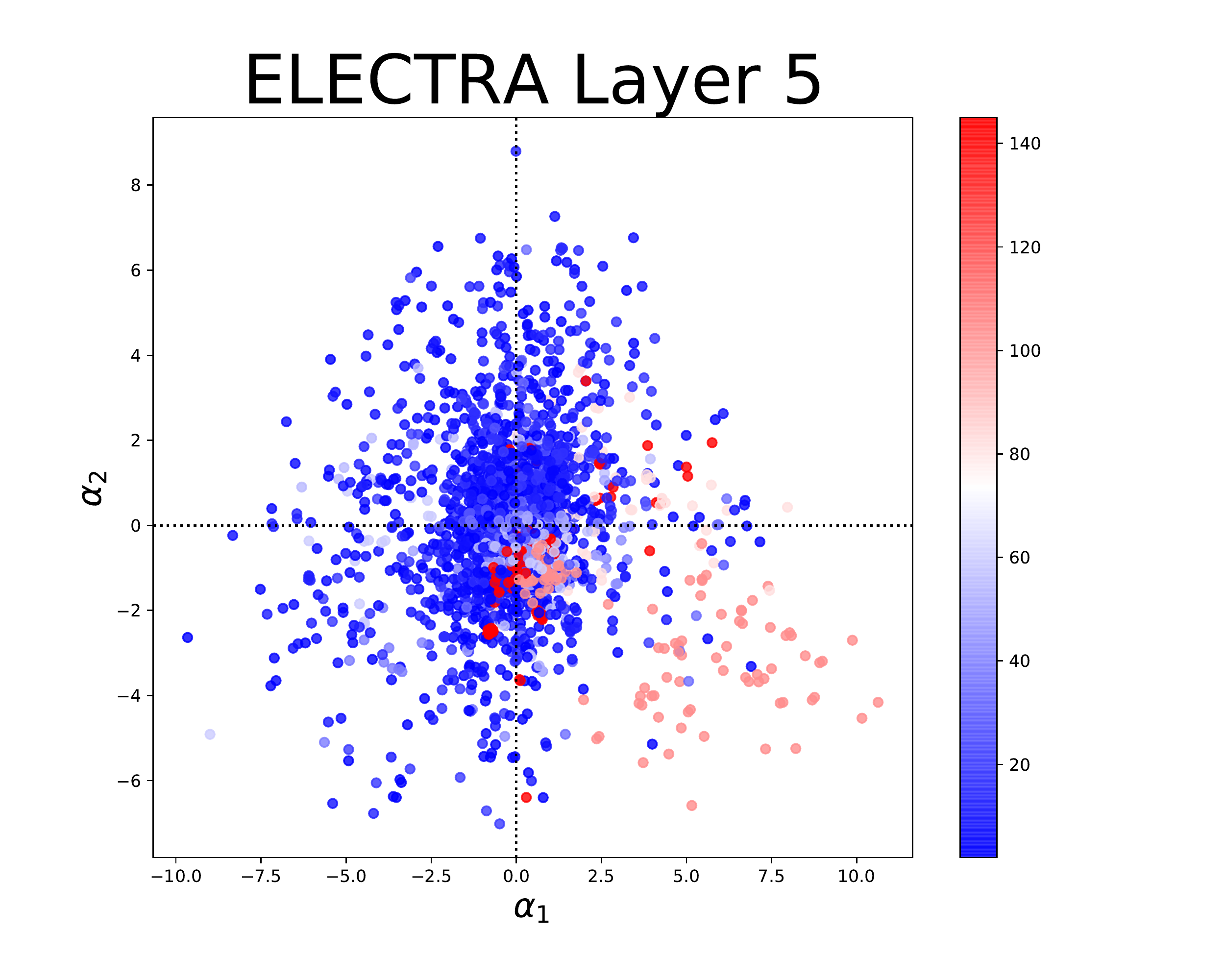}}\\
\subfloat[original]{\label{fig:m2}\includegraphics[width=0.2\linewidth]{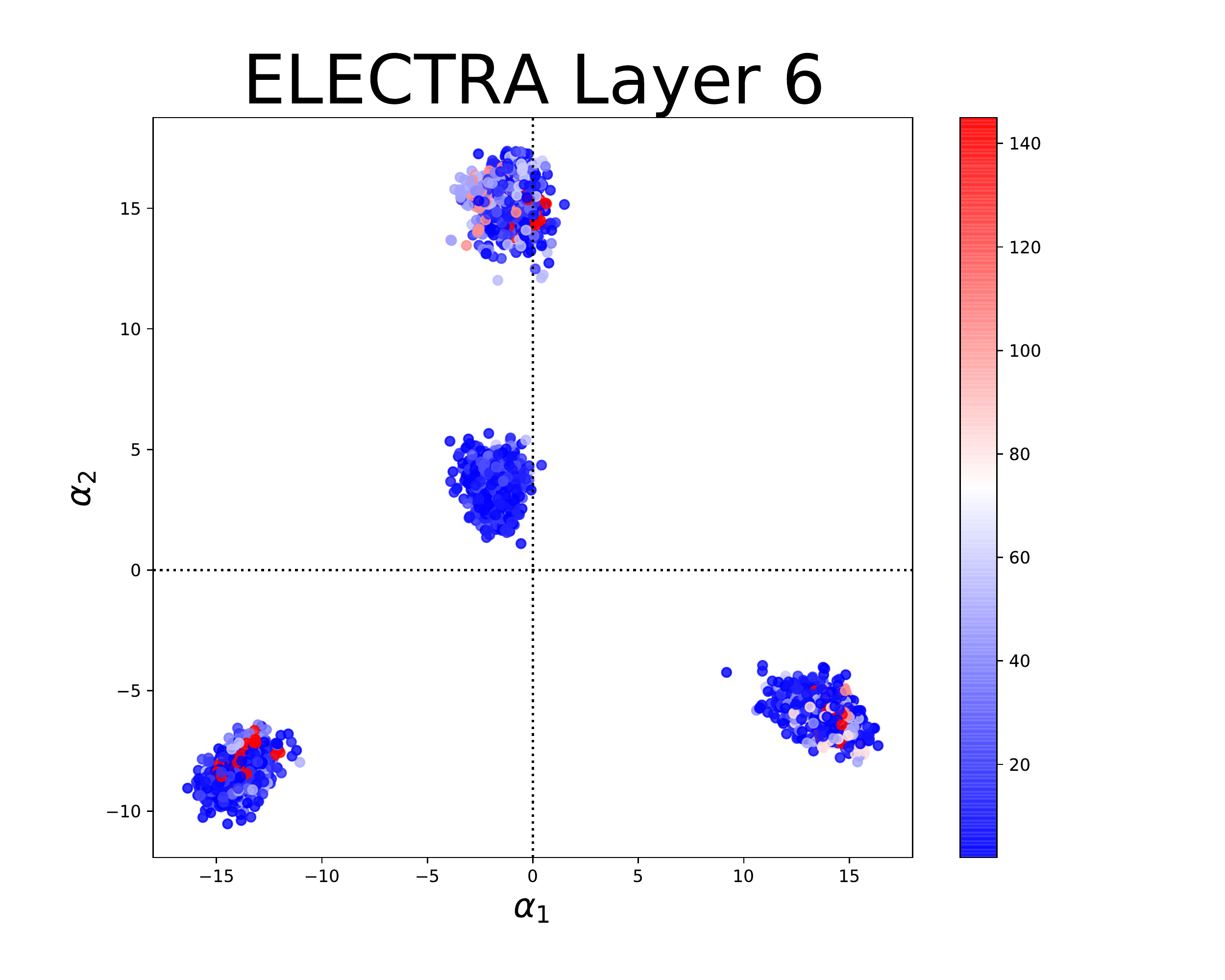}}
\subfloat[retrofitted]{\label{fig:n2}\includegraphics[width=0.2\linewidth]{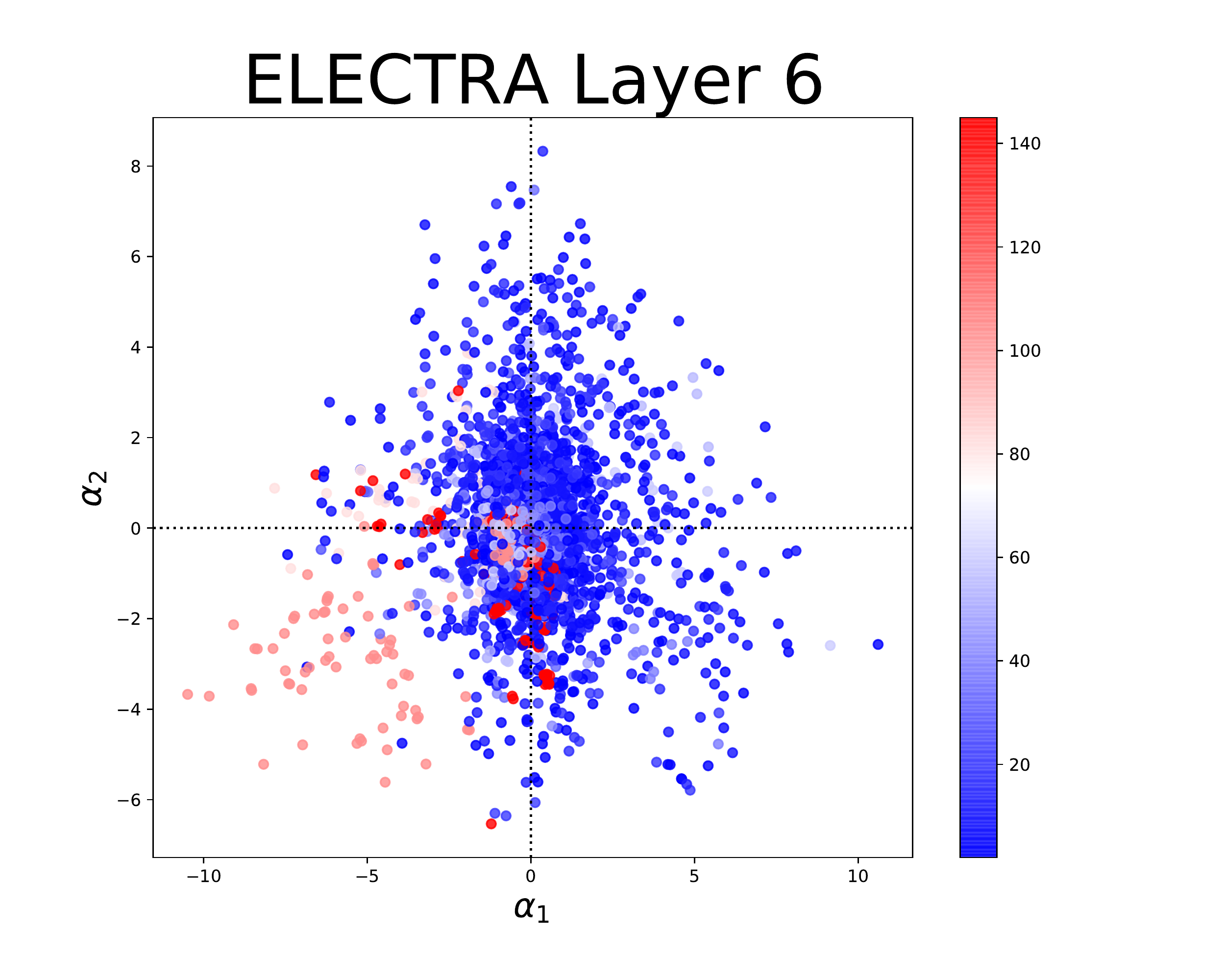}}
\subfloat[original]{\label{fig:o2}\includegraphics[width=0.2\linewidth]{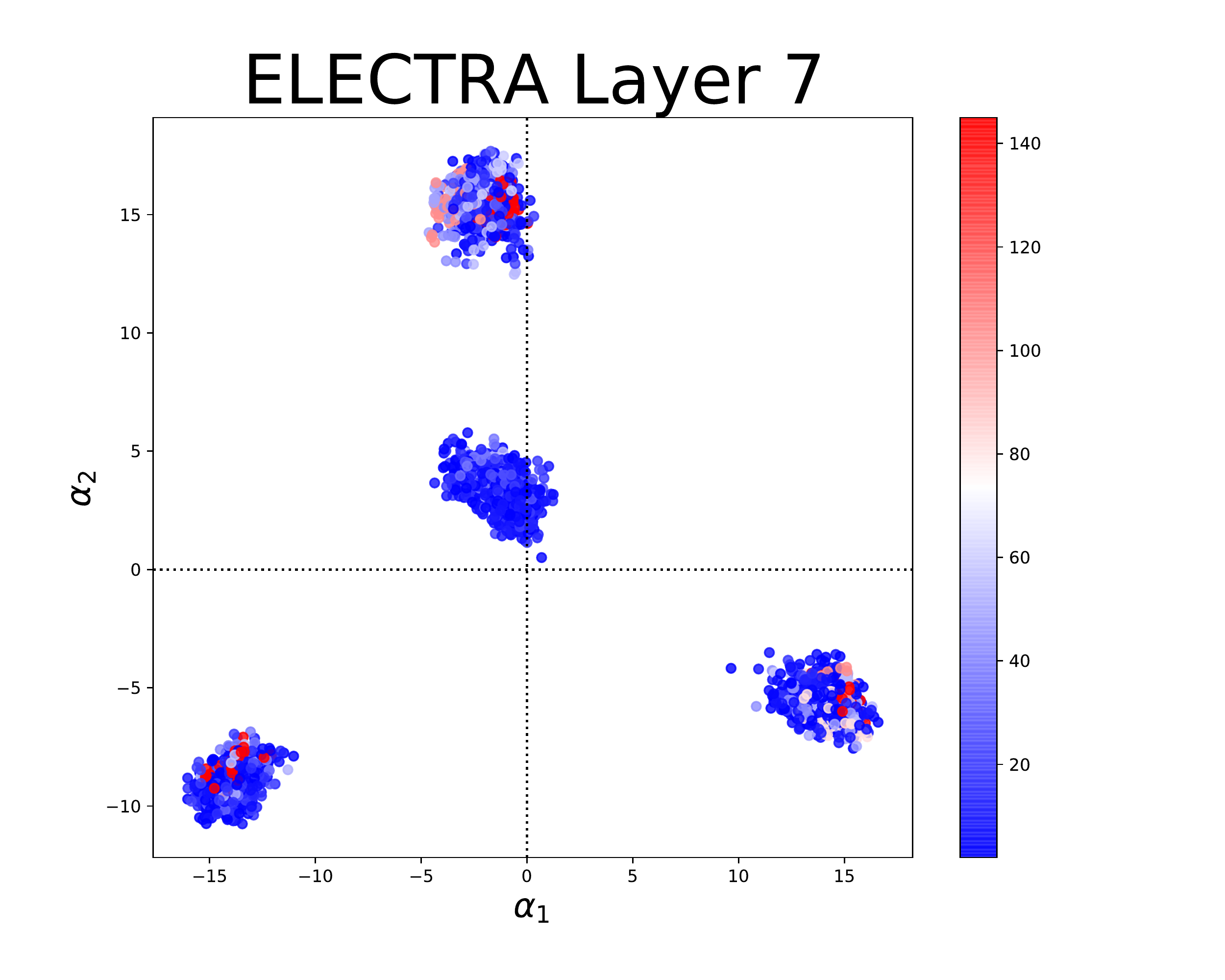}}
\subfloat[retrofitted]{\label{fig:p2}\includegraphics[width=0.2\linewidth]{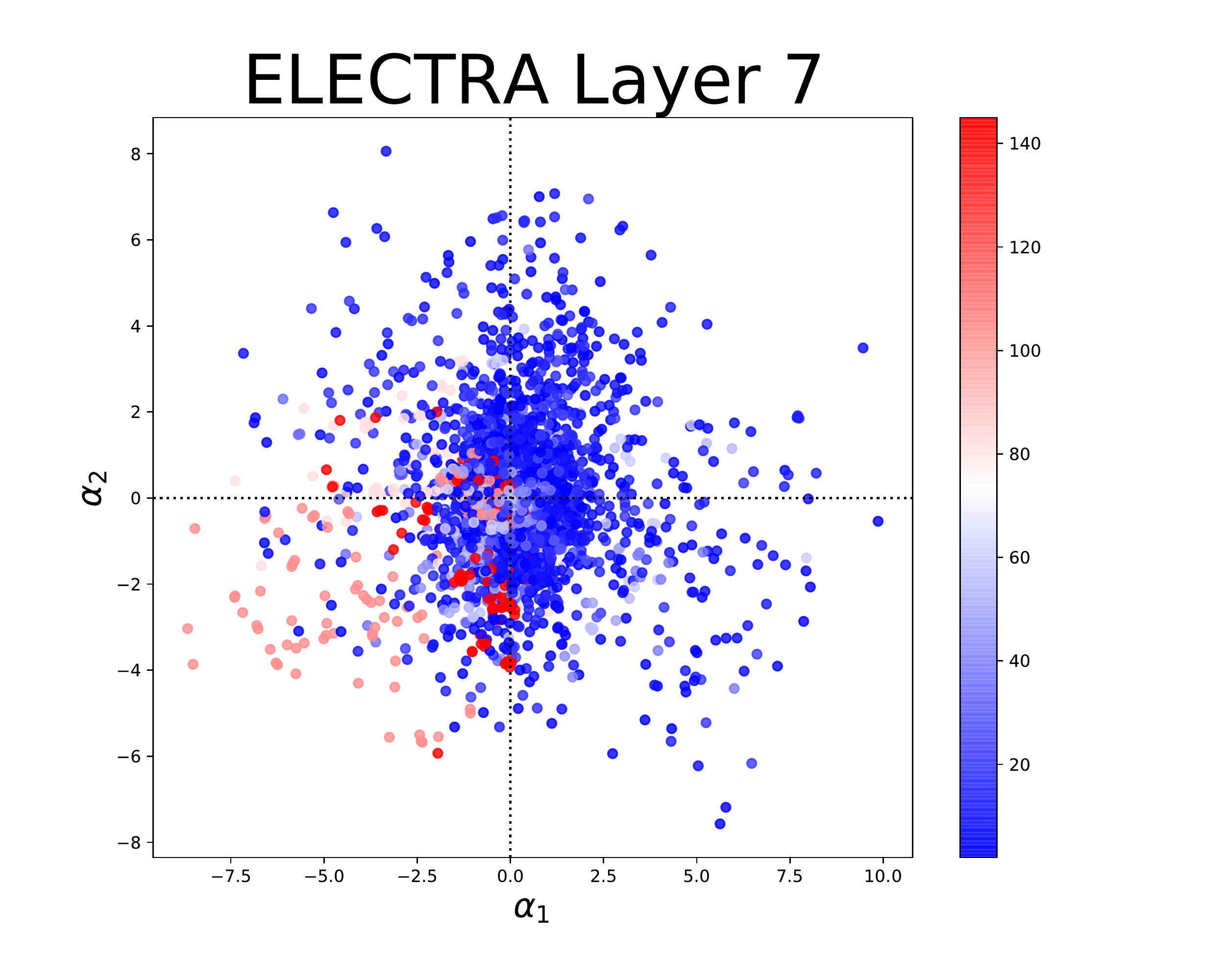}}\\
\subfloat[original]{\label{fig:q2}\includegraphics[width=0.2\linewidth]{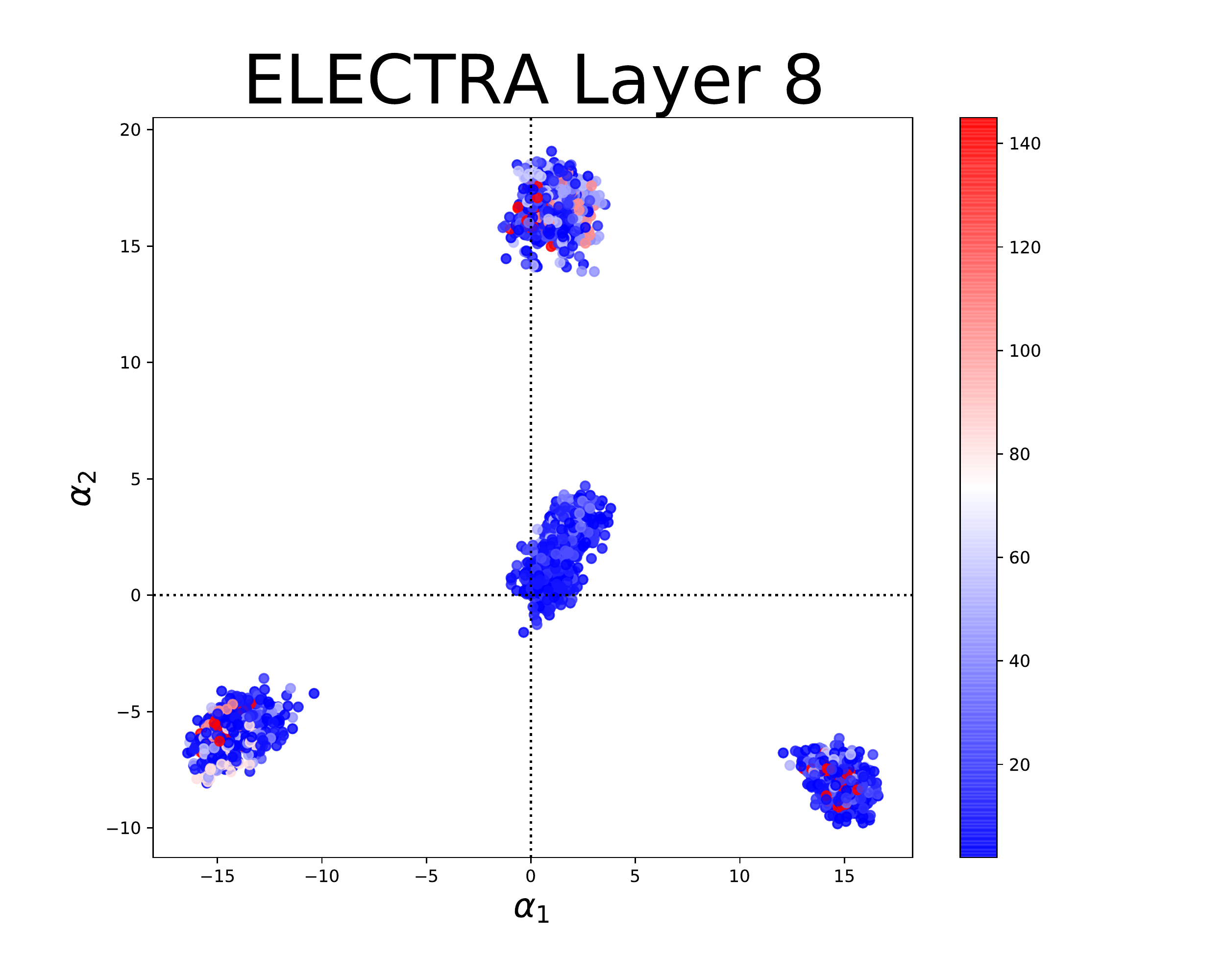}}
\subfloat[retrofitted]{\label{fig:r2}\includegraphics[width=0.2\linewidth]{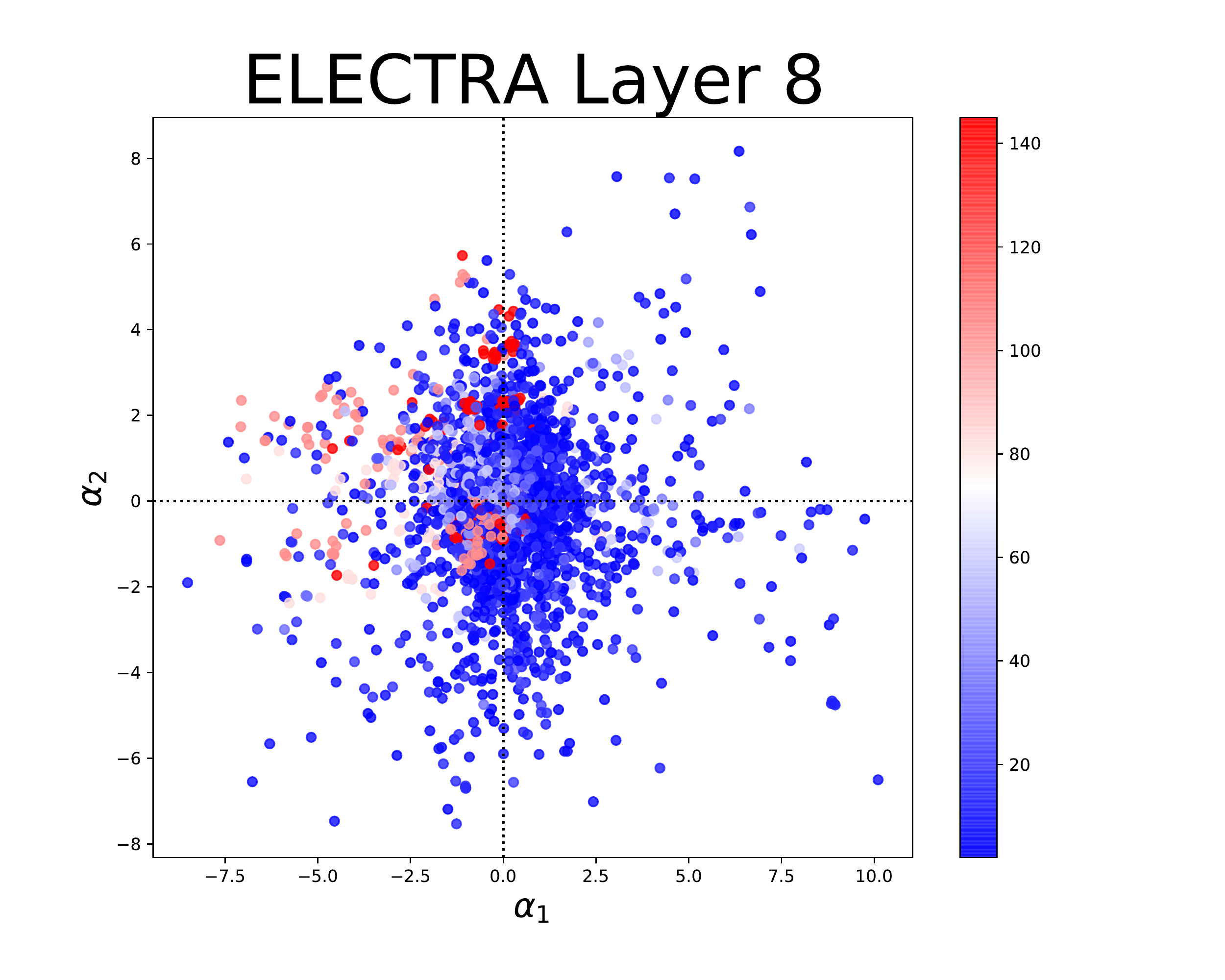}}
\subfloat[original]{\label{fig:s2}\includegraphics[width=0.2\linewidth]{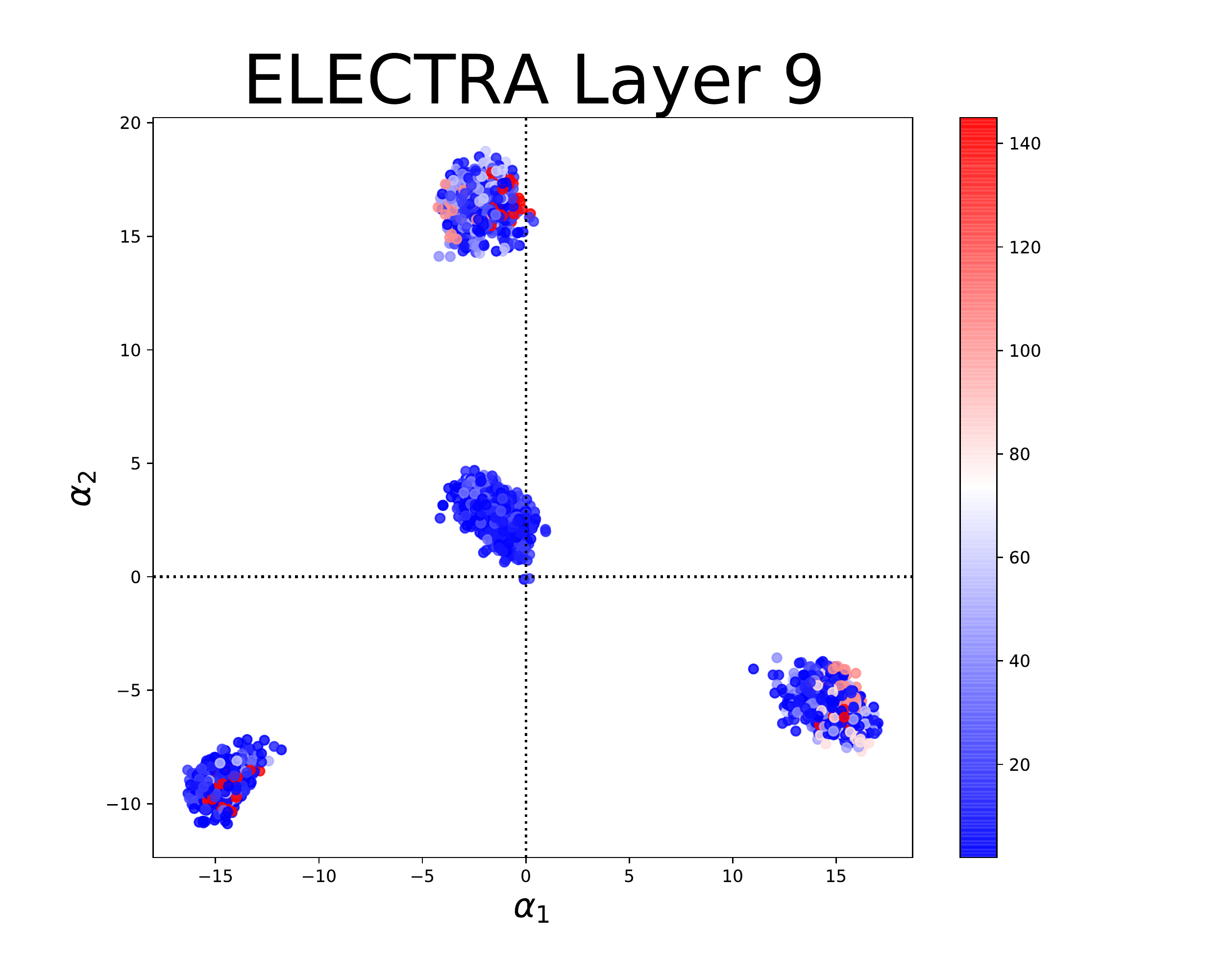}}
\subfloat[retrofitted]{\label{fig:t2}\includegraphics[width=0.2\linewidth]{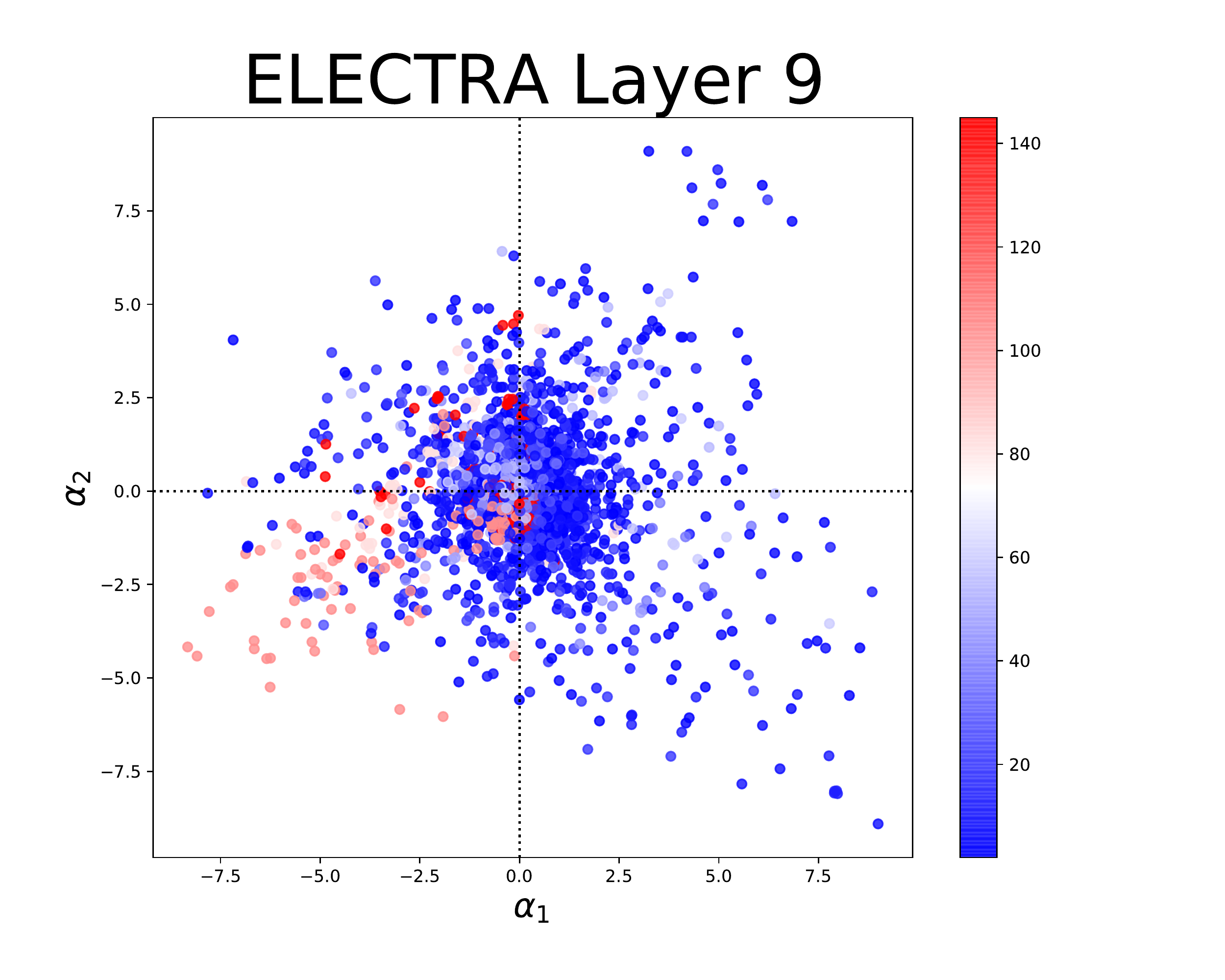}}\\
\subfloat[original]{\label{fig:u2}\includegraphics[width=0.2\linewidth]{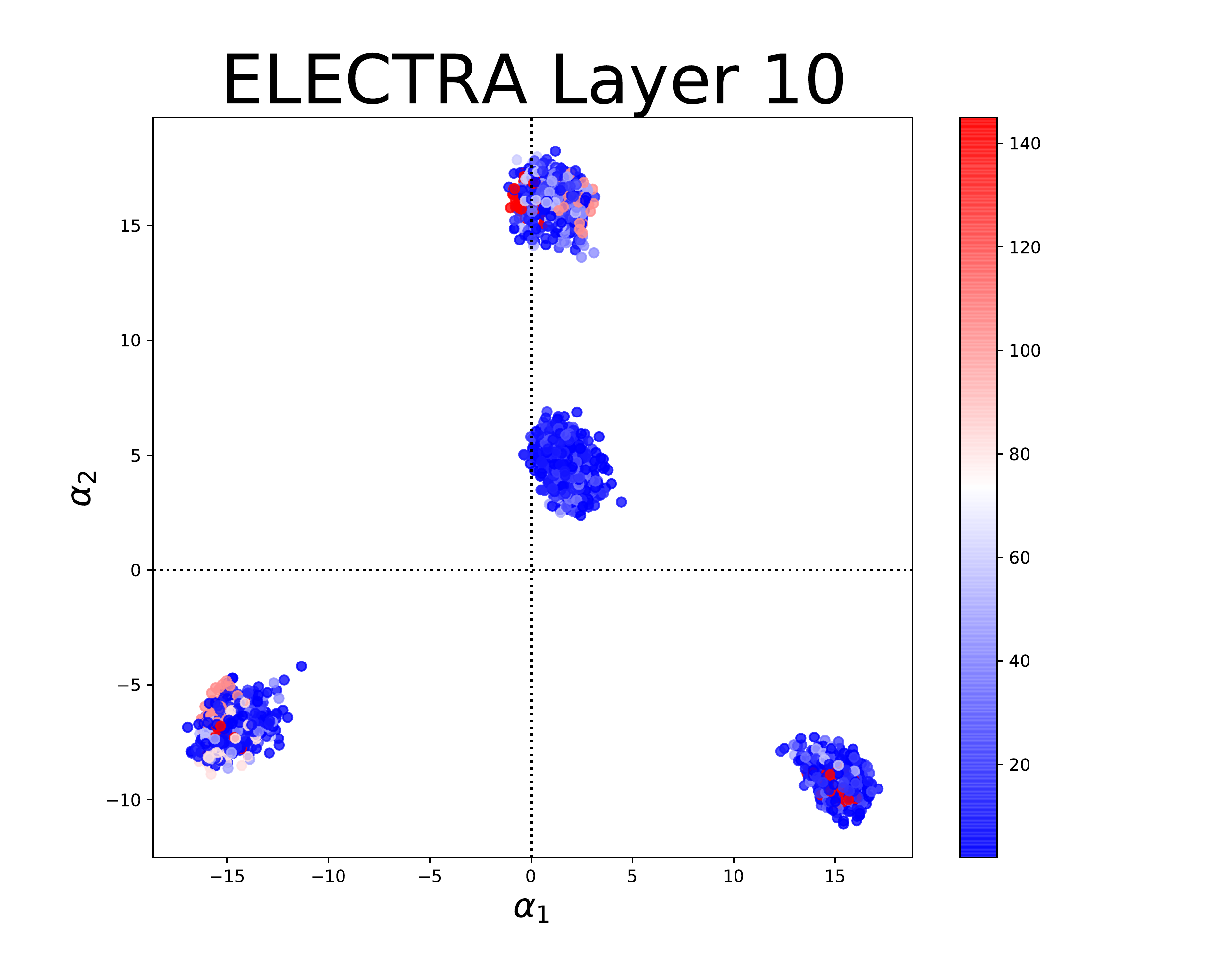}}
\subfloat[retrofitted]{\label{fig:v2}\includegraphics[width=0.2\linewidth]{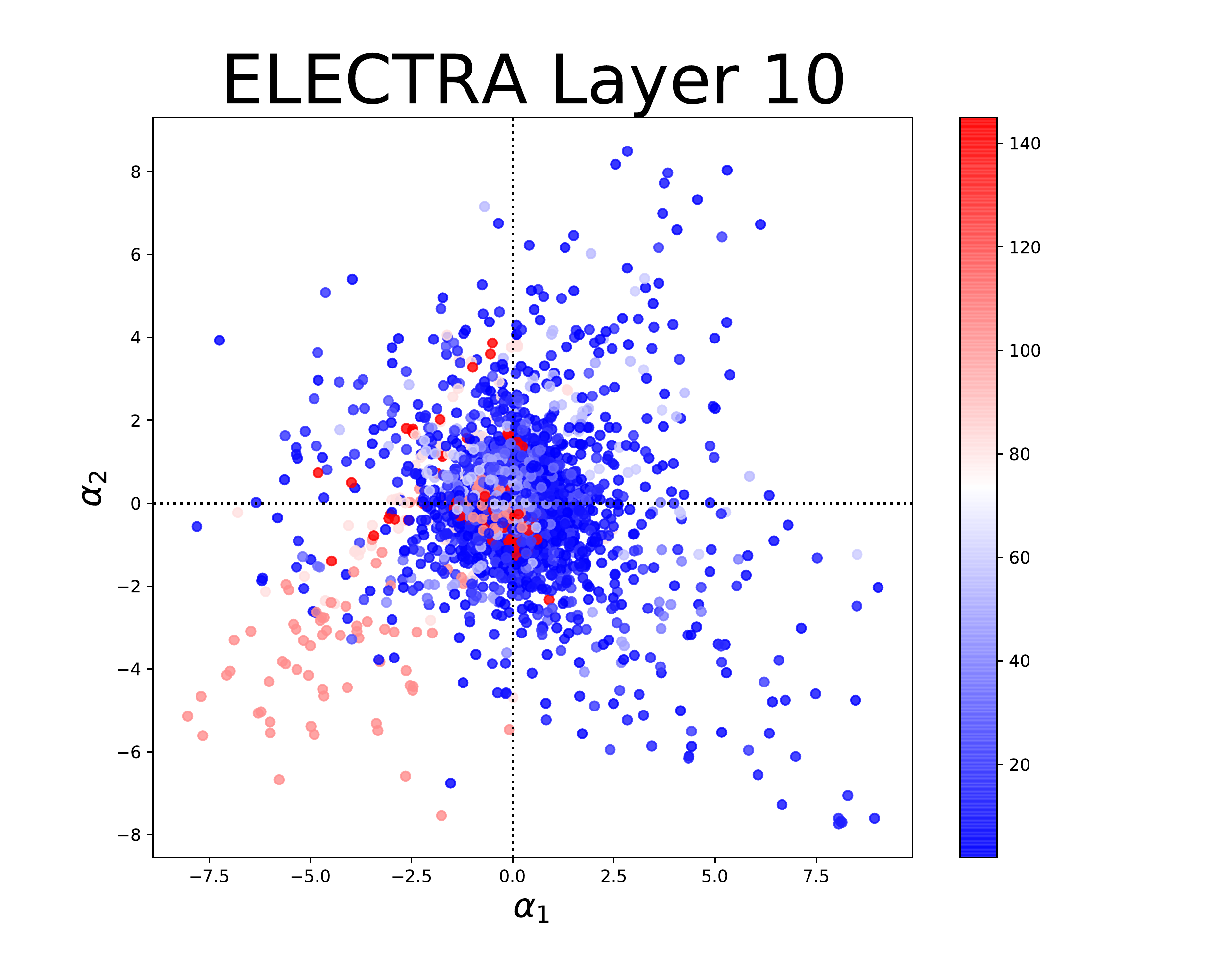}}
\subfloat[original]{\label{fig:w2}\includegraphics[width=0.2\linewidth]{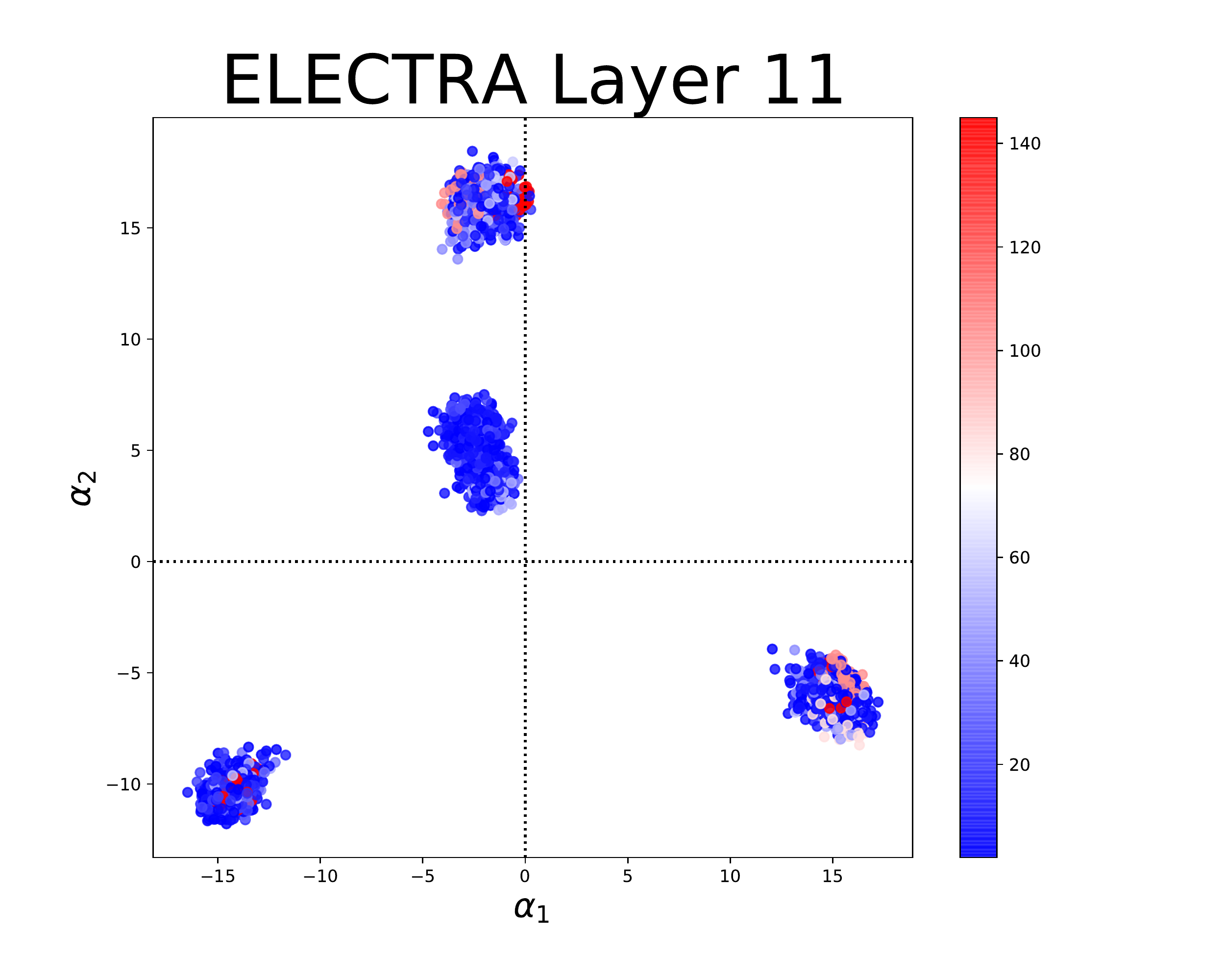}}
\subfloat[retrofitted]{\label{fig:x2}\includegraphics[width=0.2\linewidth]{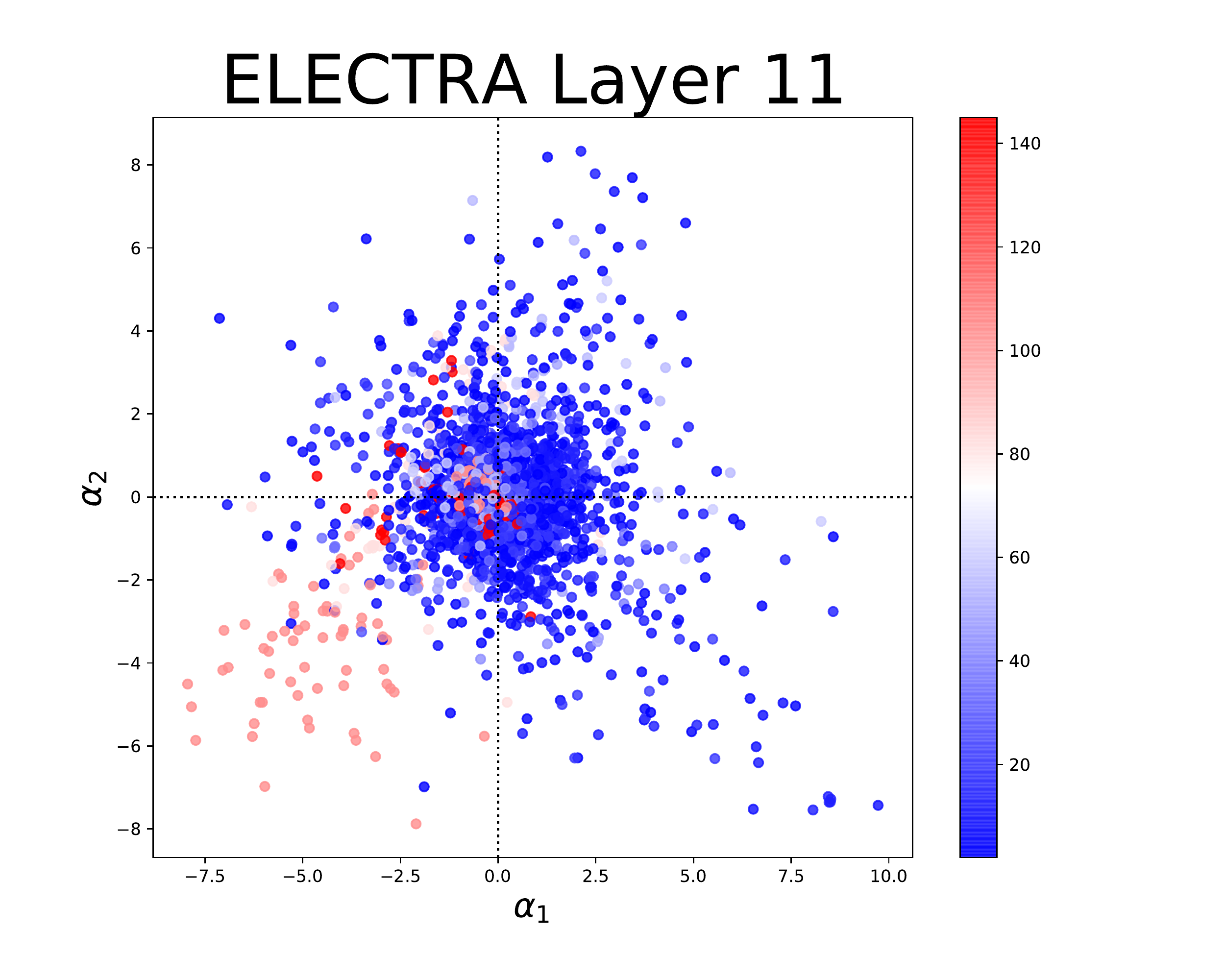}}\\
\subfloat[original]{\label{fig:y2}\includegraphics[width=0.2\linewidth]{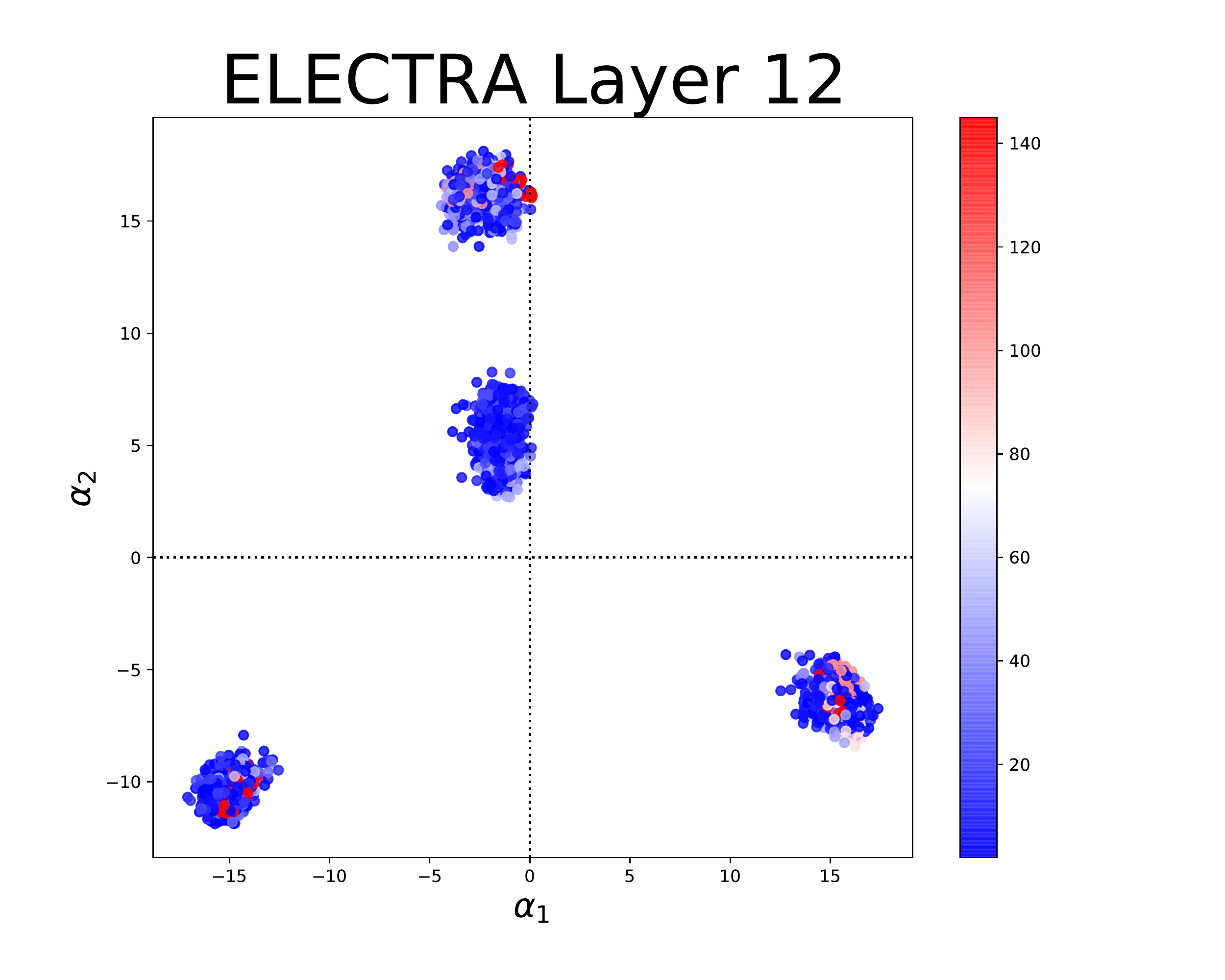}}
\subfloat[retrofitted]{\label{fig:z2}\includegraphics[width=0.2\linewidth]{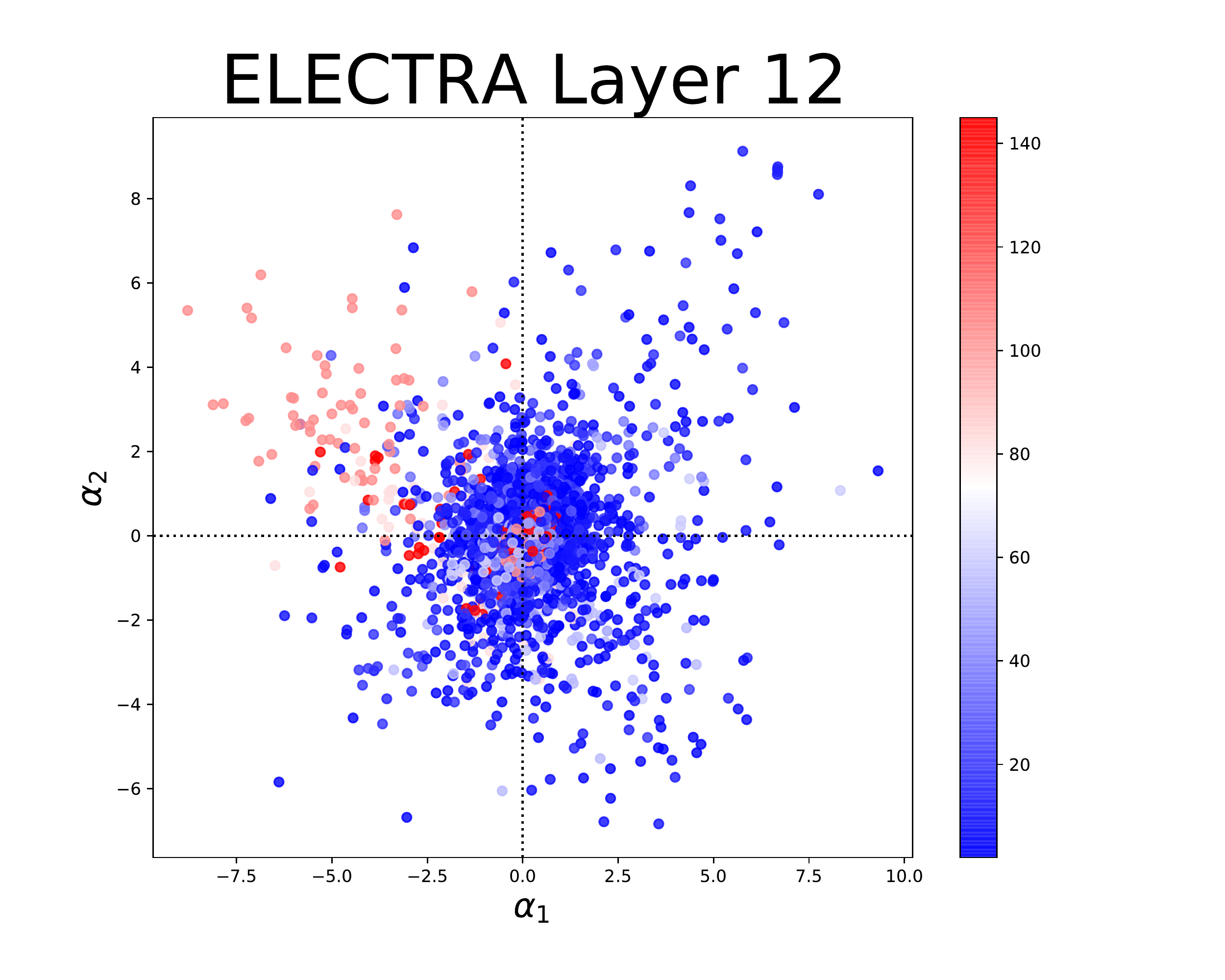}}
\caption{PCA Plots of ELECTRA Word Representations.}
\label{fig:electra_fig}
\end{figure*}

We plot distribution of word representations across the vector space, for all models across their layers. To assess whether word frequency is encoded within vector dimensions, we color code representations ranging from low frequency words (Blue) to high frequency words (Red). The plots are given in Figure \ref{fig:bert_fig} (BERT), Figure \ref{fig:gpt2_fig} (GPT-2), Figure \ref{fig:xlnet_fig} (XLNet) and Figure \ref{fig:electra_fig} (ELECTRA). We see that using LASeR post-processing ($d =1$ and hyper-parameters mentioned in the main text), anisotropy in vector space is significantly treated. For extremely anisotropic models such as GPT2 and ELECTRA, remove of the first primary component yields more uniformly spread word representations.

\end{document}